\documentclass[pdflatex,sn-nature]{sn-jnl}
\usepackage{graphicx}
\usepackage{multirow}
\usepackage{amsmath,amssymb,amsfonts}
\usepackage{amsthm}
\usepackage{comment}
\usepackage{mathrsfs}
\usepackage[ruled,vlined,linesnumbered]{algorithm2e}
\usepackage[title]{appendix}
\usepackage[table]{xcolor}
\usepackage{textcomp}
\usepackage{manyfoot}
\usepackage{booktabs}
\usepackage{algpseudocode}
\usepackage{listings}

\usepackage{bbm}
\usepackage{bibunits}
\def\BRAINFEDCOMBINED{1}
\usepackage{booktabs}
\usepackage{changepage}
\usepackage{comment}
\usepackage{longtable}
\usepackage[table]{xcolor}
\usepackage{array}
\usepackage{makecell}
\usepackage{multirow}
\usepackage{adjustbox}
\usepackage{graphicx}
\usepackage{amsmath,amssymb}
\usepackage{tabularx}
\usepackage{ragged2e}
\usepackage{tikz}
\usetikzlibrary{arrows.meta,calc,positioning}

\ifdefined\BRAINFEDCOMBINED
\else
  \usepackage[margin=1in]{geometry}
  \usepackage{caption}
  \usepackage[
    colorlinks=true,
    linkcolor=blue,
    citecolor=blue,
    urlcolor=blue,
    pdfborder={0 0 0}
  ]{hyperref}
\fi

\definecolor{cTitle}{HTML}{1F2937}
\definecolor{cAccent}{HTML}{334155}
\newcounter{dtrow}
\newcommand{\dtrown}{\stepcounter{dtrow}\thedtrow}
\definecolor{cBlue}{HTML}{2563EB}
\definecolor{cInk}{HTML}{0F172A}
\definecolor{cLine}{HTML}{CBD5E1}
\definecolor{cFill}{HTML}{F8FAFC}
\definecolor{cStep}{HTML}{EEF2FF}
\definecolor{cQC}{HTML}{ECFDF5}
\definecolor{cOut}{HTML}{FFF7ED}
\definecolor{cInput}{RGB}{246,185,59}
\definecolor{cConvA}{RGB}{168,230,207}
\definecolor{cConvB}{RGB}{129,236,236}
\definecolor{cConvC}{RGB}{120,214,198}
\definecolor{cHead}{RGB}{110,231,183}
\tikzset{
  skip/.style={dash pattern=on 3pt off 2pt, line width=0.9pt, draw=black!70},
  downarr/.style={-Latex, line width=0.9pt, draw=black!70},
  uparr/.style={-Latex, line width=1.05pt, draw=red!70},
  encnote/.style={font=\scriptsize, text=black!75, align=left},
  txt/.style={font=\scriptsize, text=black!85},
  smallbox/.style={draw=black!65, fill=white, line width=0.6pt, rounded corners=1pt,
                   inner sep=1.0pt, font=\scriptsize},
  legendbox/.style={draw=black!60, fill=black!6, rounded corners=2pt, inner sep=5pt},
}

\usepackage{tocloft}
\usepackage{multicol}
\makeatletter
\newcommand{\printfigureindex}{\@starttoc{lof}}
\newcommand{\printtableindex}{\@starttoc{lot}}
\makeatother

\ifdefined\BRAINFEDCOMBINED
\else
  \renewcommand{\figurename}{Supplementary Fig.}
  \renewcommand{\tablename}{Supplementary Table}
\fi

\usepackage{supplementary-mode}
\theoremstyle{thmstyleone}

\theoremstyle{thmstyletwo}

\theoremstyle{thmstylethree}

\begin{document}

\begin{bibunit}[sn-nature]
\newcommand{\MainBibliography}{\putbib[references]}
\title[A generalizable structural brain MRI foundation model]{A generalizable structural brain MRI foundation model via dual-priority federated pretraining}

\author[1,2,3]{\fnm{Zhen} \sur{Yu}}\email{yu\_zhen@stu.pku.edu.cn}
\author[4]{\fnm{Yang} \sur{Liu}}\email{yangliu@pku.edu.cn}
\author[5,6,7,8]{\fnm{Jianfeng} \sur{Feng}}\email{jianfeng64@gmail.com}
\author[7]{\fnm{Xiahai} \sur{Zhuang}}\email{zxh@fudan.edu.cn}
\author*[1,2,3]{\fnm{Qingchao} \sur{Chen}}\email{qingchao.chen@pku.edu.cn}
\affil[1]{\orgdiv{National Institute of Health Data Science}, \orgname{Peking University}, \orgaddress{\city{Beijing}, \postcode{100871}, \country{China}}}
\affil[2]{\orgdiv{Institute of Medical Technology}, \orgname{Peking University Health Science Center}, \orgaddress{\city{Beijing}, \postcode{100871}, \country{China}}}
\affil[3]{\orgdiv{State Key Laboratory of General Artificial Intelligence}, \orgname{Peking University}, \orgaddress{\city{Beijing}, \postcode{100871}, \country{China}}}
\affil[4]{\orgdiv{Wangxuan Institute of Computer Technology}, \orgname{Peking University}, \orgaddress{\city{Beijing}, \postcode{100871}, \country{China}}}
\affil[5]{\orgdiv{Institute of Science and Technology for Brain-Inspired Intelligence}, \orgname{Fudan University}, \orgaddress{\city{Shanghai}, \postcode{200433}, \country{China}}}
\affil[6]{\orgdiv{Key Laboratory of Computational Neuroscience and Brain-Inspired Intelligence (Fudan University)}, \orgname{Ministry of Education}, \orgaddress{\city{Shanghai}, \postcode{200433}, \country{China}}}
\affil[7]{\orgdiv{School of Data Science}, \orgname{Fudan University}, \orgaddress{\city{Shanghai}, \postcode{200433}, \country{China}}}
\affil[8]{\orgdiv{Department of Computer Science}, \orgname{University of Warwick}, \orgaddress{\city{Coventry}, \postcode{CV4 7AL}, \country{United Kingdom}}}

\abstract{
Foundation models hold promise for generalizable analysis of structural brain magnetic resonance imaging (MRI) scans across diverse populations and imaging conditions. 
However, existing models typically rely on centralized pretraining using data pooled across cohorts, which is constrained by privacy and data-governance requirements.
Moreover, larger cohorts can dominate optimization, underutilizing anatomical information from smaller cohorts and limiting generalization.
Here we present BrainFedFM, a structural brain MRI foundation model federatively pretrained on 164,707 three-dimensional scans drawn from diverse real-world data distributions and organized across 42 federated sites.
BrainFedFM uses dual-priority federated pretraining, in which spatial-priority masking emphasizes informative anatomical regions during local training, while site-priority aggregation adaptively weights contributions during global aggregation.
Across 20 downstream datasets spanning 17 classification, regression and segmentation tasks, BrainFedFM achieved the state-of-the-art performance across seven foundation models, including four centralized models, while showing consistent advantages in classification and regression tasks and strong robustness under underrepresented populations and challenging imaging conditions.
These findings demonstrate the generalizability of BrainFedFM and highlight federated pretraining as a practical strategy for developing neuroimaging foundation models from distributed data without pooling raw images.
}

\maketitle
\section{Introduction}

Structural magnetic resonance imaging (MRI) is a routine and indispensable tool for brain assessment, enabling non-invasive, high-resolution visualization of neuroanatomy~\cite{wattjes2011structural}. 
In clinical practice and neuroscience research, it is widely used to characterize structural variation across brain development and aging, as well as abnormalities associated with neurodegeneration, focal lesions and other neurological conditions~\cite{lerch2017studying,sebenius2025structural}. 
Advances in artificial intelligence have further expanded the automated analysis of structural MRI and its potential to support clinical decision-making~\cite{dorfner2025review}. 
Yet realizing this potential across real-world imaging settings remains challenging, as current deep learning approaches are constrained by scarce labelled data and limited generalizability across cohorts and sites.
Their performance can be strongly affected by differences in cohort composition, scanner hardware and acquisition protocols~\cite{marzi2024efficacy}. 
Consequently, models developed on individual datasets may capture site-specific patterns, generalize inconsistently to new cohorts and require separate training for each new task or population~\cite{leming2023challenges}. 
These limitations are acute in structural brain MRI, where clinically relevant anatomical effects are often subtle relative to normal inter-individual variation and can be obscured by systematic demographic and acquisition-related differences across sites~\cite{serre2025prediction,xiao2025mitigating}. 
More generalizable approaches are therefore needed to learn transferable representations from large, heterogeneous collections of unlabelled imaging data.

Foundation models offer a promising route toward this goal. Large-scale self-supervised pretraining can reduce reliance on extensive task-specific annotation and repeated model development while supporting transfer across diverse analysis tasks~\cite{huang2023self,wu2024voco,ma2025fully,gu2026cardiac}. 
This paradigm is well suited to structural brain MRI, where diverse analysis tasks can share a common neuroanatomical basis.
However, existing neuroimaging foundation models have largely relied on centralized pretraining, restricting participation to cohorts that can be readily pooled and shared~\cite{wald2025openmind,wang2026towards,tak2026generalizable}. Such pooling is frequently constrained by privacy, governance and data-sharing requirements~\cite{voigt2017eu,li2025challenges}. 
In addition, existing foundation model pretraining has largely emphasized scaling the volume of pooled data, often without accounting for differences in data composition or the complementary information contributed by smaller, specialized cohorts.
Consequently, when cohort sizes are highly imbalanced, optimization can be disproportionately shaped by a few large cohorts, diluting distinctive anatomical variation from underrepresented cohorts~\cite{bareja2026benchmark,zhou2026understanding}.
A generalizable foundation model should learn shared neuroanatomical structure while effectively integrating distinctive and complementary anatomical variation across distributed cohorts.

Federated learning offers a practical route to foundation model pretraining across sites without sharing raw imaging data~\cite{ren2025advances,yu2026toward}. 
However, simply combining conventional random masked pretraining with sample-size-driven aggregation of local updates is unlikely to fully exploit heterogeneous data information~\cite{yan2023label,chenunderstanding}.
Within each site, random masking treats anatomical locations as equally informative, despite pronounced regional differences in lifespan-related anatomical variation and reconstruction difficulty. Across sites, sample-size-driven aggregation can allow large cohorts to dominate global optimization, thereby underweighting updates from smaller cohorts despite their complementary anatomical information~\cite{zhou2025federated,souza2026combining}. Effective federated foundation model pretraining therefore requires coordinated prioritization at two levels: spatial-priority masking at each site directs local pretraining toward informative anatomical regions, whereas site-priority aggregation at the server adaptively weights heterogeneous site updates beyond cohort size alone.

\begin{figure}[t]
  \centering
  \includegraphics[width=1.0\textwidth]{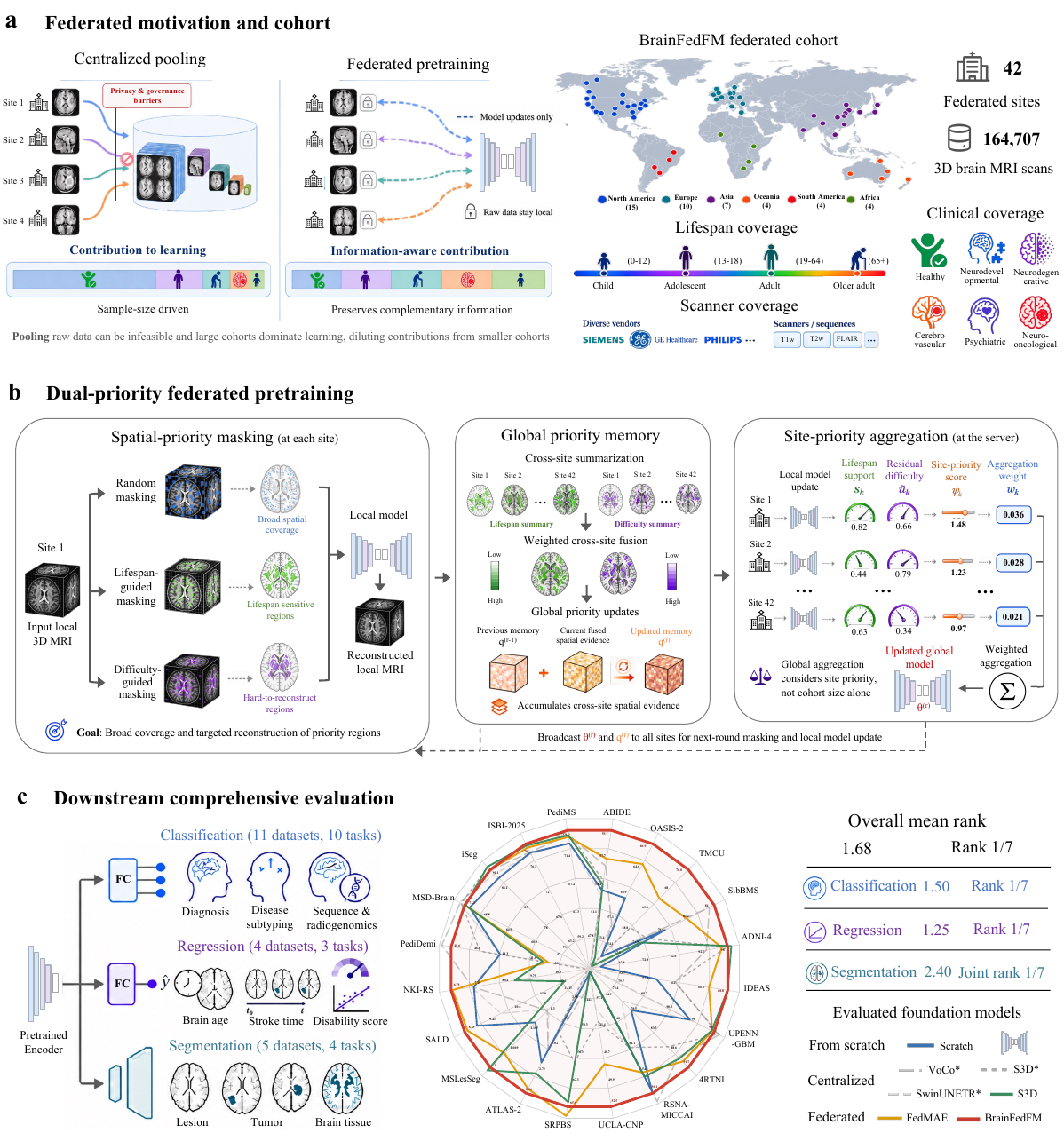}
  \vspace{0.1em}
  \caption{\textbf{Overview of BrainFedFM.}
  \label{overall}  
  \textbf{a}, Federated motivation and cohort. 
  Centralized pooling can be infeasible; BrainFedFM instead keeps raw imaging data at their local sites and exchanges only model updates, enabling federated pretraining across 42 federated sites and 164,707 brain MRI scans spanning broad geographic, lifespan, clinical and scanner diversity.
  \textbf{b}, Dual-priority federated pretraining. BrainFedFM combined spatial-priority masking at each site with site-priority aggregation at the server, linked by a global priority memory. Random, lifespan-guided and difficulty-guided masked views promote broad spatial coverage while emphasizing lifespan informative and persistently difficult regions. The server adaptively integrates site updates and returns accumulated cross-site spatial evidence to guide subsequent rounds of local pretraining.
  \textbf{c}, Downstream comprehensive evaluation. 
  The pretrained encoder was retained as the BrainFedFM foundation model and evaluated across 20 downstream datasets.
  BrainFedFM was compared with six foundation models spanning from-scratch, centralized and federated pretraining.   
  The radar plot summarizes task-wise performance, while the panel on the right reports the overall and task-family mean ranks.
  }
\end{figure}

Here we present BrainFedFM, a generalizable structural brain MRI foundation model built through dual-priority federated pretraining (Fig.~\ref{overall}). 
BrainFedFM was pretrained on 164,707 scans drawn from diverse real-world data distributions and organized across 42 federated sites, spanning substantial variation in sample scale, age distribution, clinical composition, scanner vendor and acquisition protocol.
At each site, spatial-priority masking combines lifespan-guided and difficulty-guided views to emphasize anatomically informative regions and regions that remain difficult to reconstruct, while retaining a random masking view for broad spatial coverage. 
At the server, site-priority aggregation weights site updates using complementary signals derived from lifespan support and residual reconstruction difficulty, rather than relying on cohort size alone.
Meanwhile, a global priority memory links these two levels by accumulating cross-site spatial evidence and returning it to guide subsequent rounds of local pretraining.
We benchmarked BrainFedFM across 20 downstream datasets spanning 17 classification, regression and segmentation tasks.
Among the seven foundation models spanning random initialization, centralized pretraining and federated pretraining, BrainFedFM ranked first overall (mean rank, 1.68), with a 47\% relative gain over the runner-up, and achieved the best mean ranks in classification (1.50) and regression (1.25) and segmentation (2.40).
Performance also remained strong across underrepresented populations and imaging conditions, including the best results in parkinsonism subtype classification, psychiatric differential diagnosis and pediatric MS lesion segmentation.
Together, these findings establish BrainFedFM as a generalizable foundation model and demonstrate the potential of federated pretraining to learn transferable representations from distributed data.

\section{Results}

We evaluated BrainFedFM across 20 downstream datasets representing 17 structural brain MRI analysis tasks (Supplementary Table~1). The classification benchmark comprised ABIDE~\cite{di2014autism} (autism diagnosis), OASIS-2~\cite{marcus2010open} (dementia diagnosis), TMCU~\cite{ds000174:1.0.1} (cannabis dependence classification), SibBMS~\cite{tuchinov2025sibbms} (multiple sclerosis diagnosis), ADNI-4~\cite{weiner2023increasing} (Alzheimer's disease prediction), IDEAS~\cite{taylor2025imaging} (focal epilepsy diagnosis), UPENN-GBM~\cite{bakas2022university} (MRI sequence classification), 4RTNI~\cite{WOS:000382568400386} (parkinsonism subtype classification), RSNA-MICCAI~\cite{baid2021rsna} (tumor radiogenomic prediction), UCLA-CNP~\cite{poldrack2016phenome} (psychiatric differential diagnosis) and SRPBS~\cite{tanaka2021multi} (multi-disorder psychiatric diagnosis). The regression benchmark included ATLAS-2~\cite{liew2022large} (post-stroke time estimation), MSLesSeg~\cite{guarnera2025mslesseg} (disability severity prediction), SALD~\cite{wei2018structural} (adult brain age prediction) and NKI-RS~\cite{nooner2012nki} (lifespan brain age prediction). The segmentation benchmark included PediDemi~\cite{popa2025pedidemi} (pediatric demyelinating segmentation), MSD-Brain~\cite{antonelli2022medical} (brain tumor segmentation), iSeg~\cite{sun2021multi} (infant brain tissue segmentation), ISBI-2015~\cite{carass2017longitudinal} (adult MS lesion segmentation) and PediMS~\cite{popa2025pedims} (pediatric MS lesion segmentation). 
Together, these benchmarks provide a broad evaluation of the transferability of pretrained representations across diverse downstream tasks, clinical populations and imaging settings.

To benchmark downstream transfer performance, BrainFedFM was compared with foundation models spanning publicly released centralized pretraining, data-matched centralized and federated pretraining. Scratch used randomly initialized weights with the same model architecture. SwinUNETR$^{*}$~\cite{tang2022self} (multi-task proxy learning), VoCo$^{*}$~\cite{wu2024voco} (volume-level contrastive learning) and S3D$^{*}$~\cite{wald2025revisiting} (masked image modelling) used publicly released model checkpoints from the OpenMind benchmark~\cite{wald2025openmind}.  
A separate S3D model and FedMAE~\cite{jiang2025pretraining} were pretrained on the same corpus as BrainFedFM (Supplementary Table~2), providing data-matched centralized and federated comparisons. 
All seven models used identical subject-level partitions, with 20\% of each dataset held out for testing and three-fold cross-validation performed within the remaining development set. 
For downstream task adaptation, classification and regression used newly initialized single-layer linear prediction heads, whereas segmentation used randomly initialized segmentation decoders; models underwent full-parameter fine-tuning. 
Balanced accuracy (BA), root mean squared error (RMSE) and Dice score were used as the primary metrics for classification, regression and segmentation, respectively. Detailed adaptation and evaluation procedures are described in Methods, with complete results reported in Supplementary Tables~4--23.

\subsection{Dual-priority federated pretraining supports broad downstream transfer}

\begin{figure}[h!]
  \centering
  \includegraphics[width=1.0\textwidth]{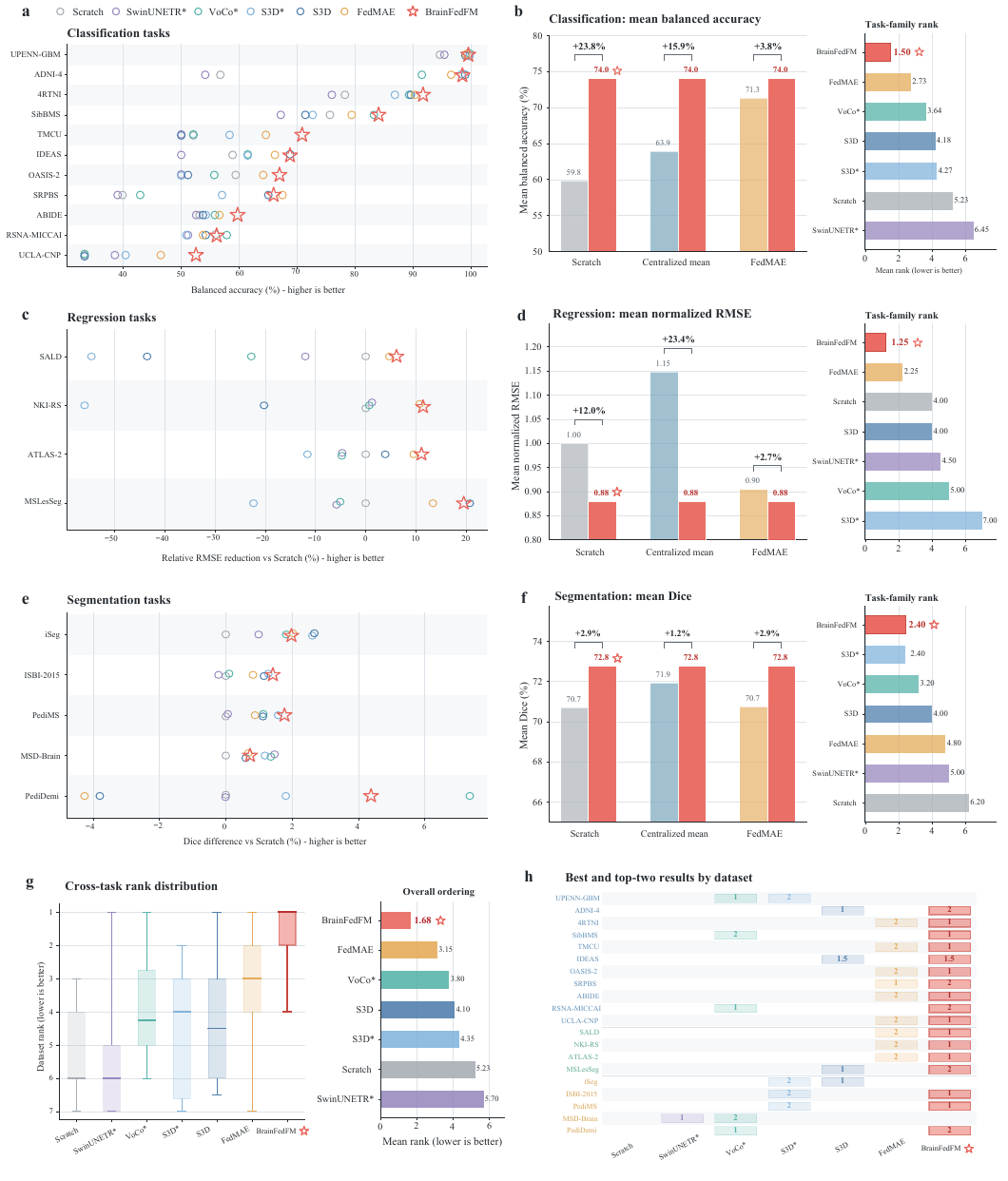}
  \caption{\textbf{Broad downstream transfer of BrainFedFM across diverse structural brain MRI analysis tasks.}
  \textbf{a}, Balanced accuracy across 11 classification datasets.
  \textbf{b}, Mean balanced accuracy relative to Scratch, the centralized-pretraining mean and FedMAE (left), and mean classification rank across all seven models (right).
  \textbf{c}, Regression performance across four datasets, expressed as relative RMSE reduction versus Scratch.
  \textbf{d}, Mean normalized RMSE relative to the three comparison groups (left) and mean regression rank (right).
  \textbf{e}, Segmentation performance across five datasets, expressed as Dice difference versus Scratch.
  \textbf{f}, Mean Dice relative to the three comparison groups (left) and mean segmentation rank (right).
  \textbf{g}, Dataset-level rank distributions across all 20 downstream datasets (left) and overall mean-rank ordering (right).
  \textbf{h}, First- and second-place results across the 20 datasets, with tied positions retained.
  The centralized-pretraining mean denotes the mean performance of SwinUNETR$^{*}$, VoCo$^{*}$, S3D$^{*}$ and data-matched S3D. 
  Higher balanced accuracy and Dice scores, and lower RMSE values and ranks, indicate better performance.}
  \label{fig_2}
\end{figure}

A central requirement of a brain MRI foundation model is that its pretrained representation transfers across clinically distinct targets and prediction paradigms.
Across the 20 dataset benchmark, BrainFedFM achieved the best overall mean rank of 1.68, followed by FedMAE (3.15), VoCo$^{*}$ (3.80), S3D (4.10), S3D$^{*}$ (4.35), Scratch (5.23) and SwinUNETR$^{*}$ (5.70) (Fig.~\ref{fig_2}g).
BrainFedFM consistently ranked among the leading foundation models across classification, regression and segmentation tasks (Fig.~\ref{fig_2}g,h), indicating that its overall advantage was not confined to a single task type. Notably, BrainFedFM maintained strong performance across clinically heterogeneous cohorts spanning diverse disease phenotypes and population groups, supporting the broad applicability of its learned representations.
BrainFedFM also outperformed the two data-matched baselines---centrally pretrained S3D and federated FedMAE---indicating that its downstream advantage was not explained by access to the pretraining corpus alone and supporting an added benefit of dual-priority federated pretraining over both centralized and conventional federated alternatives.

The most pronounced performance gains emerged in classification and regression tasks, despite substantial heterogeneity in their downstream targets. 
Classification encompassed neurological and psychiatric diagnosis, disease subtyping, MRI sequence recognition and tumour radiogenomic prediction, whereas regression included brain age estimation, post-stroke time estimation and disability severity prediction. 
Across these clinically distinct endpoints, BrainFedFM achieved the best mean rank for both classification (1.50) and regression (1.25) (Fig.~\ref{fig_2}a--d). Mean classification BA reached 74.0\%, compared with 59.8\% for Scratch, 63.9\% for the centralized-pretraining mean and 71.3\% for FedMAE, corresponding to relative gains of 23.8\%, 15.9\% and 3.8\%, respectively. 
For regression, BrainFedFM achieved a mean normalized RMSE of 0.88, compared with 1.00 for Scratch, 1.15 for the centralized-pretraining mean and 0.90 for FedMAE, corresponding to relative reductions of 12.0\%, 23.4\% and 2.7\%, respectively.
The concordant gains across categorical and continuous prediction targets indicate that the pretrained representation transferred effectively across substantially different clinical endpoints and task formulations.

Segmentation showed a narrower but still favourable performance margin across infant brain tissue, brain tumour and demyelinating lesion tasks (Fig.~\ref{fig_2}e,f). BrainFedFM achieved the highest mean Dice of 72.8\%, compared with 70.7\% for Scratch, 71.9\% for the centralized-pretraining mean and 70.7\% for FedMAE, corresponding to relative gains of 2.9\%, 1.2\% and 2.9\%, respectively, and shared the best mean rank of 2.40 with S3D$^{*}$. 
The smaller margin relative to classification and regression may reflect the different demands of dense prediction, where performance depends more directly on fine-grained spatial correspondence, boundary precision and task-specific anatomical supervision, leaving less scope for gains from transferable global representations alone. 
Nevertheless, BrainFedFM retained the strongest aggregate segmentation performance, indicating that the benefits of dual-priority federated pretraining extended beyond subject-level prediction to voxel-wise anatomical and lesion delineation. Overall, BrainFedFM led aggregate performance across all task families, with the most pronounced gains observed in classification and regression.

\subsection{Transfer advantages persist under limited labelled supervision}

\begin{figure}[h!]
  \centering
  \includegraphics[width=1.0\textwidth]{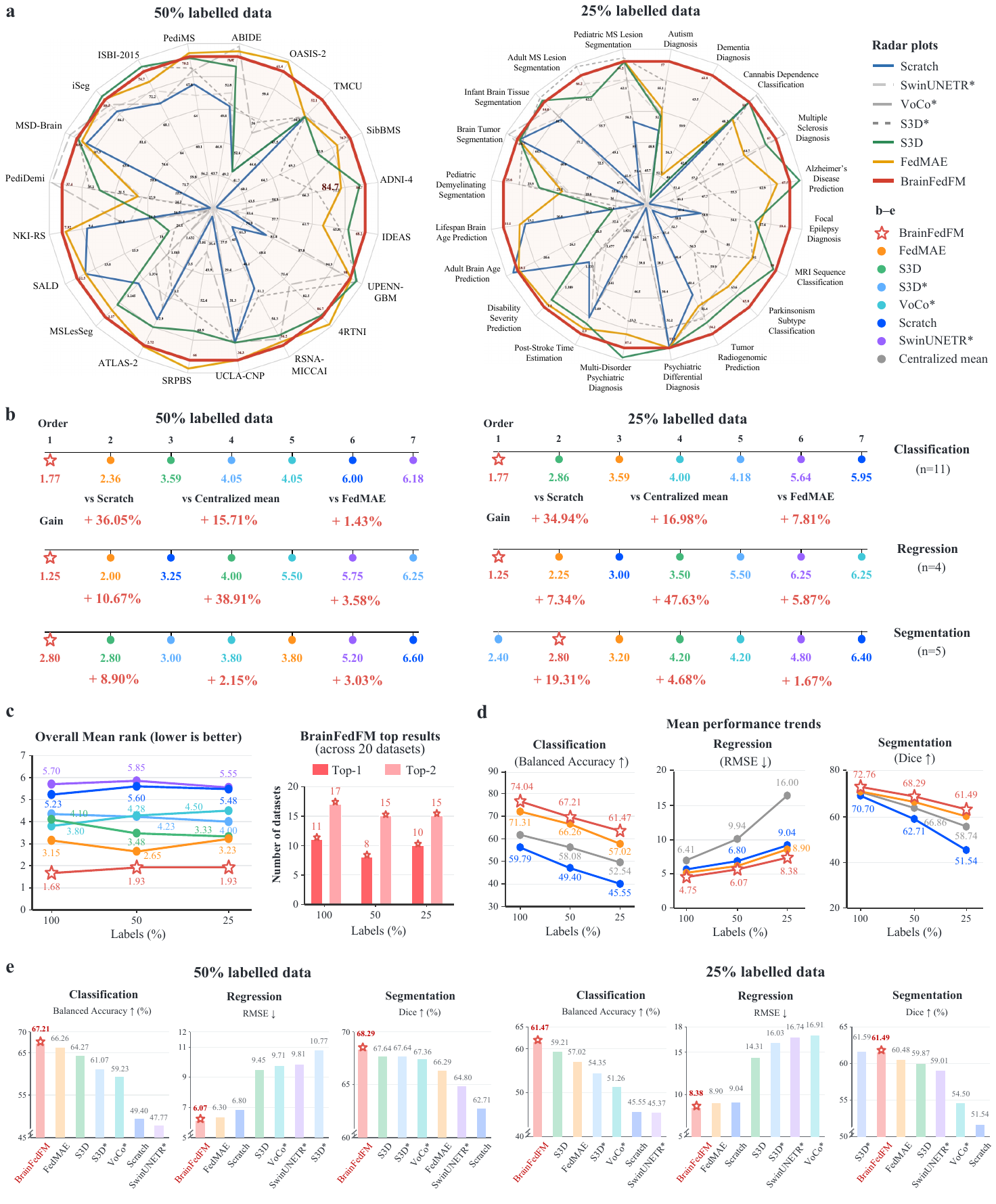}  
\caption{\textbf{BrainFedFM retains its relative advantage under reduced labelled supervision.}
\textbf{a}, Radar plots summarizing task-wise performance across the 20 downstream datasets using 50\% (left) and 25\% (right) of the labelled development data.
\textbf{b}, Mean ranks within classification, regression and segmentation at 50\% (left) and 25\% (right) labelled data. Relative improvements of BrainFedFM over Scratch, the centralized pretraining mean and FedMAE are shown below each ranking.
\textbf{c}, Overall mean ranks of all seven models across the 100\%, 50\% and 25\% labelled-data settings (left), and the numbers of datasets on which BrainFedFM ranked first or within the top two at each label fraction (right).
\textbf{d}, Changes in mean classification, regression and segmentation performance as the labelled-data fraction decreases from 100\% to 25\%.
\textbf{e}, Mean task-family performance of all seven models at 50\% (left) and 25\% (right) labelled data. The labelled-data fractions were varied only within the development set, while the held-out test sets remained fixed.}
\label{fig_3}
\end{figure}

A major promise of foundation models is to reduce reliance on costly task-specific annotation, a common challenge in medical imaging where expert-labelled data are often difficult to obtain at scale. 
To test whether BrainFedFM retained effective transfer with limited downstream supervision, we trained the downstream models using only 50\% or 25\% of the labelled development data while keeping the held-out test sets fixed. 
BrainFedFM retained the best overall rank of 1.93 at both label fractions and ranked within the top two on 15 of 20 datasets in each setting (Fig.~\ref{fig_3}c). 
The task-wise radar plots further showed that this overall advantage was maintained across classification, regression and segmentation under reduced-label settings (Fig.~\ref{fig_3}a). 
Together, these results demonstrate that BrainFedFM preserved strong downstream transfer despite substantial reductions in task-specific annotation, supporting label-efficient adaptation across heterogeneous clinical and imaging-analysis tasks.

The advantage was particularly stable for classification and regression (Fig.~\ref{fig_3}b,d,e). In classification, BrainFedFM maintained the best rank of 1.77 at both 50\% and 25\% labelled data, with mean BA values of 67.21\% and 61.47\% and corresponding mean AUROC values of 77.00\% and 71.55\%, respectively. Relative to Scratch, the centralized-pretraining mean and FedMAE, the relative BA gains were 36.05\%, 15.71\% and 1.43\% at 50\%, and 34.94\%, 16.98\% and 7.81\%, respectively, at 25\%. Regression showed an equally stable ranking, with BrainFedFM retaining the best rank of 1.25 at both label fractions. Mean RMSE was 6.07 at 50\% and 8.38 at 25\% labelled data, with corresponding mean MAE values of 4.69 and 7.01, respectively. The relative RMSE reductions were 10.67\%, 38.91\% and 3.58\% compared with Scratch, the centralized-pretraining mean and FedMAE at 50\%, and 7.34\%, 47.63\% and 5.87\%, respectively, at 25\%. 
Notably, as annotation became more restricted, the performance margins over the centralized and federated pretrained comparators were preserved and, for both task families, became more pronounced at the 25\% setting. 
These results suggest that the transferable representation learned by BrainFedFM remained particularly effective when task-specific supervision was scarce.

Voxel-wise dense prediction was more sensitive to label reduction, but the benefit of pretraining remained evident (Fig.~\ref{fig_3}b,d,e). 
BrainFedFM maintained a mean segmentation rank of 2.80 at both reduced-label settings and achieved the highest mean Dice of 68.29\% at 50\% labelled data. At 25\%, its mean Dice of 61.49\% closely matched the leading S3D$^{*}$ result of 61.59\%, while remaining above Scratch, the centralized-pretraining mean and FedMAE. The corresponding relative Dice gains were 8.90\%, 2.15\% and 3.03\% at 50\% labelled data and 19.31\%, 4.68\% and 1.67\% at 25\%, respectively. In addition, mean normalized surface Dice (NSD) was 76.17\% and 69.10\% at 50\% and 25\% labelled data, respectively, compared with 73.60\% and 67.32\% for FedMAE.
The greater sensitivity of segmentation is consistent with its stronger dependence on dense task-specific spatial annotation, yet BrainFedFM remained competitive at the most label-constrained setting. 
Collectively, these findings show that BrainFedFM preserved strong downstream transfer under substantial annotation constraints, with particularly stable advantages for clinical prediction tasks and sustained benefits for annotation-intensive dense prediction.

\subsection{Reproducible transfer across independent cohorts for corresponding clinical tasks}

\begin{figure}[h!]
\centering
\includegraphics[width=1.0\textwidth]{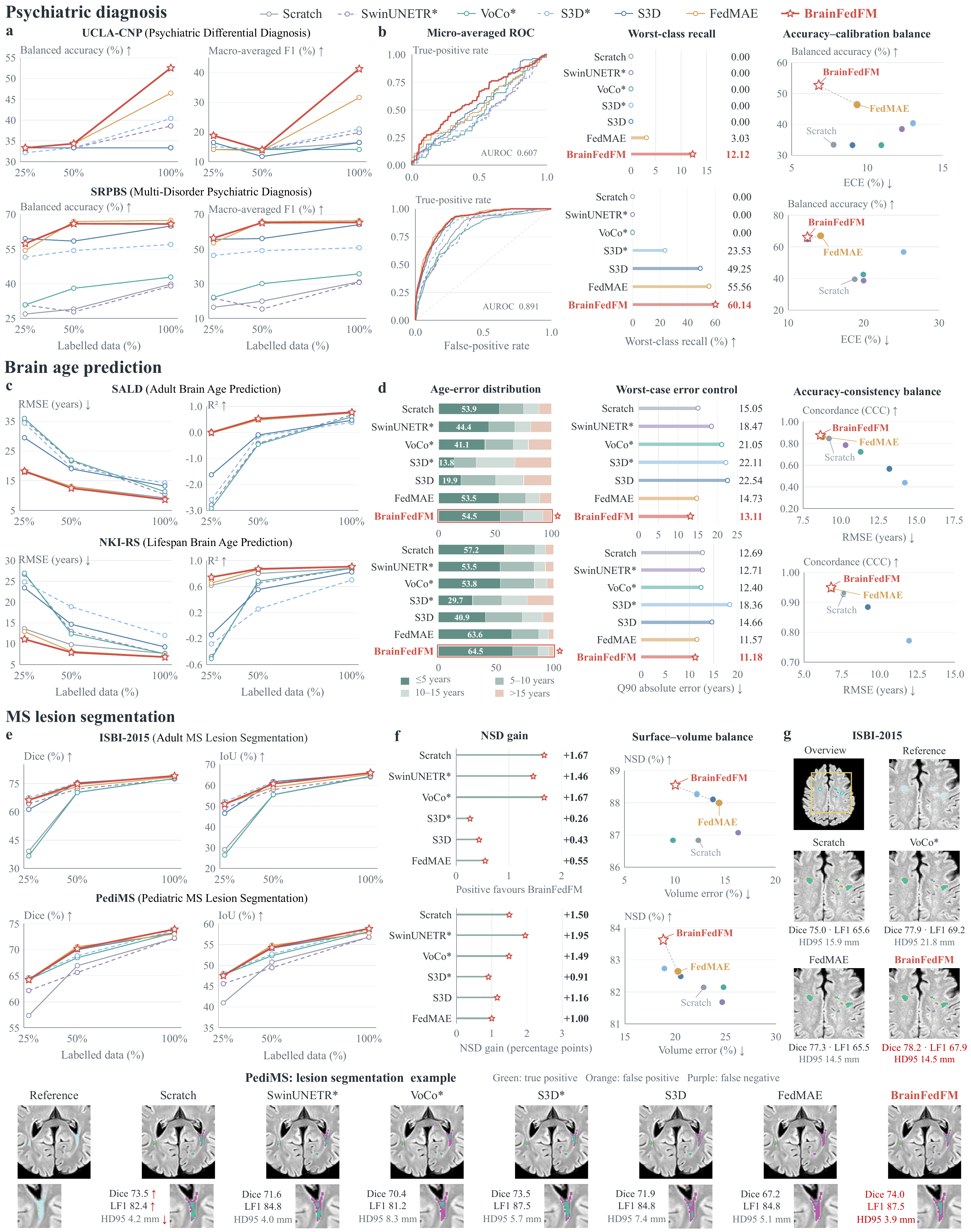} 
\par\vspace{2.5pt}
\caption{\textbf{BrainFedFM demonstrates reproducible cross-cohort transfer across corresponding clinical tasks.}
\textbf{a,b}, Psychiatric diagnosis on UCLA-CNP and SRPBS, showing balanced accuracy and macro-averaged F1, together with micro-averaged ROC curves, worst-class recall and accuracy--calibration balance.
\textbf{c,d}, Brain age prediction on SALD and NKI-RS, showing RMSE and $R^2$, together with age-error distributions, Q90 absolute error and RMSE--CCC relationships.
\textbf{e,f}, MS lesion segmentation on ISBI-2015 and PediMS, showing Dice and IoU, together with NSD gain and surface--volume balance.
\textbf{g}, Representative MS lesion segmentation results, showing reference annotations, model predictions and corresponding prediction errors.
Primary metrics are reported across all labelled-data settings; detailed analyses use 100\% labelled data.
Q90, 90th-percentile absolute error; CCC, concordance correlation coefficient; ECE, expected calibration error; NSD, normalized surface Dice; LF1, lesion F1; and HD95, 95th-percentile Hausdorff distance.}
\label{fig_4}
\end{figure}

For a foundation model to be clinically useful, performance gains observed in an aggregate benchmark should remain reproducible when corresponding clinical problems are examined in independent cohorts with different data characteristics.
We therefore evaluated three paired clinical settings: psychiatric diagnosis in UCLA-CNP and SRPBS, brain age prediction in SALD and NKI-RS, and MS lesion segmentation in ISBI-2015 and PediMS. 
Each foundation model was independently fine-tuned and evaluated within each cohort using the same protocol, allowing us to assess whether the transfer benefit of its pretrained representation was consistently reproduced. 
Beyond aggregate task-level performance, we further examined clinically informative aspects of model behaviour, including class-specific recognition and calibration for psychiatric diagnosis, the distribution and upper tail of age-prediction errors, and surface and lesion-volume agreement for MS segmentation (Fig.~\ref{fig_4}).

Psychiatric diagnosis was examined across cohorts with distinct diagnostic spectra and clinical compositions. 
UCLA-CNP distinguished attention-deficit/hyperactivity disorder, schizophrenia and bipolar disorder, whereas SRPBS included healthy controls, autism spectrum disorder, major depressive disorder and schizophrenia-spectrum disorder (Fig.~\ref{fig_4}a,b). 
At 100\% labelled data, BrainFedFM achieved the highest balanced accuracy (BA) of 52.53\% and macro-averaged F1 (mF1) of 41.23\% on UCLA-CNP, exceeding FedMAE by 6.06 and 9.66 percentage points, respectively, with a macro-averaged AUROC of 70.50\%. 
On SRPBS, BrainFedFM achieved a BA of 65.92\% and an AUROC of 87.72\%, close to the FedMAE results; under 25\% labelled data, BrainFedFM achieved the highest mF1 of 56.41\%, compared with 53.60\% for FedMAE. 
BrainFedFM also achieved the highest worst-class recall in both cohorts, reaching 12.12\% on UCLA-CNP and 60.14\% on SRPBS, compared with 3.03\% and 55.56\% for FedMAE, respectively, while yielding the lowest expected calibration error (ECE), at 6.79\% and 12.55\%. These results show that its transfer advantage persisted across distinct psychiatric diagnostic settings and extended from aggregate discrimination to class-specific recognition and calibration.

Brain age prediction examined whether the pretrained representation retained transferable age-related structural information across markedly different age ranges, from the adult SALD cohort (19--80 years) to the broader lifespan NKI-RS cohort (0--85 years) (Fig.~\ref{fig_4}c,d).
At 100\% labelled data, BrainFedFM achieved the lowest RMSE in both cohorts, at 8.65 years on SALD and 6.80 years on NKI-RS, together with the highest $R^2$ and concordance correlation coefficient (CCC). 
Its advantage extended beyond average error: 54.5\% of predictions on SALD and 64.5\% on NKI-RS fell within five years of chronological age, the largest proportions among the evaluated models. 
BrainFedFM also achieved the lowest 90th-percentile absolute error (Q90), at 13.11 and 11.18 years, compared with 14.73 and 11.57 years for FedMAE, indicating better control of large individual prediction errors. 
On NKI-RS, this advantage persisted under reduced supervision; at 25\% labelled data, BrainFedFM retained the lowest RMSE of 11.10 years, compared with 12.93 years for FedMAE and 23.42--26.94 years for the centralized-pretraining models. These findings show that BrainFedFM's advantage was reproduced across adult and lifespan populations and was evident in both overall prediction accuracy and control of upper-tail errors.

MS lesion segmentation examined whether the transfer advantage extended across distinct disease populations, from the adult ISBI-2015 cohort to the paediatric PediMS cohort, while preserving the same lesion-delineation target (Fig.~\ref{fig_4}e--g).
At 100\% labelled data, BrainFedFM achieved the highest Dice scores in both cohorts, reaching 78.97\% on ISBI-2015 and 73.93\% on PediMS, together with the highest IoU scores of 65.97\% and 58.81\%, respectively.
Its advantage extended beyond overlap accuracy to lesion-boundary agreement, with normalized surface Dice (NSD) exceeding all six comparators by 0.26--1.67 percentage points on ISBI-2015 and 0.91--1.95 percentage points on PediMS.
BrainFedFM also achieved lower absolute relative lesion-volume errors than FedMAE in both cohorts, at 10.04\% versus 14.40\% on ISBI-2015 and 18.80\% versus 20.26\% on PediMS.
The qualitative examples further illustrated the spatial pattern of residual prediction errors; in the PediMS case, BrainFedFM achieved a Dice score of 74.0\% and an HD95 of 3.9~mm, compared with 67.2\% and 5.1~mm for FedMAE (Fig.~\ref{fig_4}g).
These findings show that BrainFedFM's segmentation advantage was reproduced across adult and paediatric MS cohorts and was evident across lesion overlap, boundary agreement and volume estimation.

\subsection{Pretrained representations support parameter-efficient downstream adaptation}

In the analyses above, pretrained foundation models were adapted to downstream tasks through full-parameter fine-tuning.
For a foundation model to be practically useful, however, its pretrained representation should remain effective when downstream adaptation is constrained by limited computational resources or scarce task-specific annotations.
These constraints are particularly relevant to structural brain MRI, where new clinical applications may not support extensive task-specific optimization and may provide only a small number of expert-labelled examples.
We therefore evaluated BrainFedFM under frozen-encoder adaptation and extreme few-shot adaptation across multiple sclerosis diagnosis (SibBMS), parkinsonism subtype classification (4RTNI) and Alzheimer's disease prediction (ADNI-4) tasks.

\begin{figure}[t]
  \centering
  \includegraphics[width=1.0\textwidth]{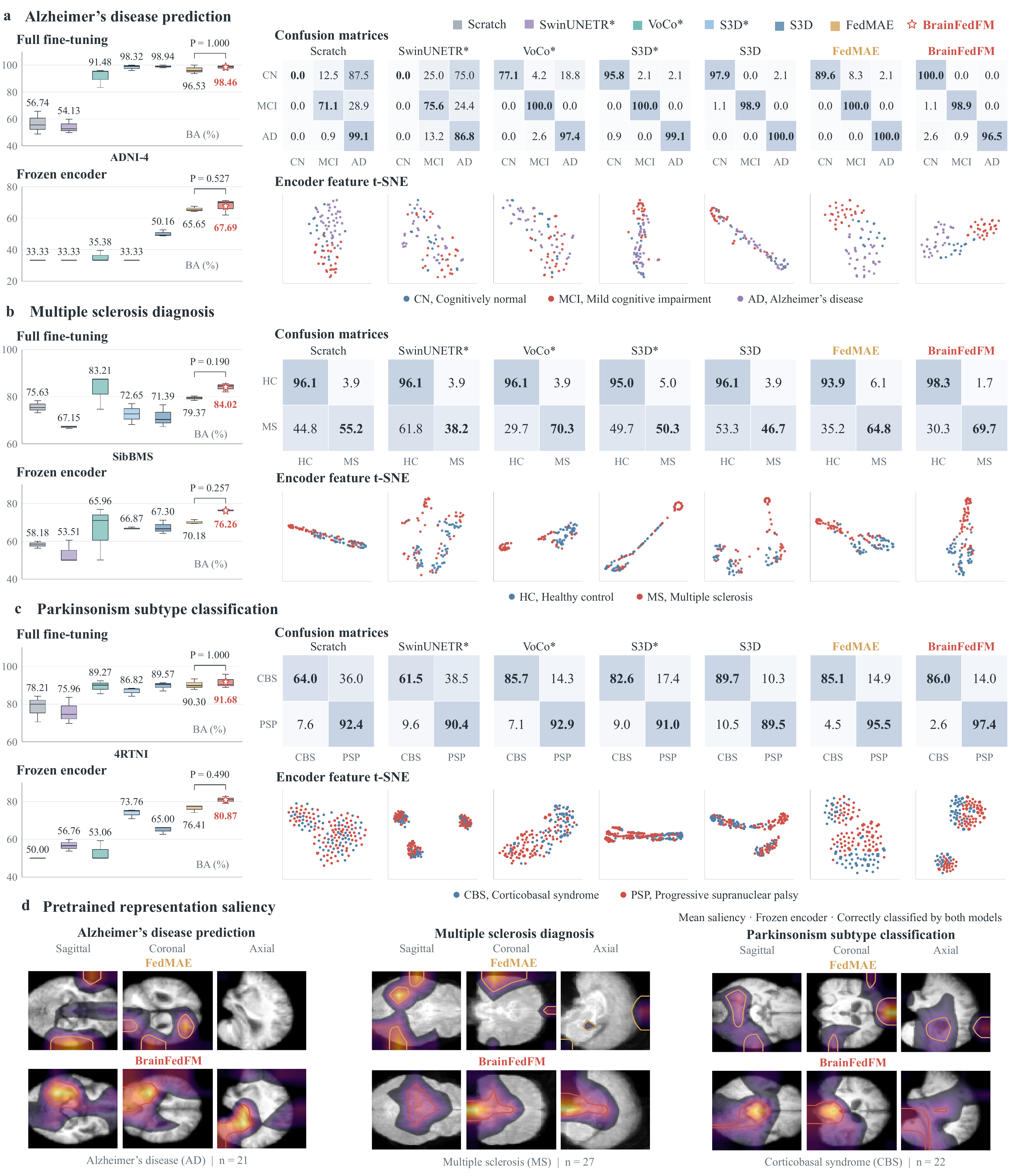}
    \par\vspace{2.5pt}
  \caption{\textbf{BrainFedFM preserves discriminative representations with parameter-efficient downstream adaptation.}
  \textbf{a}, Alzheimer's disease prediction on ADNI-4 under full fine-tuning and frozen-encoder adaptation.
  \textbf{b}, Corresponding analyses for multiple sclerosis diagnosis on SibBMS.
  \textbf{c}, Corresponding analyses for parkinsonism subtype classification on 4RTNI.
  \textbf{c}, Representative mean saliency maps under frozen-encoder adaptation for cases correctly classified by both FedMAE and BrainFedFM.
  Two-sided $P$ values were estimated using paired subject-level permutation tests with 100,000 Monte Carlo permutations and adjusted using the Holm procedure.
  Full fine-tuning panels show BA and class-normalized confusion matrices, whereas frozen-encoder panels show BA and t-SNE visualizations of encoder features.
  Despite matched prediction correctness, the two models exhibited distinct spatial organisation of task-relevant responses, with BrainFedFM showing more focused and anatomically coherent saliency. These patterns suggest that BrainFedFM pretraining establishes a more organised spatial representational basis before downstream adaptation.}
  \label{fig_5}
\end{figure}

We first examined whether BrainFedFM could support effective downstream transfer when adaptation was restricted to a small fraction of trainable parameters, thereby reducing the computational cost of task-specific fine-tuning (Fig.~\ref{fig_5}a,b,c).
With full-parameter fine-tuning as a reference, BrainFedFM achieved the highest BA among all seven models under frozen-encoder adaptation, reaching 67.69\% on ADNI-4, 76.26\% on SibBMS and 80.87\% on 4RTNI.
Under this adaptation, the pretrained encoder remained fixed apart from a narrowly restricted residual block in its final stage, with a single-layer linear prediction head operating on the 320-dimensional encoder feature.
These results indicate that BrainFedFM learned reusable structural representations that preserved task-relevant information, enabling task-specific decision boundaries to be established with minimal parameter updating.
Consistent with this interpretation, t-SNE visualizations of the encoder features showed clearer class-wise organisation for BrainFedFM.
Saliency analyses further indicated that BrainFedFM pretraining yields a more spatially organised and anatomically coherent task-relevant representation, providing a favourable representational starting point for downstream adaptation (Fig.~\ref{fig_5}d).
Together with the clearer class-wise feature organisation, these findings suggest that BrainFedFM enters downstream adaptation with a more structured and task-ready representation, enabling effective transfer with only limited parameter updating and fine-tuning cost.

We then examined whether BrainFedFM could support effective downstream adaptation under an extreme few-shot regime (Fig.~\ref{fig_6}c).
This adaptation tests whether a pretrained representation contains sufficiently reusable task-relevant information to support a new clinical task with minimal additional supervision, which is particularly important when expert annotations are costly or difficult to obtain.
Using only five labelled training examples per class, BrainFedFM achieved the highest mean BA across repeated samplings on all three tasks, reaching 65.35\% on SibBMS, 60.01\% on 4RTNI and 58.82\% on ADNI-4, compared with 63.51\%, 58.36\% and 51.45\%, respectively, for the strongest comparator on each task. The corresponding mean AUROC values for BrainFedFM were 72.43\%, 67.59\% and 76.97\%, respectively.
The persistence of this advantage under such sparse supervision indicates that BrainFedFM required comparatively little task-specific evidence to establish useful decision boundaries from its pretrained representation.
This behaviour is consistent with dual-priority federated pretraining capturing informative anatomical structure while integrating complementary variation across heterogeneous sites, thereby yielding a reusable feature basis for rapid adaptation when labelled examples are scarce.

\subsection{Diagnostic advantages over clinical assessments and practical robustness}

\begin{figure}[t]
\centering
\includegraphics[width=1.0\textwidth]{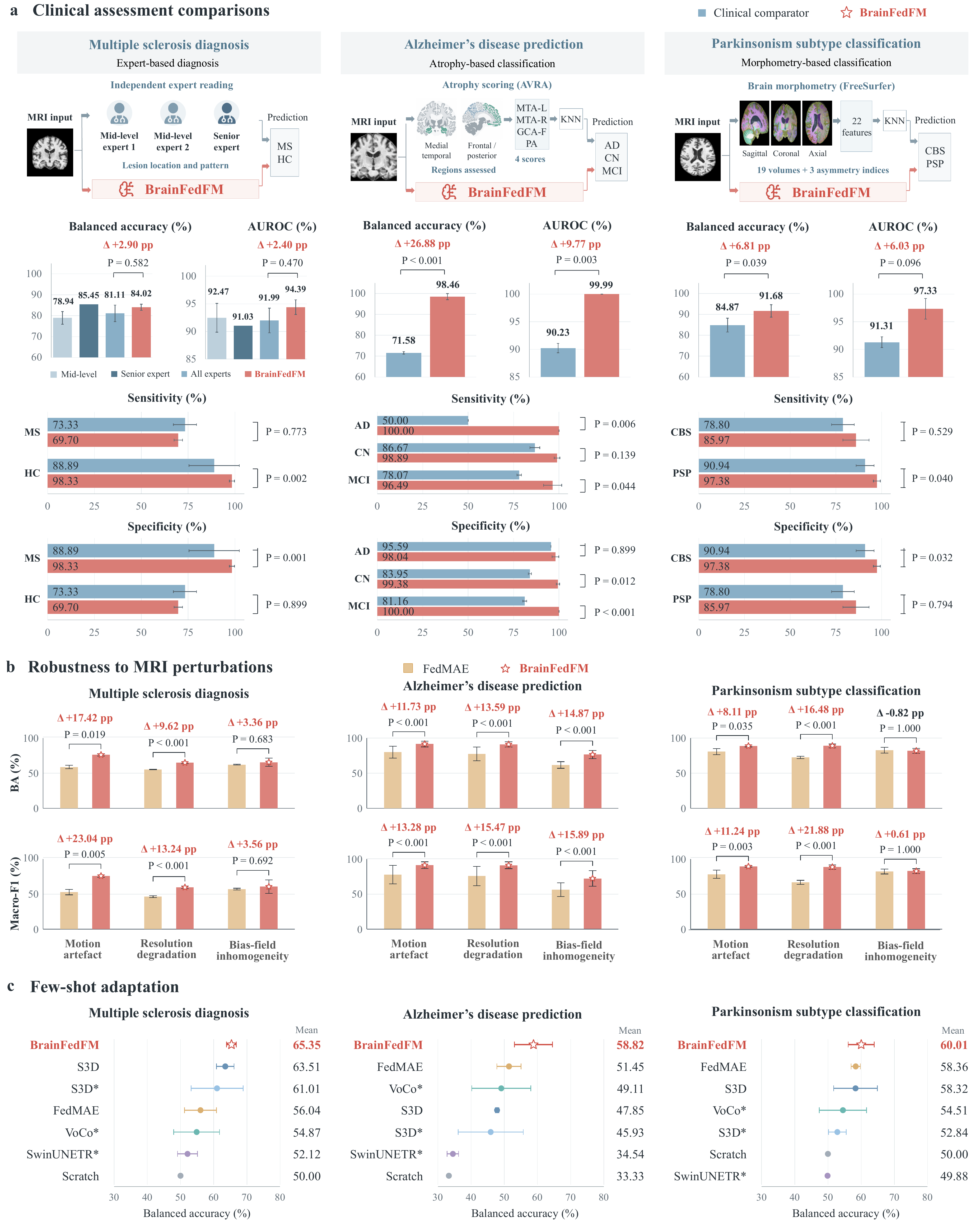}
\par\vspace{2.0pt}
\caption{\textbf{BrainFedFM demonstrates strong performance in clinical assessment comparisons and robust downstream adaptation.}
\textbf{a}, BrainFedFM was evaluated against three clinically grounded approaches: independent expert readings for multiple sclerosis diagnosis, AVRA-based atrophy scoring for Alzheimer's disease prediction, and FreeSurfer-based regional brain morphometry for parkinsonism subtype classification. The latter two approaches used KNN classifiers to generate diagnostic predictions from conventional imaging-derived features. BA, AUROC, class-specific sensitivity and specificity, performance differences and adjusted $P$ values are reported.
\textbf{b}, Robustness of downstream models fine-tuned from FedMAE and BrainFedFM to clinically plausible MRI perturbations.
The models were evaluated on held-out test images with simulated motion artefact (5$^\circ$/5\,mm rigid motion; two motion events), resolution degradation (twofold downsampling followed by restoration) or bias-field inhomogeneity (coefficient 0.20).
Values above each comparison indicate absolute performance differences in percentage points (pp).
\textbf{c}, Few-shot downstream adaptation using only five labelled training examples per class. Points and horizontal error bars summarize performance across repeated few-shot samplings, with mean values shown alongside.}
\label{fig_6}
\end{figure}

To evaluate the diagnostic advantages of BrainFedFM over established clinical assessment approaches, we compared its performance across three tasks (Fig.~\ref{fig_6}a). For multiple sclerosis diagnosis, BrainFedFM was compared with three independent clinical experts who based their assessments on lesion location, morphology and distribution: one senior expert with more than 15 years and two intermediate-level experts with 8--10 years of clinical imaging experience. 
For Alzheimer's disease prediction, AVRA-based visual atrophy scores, which characterize regional brain atrophy, served as predefined imaging features for diagnostic classification \cite{maartensson2019avra,harper2016mri}. For parkinsonism subtype classification, FreeSurfer was used to extract regional brain volumes and hemispheric asymmetry indices, which served as predefined morphometric features for classification \cite{fischl2012freesurfer,illan2022diagnostic}. Together, these comparisons tested whether BrainFedFM could achieve diagnostic performance comparable to expert assessment while improving discrimination over conventional feature-based approaches relying on predefined ratings or morphometric measures, highlighting the potential benefit of learning task-relevant representations directly from structural MRI.

Across all three tasks, BrainFedFM achieved higher BA and AUROC values than the corresponding clinical assessment comparators. For multiple sclerosis diagnosis, its BA and AUROC reached 84.02\% and 94.39\%, respectively, compared with 81.11\% and 91.99\% for the mean of the three clinical experts, indicating performance comparable to expert assessment. 
The largest improvement was observed in Alzheimer's disease prediction, where BrainFedFM achieved a BA of 98.46\% and an AUROC of 99.99\%, exceeding the AVRA-based assessment by 26.88 and 9.76 percentage points, respectively. For parkinsonism subtype classification, BrainFedFM also outperformed the FreeSurfer-based assessment, with BA and AUROC improvements of 6.81 and 6.02 percentage points, respectively. 
Class-wise analyses further revealed improved sensitivity and specificity across diagnostic categories in Alzheimer's disease prediction and parkinsonism subtype classification. For multiple sclerosis, BrainFedFM achieved higher specificity than the mean expert assessment (98.33\% vs. 88.89\%), highlighting its ability to distinguish MS cases from non-MS cases.

Beyond comparisons with established clinical assessments, robustness to variations in image quality encountered in clinical practice is another important consideration for clinical applicability. In routine structural MRI, patient motion, limited spatial resolution and intensity inhomogeneity can degrade image quality and compromise the reliability of automated predictions. We therefore examined whether the transfer advantage conferred by BrainFedFM pretraining persisted under these clinically relevant imaging challenges (Fig.~\ref{fig_6}b).
Following downstream fine-tuning, motion artefact, resolution degradation and bias-field inhomogeneity were simulated only in the held-out test images. BrainFedFM showed particularly pronounced advantages under motion and resolution degradation. On SibBMS, it exceeded FedMAE by 17.42 percentage points in BA under motion artefact and by 9.62 percentage points under resolution degradation, with corresponding mF1 gains of 23.04 and 13.24 percentage points. On 4RTNI, the largest advantage occurred under resolution degradation, with gains of 16.48 percentage points in BA and 21.88 percentage points in mF1. On ADNI-4, BrainFedFM maintained BA gains of 11.73--14.87 percentage points and mF1 gains of 13.28--15.89 percentage points across all three perturbations.
These results indicate that the transfer advantage conferred by BrainFedFM pretraining was not confined to clean MRI inputs but persisted under image degradation reflecting real-world challenges in clinical MRI acquisition.
Together with its advantages over clinical assessments, these findings highlight the potential clinical value of BrainFedFM and its robustness under challenging imaging conditions.

\section{Discussion}

BrainFedFM addresses a central challenge in neuroimaging foundation model development: how to learn broadly generalizable representations from large, heterogeneous imaging collections that cannot be readily centralized.
Despite this decentralized setting, BrainFedFM achieved stronger aggregate performance than the centralized pretraining comparators, and its advantage persisted under reduced labelled supervision, restricted parameter updating and degraded image quality.
These findings indicate that federated pretraining can support generalizable representation learning by leveraging complementary information across heterogeneous cohorts.
More broadly, the results highlight that the value of distributed imaging data depends not only on their collective scale, but also on how effectively cohort-specific information is retained and integrated into a shared representation.

BrainFedFM showed its largest gains in classification and regression, while retaining competitive performance in segmentation.
This pattern likely reflects differences in how pretrained representations are translated into downstream tasks: classification and regression couple the encoder to simple single-layer linear heads, whereas segmentation relies on a randomly initialized segmentation decoder to recover dense voxel-wise outputs.
The consistent performance across all task families indicates that the pretrained representation captures information relevant to both subject-level prediction and spatially-resolved analysis.
BrainFedFM provides a more spatially organised and anatomically coherent representational starting point for downstream adaptation. For structural brain MRI, such generalizability requires representations that preserve coherent global neuroanatomy together with localized anatomical variation.
BrainFedFM's spatial-priority masking is designed around this requirement, emphasizing lifespan-supported and reconstruction-difficult regions while retaining random masking for broad anatomical coverage.
By reinforcing informative local structure within a broad anatomical context, this strategy provides a plausible basis for the consistent performance observed across diverse downstream tasks.

BrainFedFM also showed particular strength across tasks involving clinically specialized and relatively underrepresented populations, including parkinsonism subtype classification, psychiatric differential diagnosis, cannabis dependence classification, autism diagnosis and pediatric MS lesion segmentation.
These findings highlight an important advantage of federated pretraining: smaller specialized cohorts can contribute population and imaging specific variation that may be underrepresented when centralized pretraining is dominated by much larger general-population cohort.
Conventional federated aggregation, however, often assigns greater weight to sites with more samples, which can reduce the influence of smaller cohorts despite the distinctive information they contribute.
BrainFedFM addresses this limitation through site-priority aggregation, which assigns greater aggregation weight to federated site with stronger lifespan support or higher residual reconstruction difficulty, allowing anatomically relevant or insufficiently learned information to contribute more substantially to the global model updating.
The resulting gains suggest that effective federated foundation model building depends not only on enabling distributed cohorts to participate, but also on ensuring that their complementary information is appropriately represented during global model integration.

Several limitations of the present study warrant consideration.
The current experiments used simulated synchronous cross-silo federated training with all sites participating in every communication round.
This controlled setting enabled systematic evaluation of the proposed pretraining framework but did not capture the operational heterogeneity encountered in real-world multi-institutional federation, including intermittent site participation, straggler delays, unequal computational and communication resources, and asynchronous updates.
Future multi-institutional studies should therefore assess BrainFedFM under these realistic operating conditions.
Although the pretraining federation encompassed broad variation in age, clinical population and acquisition setting, its composition was necessarily constrained by the availability of public and controlled-access research cohorts.
Coverage of rarer disorders and less commonly represented clinical populations nevertheless remains limited.
Expanding federation to these populations would further broaden the clinical scope of BrainFedFM and better leverage the advantages of distributed pretraining in settings where assembling large centralized datasets is challenging.

Overall, BrainFedFM suggests a broader strategy for scaling neuroimaging foundation models: moving beyond ever-larger centralized datasets toward effective learning across distributed and heterogeneous cohorts.
Federated pretraining is not merely a response to privacy constraints but also a means of learning from population and imaging variation that is distributed across institutions and difficult to capture through centralized pooling.
By integrating such variation into shared representations, BrainFedFM provides a practical route toward foundation models that better reflect the diversity of real-world clinical populations and imaging conditions.

\clearpage
\section{Methods}
\label{sec:methods}

\subsection{Pretraining data and preprocessing}
\label{sec:pretraining_data}

The pretraining data comprised 164,707 three-dimensional structural brain MRI scans organized across 42 federated sites.
It encompassed pediatric, adult, and older adult populations, healthy and disease-specific cohorts, and multiple imaging sequences. 
Across sites, there was substantial variation in sample scale, age distribution, clinical composition, scanner vendor, and acquisition protocol. 
Detailed data information is provided in Supplementary Table~2. 
Federated pretraining was implemented in a simulated cross-silo setting, with all sites participating in every communication round. Images and metadata remained in separate site-specific data stores. The central server received only model updates and compact block-level summaries required for site-priority aggregation and for updating the global priority memory. Raw images were neither exchanged across sites nor pooled centrally.

A standardized preprocessing workflow was applied independently at each site. 
Structural MRI series were curated to exclude localizer scans and duplicate copies of the same imaging series, and DICOM data were converted to NIfTI format using dcm2niix~\cite{li2016first} where required. Each scan underwent N4 bias-field correction~\cite{tustison2010n4itk} to reduce low-frequency intensity inhomogeneity, brain extraction using DeepBET~\cite{deepbet} to remove non-brain tissue, resampling by linear interpolation to an isotropic resolution of $1.0\times1.0\times1.0~\mathrm{mm}^3$, and rigid registration to the MNI152 reference space using FLIRT~\cite{jenkinson2002improved}. Only rigid alignment was applied to establish approximate anatomical correspondence while preserving subject-specific anatomy and disease-related structural changes. 
The MNI-aligned volumes were subsequently processed using the nnSSL framework~\cite{wald2025openmind}, which follows an nnU-Net-derived fingerprinting, planning, and preprocessing pipeline~\cite{isensee2021nnu}.
Under the 1-mm isotropic configuration, the common target spacing was retained, each volume was cropped to its nonzero image region, and per-volume $z$-score intensity normalization was applied. 
These operations reduced variation arising from image background, intensity scale, and voxel geometry while maintaining the anatomical correspondence required by the shared block-wise coordinate system used for spatial-priority masking and the global priority memory. 
Quality control evaluated image geometry, anatomical coverage, brain-extraction quality, and registration alignment; scans with evident preprocessing failures were reprocessed when possible and otherwise excluded before pretraining. The detailed preprocessing workflow is shown in Supplementary Fig.~1.

\subsection{Downstream benchmark datasets}
\label{sec:downstream_data}

The downstream benchmark comprised 20 datasets---11 classification, 4 regression, and 5 segmentation---spanning 17 distinct task definitions. 
Three of these tasks were evaluated on two datasets each.
The benchmark encompassed binary and multiclass neurological and psychiatric diagnosis, disease-subtype and radiogenomic classification, chronological-age and clinical-outcome regression, and brain tissue, tumor, and lesion segmentation. 
The pretraining and downstream cohorts were kept strictly disjoint at the subject level, with no subject appearing in both sets.
For each dataset, 20\% of the available data were reserved as a held-out test set, and three-fold cross-validation was performed within the remaining 80\% development set. 
In each cross-validation iteration, two folds were used for training and the remaining fold for validation. 
The same held-out test set and cross-validation folds were used for all comparison foundation models. 
The full-label setting used all data available within each training and validation fold, whereas the 25\% and 50\% labelled-data settings retained the corresponding proportions of the training and validation data, while leaving the held-out test set unchanged.
Detailed downstream task definitions, imaging sequences, sample splits, and data sources are provided in Supplementary Table~1.

\subsection{Overview of dual-priority federated pretraining}
\label{sec:brainfedfm_overview}

BrainFedFM used a synchronous, server-coordinated federated learning framework to learn a generalizable structural brain MRI representation while raw images remained locally stored~\cite{mcmahan2017communication}. 
Its dual-priority design operated at two complementary levels: spatial-priority masking balanced broad anatomical coverage with targeted reconstruction of informative and persistently difficult regions, whereas site-priority aggregation weighted site updates according to their priority rather than local cohort size. 
Let $K=42$ denote the number of participating sites, $\Theta^{(r)}$ the global model parameters after communication round $r$, and $q^{(r)}$ the corresponding global priority memory. At round $r$, the server broadcast $\Theta^{(r-1)}$ and $q^{(r-1)}$ to all sites; each site $k$ initialized a local copy of the global model, performed masked reconstruction on its local scans, and returned the updated parameters $\Theta_k^{(r)}$ together with compact block-level summaries. 
The server used these summaries to compute site-specific aggregation weights, combined the local parameters to obtain $\Theta^{(r)}$, and updated $q^{(r)}$ for the next round. 
Here, the global model denotes the shared encoder--decoder architecture maintained by the server during federated pretraining; after the final round, the pretrained encoder was retained as the BrainFedFM foundation model for downstream transfer.

\subsection{Pretraining architecture and masked reconstruction}
\label{sec:pretraining_architecture}

BrainFedFM adopted masked image modelling~\cite{hondru2025masked} as its self-supervised pretraining objective. 
Masked reconstruction has been reported to be more robust than contrastive learning to non-IID data in distributed self-supervised settings~\cite{chenunderstanding}, supporting its use for pretraining across heterogeneous federated sites.
For each structural MRI input, a subset of spatial blocks was withheld, and the model was trained to recover the missing image content from the visible anatomical context.
The pretraining model followed a fully convolutional 3D U-Net-style architecture based on the ResEnc-L configuration~\cite{isensee2024nnu}. 
This convolutional design was selected for its comparatively modest computational and memory requirements, facilitating repeated local optimization across sites and communication rounds.
Its multiscale encoder--decoder structure was also well suited to dense volumetric reconstruction.
Strided convolutions progressively reduced the spatial resolution to extract hierarchical volumetric representations, whereas the decoder restored the input resolution through transposed convolutions and multiscale skip concatenations from the corresponding encoder stages. The complete model architecture is shown in Supplementary Fig.~2.

Each training sample comprised a single-channel patch of
$160\times160\times160$ voxels extracted from a preprocessed MRI volume.
The patch was partitioned into non-overlapping blocks of
$16\times16\times16$ voxels, yielding a $10\times10\times10$ grid with
$B=1{,}000$ spatial blocks. Standard masked image modelling typically uses
random masking. BrainFedFM retained this random view and introduced two additional views, lifespan-guided and difficulty-guided masking, to prioritize complementary spatial regions during local pretraining.
For masking view
$v\in\{\mathrm{rand},\mathrm{life},\mathrm{dif}\}$, let the voxel-wise mask
$m_i^v$, obtained by expanding the sampled block mask, equal 1 for a visible
voxel and 0 for a masked voxel. Given the masked input $x\odot m^v$, the
model produced a full-volume reconstruction $\hat{x}^v$ with the same
spatial dimensions as the input patch. Masked reconstruction encouraged the
model to capture multiscale structural dependencies and learn transferable
representations for downstream brain MRI analysis. Although the model
generated a full-volume output, the reconstruction loss was evaluated only
over masked voxels, thereby focusing optimization on recovering missing
anatomical content rather than reproducing visible regions:
\begin{equation}
\mathcal{L}_{\mathrm{rec}}^{v}
=
\frac{
\sum_i
\left(\hat{x}_i^{v}-x_i\right)^2
\left(1-m_i^{v}\right)
}{
\sum_i
\left(1-m_i^{v}\right)
},
\quad
v\in\{\mathrm{rand},\mathrm{life},\mathrm{dif}\}.
\label{eq:masked_reconstruction}
\end{equation}

\subsection{Spatial-priority masking}
\label{sec:spatial_priority}

Spatial-priority masking was designed to maintain broad anatomical coverage while directing additional reconstruction effort toward regions exhibiting lifespan-associated variation and regions that remained persistently difficult to reconstruct.
At each federated site, every input patch was converted into three stochastic masked views with the same 50\% masking ratio. The random view sampled spatial blocks uniformly, 
the lifespan-guided view preferentially masked blocks showing greater lifespan-associated variation across the age intervals represented locally, and the difficulty-guided view preferentially masked blocks that remained difficult for the evolving model to reconstruct. All three views shared the same model parameters and reconstruction target and differed only in their block-sampling distributions.

The lifespan prior $l_k$ was estimated once before federated pretraining from scans with valid age metadata, with lifespan-associated variation calculated using comparisons within the same imaging sequence to reduce sequence-dependent contrast effects.
Its contribution was scaled by a site-specific support score $s_k$ that reflected the availability and distribution of local age information; age metadata were used only to construct this masking prior and were not provided to the model. In parallel, each site maintained a dynamic difficulty map $h_k^{(r)}$ based on the moving average of masked-block reconstruction errors. At communication round $r$, the local maps were combined with the global priority memory $q^{(r-1)}$:
\[
\begin{aligned}
p_k^{\mathrm{life},(r)}
&=
\operatorname{Norm}\left[
\alpha_l s_k\mathcal{R}(l_k)
+
\alpha_g\mathcal{R}\left(q^{(r-1)}\right)
\right],\\
p_k^{\mathrm{dif},(r)}
&=
\operatorname{Norm}\left[
\alpha_d\mathcal{R}\left(h_k^{(r-1)}\right)
+
\alpha_g\mathcal{R}\left(q^{(r-1)}\right)
\right],
\end{aligned}
\]
where $\mathcal{R}(\cdot)$ denotes robust normalization and $\operatorname{Norm}(\cdot)$ normalizes the resulting block-priority scores to a common scale.
The coefficients $\alpha_l$, $\alpha_d$, and $\alpha_g$ control the relative contributions of the lifespan prior, local difficulty map, and global priority memory. 
Guided priorities were converted into stochastic sampling probabilities with a nonzero baseline for every block, such that blocks with higher priorities were more likely to be masked without imposing fixed anatomical masks.

The three views were optimized jointly during each local update. Their masked-voxel mean-squared reconstruction losses were averaged, and a low-weight consistency term encouraged the full-volume reconstructions generated from different visible anatomical contexts to remain similar, thereby promoting stable reconstruction of the same underlying anatomy across different masking patterns:
\[
\begin{aligned}
\mathcal{L}_{\mathrm{local}}^{(r)}
={}&
\frac{1}{3}
\left(
\mathcal{L}_{\mathrm{rec}}^{\mathrm{rand}}
+
\mathcal{L}_{\mathrm{rec}}^{\mathrm{life}}
+
\mathcal{L}_{\mathrm{rec}}^{\mathrm{dif}}
\right)+
\lambda_{\mathrm{con}}g_c(r)
\left(
\mathcal{L}_{\mathrm{con}}^{\mathrm{life}}
+
\mathcal{L}_{\mathrm{con}}^{\mathrm{dif}}
\right),
\end{aligned}
\]
where the $\mathcal{L}_{\mathrm{con}}$ terms denote consistency losses, $\lambda_{\mathrm{con}}$ sets their maximum weight, and $g_c(r)$ gradually increases their contribution from zero to full strength over training.

Together, the three masking views balanced broad anatomical coverage with targeted reconstruction. The random view ensured continued sampling across the full patch, the lifespan-guided view emphasized blocks showing greater lifespan-associated variation across locally represented age intervals, and the difficulty-guided view revisited blocks with persistently high reconstruction error. Combining these local priority maps with the global priority memory allowed each federated site to preserve site-specific spatial information while benefiting from priorities accumulated across participating sites.
Further methodological details are provided in Supplementary Methods. Site-specific age distributions and representative lifespan-guided priority maps are shown in Supplementary Fig.~3, and examples of the three masked views and their reconstructions are shown in Supplementary Fig.~4.

\subsection{Site-priority aggregation}
\label{sec:site_priority}

Site-priority aggregation was designed to make each global model update reflect complementary anatomical information across federated sites, rather than being determined directly by local cohort size. 
All sites performed the same fixed number of local gradient-update steps in each communication round, without requiring a full pass over all locally available samples. 
BrainFedFM therefore prioritized site updates using lifespan support and residual reconstruction difficulty. Greater aggregation weights were assigned to sites with more reliable lifespan priors or greater reconstruction difficulty in globally prioritized regions, helping the global model integrate lifespan-related variation and unresolved reconstruction patterns across sites.

At communication round $r$, each site initialized its local model from the broadcast global parameters $\Theta^{(r-1)}$, performed local masked reconstruction, and returned the updated parameters $\Theta_k^{(r)}$ to the server. Before federated aggregation, the server calculated a site-priority weight from two complementary quantities. The lifespan support score $s_k$ reflected the reliability of the local lifespan prior, whereas the residual-difficulty score $u_k^{(r)}$ summarized the remaining reconstruction difficulty at site $k$, weighted by the corresponding global block priorities. These quantities were combined into the site-priority score $\psi_k^{(r)}$, which was converted into the aggregation weight $w_k^{(r)}$ used to construct the next global model:
\[
\begin{aligned}
u_k^{(r)}
=
\frac{1}{B}
\sum_{b=1}^{B}
\hat{h}_{k,b}^{(r-1)}
\left(1+\hat{q}_{b}^{(r-1)}\right)&, \ \ \
\psi_k^{(r)}
=
\lambda_{\mathrm{sup}}s_k
+
\lambda_{\mathrm{dif}}\tilde{u}_k^{(r)},\\
w_k^{(r)}
=
\frac{\exp\left(\psi_k^{(r)}\right)}
{\sum_{j=1}^{K}\exp\left(\psi_j^{(r)}\right)}&, \ \ \
\Theta^{(r)}
=
\sum_{k=1}^{K}
w_k^{(r)}\Theta_k^{(r)},
\end{aligned}
\]
where $\hat{h}_k^{(r-1)}$ is the normalized local difficulty map of site $k$, $\hat{q}^{(r-1)}$ is the normalized global priority memory, and $\tilde{u}_k^{(r)}$ is the min--max normalization of $u_k^{(r)}$ across sites. The factor $1+\hat{q}_{b}^{(r-1)}$ preserved difficulty information from the full spatial grid while assigning greater importance to residual difficulty in globally prioritized blocks. 
The lifespan-support and residual-difficulty terms contributed equally with $\lambda_{\mathrm{sup}}=\lambda_{\mathrm{dif}}=1$. 
Softmax normalization allowed every site to contribute while increasing the influence of updates with stronger lifespan support or greater residual difficulty in globally emphasized regions.

The global priority memory linked site-priority aggregation back to spatial-priority masking in subsequent rounds. It was initialized from a support-weighted combination of the local lifespan priors. After each round, the current site-priority weights were used to form normalized weighted summaries of the local lifespan and difficulty maps, denoted by $\bar{l}^{(r)}$ and $\bar{h}^{(r)}$, respectively. The memory was updated as
\[
q^{(r)}
=
\operatorname{Norm}\left\{
(1-\eta)q^{(r-1)}
+
\eta
\left[
\beta_r\bar{l}^{(r)}
+
(1-\beta_r)\bar{h}^{(r)}
\right]
\right\},
\ \ 
\eta=0.2,
\]
where $\eta$ controlled the memory update rate and $\beta_r$ balanced the lifespan and difficulty summaries. The coefficient $\beta_r$ decreased linearly from 0.75 to 0.50 during pretraining, increasing the contribution of the difficulty summary from 0.25 to 0.50. 
The memory therefore emphasized the more stable lifespan priors during early pretraining and progressively incorporated greater difficulty information as the reconstruction-based difficulty estimates became more reliable. 
The updated memory was broadcast with the global model at the beginning of the next round and contributed to both guided masking views, allowing spatial priorities identified locally to be accumulated across sites and reused during subsequent local pretraining.

\subsection{Federated pretraining implementation}
\label{sec:pretraining_implementation}

BrainFedFM was pretrained for 150 synchronous communication rounds, with all 42 federated sites participating in every round. 
Each site initialized its local model from the global parameters, performed 200 local gradient-update steps, and returned the updated parameters for aggregation. 
Local optimization used stochastic gradient descent with Nesterov momentum, a constant learning rate of $1\times10^{-2}$, momentum of 0.99, and weight decay of $3\times10^{-5}$. Automatic mixed precision was used, and the maximum gradient norm was set to 12. 
To stabilize priority estimates, lifespan and global-memory guidance were activated after the first 10\% of pretraining, whereas difficulty guidance was activated after the first 20\%. Cross-view consistency was activated after the first 10\% of rounds, with its weight increasing linearly from zero to $\lambda_{\mathrm{con}}=0.03$ by 50\% of the pretraining schedule. 
After each round, the global model was evaluated on the validation partition of every site using the equally weighted mean of the three masked-reconstruction losses, and the resulting site-level losses were averaged uniformly. 
Cross-view consistency was excluded from model selection, and the checkpoint with the lowest federated validation loss was retained for downstream transfer. 
A single pretraining configuration was used across all sites, and detailed settings are provided in Supplementary Table~3.

\subsection{Adaptation to downstream tasks}
\label{sec:downstream_transfer}

After federated pretraining, the reconstruction decoder was discarded and the pretrained encoder was retained as the BrainFedFM foundation model. 
For each downstream task, the pretrained encoder served as the backbone and all primary evaluations employed full-parameter fine-tuning. For classification and regression, the encoder was jointly optimized with a newly initialized single-layer linear prediction head, whose output dimension matched the number of classes or continuous targets, respectively. For segmentation, the encoder was jointly optimized with a randomly initialized segmentation decoder. Within each dataset, all evaluated models used the same task-specific head or decoder architecture and followed an identical downstream adaptation protocol.
Transfer performance was evaluated using 25\%, 50\%, and 100\% of the labelled training and validation data within each predefined subject-level fold, while the held-out test set remained fixed. At each label fraction, the same subjects, data augmentations, optimization settings, and model-selection procedure were used across models. The checkpoint achieving the best validation performance was selected for evaluation on the held-out test set. Detailed task-specific adaptation settings are provided in Supplementary Tables~4--23.

\subsection{Comparison foundation models}
\label{sec:comparison}

BrainFedFM was compared with six models spanning random initialization, publicly released centralized pretraining, data-matched centralized pretraining, and federated pretraining.
Scratch used randomly initialized weights with the same downstream architecture. 
SwinUNETR$^{\ast}$~\cite{tang2022self} (multi-task proxy learning), VoCo$^{\ast}$~\cite{wu2024voco} (volume-level contrastive learning), and S3D$^{\ast}$~\cite{wald2025revisiting} (masked image modelling) used publicly released checkpoints from the OpenMind benchmark~\cite{wald2025openmind}.
A separate S3D model was centrally pretrained on the same corpus used for BrainFedFM, providing a data-matched centralized comparison. FedMAE~\cite{jiang2025pretraining} provided the corresponding federated comparison. It matched BrainFedFM in pretraining data, site organization, communication schedule, and local update budget, but used random masking and standard Federated Averaging (FedAvg) without spatial-priority masking, site-priority aggregation, or the global priority memory. All foundation models were compared using the same downstream adaptation protocol described above.

\subsection{Evaluation metrics}
\label{sec:evaluation_metrics}

Balanced accuracy (BA), root mean squared error (RMSE), and dice similarity coefficient (Dice) were prespecified as the primary metrics~\cite{maier2024metrics} for classification, regression, and segmentation, respectively. BA accounts for class imbalance, RMSE quantifies prediction error with greater sensitivity to large errors, and Dice measures spatial overlap between predicted and reference labels. Fold-averaged primary metrics were used for dataset-level comparisons and overall model ranking, with higher BA and Dice and lower RMSE indicating better performance. We additionally reported macro F1 score (mF1) and macro one-versus-rest ROC area under the curve (AUC) for classification, mean absolute error (MAE) and the coefficient of determination ($R^2$) for regression, and intersection-over-union (IoU) and normalized surface Dice (NSD) for segmentation. More task-specific metrics and complementary visual analyses are provided in Supplementary Tables~4--23 and their associated panels.

\subsection{Computing software and hardware}
\label{sec:computing_resources}

All experiments were implemented in Python (v3.12.12) using PyTorch (v2.9.1)~\cite{paszke2019pytorch}. Federated pretraining was implemented by extending the nnSSL framework with the proposed federated optimization and dual-priority modules~\cite{wald2025openmind}, and downstream segmentation was implemented based on nnU-Net v2.6.2~\cite{isensee2021nnu}. Classification and regression experiments were implemented using Lightning (v2.4.0). NumPy (v2.3.5)~\cite{harris2020array} and pandas (v2.3.3)~\cite{mckinney2010data} were used for numerical processing and result aggregation, scikit-learn (v1.8.0)~\cite{pedregosa2011scikit} and SciPy (v1.16.3)~\cite{virtanen2020scipy} for metric computation and statistical analysis, and Matplotlib (v3.10.8) and NiBabel (v5.3.3) for visualization and neuroimaging file handling. The complete BrainFedFM pretraining network contained approximately 102.35 million parameters. Federated pretraining was performed on four NVIDIA H800 GPUs with 80~GB of memory each and required 26.3 days, corresponding to 2,525 H800 GPU-hours, whereas downstream adaptation and inference were performed on a single NVIDIA~3090 GPU with 24~GB of memory.

\subsection{Statistical analysis}
\label{sec:statistics}

For each model and downstream dataset, test performance was reported as the mean $\pm$ standard deviation across the three fold-specific models obtained from cross-validation within the development set and evaluated on the held-out test set.
BrainFedFM was compared with each of the six comparison models using paired predictions from the same test subjects.
Two-sided $P$ values for between-model differences were estimated using paired subject-level permutation tests with 100,000 Monte Carlo permutations~\cite{nichols2002nonparametric}.
The three fold-specific predictions for each test subject were retained as a single cluster throughout permutation testing. $P$ values were adjusted across the comparisons within each dataset using the Holm procedure~\cite{holm1979simple}.
For overall ranking, evaluated models were ranked within each dataset using the fold-averaged primary metric---BA for classification, RMSE for regression, and Dice for segmentation---and mean ranks across the 20 datasets were calculated.

\section*{Data availability}
All imaging data used in this study are available from publicly accessible or controlled-access repositories maintained by the original data providers. Dataset sources, cohort characteristics, imaging sequences and access information for the pretraining cohorts organized across 42 federated sites are provided in Supplementary Table~2. Dataset sources, task definitions, imaging sequences, sample splits and access information for the 20 downstream benchmark datasets are provided in Supplementary Table~1. The study does not redistribute raw imaging data, and access remains subject to the registration requirements, licences and data-use agreements specified by the corresponding repositories or data providers.

\section*{Code availability}
The source code for data preprocessing, federated pretraining, downstream fine-tuning, inference evaluation across classification, regression, and segmentation tasks, together with all pretrained foundation model weights, is publicly available at \url{https://github.com/SPIresearch/BrainFedFM}.

\MainBibliography

\section*{Acknowledgements}
We sincerely thank the clinical experts involved in this study for their valuable contributions to the clinical evaluation and interpretation of our findings. We also gratefully acknowledge the broader research community for sharing neuroimaging datasets, pretrained model weights, and open-source software frameworks, as well as the study participants who made these datasets possible.

\section*{Author contributions}
Z.Y. collected and curated the data, developed the methodology and software, conducted the experiments and analyses, prepared the figures and tables, and drafted the manuscript.
Y.L. contributed to methodology development, research supervision and manuscript revision.
J.F. contributed part of the data resources and provided strategic guidance on the overall study design and methodological development.
X.Z. contributed part of the data resources offered scientific advice and supervision throughout the study.
Q.C. conceived and designed the study, provided overall scientific direction, supervised the research and critically revised the manuscript.
All authors reviewed and approved the final manuscript.

\section*{Funding}
This work was supported by the National Natural Science Foundation of China under Grant Nos. 62571009.

\section*{Competing interests}
The authors declare no competing interests.

\end{bibunit}

\BeginSupplementaryMode

\begin{bibunit}[unsrt]
\newcommand{\SupplementaryBibliography}{\putbib[supprefs]}
\title{\textbf{Supplementary Information for}\\[0.8em]
A generalizable structural brain MRI foundation model via dual-priority federated pretraining}
\date{}
\maketitle

\vspace{-2.5em}

\begin{center}
{\large\bfseries Supplementary contents}
\end{center}

\vspace{0.35em}
\begingroup
\footnotesize
\setlength{\parindent}{0pt}
\setlength{\parskip}{0pt}

\noindent
\textbf{\footnotesize\bfseries Supplementary Methods}
\dotfill \pageref*{sec:supp_spatial_priority}

\vspace{0.55em}

\noindent{\footnotesize\bfseries Supplementary Figures}\par
\vspace{0.4em}

\noindent
 \scriptsize Supplementary Fig. 1\quad Preprocessing workflow
\dotfill
\pageref*{suppfig:preprocess}
\par \vspace{0.28em}

\noindent
Supplementary Fig. 2\quad Pretraining backbone architecture
\dotfill \pageref*{fig:backbone}\par
\vspace{0.28em}

\noindent
Supplementary Fig. 3\quad Site-specific lifespan distribution
\dotfill \pageref*{fig:lifespan_prior}\par
\vspace{0.28em}

\noindent
Supplementary Fig. 4\quad Masked-reconstruction examples
\dotfill \pageref*{fig:reconstruction}\par
\vspace{0.28em}

\noindent
Supplementary Fig. 5\quad Full-label benchmark performance
\dotfill \pageref*{fig:radar_full}\par
\vspace{0.28em}

\noindent
Supplementary Fig. 6\quad 50\% labelled-data benchmark performance
\dotfill \pageref*{fig:radar_50}\par
\vspace{0.24em}

\noindent
Supplementary Fig. 7\quad 25\% labelled-data benchmark performance
\dotfill \pageref*{fig:radar_25}\par
\vspace{0.24em}

\vspace{0.6em}

\noindent{\footnotesize\bfseries Supplementary Tables}\par
\vspace{-0.72em}

\begin{NoHyper}
\begin{multicols}{2}
\raggedcolumns
\scriptsize
\setlength{\parskip}{0.22em}
\printtableindex
\end{multicols}
\end{NoHyper}

\endgroup

\clearpage
\begin{table}[h]
\centering
\caption[Downstream benchmark datasets]{\textbf{Downstream benchmark datasets.}
Summary of the downstream tasks used to evaluate brain MRI foundation models. 
The benchmarks span classification, regression, tissue and lesion segmentation, and radiogenomic prediction across neurodevelopmental, neurodegenerative, psychiatric, demyelinating, cerebrovascular, and neuro-oncological settings. For each dataset, we report the task definition, imaging sequences, sample splits, and data source.}
\label{supptab:downstream}

\renewcommand{\arraystretch}{2.138}
\resizebox{1.0\linewidth}{!}{
\setcounter{dtrow}{0} 
\begin{tabular}{c l l l l l l}
\toprule

\textbf{No.} & \textbf{DownTask} & Source Dataset & Task Description & Sequences & Sample Splits & Source Link \\
\midrule

\textbf{\dtrown} 
& \hyperref[tab:result_autism]{%
    \textcolor{blue}{{\textbf{Autism Diagnosis}}}
} & ABIDE~\cite{di2014autism} & Classification (ASD/Control) & T1w & 1180 (755/189/236) & \href{http://fcon_1000.projects.nitrc.org/indi/abide}{ABIDE} \\

\textbf{\dtrown} 
& \hyperref[tab:result_autism]{%
    \textcolor{blue}{\textbf{Post-Stroke Time Estimation}}
} & ATLAS-2~\cite{liew2022large} & Regression (Stroke Onset Time) & T1w & 549 (351/88/110) & \href{https://atlas.grand-challenge.org}{ATLAS} \\

\textbf{\dtrown}   
& \hyperref[tab:result_oasis]{
    \textcolor{blue}{\textbf{Dementia Diagnosis}}}
& OASIS-2~\cite{marcus2010open} & Classification (Demented/Nondemented) & T1w & 336 (215/54/67) & \href{https://sites.wustl.edu/oasisbrains/home/oasis-2/}{OASIS} \\

\textbf{\dtrown} 
& \hyperref[tab:result_tmcu]{\textcolor{blue}{\textbf{Cannabis Dependence Classification}}}
& TMCU~\cite{ds000174:1.0.1} & Classification (CB/HC) & T1w, T2w & 
 84 (54/14/16) & \href{https://openneuro.org/datasets/ds000174/versions/1.0.1}{OpenNeuro} \\

\textbf{\dtrown} 
 & \hyperref[tab:result_demy]{\textcolor{blue}{\textbf{Pediatric Demyelinating Segmentation}}}
& PediDemi~\cite{popa2025pedidemi} & Segmentation (Demyelinating Lesions) & T1w, T2w, FLAIR & 13 (6/4/3) & \href{https://figshare.com/articles/dataset/PediDemi/28694435}{Figshare} \\

\textbf{\dtrown}  & \hyperref[tab:result_mms]{\textcolor{blue}{\textbf{Multiple Sclerosis Diagnosis}}}
& SibBMS~\cite{tuchinov2025sibbms} & Classification (MS/HC) & T1w, T2w, FLAIR & 572 (365/92/115)& \href{https://www.melba-journal.org/papers/2025:046.html}{MELBA}\\

\textbf{\dtrown}  & \hyperref[tab:result_adni]{\textcolor{blue}{\textbf{Alzheimer’s Disease Prediction}}}
& ADNI-4~\cite{weiner2023increasing} & Classification (CN/MCI/AD) & T1w & 419 (268/67/84) & \href{https://adni.loni.usc.edu/}{ADNI} \\

\textbf{\dtrown}  & \hyperref[tab:result_ideas]{\textcolor{blue}{\textbf{Focal Epilepsy Diagnosis}}}
& IDEAS~\cite{taylor2025imaging} & Classification (Epilepsy/HC) & T1w, FLAIR & 320 (205/51/64) & \href{https://sites.google.com/view/cnnp-lab/ideas-data}{CNNP} \\

\textbf{\dtrown}  & \hyperref[tab:result_msd]{\textcolor{blue}{\textbf{Brain Tumor Segmentation}}}
& MSD-Brain~\cite{antonelli2022medical} & Segmentation (Tumor Subregions) & T1w, T1ce, T2w, FLAIR & 484 (258/129/97) & \href{http://medicaldecathlon.com}{MSD} \\

\textbf{\dtrown}  & \hyperref[tab:result_msles]{\textcolor{blue}{\textbf{Disability Severity Prediction}}}
& MSLesSeg~\cite{guarnera2025mslesseg} & Regression (MS EDSS Score) & T1w, T2w, FLAIR &  278 (176/45/57)& 
\href{https://figshare.com/articles/dataset/MSLesSeg_baseline_and_benchmarking_of_a_new_Multiple_Sclerosis_Lesion_Segmentation_dataset/27919209}{Figshare}  \\

\textbf{\dtrown}  & \hyperref[tab:result_iseg]{\textcolor{blue}{\textbf{Infant Brain Tissue Segmentation}}}
& iSeg~\cite{sun2021multi} & Segmentation (WM/GM/CSF) & T1w, T2w & 10 (5/3/2) & \href{https://iseg2019.web.unc.edu/data}{iSeg} \\

\textbf{\dtrown} & \hyperref[tab:result_upenn]{\textcolor{blue}{\textbf{MRI Sequence Classification}}}
& UPENN-GBM~\cite{bakas2022university} & Classification (Sequence Type) & T1w, T1ce, T2w, FLAIR  & 2584 (1653/414/517)& \href{https://www.cancerimagingarchive.net/collection/upenn-gbm}{TCIA} \\

\textbf{\dtrown}  & \hyperref[tab:result_4rnti]{\textcolor{blue}{\textbf{Parkinsonism Subtype Classification}}}
& 4RTNI~\cite{WOS:000382568400386} & Classification (CBS/PSP) & T1w, T2w, FLAIR & 1464 (937/234/293) & \href{https://neurodegenerationresearch.eu/survey/the-four-repeat-tauopathy-neuroimaging-initiative}{4RTNI} \\

\textbf{\dtrown} & \hyperref[tab:result_rsna]{\textcolor{blue}{\textbf{Tumor Radiogenomic Prediction}}}
& RSNA-MICCAI~\cite{baid2021rsna} & Classification (Methylated/Unmethylated) & T1w, T1ce, T2w, FLAIR & 2308 (1476/372/460) & \href{https://www.kaggle.com/competitions/rsna-miccai-brain-tumor-radiogenomic-classification/data}{Kaggle} \\

\specialrule{0.5pt}{3.5pt}{1.5pt}

\textbf{\dtrown}  & \hyperref[tab:result_sald]{\textcolor{blue}{\textbf{Adult Brain Age Prediction}}}
& SALD~\cite{wei2018structural} & Regression (Chronological Age) & T1w & 494 (316/79/99) & \href{https://fcon_1000.projects.nitrc.org/indi/retro/sald.html}{SALD} \\

\textbf{\dtrown}  & \hyperref[tab:result_nki]{\textcolor{blue}{\textbf{Lifespan Brain Age Prediction}}}
& NKI-RS~\cite{nooner2012nki} & Regression (Chronological Age) & T1w & 1326 (849/212/265) & 
\href{https://rocklandsample.org/}{NKI-RS} \\

\textbf{\dtrown} & \hyperref[tab:result_isbi]{\textcolor{blue}{\textbf{Adult MS Lesion Segmentation}}}
& ISBI-2015~\cite{carass2017longitudinal} & Segmentation (MS Lesions) & T1w, T2w, PDw, FLAIR & 21 (11/6/4) & \href{https://www.kaggle.com/datasets/soumikrakshit/isbi-challenge-dataset}{Kaggle} \\

\textbf{\dtrown} & \hyperref[tab:result_pediMS]{\textcolor{blue}{\textbf{Pediatric MS Lesion Segmentation}}}
& PediMS~\cite{popa2025pedims} & Segmentation (MS Lesions) & T1w, T2w, FLAIR & 28 (14/8/6) & \href{https://figshare.com/articles/dataset/PediMS_A_Pediatric_Multiple_Sclerosis_Lesion_Segmentation_Dataset/28701065}{Figshare} \\

\textbf{\dtrown}
& \hyperref[tab:result_ucla]{\textcolor{blue}{\textbf{Psychiatric Differential Diagnosis}}}
& UCLA-CNP~\cite{poldrack2016phenome} & Classification (ADHD/SCHZ/BIPOLAR) & T1w & 140 (89/23/28) & \href{https://openneuro.org/datasets/ds000030/versions/1.0.0}{OpenNeuro} \\

\textbf{\dtrown} 
& \hyperref[tab:resultsrpbs]{\textcolor{blue}{\textbf{Multi-Disorder Psychiatric Diagnosis}}}
& SRPBS~\cite{tanaka2021multi} & Classification (HC/ASD/MDD/SSD) & T1w & 765 (489/123/153) & \href{https://bicr-resource.atr.jp/srpbsopen/}{SRPBS} \\
\bottomrule
\end{tabular}}

\vspace{6.2pt}
{\tiny\raggedright
\leftskip=0pt
\textit{Abbreviations:}
PSP, progressive supranuclear palsy;
CBS, corticobasal syndrome;
ASD, autism spectrum disorder;
WM, white matter;
GM, gray matter;
CSF, cerebrospinal fluid;
MCI, mild cognitive impairment;
AD, Alzheimer's disease;
ADHD, attention-deficit/hyperactivity disorder;
SCHZ, schizophrenia;
MDD, major depressive disorder;
SSD, schizophrenia spectrum disorder;
CB, cannabis users;
HC, healthy controls;
CN, cognitively normal;
MS, multiple sclerosis;
EDSS, Expanded Disability Status Scale;
\par}
\end{table}

\begin{table}[h]
\centering
\caption[Pretraining datasets]{\textbf{Pretraining datasets.}
Summary of the 3D structural brain MRI datasets used for the federated pretraining of BrainFedFM. The pretraining cohort comprises 164,707 scans from 42 sites and spans diverse age groups, scanner vendors, acquisition protocols, and clinical conditions. This multi-site organization reflects a realistic cross-silo federated setting in which heterogeneous institutions, including small or specialized sites, can participate without sharing raw imaging data. For each cohort, we report participant characteristics, scanner and sequence information, the number of 3D scans, and data-access links.}

\setcounter{dtrow}{0}
\renewcommand{\arraystretch}{1.53}
\resizebox{1.0\linewidth}{!}{
\begin{tabular}{c l l l l l}
\toprule
\textbf{Site.} & \textbf{Source Dataset} & Subject Description & Scanners Sequences & 3D Scans & Data Link \\
\midrule

\textbf{\dtrown} & \textbf{Calgary}~\cite{reynolds2020calgary} & Children (Healthy, 2--7 Years, Canada) &  GE Healthcare MR750 (T1w) & 278 & \href{https://osf.io/axz5r}{OSF} \\

\textbf{\dtrown} & \textbf{AIBL}~\cite{ellis2009australian} & Elderly (Alzheimer, 50--90 Years, Australia) & Siemens \& GE (T1w, T2w, FLAIR) & 1{,}903 & \href{https://aibl.csiro.au}{AIBL} \\

\textbf{\dtrown} & \textbf{BCAS}~\cite{zhu2022chinese} & Adults (Healthy, 22--79 Years, China) & Siemens Magnetom TrioTim (T1w) & 180 & \href{https://www.nitrc.org/projects/adultatlas}{NITRC} \\

\textbf{\dtrown} & \textbf{ABVIB}~\cite{rodriguez2019psychometric2} 
& Elderly (Alzheimer, 60--91 Years, USA) 
& Siemens Prisma (T1w, T2w, FLAIR) 
& 801 
& \href{https://adrc.usc.edu}{ADRC} \\

\textbf{\dtrown} & \textbf{ISYB}~\cite{gao2022chinese}  
& Adults (Healthy, 18--30 Years, China) 
&  GE Healthcare MR750 (T1w) 
& 200
& \href{https://ccnp.scidb.cn/detail?dataSetId=826407529641672704}{SciDB} \\

\textbf{\dtrown} & \textbf{DLBS}~\cite{park2024dallas} 
& Adults (Healthy, 20--90 Years, USA) 
& Siemens \& GE \& Philips (T1w) 
& 461 
& \href{https://openneuro.org/datasets/ds004856}{OpenNeuro} \\

\textbf{\dtrown} & \textbf{ICBM}~\cite{mazziotta2001probabilistic} 
& Adults (Healthy, 18--90 Years, Worldwide) 
& Siemens \& GE \& Philips (T1w, FLAIR) 
& 1{,}408 
& \href{https://ida.loni.usc.edu/home/projectPage.jsp?project=ICBM}{LONI IDA} \\

\textbf{\dtrown} & \textbf{DFCI}~\cite{zapaishchykova2024dfci} 
& Children (Glioma, 5--10 Years, USA) 
& Siemens \& GE (T1w, T1ce, T2w, FLAIR) 
& 70 
& \href{https://www.cancerimagingarchive.net/collection/dfci-bch-bwh-peds-hgg}{TCIA} \\

\textbf{\dtrown} & \textbf{MSCA}~\cite{roberts2008mayo} 
& Elderly (Alzheimer, 49--89 Years, USA)
& Siemens \& GE (T1w) 
& 1,791
& \href{https://www.mayo.edu/research/centers-programs/alzheimers-disease-research-center/research-activities/mayo-clinic-study-aging}{Mayo} \\

\textbf{\dtrown} & \textbf{MR-ART}~\cite{narai2022movement}  
& Adults (Healthy, 18--75 Years, Hungary) 
& Siemens Magnetom Prisma (T1w) 
& 436 
& \href{https://openneuro.org/datasets/ds0041734}{OpenNeuro} \\

\textbf{\dtrown} & \textbf{HABS-HD}~\cite{petersen2025health}  
& Adults (Alzheimer, 25--91 Years, USA)
& Siemens \& GE \& Philips (T1w, T2w, FLAIR) 
& 9,950 
& \href{https://apps.unthsc.edu/itr/habs-hd}{HABS-HD} \\

\textbf{\dtrown} & \textbf{LONG579}~\cite{wang2022longitudinal}  
& Children (Healthy, 5--9 Years, USA) 
& Siemens Magnetom Skyra (T1w) 
& 827
& \href{https://openneuro.org/datasets/ds003604}{OpenNeuro} \\

\textbf{\dtrown} & \textbf{PIXAR}~\cite{richardson2018development}  
& Adults (Healthy, 3--12 \& 18--39 Years, USA) 
& Siemens Magnetom TrioTim (T1w) 
& 155
& \href{https://openneuro.org/datasets/ds000228}{OpenNeuro} \\

\textbf{\dtrown} & \textbf{IXI}~\cite{robinson2010identifying}  
& Adults (Healthy, 20--86 Years, UK) 
& Philips Intera \& GE (T1w, T2w, PDw) 
& 1,739
& \href{https://brain-development.org/ixi-dataset}{IXI} \\

\textbf{\dtrown} & \textbf{MIRIAD}~\cite{malone2013miriad}  
& Elderly (Alzheimer, 62--76 Years, UK)
& GE Signa HDxt (T1w) 
& 708
& \href{https://www.ucl.ac.uk/drc/research-clinical-trials/minimal-interval-resonance-imaging-alzheimers-disease-miriad}{UCL} \\

\textbf{\dtrown} & \textbf{HCP}~\cite{bastiani2019automated}  
& Young adults (Healthy, 22--35 Years, USA) 
& Siemens Magnetom Skyra (T1w, T2w, FLAIR) 
& 733
& \href{https://www.humanconnectome.org/study/hcp-young-adult}{HCP} \\

\textbf{\dtrown} & \textbf{PETfrog}~\cite{ds002385:1.1.0}  
& Young adults (Healthy, 12--32 Years, USA) 
& Siemens Biograph mMR (T1w) 
& 254
& \href{https://openneuro.org/datasets/ds002385}{OpenNeuro} \\

\textbf{\dtrown} &  \textbf{SCAN}~\cite{mormino2025consortium}  
& Elderly (Alzheimer, 64--80 Years, USA)
& Siemens \& GE \& Philips (T1w) 
& 4,007
& \href{https://scan.naccdata.org}{SCAN} \\ 

\textbf{\dtrown} & \textbf{A4}~\cite{ngolab2021feasibility} 
& Elderly (Alzheimer, 65--85 Years, USA)
& Siemens \& GE \& Philips  (T1w) 
& 1,880
& \href{https://a4study.org}{A4} \\

\textbf{\dtrown} & \textbf{SLIM}~\cite{liu2017longitudinal}  
& Young adults (Healthy, 17--27 Years, China)
& Siemens Trio (T1w) 
& 1,038
& \href{https://fcon_1000.projects.nitrc.org/indi/retro/southwestuni_qiu_index.html}{SLIM} \\ 

\textbf{\dtrown} & \textbf{PPMI}~\cite{marek2011parkinson}  
& Adults (Parkinson, 30--93 Years, Worldwide) 
& Siemens \& GE \& Philips (T1w, T2w, FLAIR) 
& 5,688
& \href{http://www.ppmi-info.org}{PPMI} \\ 

\textbf{\dtrown} & \textbf{SOOP}~\cite{absher2024stroke} 
& Adults (Stroke, 16--90 Years, USA)
& GE Healthcare MR750 (T1w, T2w, FLAIR) 
& 2,576
& \href{https://openneuro.org/datasets/ds004889/versions/1.1.2}{OpenNeuro} \\

\textbf{\dtrown} & \textbf{OASIS-1}~\cite{marcus2007open} 
& Adults (Alzheimer, 18--96 Years, USA)
& Siemens Vision (T1w) 
& 436
& \href{https://sites.wustl.edu/oasisbrains/home/oasis-1}{OASIS} \\

\textbf{\dtrown} & \textbf{OASIS-3}~\cite{lamontagne2019oasis} 
& Adults (Cognitive decline, 42--95 Years, USA)
& Siemens Vision \& TrioTim (T1w, T2w) 
& 5,748
& \href{https://sites.wustl.edu/oasisbrains/home/oasis-3}{OASIS} \\

\textbf{\dtrown} & \textbf{AOMIC-ID1000}~\cite{ds003097:1.2.1}
& Young adults (Healthy, 18--26 Years, Netherlands)
& Philips Intera (T1w) 
& 2,768
& \href{https://openneuro.org/datasets/ds003097}{OpenNeuro} \\ 

\textbf{\dtrown} & \textbf{AOMIC-PIOP1}~\cite{ds002785:2.0.0}
& Young adults (Healthy, 18--26 Years, Netherlands)
& Philips Achieva (T1w) 
& 215
& \href{https://openneuro.org/datasets/ds002785}{OpenNeuro} \\ 

\textbf{\dtrown} & \textbf{AOMIC-PIOP2}~\cite{ds002790:2.0.0} 
& Young adults (Healthy, 18--26 Years, Netherlands)
& Philips Achieva dStream (T1w) 
& 225
& \href{https://openneuro.org/datasets/ds002790}{OpenNeuro} \\ 

\textbf{\dtrown} & \textbf{UKB-1}~\cite{sudlow2015uk} &  Adults (Healthy, 40--69 Years, UK) & Siemens Magnetom Skyra (T1w, T2w, FLAIR) & 16,048 & \href{https://ams.ukbiobank.ac.uk}{UKBiobank} \\

\textbf{\dtrown} & \textbf{UKB-2}~\cite{sudlow2015uk} &  Adults (Healthy, 40--69 Years, UK) & Siemens Magnetom Skyra (T1w, T2w, FLAIR) & 18,215 & \href{https://ams.ukbiobank.ac.uk}{UKBiobank} \\

\textbf{\dtrown} & \textbf{UKB-3}~\cite{sudlow2015uk} &  Adults (Healthy, 40--69 Years, UK) & Siemens Magnetom Skyra (T1w, T2w, FLAIR) & 18,682 & \href{https://ams.ukbiobank.ac.uk}{UKBiobank} \\

\textbf{\dtrown} & \textbf{UKB-4}~\cite{sudlow2015uk} &  Adults (Healthy, 40--69 Years, UK) & Siemens Magnetom Skyra (T1w, T2w, FLAIR) & 17,335 & \href{https://ams.ukbiobank.ac.uk}{UKBiobank} \\

\textbf{\dtrown} & \textbf{UKB-5}~\cite{sudlow2015uk} &  Adults (Healthy, 40--69 Years, UK) & Siemens Magnetom Skyra (T1w, T2w, FLAIR) & 16,715 & \href{https://ams.ukbiobank.ac.uk}{UKBiobank} \\

\textbf{\dtrown} & \textbf{ADNI-1}~\cite{jack2008alzheimer} 
& Elderly (Cognitive decline, 55--90 Years, America)
& Siemens \& GE \& Philips (T1w, T2w, FLAIR) 
& 9,275
& \href{http://adni.loni.usc.edu}{ADNI} \\

\textbf{\dtrown} & \textbf{ADNI-2}~\cite{jack2015magnetic} 
& Elderly (Cognitive decline, 55--90 Years, America)
& Siemens \& GE \& Philips (T1w, T2w, FLAIR) 
& 4,427
& \href{http://adni.loni.usc.edu}{ADNI} \\

\textbf{\dtrown} & \textbf{ADNI-3}~\cite{weiner2017alzheimer} 
& Elderly (Cognitive decline, 55--90 Years, America)
& Siemens \& GE \& Philips (T1w, T2w, FLAIR) 
& 2,717
& \href{http://adni.loni.usc.edu}{ADNI} \\

\textbf{\dtrown} & \textbf{ADNIGo}~\cite{weiner2013alzheimer} 
& Elderly (Early MCI, 55--90 Years, America)
& Siemens \& GE \& Philips (T1w, T2w, FLAIR) 
& 1,159
& \href{http://adni.loni.usc.edu}{ADNI} \\

\textbf{\dtrown} & \textbf{ADNIDOD}~\cite{clark2021higher} 
& Elderly (Vietnam veterans, 50--90 Years, USA)
& Siemens \& GE \& Philips (T1w, T2w, FLAIR) 
& 1,051
& \href{http://adni.loni.usc.edu}{ADNI} \\

\textbf{\dtrown} & \textbf{BraTS23-GLI}~\cite{li2024brain}
& Adults (Glioma, 17--94 Years, Worldwide) 
& Siemens \& GE \& Philips (T1w, T1ce, T2w, FLAIR) 
& 5,892
& \href{https://www.synapse.org/Synapse:syn51156910/wiki}{Synapse} \\

\textbf{\dtrown} & \textbf{BraTS23-MEN}~\cite{labella2023asnr}
& Adults (Meningioma, 17--94 Years, Worldwide) 
& Siemens \& GE \& Philips (T1w, T1ce, T2w, FLAIR) 
& 4,759
& \href{https://www.synapse.org/Synapse:syn51156910/wiki}{Synapse} \\ 

\textbf{\dtrown} & \textbf{BraTS23-MET}~\cite{moawad2024brain}
& Adults (Metastases, 17--94 Years, Worldwide) 
& Siemens \& GE \& Philips (T1w, T1ce, T2w, FLAIR) 
& 1,077
& \href{https://www.synapse.org/Synapse:syn51156910/wiki}{Synapse} \\

\textbf{\dtrown} & \textbf{BraTS23-PED}~\cite{kazerooni2024brain}
& Children (Tumor, 1--20 Years, Worldwide) 
& Siemens \& GE \& Philips (T1w, T1ce, T2w, FLAIR) 
& 580
& \href{https://www.synapse.org/Synapse:syn51156910/wiki}{Synapse} \\ 

\textbf{\dtrown} & \textbf{BraTS23-SSA}~\cite{adewole2023brain}
& Adults (Tumor, 17--94 Years, Africa) 
& Siemens \& GE \& Philips (T1w, T1ce, T2w, FLAIR) 
& 300
& \href{https://www.synapse.org/Synapse:syn51156910/wiki}{Synapse} \\ 

\bottomrule
\end{tabular}}

\vspace{4.0pt}
{\tiny\raggedright
\leftskip=0pt
\textit{Abbreviations:}
MRI, magnetic resonance imaging;
T1w, T1-weighted imaging;
T1ce, contrast-enhanced T1-weighted imaging;
T2w, T2-weighted imaging;
PDw, proton-density-weighted imaging;
FLAIR, fluid-attenuated inversion recovery;
MCI, mild cognitive impairment.
\par}
\end{table}
\hyphenpenalty=10000
\exhyphenpenalty=10000
\definecolor{cTitle}{HTML}{1F2937}  
\definecolor{cAccent}{HTML}{334155} 
\definecolor{cLine}{HTML}{D1D5DB}  
\definecolor{cProc}{HTML}{F9FAFB}   
\definecolor{cQC}{HTML}{F0FDF4}     
\definecolor{cOut}{HTML}{FFF7ED}   
\definecolor{cTitle}{HTML}{1F2933}
\definecolor{cAccent}{HTML}{4B6478}
\definecolor{cLine}{HTML}{CCD2D8}
\definecolor{cProc}{HTML}{F7F8F9}
\definecolor{cQC}{HTML}{F3F6F8}
\definecolor{cOut}{HTML}{F8F5EF}
\newcommand{\Ttl}[1]{{\color{cTitle}\bfseries #1}}
\newcommand{\Sm}[1]{{\scriptsize\color{black!68}#1}}
\newcommand{\Bt}[1]{{\scriptsize\textbullet\hspace{0.55em}#1}}

\begin{figure}[h]
\centering
\footnotesize
\begin{tikzpicture}[
  font=\footnotesize,
  node distance=5.8mm and 8mm,
  >={Stealth[length=2.0mm]},
  proc/.style={
    draw=cLine, line width=0.6pt, fill=cProc,
    rounded corners=2.2mm,
    minimum height=10mm, text width=76mm,
    align=left, inner xsep=3.8mm, inner ysep=2.6mm,
    execute at begin node=\RaggedRight
  },
  qcbox/.style={
    draw=cLine, line width=0.6pt, fill=cQC,
    rounded corners=2.2mm,
    minimum height=10mm, text width=58mm,
    align=left, inner xsep=3.8mm, inner ysep=2.6mm,
    execute at begin node=\RaggedRight
  },
  outbox/.style={
    draw=cLine, line width=0.6pt, fill=cOut,
    rounded corners=2.2mm,
    minimum height=10mm, text width=76mm,
    align=left, inner xsep=3.8mm, inner ysep=2.6mm,
    execute at begin node=\RaggedRight
  },
  arrow/.style={->, thick, draw=cAccent},
  dashedarrow/.style={->, thick, draw=cAccent, dashed},
  legend/.style={
  font=\scriptsize,
  text=black!65,
  align=left,
  text width=176mm
}
]

\node[proc] (s1) {\Ttl{(1) Ingestion \& Curation}\\[-0.3mm]
\Bt{Structural MRI only; remove localizers/duplicates}\\
\Bt{Retain metadata (vendor, field strength, spacing)}\\
\Sm{Outputs: curated scan list + metadata table}
};

\node[proc, below=of s1] (s2) {\Ttl{(2) DICOM $\rightarrow$ NIfTI (BIDS-like)}\\[-0.3mm]
\Bt{dcm2niix conversion; consistent IDs}\\
\Bt{Export sidecar JSON for auditing}\\
\Sm{Outputs: NIfTI (.nii.gz) + JSON}
};

\node[proc, below=of s2] (s3) {\Ttl{(3) Bias-field Correction \& Normalization}\\[-0.3mm]
\Bt{N4 bias correction (SimpleITK)}\\
\Bt{Per-scan intensity normalization (stable SSL)}\\
\Sm{Outputs: intensity-standardized NIfTI}
};

\node[proc, below=of s3] (s4) {\Ttl{(4) Skull Stripping (DeepBET~\cite{deepbet})}\\[-0.3mm]
\Bt{Generate brain mask; remove non-brain tissues}\\
\Bt{Mask QC: over-strip / holes / coverage flags}\\
\Sm{Outputs: brain-only NIfTI + brain mask}
};

\node[proc, below=of s4] (s5) {\Ttl{(5) Resampling (Geometry Harmonization)}\\[-0.3mm]
\Bt{Resample to 1$\times$1$\times$1 mm$^3$ (linear)}\\
\Bt{Consistent voxel geometry for 3D patching}\\
\Sm{Outputs: isotropic brain-only NIfTI}
};

\node[proc, below=of s5] (s6) {\Ttl{(6) Rigid MNI Registration}\\[-0.3mm]
\Bt{Rigid registration to MNI atlas (6-DoF)}\\
\Bt{Similarity metric: mutual information}\\
\Sm{Outputs: MNI-aligned NIfTI + rigid transform}
};

\node[outbox, below=10mm of s6] (out) {\Ttl{Output: Pretraining-ready Corpus}\\[-0.3mm]
\Bt{MNI-aligned volumes + brain masks}\\
\Bt{Transforms + metadata + QC logs}\\
\Sm{Deliverables: dataset package + summary report}
};
\draw[arrow] (s1) -- (s2);
\draw[arrow] (s2) -- (s3);
\draw[arrow] (s3) -- (s4);
\draw[arrow] (s4) -- (s5);
\draw[arrow] (s5) -- (s6);
\draw[arrow] (s6) -- (out);

\node[qcbox, right=20mm of s3] (q1) {\Ttl{(Q1) Automatic QC}\\[-0.3mm]
\Bt{Header/orientation; spacing sanity}\\
\Bt{Coverage/resolution; artifact heuristics}\\
\Sm{Outputs: QC flags per scan}
};

\node[qcbox, below=of q1] (q2) {\Ttl{(Q2) MNI Registration QC}\\[-0.3mm]
\Bt{Similarity score + outlier detection}\\
\Bt{Failure handling: rerun / exclude}\\
\Sm{Outputs: pass/fail + scores}
};

\node[qcbox, below=of q2] (q3) {\Ttl{(Q3) Dataset Audit \& Release}\\[-0.3mm]
\Bt{Per-scan QC table; cohort summaries}\\
\Bt{Manifest files; provenance logs (versioned)}\\
\Sm{Outputs: release CSVs + summary reports}
};

\draw[dashedarrow] (s2.east) -- ++(10mm,0) |- (q1.west);
\draw[dashedarrow] (s4.east) -- ++(10mm,0) |- (q1.west);
\draw[dashedarrow] (s6.east) -- ++(10mm,0) |- (q2.west);
\draw[dashedarrow] (q1) -- (q2);
\draw[dashedarrow] (q2) -- (q3);
\draw[dashedarrow] (q3.south) |- (out.east);

\node[legend, anchor=north west]
at ($(out.south west)+(0,-5mm)$)
{Solid arrows: local preprocessing steps performed independently at each site; dashed arrows: local QC gating, auditing, and logging.};

\end{tikzpicture}
\caption{\textbf{Standardized site-level preprocessing workflow.}
The same pipeline is applied locally and independently at each federated site to prepare structural brain MRI scans and quality-control records for federated pretraining. This workflow standardizes preprocessing across sites without requiring raw-data sharing or centralized data pooling.}
\label{suppfig:preprocess}
\end{figure}

\begin{table}[t]
\centering
\caption[Key configurations in BrainFedFM]{
Key configurations for data preprocessing, federated pretraining, optimization, and computational resources in BrainFedFM.}
\label{tab:brainfedfm_pretraining_configuration}
\small
\renewcommand{\arraystretch}{1.0888}
\setlength{\tabcolsep}{6pt}

\begin{tabularx}{0.94\linewidth}
{
>{\raggedright\arraybackslash}p{0.28\linewidth}
>{\raggedright\arraybackslash}X
}
\toprule
\textbf{Item} & \textbf{Setting} \\
\midrule

\multicolumn{2}{l}{\textbf{Pre-processing}} \\

Spatial resampling
& $1.0 \times 1.0 \times 1.0~\mathrm{mm}^{3}$ \\

Intensity normalization
& Per-volume z-score normalization \\

Input patch size
& $160 \times 160 \times 160$ voxels \\

\addlinespace[2pt]
\midrule
\multicolumn{2}{l}{\textbf{Federated pretraining}} \\

Federated setting
& Simulated cross-silo federated pretraining with all 42 sites participating in each communication round \\

Local training budget
& 200 local gradient-update steps per site per communication round \\

Pretraining views
& Random, lifespan-guided, and difficulty-guided masked views \\

View loss weights
& Equal weighting of the three masked-reconstruction losses \\

Communication rounds
& 150 \\

Masking ratio
& 50\% for each view \\

Mask block size
& $16 \times 16 \times 16$ voxels \\

Lifespan age bins
& $[0,12)$, $[12,20)$, $[20,40)$, $[40,65)$, and $[65,120]$ years \\

Reconstruction objective
& Mean squared error computed over masked voxels \\

Cross-view consistency
& Smooth-$L_{1}$ consistency loss activated after 10\% of rounds, with its weight increased linearly from zero to $\lambda_{\mathrm{con}}=0.03$ by 50\% of the pretraining schedule \\

Site aggregation
& Priority-aware aggregation based on lifespan support and residual reconstruction difficulty \\

Aggregation coefficients
& $\lambda_{\mathrm{sup}}=1.0$ and $\lambda_{\mathrm{dif}}=1.0$ \\

Global priority memory
& Updated using an EMA with $\eta=0.2$ and a round-dependent mixing coefficient $\beta_{r}$ decreasing linearly from $0.75$ to $0.50$ \\

Guidance schedule
& Lifespan and global-memory guidance were activated after 10\% of rounds, and difficulty guidance after 20\%; each was linearly ramped from 0.25 to 1.0 over the following ten rounds \\

Checkpoint selection
& Lowest uniformly averaged site-level validation reconstruction loss, with cross-view consistency excluded \\

\addlinespace[2pt]
\midrule
\multicolumn{2}{l}{\textbf{Optimization resources}} \\

Optimizer 
& Stochastic gradient descent with Nesterov momentum \\

Learning rate
& $1 \times 10^{-2}$ \\

Momentum
& $0.99$ \\

Weight decay
& $3 \times 10^{-5}$ \\

Learning-rate schedule
& Constant learning rate during federated pretraining \\

Numerical precision
& Automatic mixed-precision training \\

Gradient clipping
& Maximum gradient norm of 12 \\

Pretraining resources
& $4\times$ NVIDIA H800 GPUs (80~GB each) \\

Inference resources
& $1\times$ NVIDIA GeForce RTX 3090 GPU (24~GB) \\

\bottomrule
\end{tabularx}
\end{table}

\begin{figure}[t]
\centering
\includegraphics[width=1.0\textwidth]{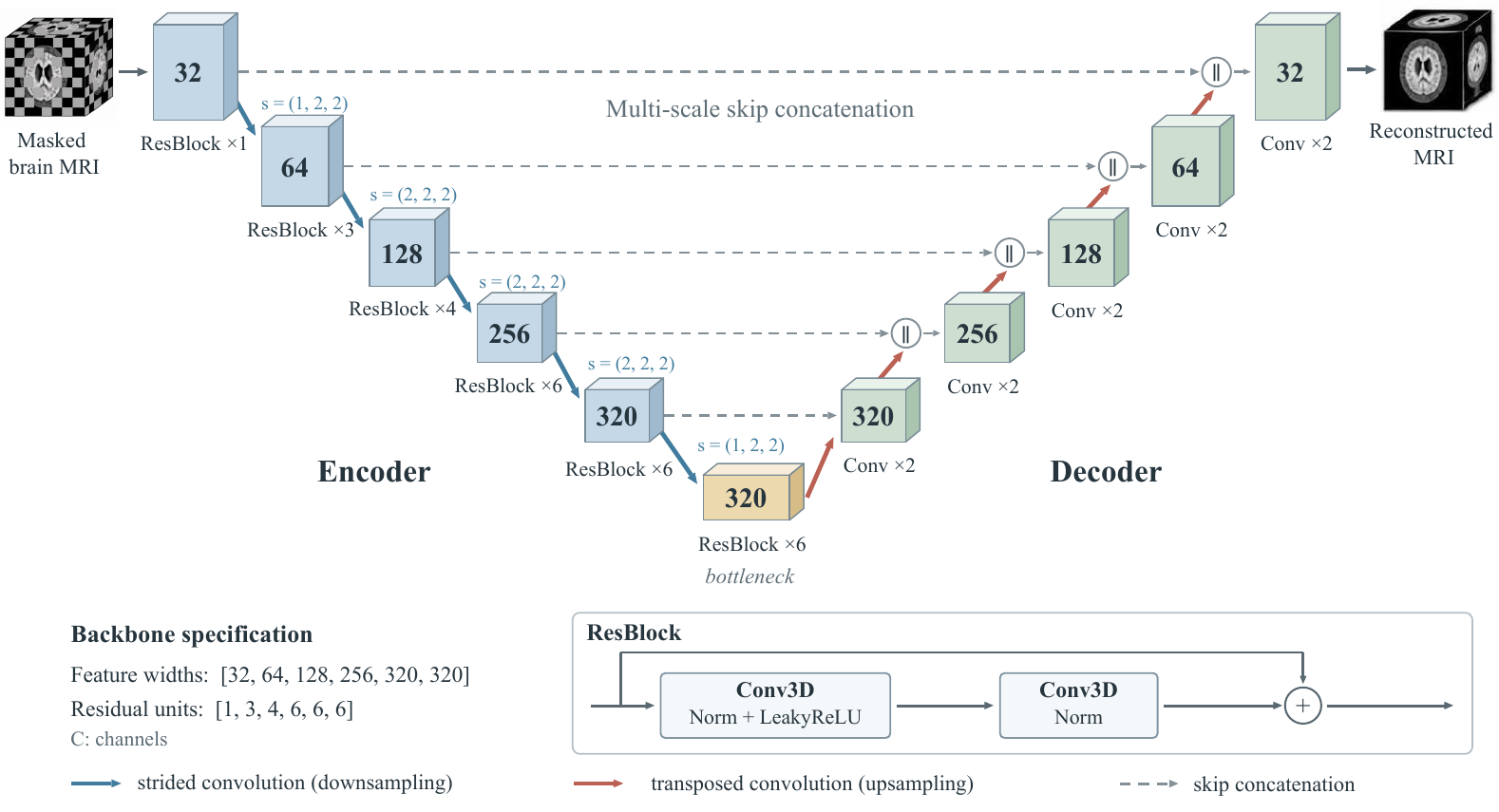}
\par\vspace{1pt}
\caption{\textbf{Pretraining backbone architecture.} Our pretraining backbone follows a 3D U-Net-style encoder--decoder architecture~\cite{isensee2024nnu}. It comprises a 3D ResEnc-L residual encoder, a 3D U-Net decoder, and multi-scale skip concatenations between corresponding stages. The encoder hierarchically extracts volumetric features through residual convolutional blocks and strided downsampling, while the decoder progressively reconstructs high-resolution representations via transposed convolutions and encoder--decoder feature fusion. Compared with heavier transformer-style alternatives, this fully convolutional backbone provides a practical and robust choice for structural brain MRI, with strong locality modeling and favorable suitability for federated pretraining across heterogeneous sites.}
\label{fig:backbone}
\end{figure}

\begin{figure}[t]
  \centering
\includegraphics[width=1.0\textwidth]{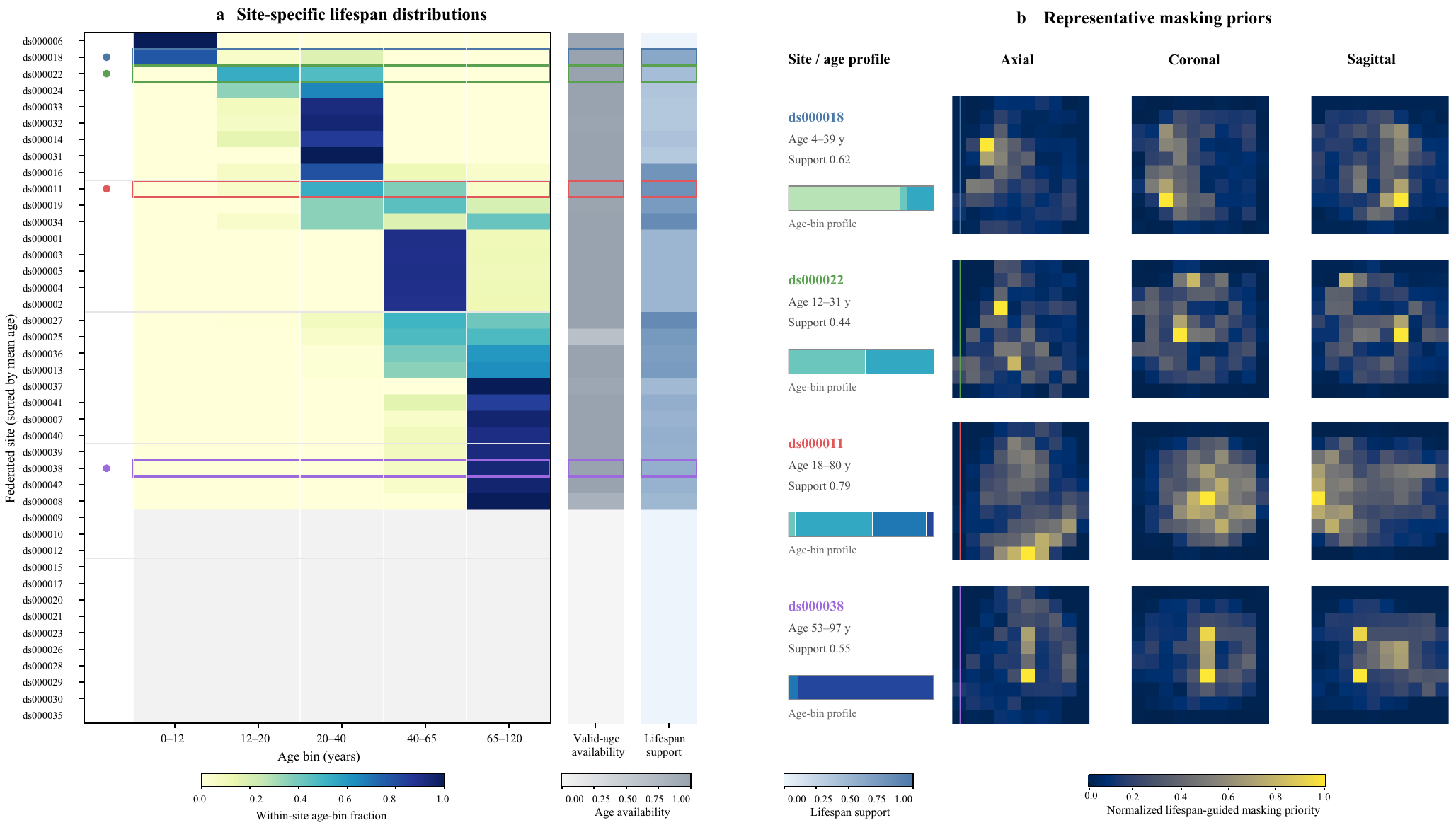}
\caption{
\textbf{Site-specific lifespan distributions and lifespan-guided masking priors.}
\textbf{a}, Distribution of subjects across five age bins at each federated site, ordered by mean age. The adjacent columns report age availability and lifespan support. The lifespan-support score integrates valid-age availability, normalized age-span coverage, occupied age-bin coverage, and lower-quartile sample density across non-empty bins. Colored outlines identify the representative sites shown in panel b.
\textbf{b}, Representative site-specific masking-priority maps used to construct the lifespan-guided view. Axial, coronal, and sagittal maximum-intensity projections are shown, with higher intensities indicating spatial blocks assigned greater masking priority.
}
\label{fig:lifespan_prior}
\end{figure}

\begin{figure}[t]
  \centering
\includegraphics[width=1.0\textwidth]{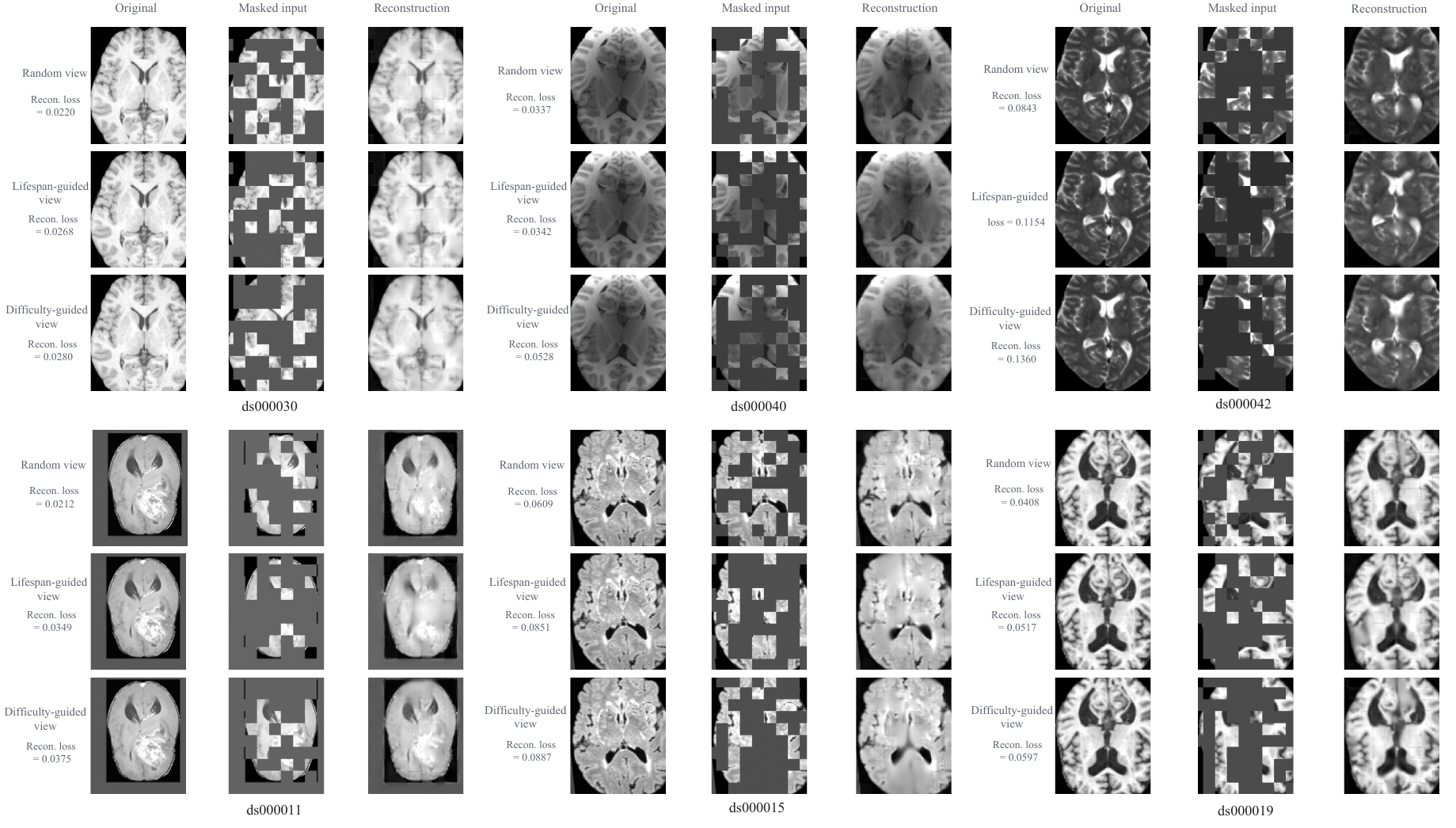}
\caption{
\textbf{Representative reconstruction examples during federated pretraining.}
Representative axial examples from six selected federated sites. Columns show the original image, masked input, and reconstructed image, while rows compare the three views used in spatial-priority masking on the same image. The random view provides broad spatial coverage, the lifespan-guided view emphasizes regions informed by lifespan-associated variation, and the difficulty-guided view prioritizes regions with high reconstruction difficulty at each site. Together, these views provide complementary broad-coverage, lifespan-associated, and reconstruction-difficulty signals for masked pretraining of structural brain MRI.
}
\label{fig:reconstruction}
\end{figure}

\begin{figure}[t]
  \centering
\includegraphics[width=1.0\textwidth]{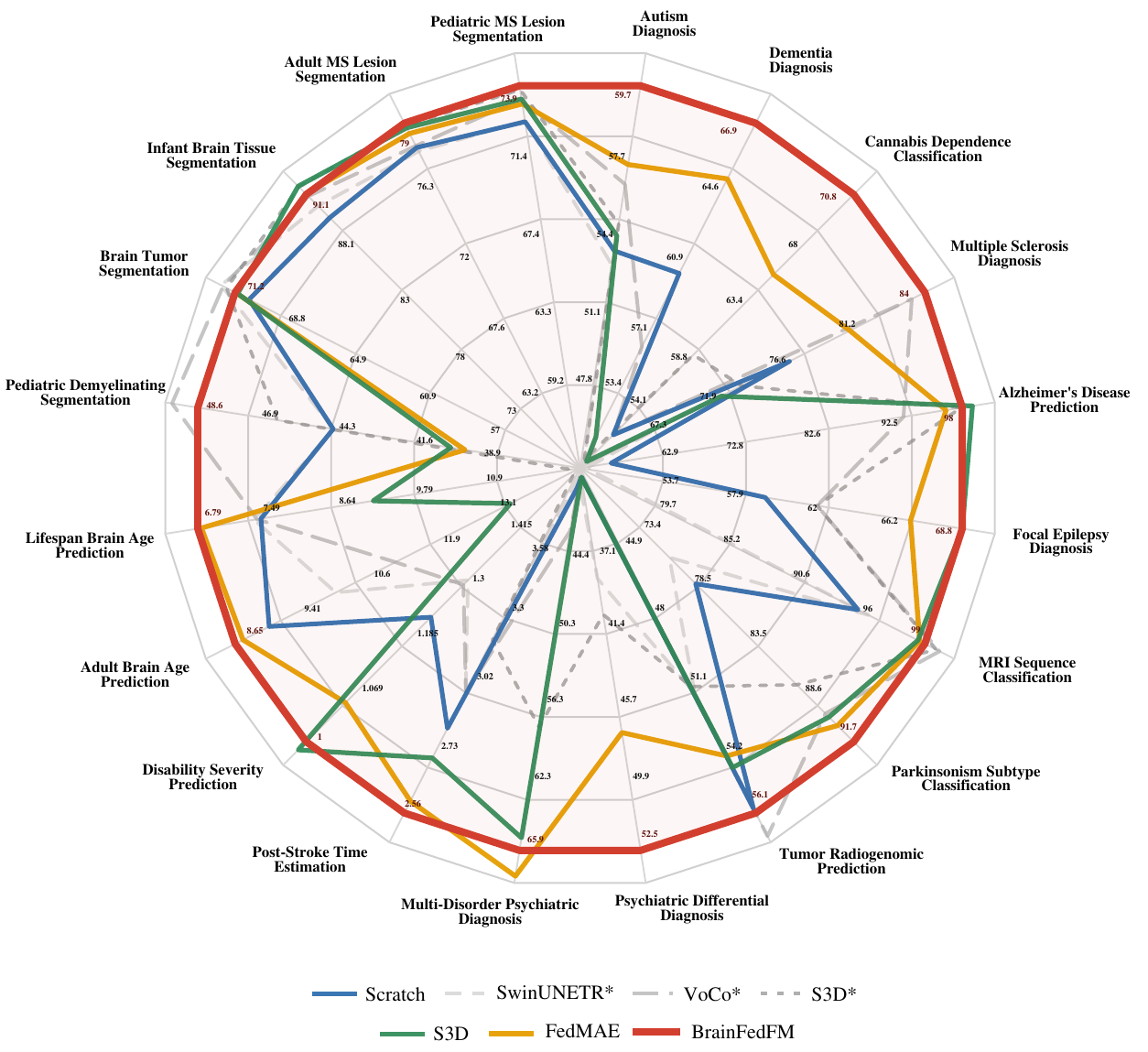}
\caption{
\textbf{Downstream performance across 20 benchmarks under the full-label setting.}
The radar plot compares BrainFedFM with the comparison foundation models using all labelled training and validation data.
Each spoke represents one dataset--task benchmark and its primary evaluation metric.
BrainFedFM forms the red reference contour; models outside or inside this contour perform better or worse, respectively, on the corresponding benchmark.
Larger radial positions indicate better performance, and tick labels report the original metric values.
}
\label{fig:radar_full}
\end{figure}

\begin{figure}[t]
  \centering
\includegraphics[width=1.0\textwidth]{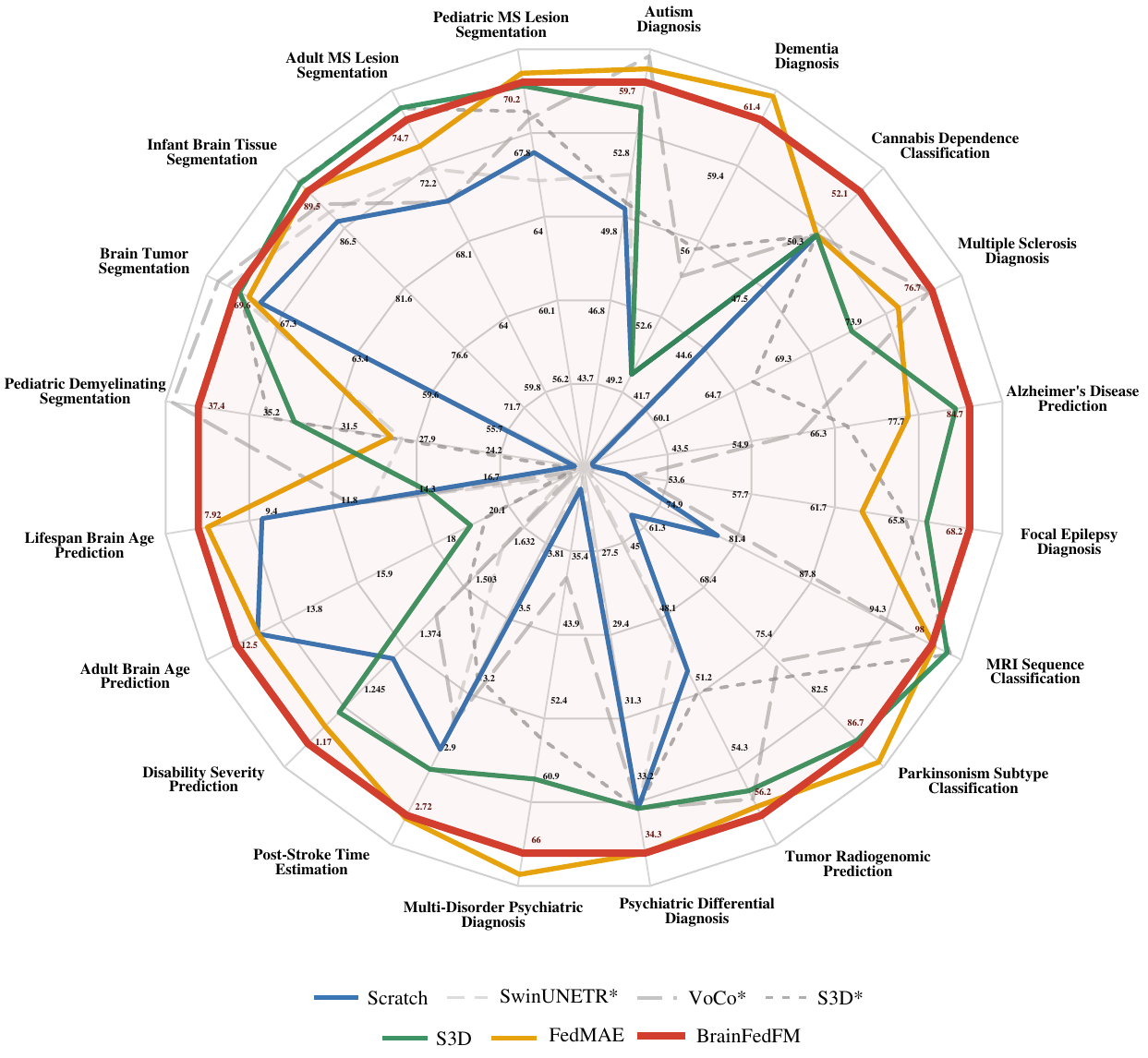}
\caption{
\textbf{Downstream performance across 20 benchmarks with 50\% labelled data.}
The radar plot compares BrainFedFM with the comparison foundation models using 50\% of the original training and validation subjects, while retaining the same test sets as in the full-label setting.
Each spoke represents one dataset--task benchmark and its primary evaluation metric.
BrainFedFM forms the red reference contour; models outside or inside this contour perform better or worse, respectively, on the corresponding benchmark.
Larger radial positions indicate better performance, and tick labels report the original metric values.
}
\label{fig:radar_50}
\end{figure}

\begin{figure}[t]
  \centering
\includegraphics[width=1.0\textwidth]{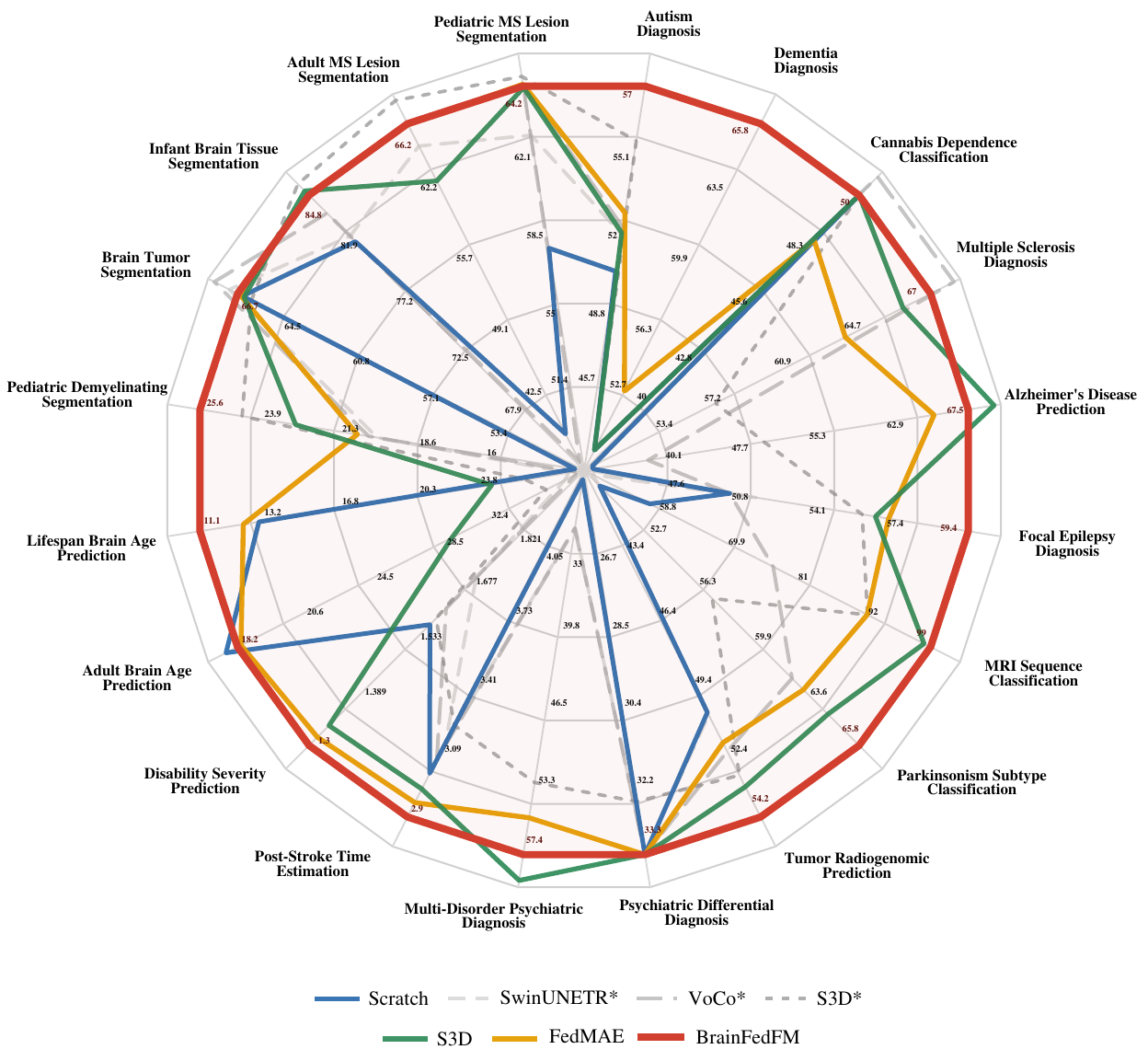}
\caption{
\textbf{Downstream performance across 20 benchmarks with 25\% labelled data.}
The radar plot compares BrainFedFM with the comparison foundation models using 25\% of the original training and validation subjects, while retaining the same test sets as in the full-label setting.
Each spoke represents one dataset--task benchmark and its primary evaluation metric.
BrainFedFM forms the red reference contour; models outside or inside this contour perform better or worse, respectively, on the corresponding benchmark.
Larger radial positions indicate better performance, and tick labels report the original metric values.
}
\label{fig:radar_25}
\end{figure}

\clearpage

\setcounter{section}{4}
\setcounter{subsection}{4}
\setcounter{secnumdepth}{3}

\subsection{Detailed formulation of spatial-priority masking}
\label{sec:supp_spatial_priority}

At each federated site, spatial-priority masking generated three views of every input patch: random, lifespan-guided, and difficulty-guided. All views used the same model parameters, reconstruction target, and masking ratio. They differed only in their block-sampling distributions. Random masking sampled uniformly from the $B=1{,}000$ spatial blocks, whereas each guided view combined local spatial evidence with the global priority memory.

\subsubsection{Lifespan-guided priority}

The local lifespan prior was estimated independently at each site before federated pretraining, with lifespan-associated variation calculated using comparisons within the same imaging sequence to reduce sequence-dependent contrast effects. For notational simplicity, the sequence index is omitted below.
Scans with valid age metadata were assigned to five common intervals indexed by $m$: $[0,12)$, $[12,20)$, $[20,40)$, $[40,65)$, and $[65,120]$ years. For block $b$ in volume $x$, its local image appearance was summarized by the mean absolute intensity and within-block intensity variation: 
\[
d_b(x)
=
\operatorname{Mean}_{i\in\mathcal{V}_b}|x_i|
+
\frac{1}{2}
\operatorname{Std}_{i\in\mathcal{V}_b}(x_i),
\]
where $\mathcal{V}_b$ denotes the voxels within block $b$. The descriptor was then averaged across all local scans assigned to age interval $m$:
\[
\mu_{k,m,b}
=
\frac{1}{|\mathcal{I}_{k,m}|}
\sum_{x\in\mathcal{I}_{k,m}} d_b(x),
\]
where $\mathcal{I}_{k,m}$ contains the scans from site $k$ in interval $m$ within the corresponding imaging sequence.
The local lifespan priority was defined as the normalized variation in these interval-specific means:
\[
l_{k,b}
=
\operatorname{Norm}\left(
\operatorname{Var}_{m:\,|\mathcal{I}_{k,m}|>0}
\left[\mu_{k,m,b}\right]
\right),
\]
where $\operatorname{Norm}(\cdot)$ denotes normalization of the block-wise scores to a common scale.
A larger $l_{k,b}$ indicated a greater difference in the block-level intensity descriptor across the age intervals represented at site $k$. To avoid unstable estimates, intervals containing at least three scans were used whenever at least two such intervals were available; otherwise, all nonempty intervals were used if at least two intervals were represented. The local lifespan prior was computed once before federated pretraining and remained fixed throughout training. 

Because the reliability of the local lifespan prior depended on the completeness and distribution of age metadata, its contribution was modulated by a site-specific support score. The score incorporated valid-age availability, age-interval coverage, observed age span, and sample density across represented intervals:
\[
a_k
=
\frac{N_k^{\mathrm{valid}}}{N_k},
\qquad
o_k
=
\frac{B_k^{\mathrm{nonempty}}-1}
{B_{\mathrm{age}}-1},
\]

\[
r_k
=
\min\left(
1,
\frac{\max(\mathcal{A}_k)-\min(\mathcal{A}_k)}{120}
\right),
\quad
d_k
=
\min\left(
1,
\frac{
Q_{25}\left(\{n_{k,m}:n_{k,m}>0\}\right)
}{8}
\right),
\]
where $B_{\mathrm{age}}=5$, $N_k^{\mathrm{valid}}$ and $N_k$ denote the numbers of scans with valid age metadata and total scans at site $k$, respectively; $B_k^{\mathrm{nonempty}}$ is the number of represented age intervals; $\mathcal{A}_k$ is the set of valid ages; $n_{k,m}$ is the number of scans in interval $m$; and $Q_{25}$ denotes the lower quartile of scan counts across nonempty intervals. The four components were combined using their geometric mean:
\[
s_k
=
\left(a_k o_k r_k d_k\right)^{1/4}.
\]
Incomplete age metadata, narrow age coverage, or sparsely populated intervals reduced $s_k$ and consequently weakened the contribution of the local lifespan prior. 
If age metadata were unavailable or fewer than two age intervals were represented, the site still generated a lifespan-guided masked view, but its block-sampling priorities were determined solely by the global priority memory accumulated across the participating sites. 
Age metadata were used only to construct the local masking prior and support score and were not provided as input to the model.

\subsubsection{Difficulty-guided priority}

In parallel, each site maintained a dynamic difficulty map
$h_k^{(r)}\in\mathbb{R}^{B}$ to identify spatial regions that remained
difficult to reconstruct during local pretraining. For each block that was
masked, squared voxel-wise reconstruction errors were averaged within the
block and across local mini-batches. Let $e_{k,b}^{(r)}$ denote the resulting
error for block $b$ at communication round $r$. The difficulty map was updated
using an exponential moving average:
\[
h_{k,b}^{(r)}
=
\operatorname{Norm}\left[
\zeta h_{k,b}^{(r-1)}
+
(1-\zeta)e_{k,b}^{(r)}
\right],
\quad
\zeta=0.9,
\]
the coefficient $\zeta$ retained 90\% of the previous difficulty estimate
and incorporated 10\% of the current reconstruction error. This smoothing
assigned higher priority to blocks that remained difficult across repeated
observations, while limiting the influence of occasional high errors.

\subsubsection{Integration with the global priority memory}

The local lifespan and difficulty maps were combined with the global priority memory before guided block sampling. To reduce the influence of extreme values, each map was clipped at its 5th and 95th percentiles and then
min--max normalized. This robust normalization operation is denoted by $\mathcal{R}(\cdot)$. At communication round $r$, the two guided priority
maps were defined as
\[
\begin{aligned}
p_k^{\mathrm{life},(r)}
&=
\operatorname{Norm}\left[
\alpha_l s_k\mathcal{R}(l_k)
+
\alpha_g\mathcal{R}\left(q^{(r-1)}\right)
\right],\\
p_k^{\mathrm{dif},(r)}
&=
\operatorname{Norm}\left[
\alpha_d\mathcal{R}\left(h_k^{(r-1)}\right)
+
\alpha_g\mathcal{R}\left(q^{(r-1)}\right)
\right],
\end{aligned}
\]
where $\alpha_l$ and $\alpha_d$ controlled the contributions of the local
lifespan and difficulty maps, respectively, whereas $\alpha_g$ controlled
the contribution of the global priority memory. 
We used $\alpha_l=1.0$, $\alpha_d=1.0$, and $\alpha_g=0.5$.
The lifespan support score $s_k$ scaled only the local lifespan prior. Consequently, sites with incomplete or narrowly distributed age metadata received less influence from their local lifespan maps, while the contribution of the global priority memory remained unchanged.
For guided view $v\in\{\mathrm{life},\mathrm{dif}\}$, the block priorities were converted into sampling weights:
\[
\omega_{k,b}^{v,(r)}
=
1+
\kappa g_v(r)
\left(p_{k,b}^{v,(r)}\right)^{\gamma},
\quad
\kappa=2.0,
\ \
\gamma=1.5,
\]
where $g_v(r)$ denotes the round-dependent guidance schedule. For each guided view, the blocks to be masked were sampled without replacement, with selection probabilities proportional to $\omega_{k,b}^{v,(r)}$. The additive baseline of one preserved a nonzero sampling probability for every block, whereas the priority-dependent term increased the likelihood of masking regions highlighted by local lifespan-associated variation, persistent reconstruction difficulty, or spatial evidence accumulated in the global priority memory.
Thus, the guidance shifted sampling toward prioritized regions while retaining stochastic coverage across the entire brain, rather than enforcing a fixed anatomical mask.

\subsubsection{Joint local-pretraining objective}

The random, lifespan-guided, and difficulty-guided views were optimized jointly during each local update. Their masked-reconstruction losses were averaged, and a low-weight consistency term was added to reduce disagreement among the full-volume reconstructions generated from different visible anatomical contexts, thereby promoting stable reconstruction of the same underlying anatomy across different masking patterns:
\[
\begin{aligned}
\mathcal{L}_{\mathrm{local}}^{(r)}
={}&
\frac{1}{3}
\left(
\mathcal{L}_{\mathrm{rec}}^{\mathrm{rand}}
+
\mathcal{L}_{\mathrm{rec}}^{\mathrm{life}}
+
\mathcal{L}_{\mathrm{rec}}^{\mathrm{dif}}
\right)+
\lambda_{\mathrm{con}}g_c(r)
\left(
\mathcal{L}_{\mathrm{con}}^{\mathrm{life}}
+
\mathcal{L}_{\mathrm{con}}^{\mathrm{dif}}
\right),
\end{aligned}
\]
where $\lambda_{\mathrm{con}}=0.03$ and $g_c(r)$ denotes the round-dependent consistency schedule. The random-view reconstruction served
as the reference for the lifespan-guided view, whereas the
difficulty-guided reconstruction was regularized toward the mean of the random and lifespan-guided reconstructions:
\[
\begin{aligned}
\mathcal{L}_{\mathrm{con}}^{\mathrm{life}}
&=
\operatorname{SmoothL1}\left(
\hat{x}^{\mathrm{life}},
\operatorname{sg}\left[\hat{x}^{\mathrm{rand}}\right]
\right),\\
\mathcal{L}_{\mathrm{con}}^{\mathrm{dif}}
&=
\operatorname{SmoothL1}\left(
\hat{x}^{\mathrm{dif}},
\frac{1}{2}
\operatorname{sg}\left[
\hat{x}^{\mathrm{rand}}
+
\hat{x}^{\mathrm{life}}
\right]
\right),
\end{aligned}
\]
where $\operatorname{sg}$ denotes the stop-gradient operation. Reconstruction losses were evaluated only over the voxels masked in each view, whereas consistency losses were evaluated over the corresponding full-volume model outputs. 
$\operatorname{sg}$ prevented the reference reconstructions from receiving gradients from the consistency terms, thereby providing stable targets for the guided views.

\definecolor{pct100}{gray}{0.97}
\definecolor{pct75}{gray}{1.0}
\definecolor{pct50}{gray}{0.99}
\definecolor{pct25}{gray}{1.0}
\definecolor{overallgray}{gray}{1.0}
\definecolor{ourscol}{RGB}{244,248,252}

\begin{table*}[t]
\centering
\tiny
\caption[Autism diagnosis task]{\textbf{Autism diagnosis task.}
We evaluate foundation model performance on the ABIDE benchmark for autism spectrum disorder (ASD) classification from structural brain MRI. This task assesses whether MRI representations can capture imaging patterns useful for distinguishing ASD from control subjects.
The table summarizes performance under 25\%, 50\%, and 100\% labelled-data settings, which indicate the proportion of the labelled training--validation pool used for downstream fine-tuning and model selection; the held-out test set remains unchanged across all settings.
We compare a randomly initialized scratch baseline, centralized pretrained foundation models, and federated pretrained foundation models.}
\label{tab:result_autism}

\renewcommand{\arraystretch}{3.85}
{\scriptsize\bfseries Panel A. Summary under different labelled-data percentages.}
\vspace{5pt}

{\setlength{\tabcolsep}{3.38pt}

\resizebox{1.0\textwidth}{!}{
\begin{tabular}{
l c c c c c c
>{\columncolor{ourscol}\centering\arraybackslash}p{1.62cm}
}
\toprule
\multicolumn{2}{c}{\textbf{Brain sMRI FMs }}
& \multicolumn{4}{c}{Centralized Pretraining}
& \multicolumn{2}{c}{Federated Pretraining} \\
\cmidrule(lr){1-2}\cmidrule(lr){3-6}\cmidrule(lr){7-8}
\makecell[c]{Metric/Methods}
& Scratch
& \makecell[c]{SwinUNETR$^{\ast}$}  &\makecell[c]{VoCo$^{\ast}$} & \makecell[c]{S3D$^{\ast}$} & \makecell[c]{S3D}
& \makecell[c]{FedMAE} & \makecell[c]{BrainFedFM} \\
\midrule

\rowcolor{pct100}
\textbf{100\%  BA $\uparrow$}
&  53.15($\pm$2.024) & 52.57($\pm$1.571) &  55.80($\pm$3.829) &  54.22($\pm$4.507) & 53.75($\pm$1.379)  & 56.58($\pm$0.807)
& \cellcolor{ourscol}59.71($\pm$1.597) \\
\rowcolor{pct100}
\textbf{100\%   mF1 $\uparrow$}
& 49.36($\pm$4.069) & 49.97($\pm$5.052)  & 52.51($\pm$7.893) & 48.45($\pm$8.423) & 51.60($\pm$2.528)  & 53.93($\pm$1.273)
& \cellcolor{ourscol}59.07($\pm$1.443) \\
\rowcolor{pct100}
\textbf{100\% AUC $\uparrow$}
& 56.44($\pm$1.539) & 54.93($\pm$0.344) & 60.27($\pm$2.188) & 56.59($\pm$2.316) & 54.68($\pm$3.098) & 60.34($\pm$0.449)
& \cellcolor{ourscol}64.46($\pm$0.922) \\
\midrule

\rowcolor{pct50}
\textbf{50\%  BA $\uparrow$}
& 50.06($\pm$0.835) & 51.30($\pm$1.250) &  56.26($\pm$1.240) & 50.28($\pm$0.395) & 53.71($\pm$2.070)  & 55.46($\pm$2.243)
& \cellcolor{ourscol} 54.63($\pm$1.793) \\
\rowcolor{pct50}
\textbf{50\%  mF1 $\uparrow$}
& 42.57($\pm$5.598) & 42.62($\pm$6.629) & 55.02($\pm$2.089) & 36.89($\pm$3.722)  & 51.15($\pm$5.023)  & 54.40($\pm$3.338)
& \cellcolor{ourscol}52.23($\pm$4.930) \\
\rowcolor{pct50}
\textbf{50\% AUC $\uparrow$}
&  51.80($\pm$1.439) & 54.46($\pm$1.259) & 58.56($\pm$0.976) & 57.51($\pm$1.064) & 56.90($\pm$1.771)  &  58.31($\pm$0.414)
& \cellcolor{ourscol}58.26($\pm$0.367)\\
\midrule

\rowcolor{pct25}
\textbf{25\%  BA $\uparrow$}
&  50.03($\pm$0.030) & 51.43($\pm$2.028) & 51.70($\pm$1.241) & 54.95($\pm$2.043) &51.48($\pm$2.037)  &  52.24($\pm$1.459)
& \cellcolor{ourscol}57.02($\pm$0.914) \\
\rowcolor{pct25}
\textbf{25\%  mF1 $\uparrow$}
& 39.94($\pm$6.976) & 39.03($\pm$6.745) & 41.16($\pm$4.077) &  53.61($\pm$2.637) & 41.12($\pm$8.641)  & 47.50($\pm$5.098)
& \cellcolor{ourscol}56.89($\pm$0.965) \\
\rowcolor{pct25}
\textbf{25\% AUC $\uparrow$}
& 51.70($\pm$1.506) & 52.92($\pm$1.509) & 54.77($\pm$2.164) & 56.16($\pm$2.246) & 54.33($\pm$1.177)  &  55.90($\pm$0.810)
& \cellcolor{ourscol}57.73($\pm$1.314) \\
\bottomrule
\end{tabular}
}}

\vspace{5pt}

\begin{minipage}{1.0\textwidth}
\scriptsize
\textit{Notes. Values are mean($\pm$std) over three folds. BA, balanced accuracy; mF1, macro F1-score; AUC, macro one-vs-rest ROC AUC. Scratch uses random initialization; SwinUNETR$^{\ast}$, VoCo$^{\ast}$, and S3D$^{\ast}$ use publicly released centralized foundation model weights; S3D uses centralized pretraining on our pretraining corpus, FedMAE and BrainFedFM use federated pretraining.
Fine-tuning uses 25 epochs of AdamW, learning rate $1\times10^{-6}$, weight decay $1\times10^{-2}$, cosine scheduling with a 3-epoch warm-up, and label smoothing of 0.2. The best checkpoint is selected according to validation AUC.}
\end{minipage}
\end{table*}

\clearpage

\noindent{\scriptsize\bfseries Panel B. Labelled-data scaling and aggregate metric summaries.}
\vspace{2pt}
{\tiny\raggedright
Summarizes method behavior across labelled-data settings.
\textbf{B1}, fold-averaged trends for classification and calibration metrics; \textbf{B2}, extended metric heatmaps separating higher-is-better classification metrics from lower-is-better calibration/uncertainty metrics; \textbf{B3}, the accuracy--calibration trade-off, where points closer to the upper-left indicate stronger discrimination with better calibration.\par}
\vspace{-3pt}

\begin{center}
\includegraphics[width=1.0\textwidth,keepaspectratio]{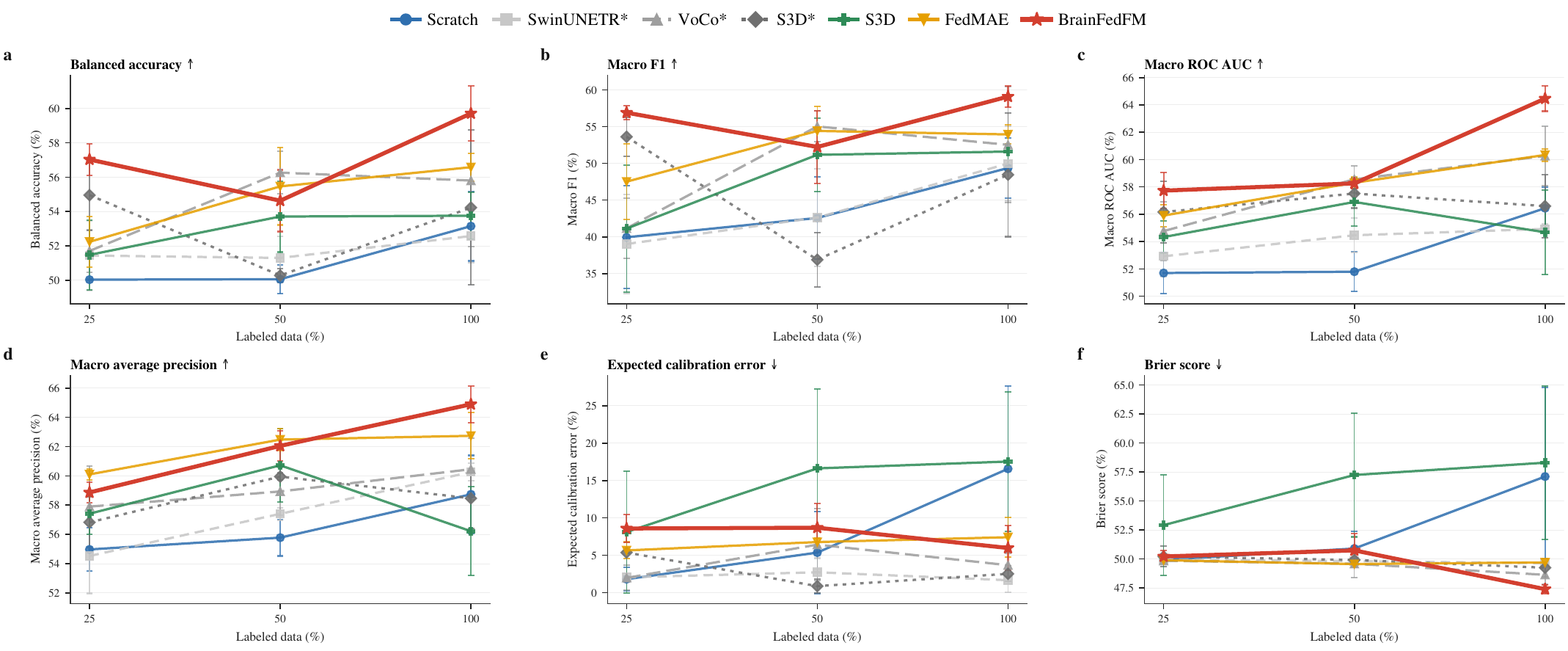} \par\vspace{3pt}
\includegraphics[width=1.0\textwidth,keepaspectratio]{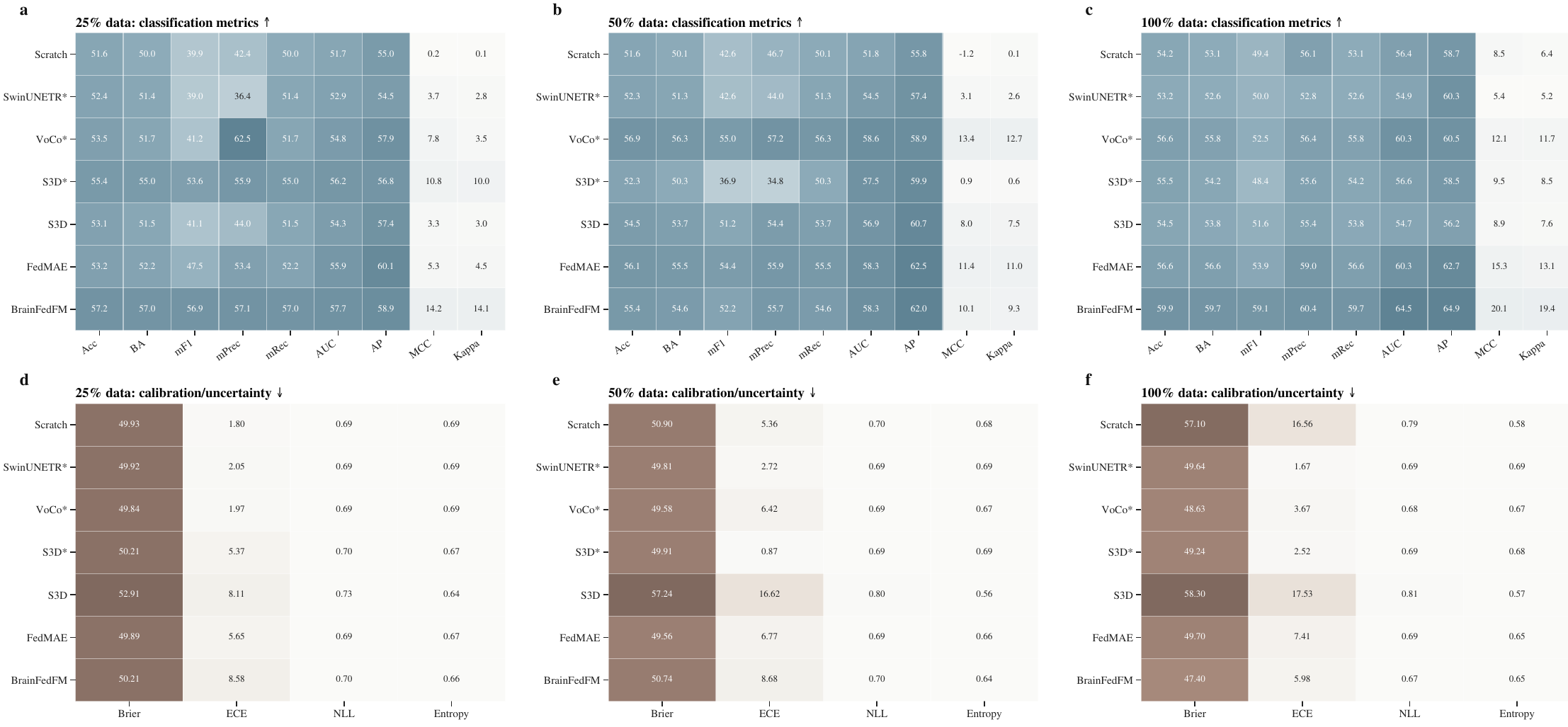} \par\vspace{5pt}
\includegraphics[width=0.935\textwidth,keepaspectratio]{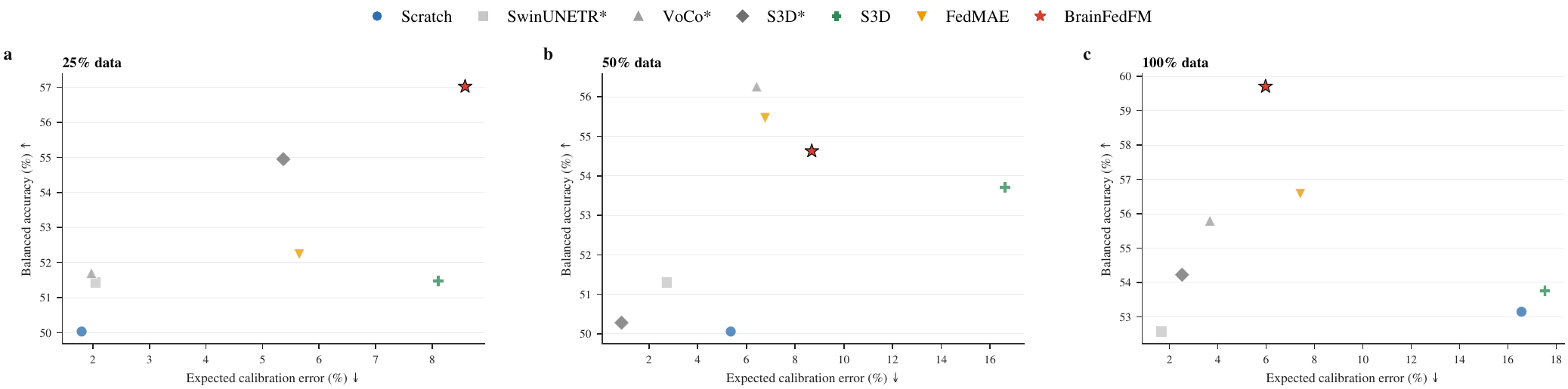}
\end{center}

\vspace{-2pt}

\noindent{\scriptsize\bfseries Panel C. Fold stability and full-data performance analyses.}
\vspace{2pt}
{\tiny\raggedright
Examines robustness and error modes beyond the aggregate scores. \textbf{C1}, fold-to-fold variability and average method ranking across balanced accuracy, macro F1, and macro ROC AUC; \textbf{C2}, probability-level analysis at 100\% labelled data, including micro-averaged ROC/precision--recall curves, reliability curves, and calibration scores; \textbf{C3}, row-normalized confusion matrices at 100\% labelled data, with rows denoting true classes and columns denoting predicted classes.\par}

\clearpage

\begin{center}
\includegraphics[width=0.84\textwidth,keepaspectratio]{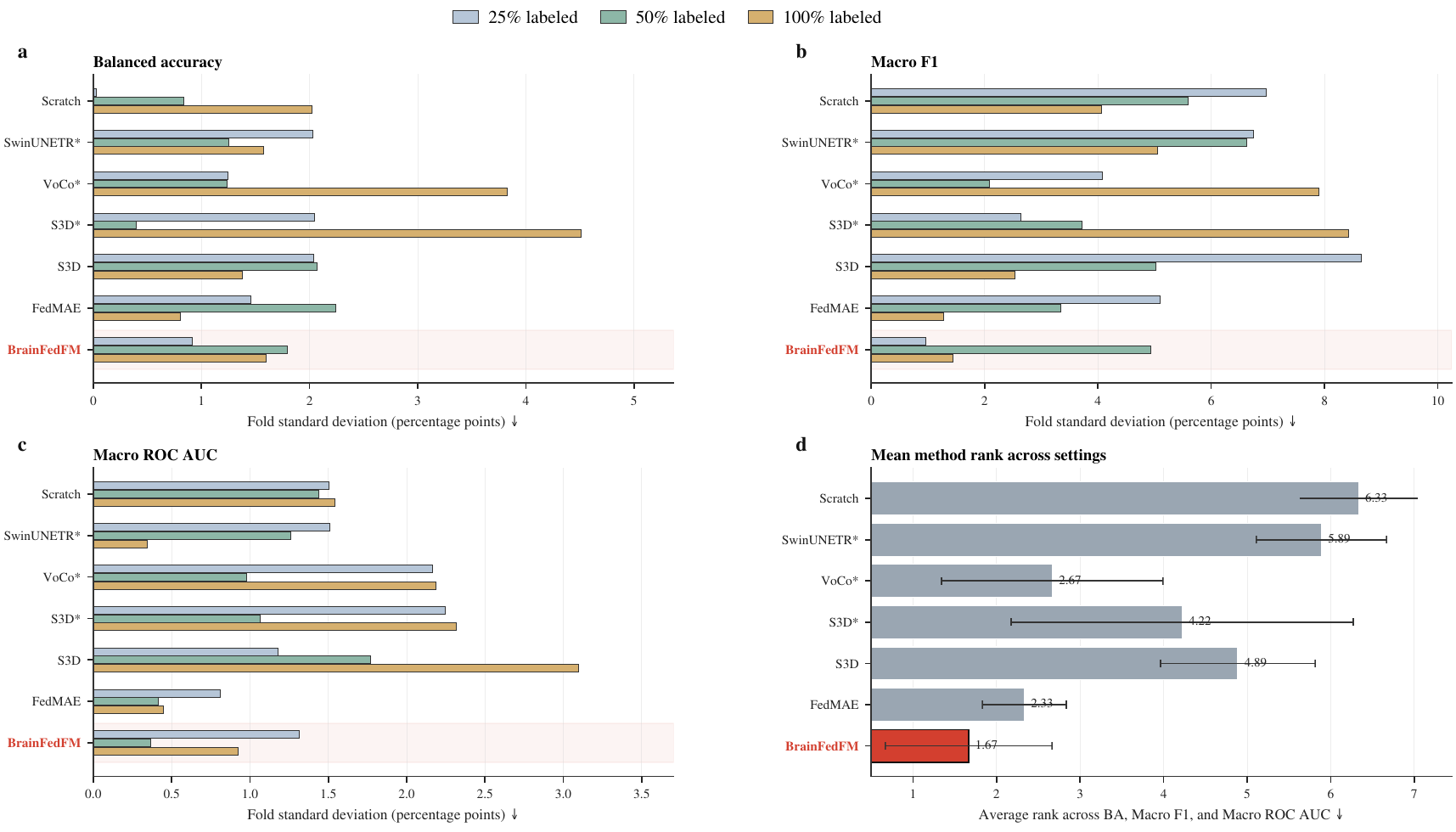}  \par\vspace{6pt}
\includegraphics[width=0.785\textwidth,keepaspectratio]{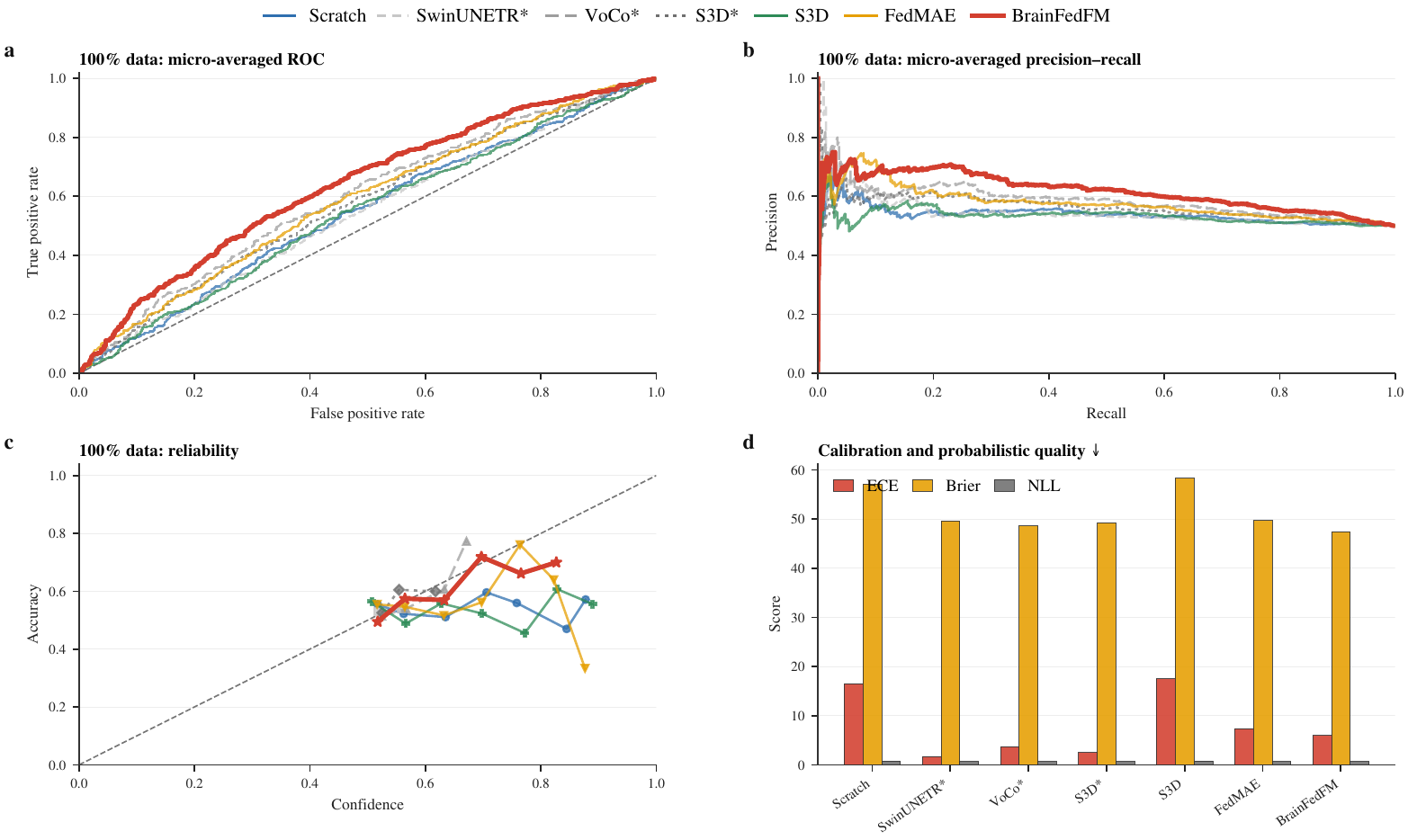}    \par\vspace{4pt}
\includegraphics[width=0.83\textwidth,keepaspectratio]{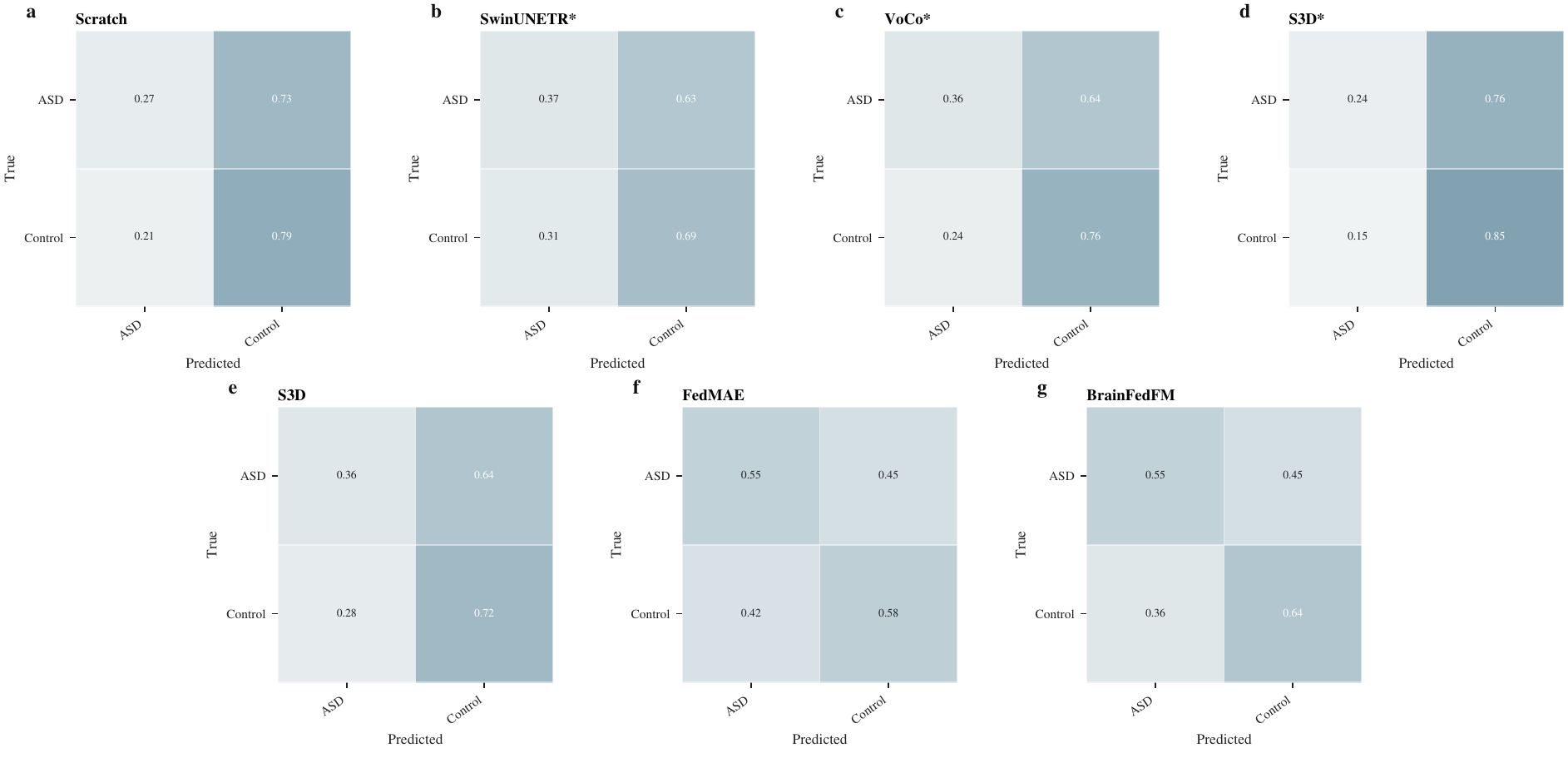}
\end{center}

\definecolor{pct100}{gray}{0.97}
\definecolor{pct75}{gray}{1.0}
\definecolor{pct50}{gray}{0.99}
\definecolor{pct25}{gray}{1.0}
\definecolor{overallgray}{gray}{1.0}
\definecolor{ourscol}{RGB}{244,248,252}

\begin{table*}[t]
\centering
\tiny
\caption[Post-stroke time estimation task]{\textbf{Post-stroke time estimation task.}
We evaluate foundation model performance on ATLAS-2 for estimating time since stroke onset from structural brain MRI.
This task models stroke-related neuroimaging changes over time and provides a representative regression setting in cerebrovascular disease. The table summarizes performance under 25\%, 50\%, and 100\% labelled-data settings, which indicate the proportion of the labelled training--validation pool used for downstream fine-tuning and model selection; the held-out test set remains unchanged across all settings.
We compare a randomly initialized scratch baseline, centralized pretrained foundation models, and federated pretrained foundation models.}
\label{tab:result_atlas}

\renewcommand{\arraystretch}{3.85}
{\scriptsize\bfseries Panel A. Summary under different labelled-data percentages.}
\vspace{5pt}

{\setlength{\tabcolsep}{3.38pt}
\resizebox{1.0\textwidth}{!}{
\begin{tabular}{
l c c c c c c
>{\columncolor{ourscol}\centering\arraybackslash}p{1.62cm}
}
\toprule
\multicolumn{2}{c}{\textbf{Brain sMRI FMs }}
& \multicolumn{4}{c}{Centralized Pretraining}
& \multicolumn{2}{c}{Federated Pretraining} \\
\cmidrule(lr){1-2}\cmidrule(lr){3-6}\cmidrule(lr){7-8}
\makecell[c]{Metric/Methods}
& Scratch
& \makecell[c]{SwinUNETR$^{\ast}$}  &\ VoCo$^{\ast}$ \ & \ S3D$^{\ast}$  & \ S3D
& \makecell[c]{FedMAE} & \makecell[c]{BrainFedFM} \\
\midrule

\rowcolor{pct100}
\textbf{100\%  RMSE $\downarrow$}
&  2.882($\pm$0.034) & 3.017($\pm$0.005) & 3.019($\pm$0.025) & 3.216($\pm$0.021) & 2.770($\pm$0.081) & 2.607($\pm$0.058)
& \cellcolor{ourscol}2.562($\pm$0.012) \\
\rowcolor{pct100}
\textbf{100\%   MAE $\downarrow$}
& 1.837($\pm$0.028) & 1.978($\pm$0.055) & 1.856($\pm$0.016) & 1.850($\pm$0.008) & 1.834($\pm$0.056) & 1.791($\pm$0.059)
& \cellcolor{ourscol}1.697($\pm$0.042) \\
\rowcolor{pct100}
\textbf{100\%   R$^2$ $\uparrow$}
& 0.076($\pm$0.022) & -0.013($\pm$0.003) & -0.014($\pm$0.017) & -0.150($\pm$0.015) & 0.146($\pm$0.050) & 0.244($\pm$0.034)
& \cellcolor{ourscol}0.270($\pm$0.007) \\
\midrule

\rowcolor{pct50}
\textbf{50\%  RMSE $\downarrow$}
& 2.983($\pm$0.012) & 3.065($\pm$0.023) & 3.094($\pm$0.007) & 3.273($\pm$0.017) & 2.904($\pm$0.022) & 2.710($\pm$0.022)
& \cellcolor{ourscol}2.721($\pm$0.028) \\
\rowcolor{pct50}
\textbf{50\%   MAE $\downarrow$}
& 1.915($\pm$0.016) & 1.932($\pm$0.032) & 1.890($\pm$0.008) & 1.876($\pm$0.006) & 2.052($\pm$0.088) & 1.946($\pm$0.119)
& \cellcolor{ourscol}2.015($\pm$0.068) \\
\rowcolor{pct50}
\textbf{50\%  R$^2$ $\uparrow$}
& 0.010($\pm$0.008) & -0.045($\pm$0.016) & -0.065($\pm$0.005) & -0.191($\pm$0.012) & 0.062($\pm$0.014) & 0.183($\pm$0.013)
& \cellcolor{ourscol}0.176($\pm$0.017) \\
\midrule

\rowcolor{pct25}
\textbf{25\%  RMSE $\downarrow$}
& 3.081($\pm$0.027) &  3.213($\pm$0.020) & 3.140($\pm$0.008) & 3.286($\pm$0.020) & 3.015($\pm$0.008) & 2.957($\pm$0.087)
& \cellcolor{ourscol}2.896($\pm$0.167) \\
\rowcolor{pct25}
\textbf{25\% MAE $\downarrow$}
& 1.903($\pm$0.037) & 1.852($\pm$0.008) & 1.840($\pm$0.009) & 1.881($\pm$0.007) & 2.362($\pm$0.017) & 2.356($\pm$0.108)
& \cellcolor{ourscol}2.272($\pm$0.225) \\
\rowcolor{pct25}
\textbf{25\% R$^2$ $\uparrow$}
& -0.056($\pm$0.018) & -0.148($\pm$0.014) & -0.097($\pm$0.006) & -0.201($\pm$0.015) & -0.011($\pm$0.006) & 0.027($\pm$0.058)
& \cellcolor{ourscol}0.064($\pm$0.109) \\
\bottomrule
\end{tabular}
}}

\vspace{5pt}

\begin{minipage}{1.0\textwidth}
\scriptsize
\textit{Notes. RMSE, root mean squared error; MAE, mean absolute error; R$^2$, coefficient of determination. RMSE and MAE are reported in years; lower values indicate better performance, higher R$^2$ indicates better performance. Scratch uses random initialization; SwinUNETR$^{\ast}$, VoCo$^{\ast}$, S3D$^{\ast}$ use publicly released centralized foundation models; S3D uses centralized pretraining on our pretraining corpus, FedMAE and BrainFedFM use federated pretraining. Fine-tuning uses 25 epochs of AdamW, a learning rate of $1\times10^{-6}$, weight decay of $1\times10^{-2}$, and cosine scheduling with a 3-epoch warm-up.}
\end{minipage}
\end{table*}

\clearpage

\begin{figure*}[p]
\noindent{\scriptsize\bfseries Panel B. Labelled-data scaling and full-data regression analyses.}
{\tiny\raggedright
Characterizes regression performance from complementary perspectives. \textbf{B1}, fold-averaged RMSE, MAE, R$^2$, Pearson correlation, Spearman correlation, and concordance correlation coefficient across the 25\%, 50\%, and 100\% labelled-data settings. \textbf{B2}, cross-method analyses including error magnitude, association and agreement, signed-error distributions, and MAE across the observed target range. \textbf{B3}, the joint RMSE--CCC performance landscape across labelled-data settings, where lower RMSE and higher CCC indicate a more favorable error--agreement trade-off.\par}

\par\vspace{8pt}

\includegraphics[width=\textwidth,
    keepaspectratio
]{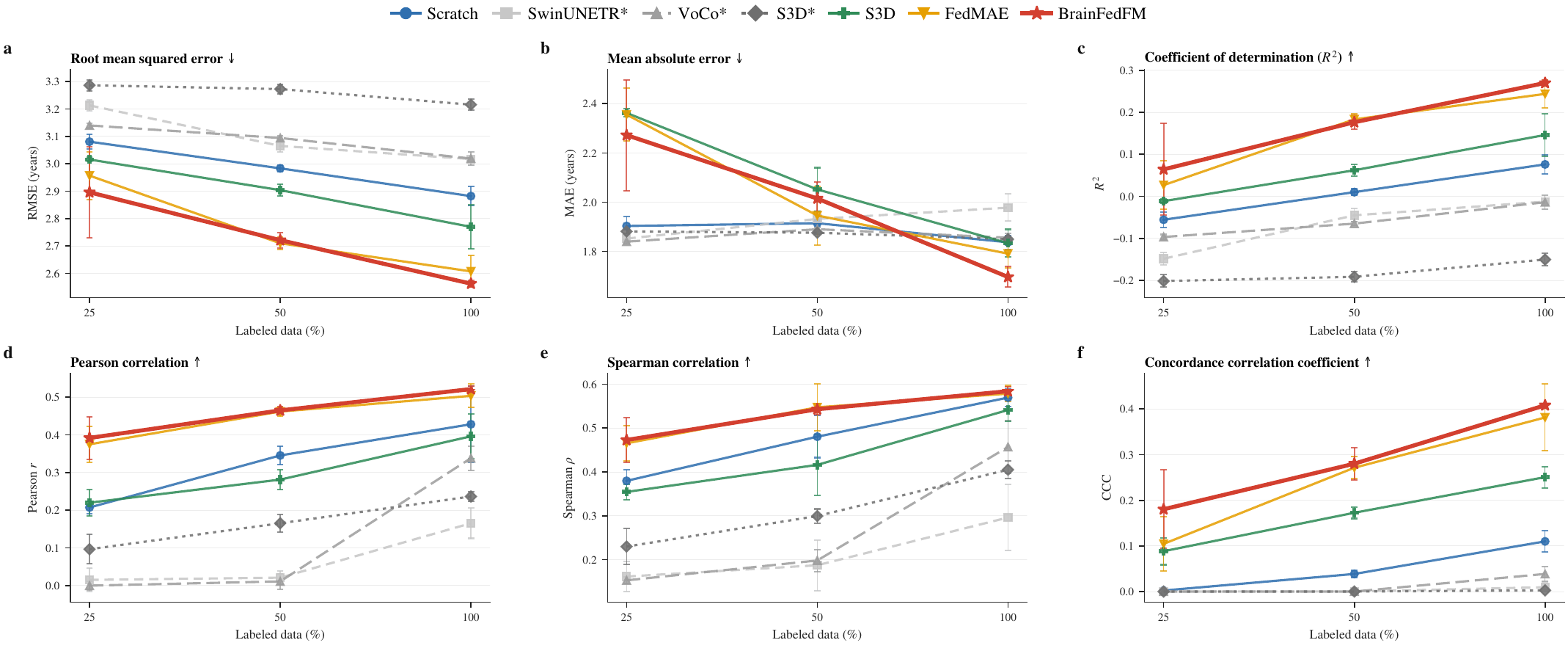}

\par\vspace{3pt}

\includegraphics[
    width=0.86\textwidth,
    keepaspectratio
]{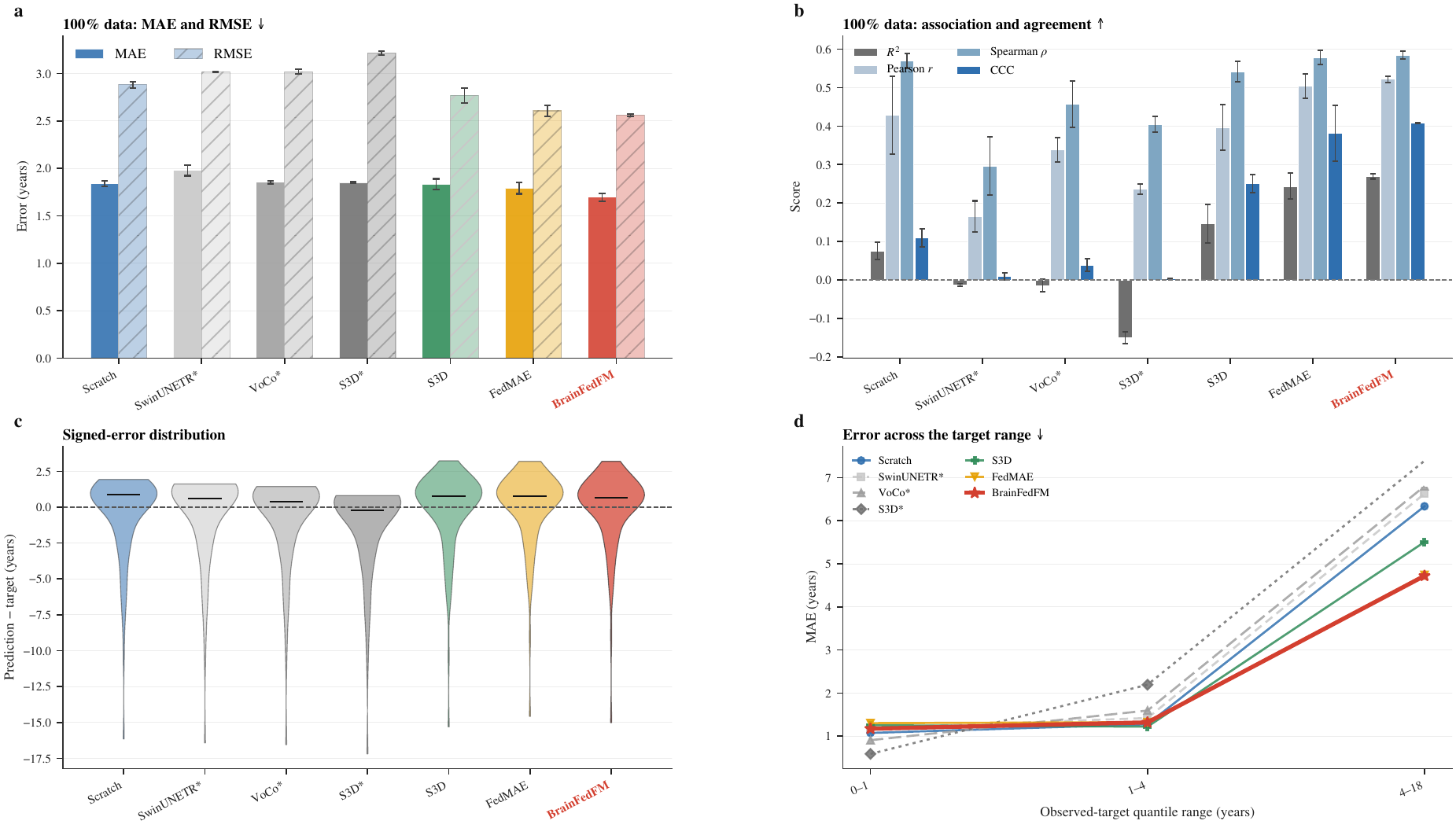}

\par\vspace{3pt}

\includegraphics[
    width=0.96\textwidth,
    keepaspectratio
]{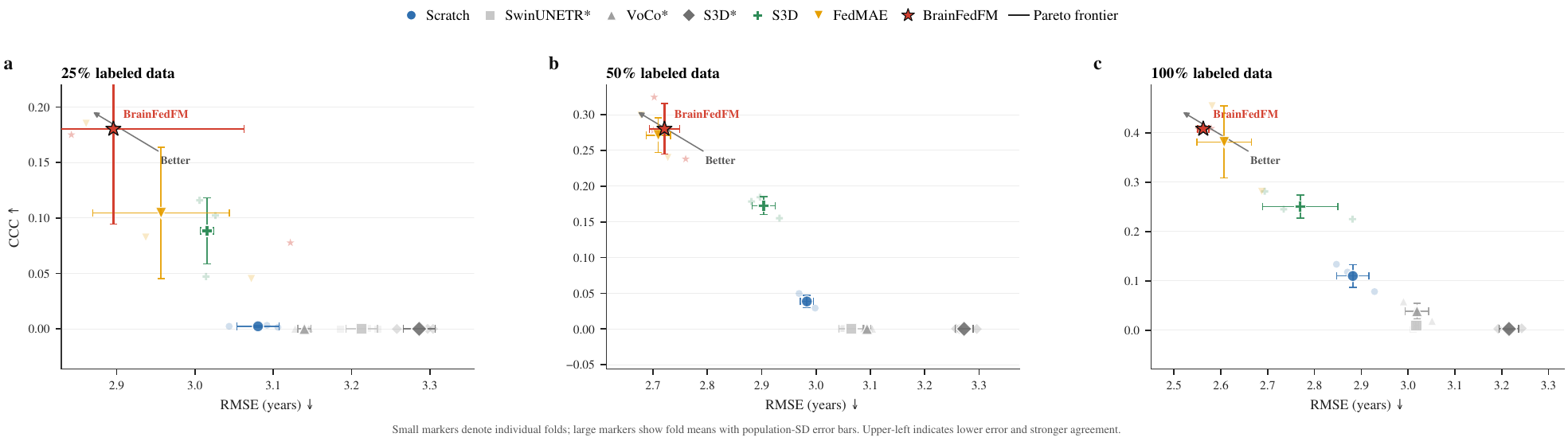}

\end{figure*}

\clearpage

\begin{figure*}[p]
\noindent{\scriptsize\bfseries
Panel C. Relative ranking, fold stability, and transfer-benefit analyses.}
{\tiny\raggedright
Summarizes comparative performance, robustness, and transfer benefit beyond the aggregate regression scores.
\textbf{C1}, relative method ranks across all labelled-data settings and six regression metrics, with rank 1 denoting the best result within each setting--metric combination.
\textbf{C2}, fold-to-fold variability in RMSE, MAE, and R$^2$, together with
the average method rank across the core regression metrics.
\textbf{C3}, performance changes relative to the matched Scratch baseline across labelled-data settings, where positive values denote RMSE or MAE reduction and improvement in R$^2$.\par}

\vspace{5.5pt}

\includegraphics[
    width=\textwidth,
    keepaspectratio
]{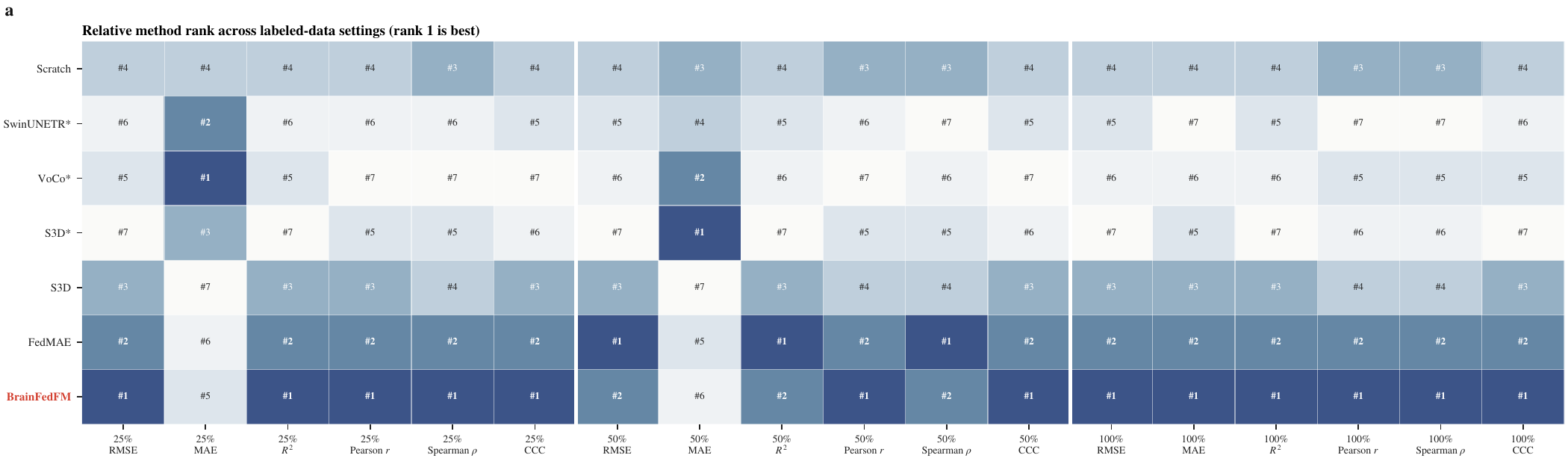}

\vspace{6pt}

\includegraphics[
    width=\textwidth,
    keepaspectratio
]{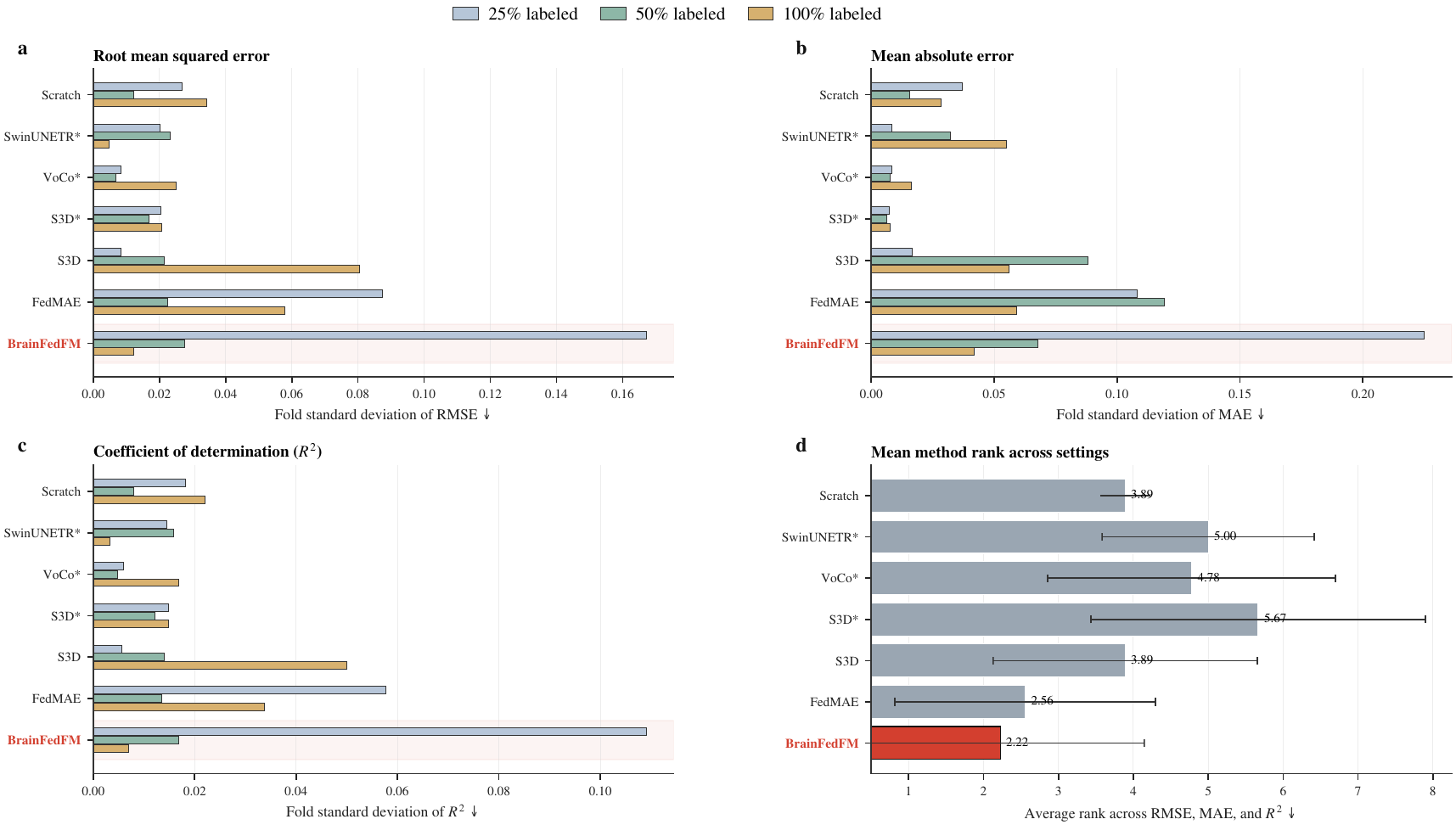}

\vspace{6pt}

\includegraphics[
    width=\textwidth,
    keepaspectratio
]{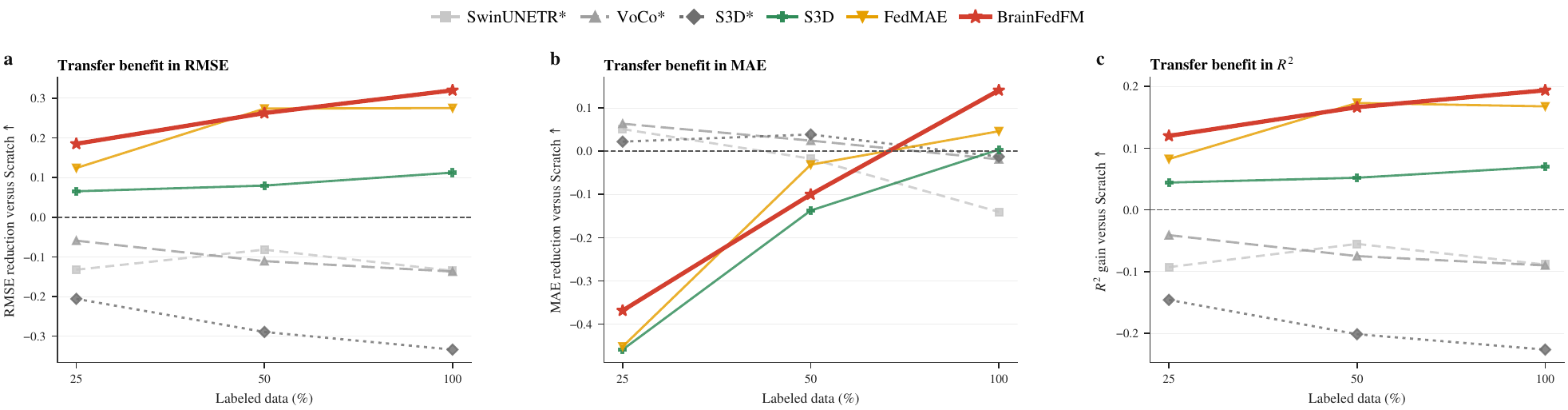}

\end{figure*}

\definecolor{pct100}{gray}{0.97}
\definecolor{pct75}{gray}{1.0}
\definecolor{pct50}{gray}{0.99}
\definecolor{pct25}{gray}{1.0}
\definecolor{overallgray}{gray}{1.0}
\definecolor{ourscol}{RGB}{244,248,252}

\begin{table*}[t]
\centering
\tiny
\caption[Dementia diagnosis task]{\textbf{Dementia diagnosis task.}
We evaluate foundation model performance on the OASIS-2 benchmark for dementia classification. This task aims to classify structural MRI scans into demented and nondemented categories.
Accurate dementia prediction is important for neurodegenerative disease analysis, supporting subject stratification, cohort curation, and downstream evaluation in dementia-related studies.
The table summarizes performance under 25\%, 50\%, and 100\% labelled-data settings, which indicate the proportion of the labelled training--validation pool used for downstream fine-tuning and model selection; the held-out test set remains unchanged across all settings.
We compare a randomly initialized scratch baseline, centralized pretrained foundation models, and federated pretrained foundation models.}
\label{tab:result_oasis}

\renewcommand{\arraystretch}{3.85}
{\scriptsize\bfseries Panel A. Summary under different labelled-data percentages.}
\vspace{5pt}

{\setlength{\tabcolsep}{3.38pt}

\resizebox{1.0\textwidth}{!}{
\begin{tabular}{
l c c c c c c
>{\columncolor{ourscol}\centering\arraybackslash}p{1.62cm}
}
\toprule
\multicolumn{2}{c}{\textbf{Brain sMRI FMs }}
& \multicolumn{4}{c}{Centralized Pretraining}
& \multicolumn{2}{c}{Federated Pretraining} \\
\cmidrule(lr){1-2}\cmidrule(lr){3-6}\cmidrule(lr){7-8}
\makecell[c]{Metric/Methods}
& Scratch
& \makecell[c]{SwinUNETR$^{\ast}$}  &\ VoCo$^{\ast}$ \ & \ S3D$^{\ast}$  & \ S3D 
& \makecell[c]{FedMAE} & \makecell[c]{BrainFedFM} \\
\midrule

\rowcolor{pct100}
\textbf{100\%  BA $\uparrow$}
& 59.38($\pm$13.26) & 50.00($\pm$0.000) & 55.69($\pm$8.048) & 50.00($\pm$0.000) & 51.16($\pm$1.345) & 64.13($\pm$12.12)
& \cellcolor{ourscol}66.93($\pm$11.41) \\
\rowcolor{pct100}
\textbf{100\%   mF1 $\uparrow$}
& 46.65($\pm$22.09) & 33.88($\pm$0.307) & 42.11($\pm$11.50) & 31.98($\pm$2.856) & 45.16($\pm$7.859) & 56.50($\pm$21.27)
& \cellcolor{ourscol}61.60($\pm$17.18) \\
\rowcolor{pct100}
\textbf{100\%   AUC $\uparrow$}
& 73.46($\pm$15.69) & 70.93($\pm$3.569) & 70.56($\pm$9.494) & 73.70($\pm$12.879) & 53.47($\pm$2.458) & 81.71($\pm$5.860)
& \cellcolor{ourscol}80.80($\pm$6.639) \\
\midrule

\rowcolor{pct50}
\textbf{50\%  BA $\uparrow$}
& 50.00($\pm$0.000) & 50.00($\pm$0.000) & 54.41($\pm$6.235) & 55.63($\pm$7.960) & 50.00($\pm$0.000) & 65.27($\pm$5.926)
& \cellcolor{ourscol}61.44($\pm$6.032) \\
\rowcolor{pct50}
\textbf{50\%  mF1 $\uparrow$}
& 31.98($\pm$2.856) & 31.98($\pm$2.856) & 41.24($\pm$10.25) & 42.91($\pm$12.63) & 31.98($\pm$2.856) & 62.68($\pm$9.788)
& \cellcolor{ourscol}59.81($\pm$7.372) \\
\rowcolor{pct50}
\textbf{50\% AUC $\uparrow$}
& 51.16($\pm$8.077) & 57.58($\pm$6.060) & 70.19($\pm$6.590) & 71.99($\pm$15.50) & 68.85($\pm$8.983) & 70.01($\pm$6.617)
& \cellcolor{ourscol}71.48($\pm$5.900) \\
\midrule

\rowcolor{pct25}
\textbf{25\%  BA $\uparrow$}
& 50.00($\pm$0.000) & 50.00($\pm$0.000) & 50.00($\pm$0.000) & 50.00($\pm$0.000) & 50.00($\pm$0.000) & 52.85($\pm$4.024)
& \cellcolor{ourscol}65.75($\pm$10.46) \\
\rowcolor{pct25}
\textbf{25\%  mF1 $\uparrow$}
& 31.98($\pm$2.856) & 31.31($\pm$2.436) & 31.98($\pm$2.856) & 32.19($\pm$2.997) & 31.98($\pm$2.856) & 37.60($\pm$5.111)
& \cellcolor{ourscol}61.47($\pm$16.06) \\
\rowcolor{pct25}
\textbf{25\% AUC $\uparrow$}
& 47.58($\pm$5.254) & 55.02($\pm$7.271) & 61.76($\pm$12.75) & 74.03($\pm$5.564) & 54.65($\pm$13.82) & 69.76($\pm$4.560)
& \cellcolor{ourscol}75.41($\pm$8.094) \\
\bottomrule
\end{tabular}
}}

\vspace{5pt}

\begin{minipage}{1.0\textwidth}
\scriptsize
\textit{Notes. Values are mean($\pm$std) over three folds. BA, balanced accuracy; mF1, macro F1-score; AUC, macro one-vs-rest ROC AUC. Scratch uses random initialization; SwinUNETR$^{\ast}$, VoCo$^{\ast}$, and S3D$^{\ast}$ use publicly released centralized foundation model weights; S3D uses centralized pretraining on our pretraining corpus, FedMAE and BrainFedFM use federated pretraining.
Fine-tuning uses 10 epochs of AdamW, learning rate $1\times10^{-6}$, weight decay $1\times10^{-2}$, cosine scheduling with a 1-epoch warm-up, and label smoothing of 0.2. The best checkpoint is selected according to validation AUC.}
\end{minipage}
\end{table*}

\clearpage

\noindent{\scriptsize\bfseries Panel B. Labelled-data scaling and aggregate metric summaries.}
\vspace{2pt}
{\tiny\raggedright
Summarizes method behavior across labelled-data settings.
\textbf{B1}, fold-averaged trends for classification and calibration metrics; \textbf{B2}, extended metric heatmaps separating higher-is-better classification metrics from lower-is-better calibration/uncertainty metrics; \textbf{B3}, the accuracy--calibration trade-off, where points closer to the upper-left indicate stronger discrimination with better calibration.\par}
\vspace{-3pt}

\begin{center}
\includegraphics[width=1.0\textwidth,keepaspectratio]{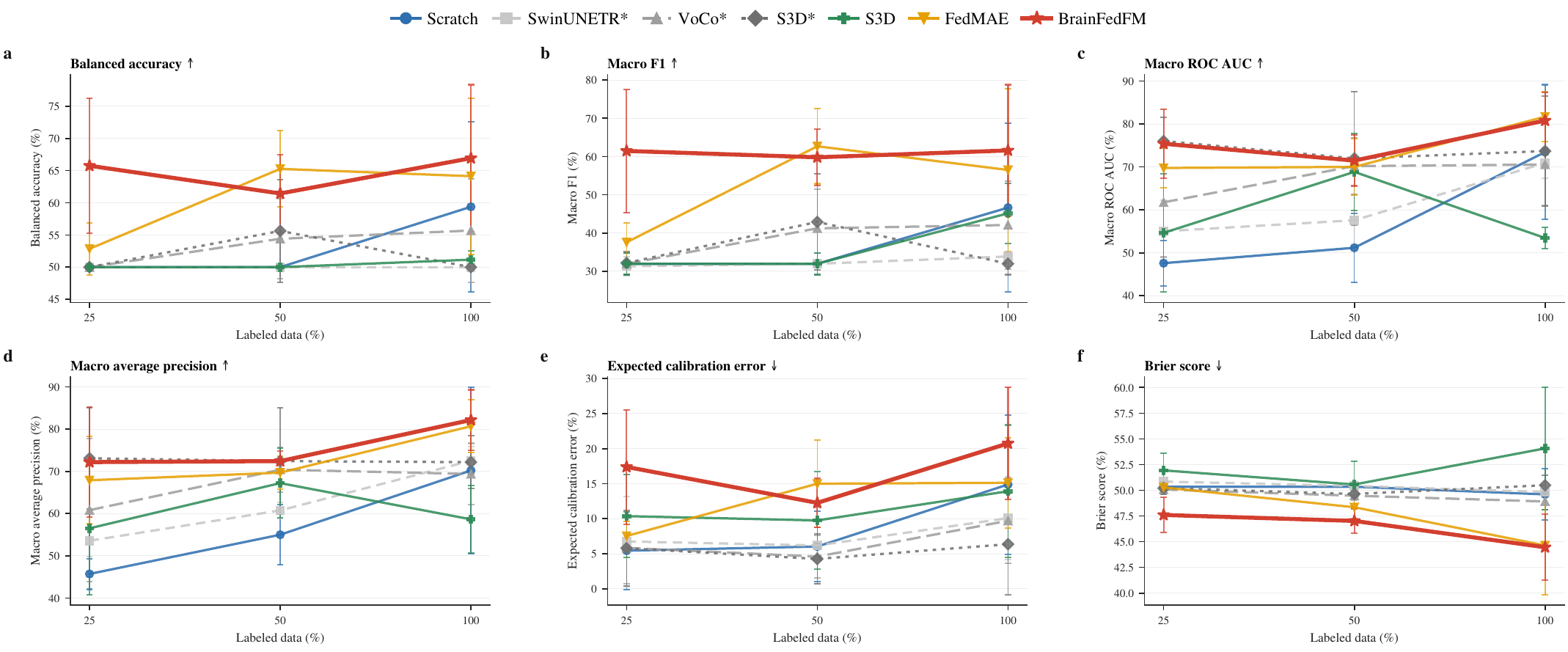} \par\vspace{3pt}
\includegraphics[width=1.0\textwidth,keepaspectratio]{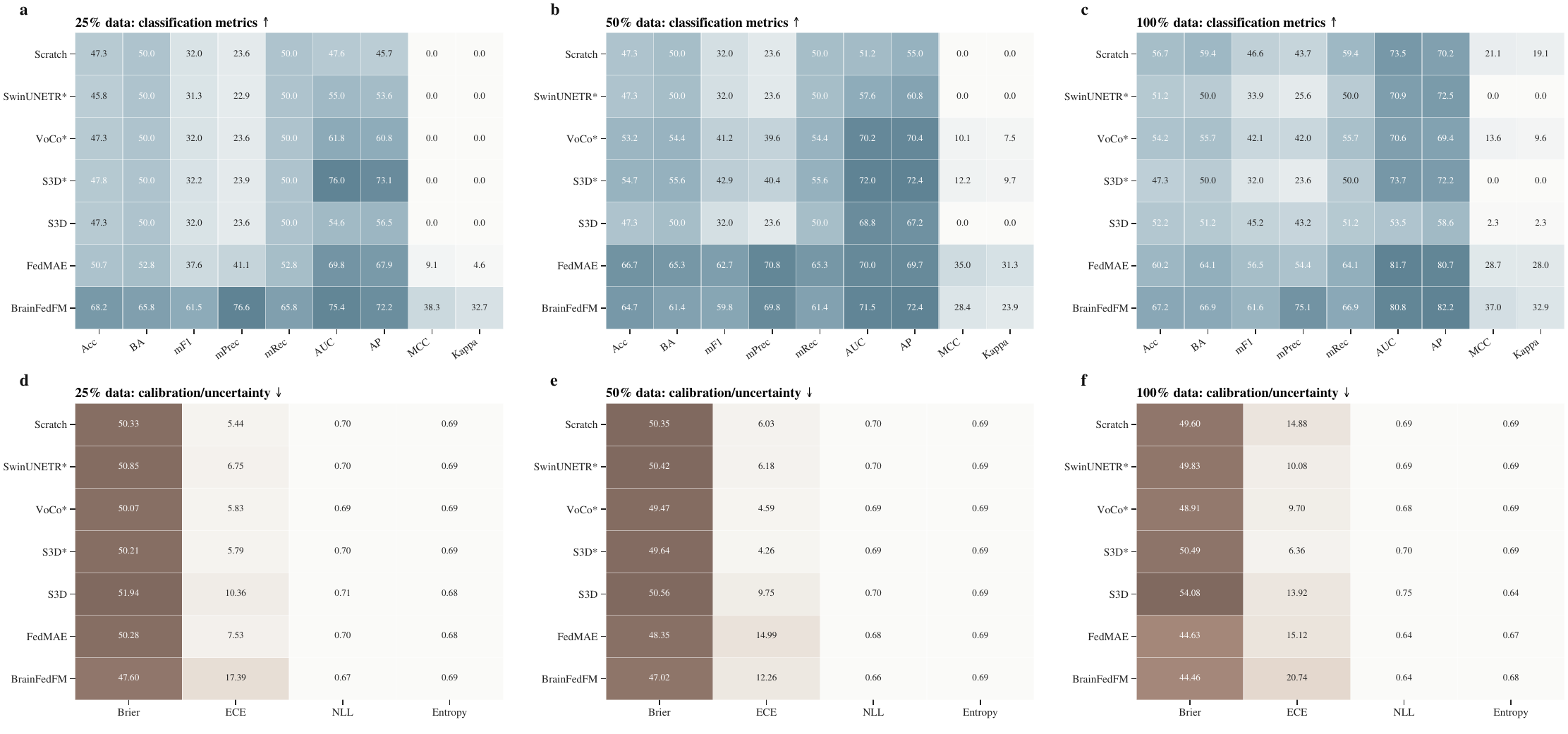} \par\vspace{5pt}
\includegraphics[width=0.935\textwidth,keepaspectratio]{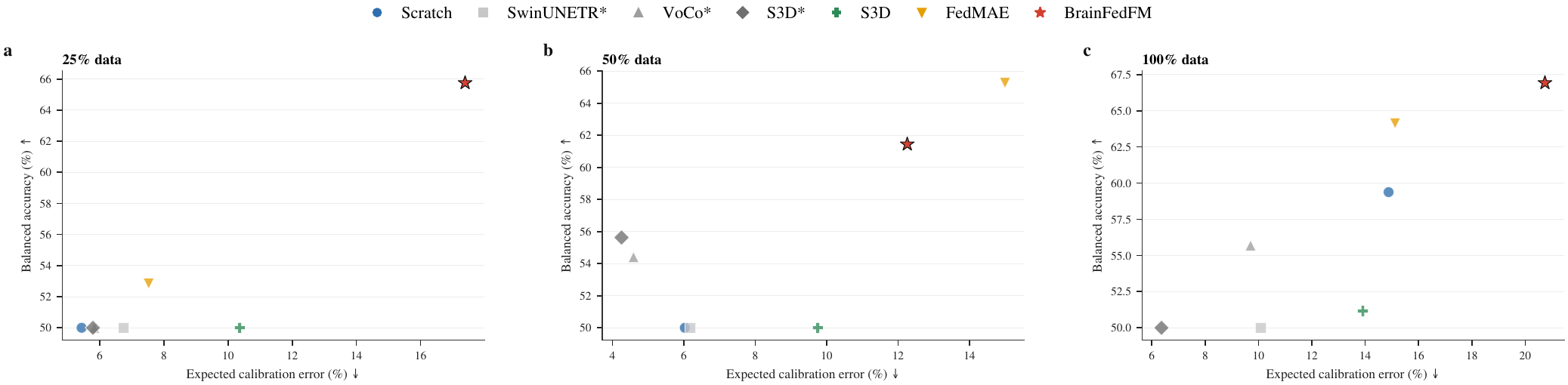}
\end{center}

\vspace{-2pt}

\noindent{\scriptsize\bfseries Panel C. Fold stability and full-data performance analyses.}
\vspace{2pt}
{\tiny\raggedright
Examines robustness and error modes beyond the aggregate scores. \textbf{C1}, fold-to-fold variability and average method ranking across balanced accuracy, macro F1, and macro ROC AUC; \textbf{C2}, probability-level analysis at 100\% labelled data, including micro-averaged ROC/precision--recall curves, reliability curves, and calibration scores; \textbf{C3}, row-normalized confusion matrices at 100\% labelled data, with rows denoting true dementia classes and columns denoting predicted classes.\par}

\clearpage

\begin{center}
\includegraphics[width=0.84\textwidth,keepaspectratio]{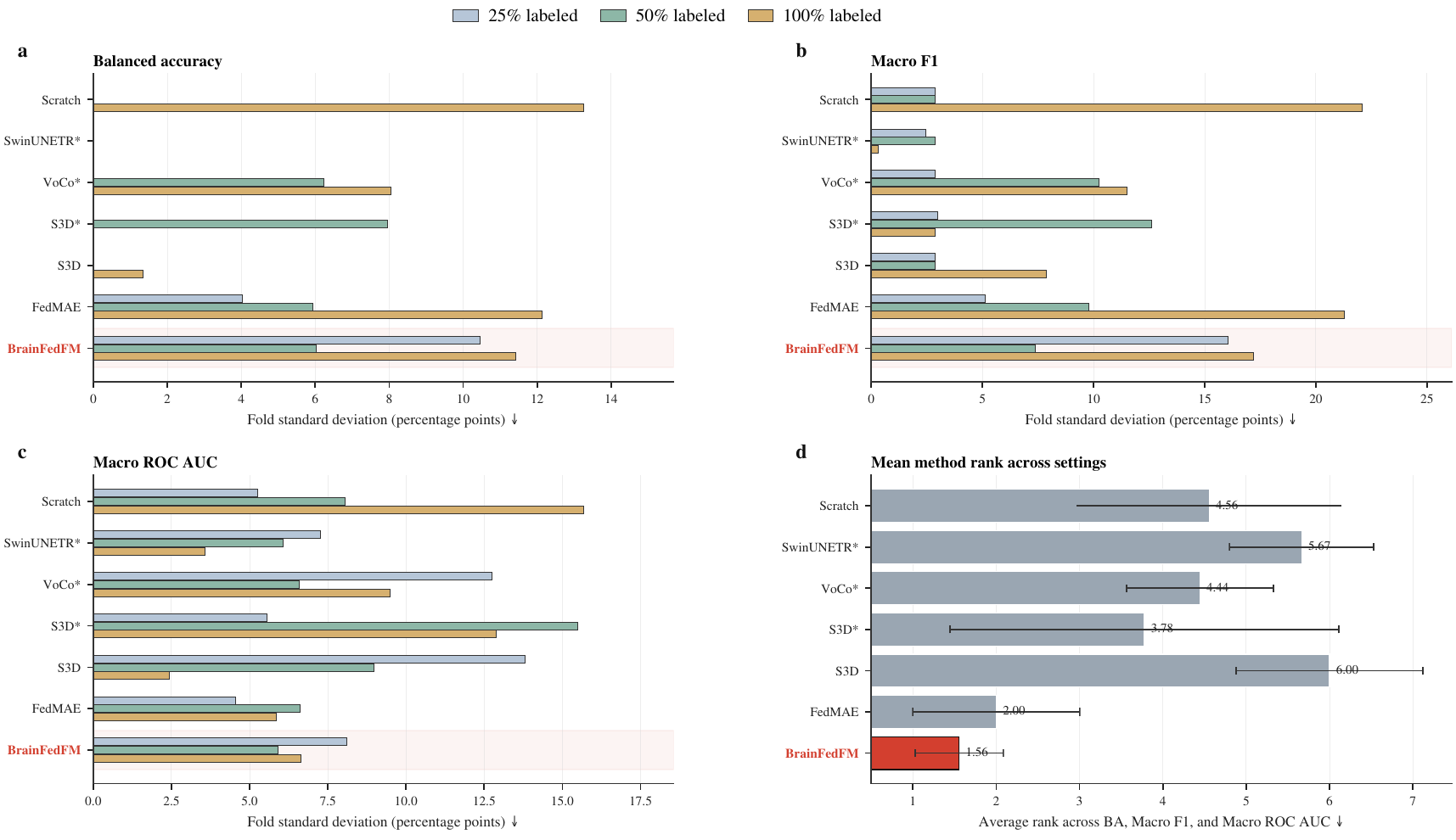}   \par\vspace{6pt}
\includegraphics[width=0.785\textwidth,keepaspectratio]{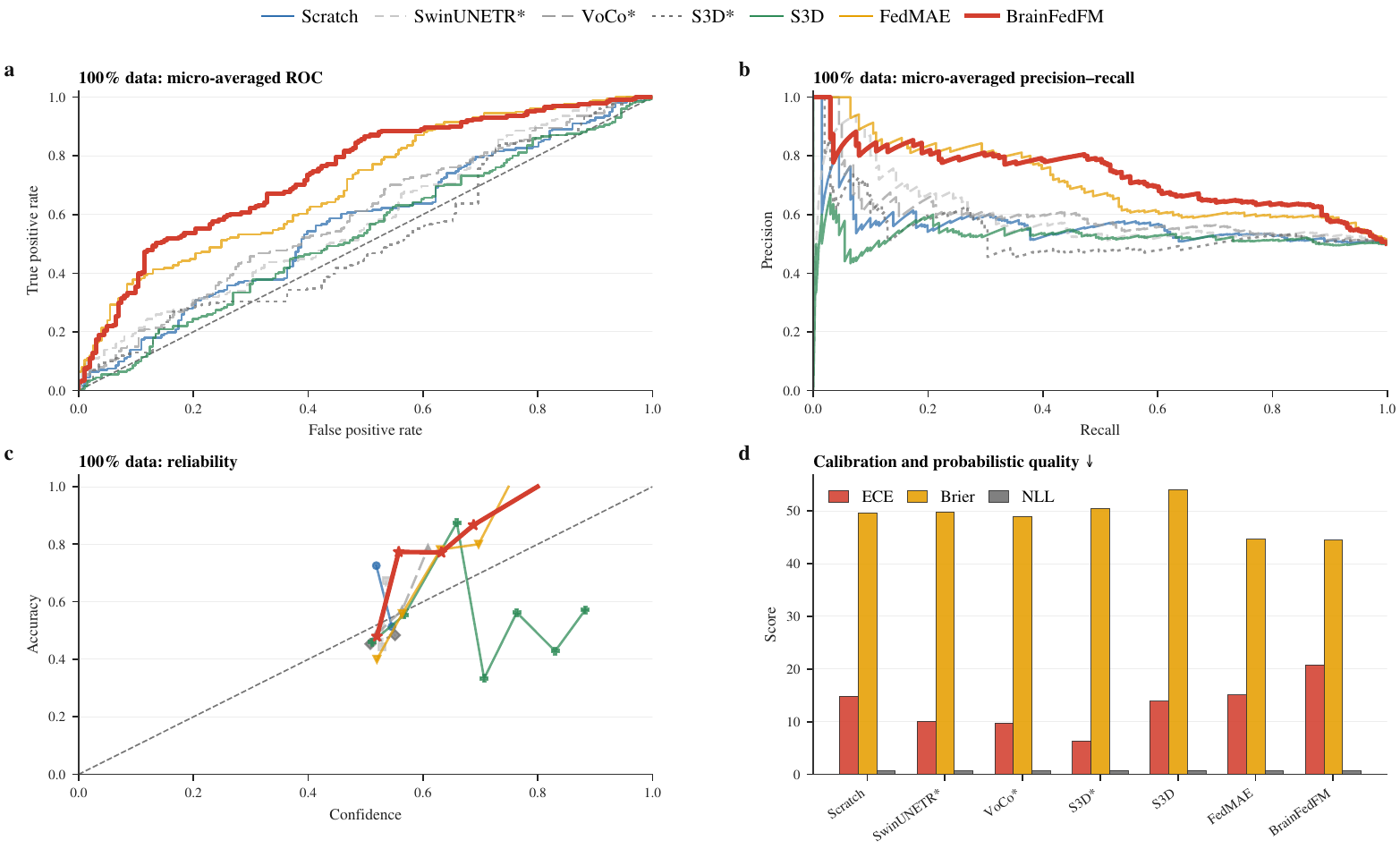}   \par\vspace{4pt}
\includegraphics[width=0.838\textwidth,keepaspectratio]{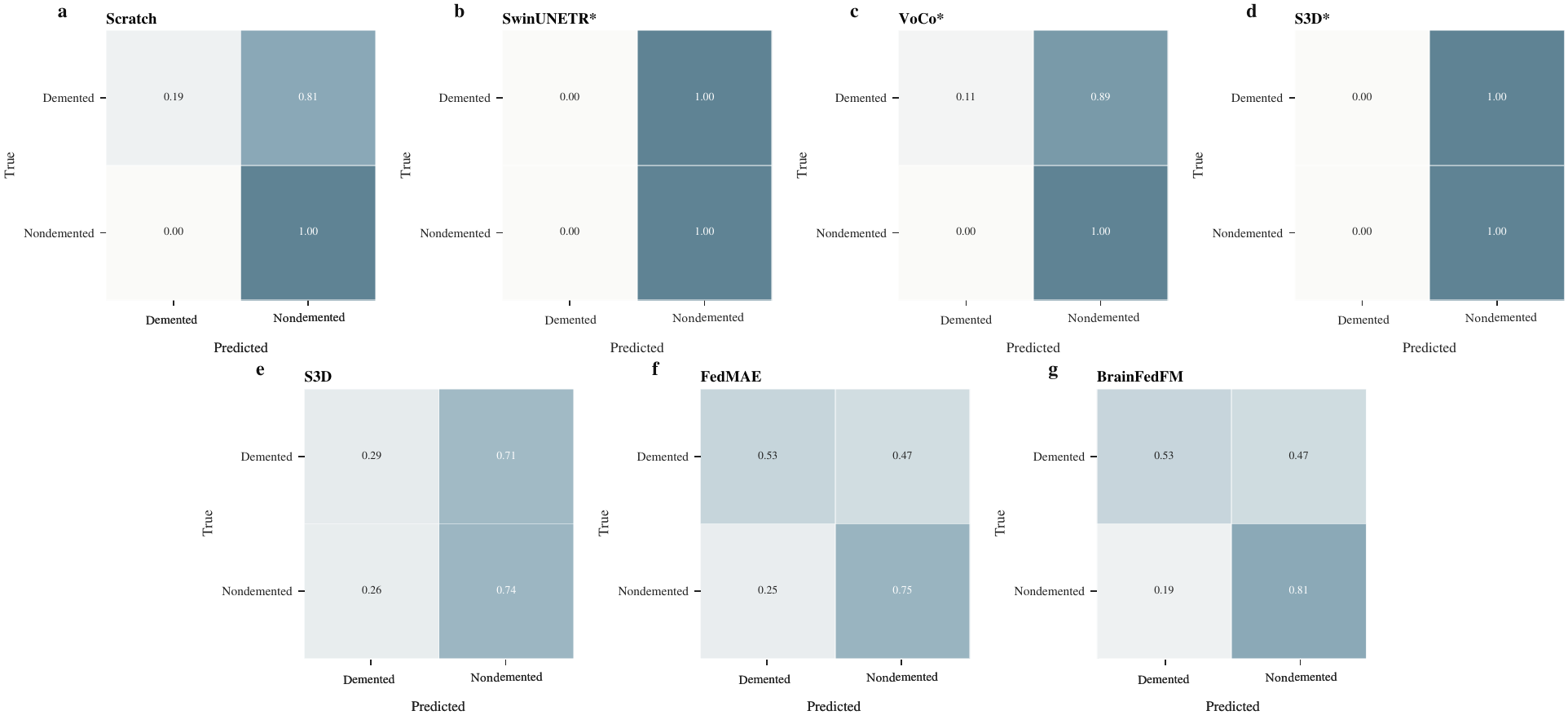}
\end{center}
\definecolor{pct100}{gray}{0.97}
\definecolor{pct75}{gray}{1.0}
\definecolor{pct50}{gray}{0.99}
\definecolor{pct25}{gray}{1.0}
\definecolor{overallgray}{gray}{1.0}
\definecolor{ourscol}{RGB}{244,248,252}

\begin{table*}[t]
\centering
\tiny
\caption[Cannabis dependence classification task]{\textbf{Cannabis dependence classification task.}
We evaluate foundation model performance on the TMCU benchmark for cannabis dependence classification. This task aims to classify structural MRI scans into cannabis-dependent and healthy-control categories. Accurate cannabis dependence classification is useful for identifying structural brain patterns associated with cannabis use and for evaluating MRI-based biomarkers in substance-use-related neuroimaging studies.
The table summarizes performance under 25\%, 50\%, and 100\% labelled-data settings, which indicate the proportion of the labelled training--validation pool used for downstream fine-tuning and model selection; the held-out test set remains unchanged across all settings. We compare a randomly initialized scratch baseline, centralized pretrained foundation models, and federated pretrained foundation models.}
\label{tab:result_tmcu}

\renewcommand{\arraystretch}{3.85}
{\scriptsize\bfseries Panel A. Summary under different labelled-data percentages.}
\vspace{5pt}

{\setlength{\tabcolsep}{3.38pt}

\resizebox{1.0\textwidth}{!}{
\begin{tabular}{
l c c c c c c
>{\columncolor{ourscol}\centering\arraybackslash}p{1.62cm}
}
\toprule
\multicolumn{2}{c}{\textbf{Brain sMRI FMs }}
& \multicolumn{4}{c}{Centralized Pretraining}
& \multicolumn{2}{c}{Federated Pretraining} \\
\cmidrule(lr){1-2}\cmidrule(lr){3-6}\cmidrule(lr){7-8}
\makecell[c]{Metric/Methods}
& Scratch
& \makecell[c]{SwinUNETR$^{\ast}$}  &\ VoCo$^{\ast}$ \ & \ S3D$^{\ast}$  & \ S3D 
& \makecell[c]{FedMAE} & \makecell[c]{BrainFedFM} \\
\midrule

\rowcolor{pct100}
\textbf{100\%  BA $\uparrow$}
& 52.08($\pm$2.946) & 50.00($\pm$0.000) & 52.08($\pm$2.946) & 58.33($\pm$11.79) & 50.00($\pm$0.000) & 64.58($\pm$25.17)
& \cellcolor{ourscol}70.83($\pm$12.84) \\
\rowcolor{pct100}
\textbf{100\%   mF1 $\uparrow$}
& 37.52($\pm$5.921) & 33.33($\pm$0.000) & 37.52($\pm$5.921) & 47.22($\pm$19.64) & 33.33($\pm$0.000) & 54.59($\pm$32.13)
& \cellcolor{ourscol}70.83($\pm$12.84) \\
\rowcolor{pct100}
\textbf{100\%   AUC $\uparrow$}
& 59.90($\pm$7.026) & 66.15($\pm$15.85) & 62.50($\pm$4.600) & 57.29($\pm$18.10) & 59.90($\pm$13.40) & 72.40($\pm$19.78)
& \cellcolor{ourscol}72.40($\pm$19.53) \\
\midrule

\rowcolor{pct50}
\textbf{50\%  BA $\uparrow$}
& 50.00($\pm$0.000) & 50.00($\pm$0.000) & 50.00($\pm$0.000) & 50.00($\pm$0.000) & 50.00($\pm$0.000) & 50.00($\pm$0.000)
& \cellcolor{ourscol}52.08($\pm$2.946) \\
\rowcolor{pct50}
\textbf{50\%  mF1 $\uparrow$}
& 33.33($\pm$0.000) & 33.33($\pm$0.000) & 33.33($\pm$0.000) & 33.33($\pm$0.000) & 33.33($\pm$0.000) & 36.16($\pm$4.000)
& \cellcolor{ourscol}40.44($\pm$10.05) \\
\rowcolor{pct50}
\textbf{50\% AUC $\uparrow$}
& 63.96($\pm$7.783) & 45.31($\pm$7.967) & 48.44($\pm$13.44) & 45.83($\pm$6.421) & 43.23($\pm$13.64) & 64.06($\pm$10.13)
& \cellcolor{ourscol}60.42($\pm$14.00) \\
\midrule

\rowcolor{pct25}
\textbf{25\%  BA $\uparrow$}
& 50.00($\pm$0.000) & 50.00($\pm$0.000) & 52.08($\pm$2.946) & 50.00($\pm$0.000) & 50.00($\pm$0.000) & 47.92($\pm$2.946)
& \cellcolor{ourscol}50.00($\pm$0.000) \\
\rowcolor{pct25}
\textbf{25\%  mF1 $\uparrow$}
& 33.33($\pm$0.000) & 33.33($\pm$0.000) & 39.39($\pm$8.571) & 33.33($\pm$0.000) & 33.33($\pm$0.000) & 32.37($\pm$1.366)
& \cellcolor{ourscol}33.33($\pm$0.000) \\
\rowcolor{pct25}
\textbf{25\% AUC $\uparrow$}
& 52.60($\pm$3.211) & 50.00($\pm$9.632) & 48.96($\pm$7.026) & 48.44($\pm$8.366) & 49.48($\pm$3.683) & 44.27($\pm$5.156)
& \cellcolor{ourscol}65.10($\pm$17.23) \\
\bottomrule
\end{tabular}
}}

\vspace{5pt}

\begin{minipage}{1.0\textwidth}
\scriptsize
\textit{Notes. Values are mean($\pm$std) over three folds. BA, balanced accuracy; mF1, macro F1-score; AUC, macro one-vs-rest ROC AUC. Scratch uses random initialization; SwinUNETR$^{\ast}$, VoCo$^{\ast}$, and S3D$^{\ast}$ use publicly released centralized foundation model weights; S3D uses centralized pretraining on our pretraining corpus, FedMAE and BrainFedFM use federated pretraining.
Fine-tuning uses 10 epochs of AdamW, learning rate $1\times10^{-6}$, weight decay $1\times10^{-2}$, cosine scheduling with a 1-epoch warm-up, and label smoothing of 0.2. The best checkpoint is selected according to validation AUC.}
\end{minipage}
\end{table*}

\clearpage

\noindent{\scriptsize\bfseries Panel B. Labelled-data scaling and aggregate metric summaries.}
\vspace{2pt}
{\tiny\raggedright
Summarizes method behavior across labelled-data settings.
\textbf{B1}, fold-averaged trends for classification and calibration metrics; \textbf{B2}, extended metric heatmaps separating higher-is-better classification metrics from lower-is-better calibration/uncertainty metrics; \textbf{B3}, the accuracy--calibration trade-off, where points closer to the upper-left indicate stronger discrimination with better calibration.\par}
\vspace{-3pt}

\begin{center}
\includegraphics[width=1.0\textwidth,keepaspectratio]{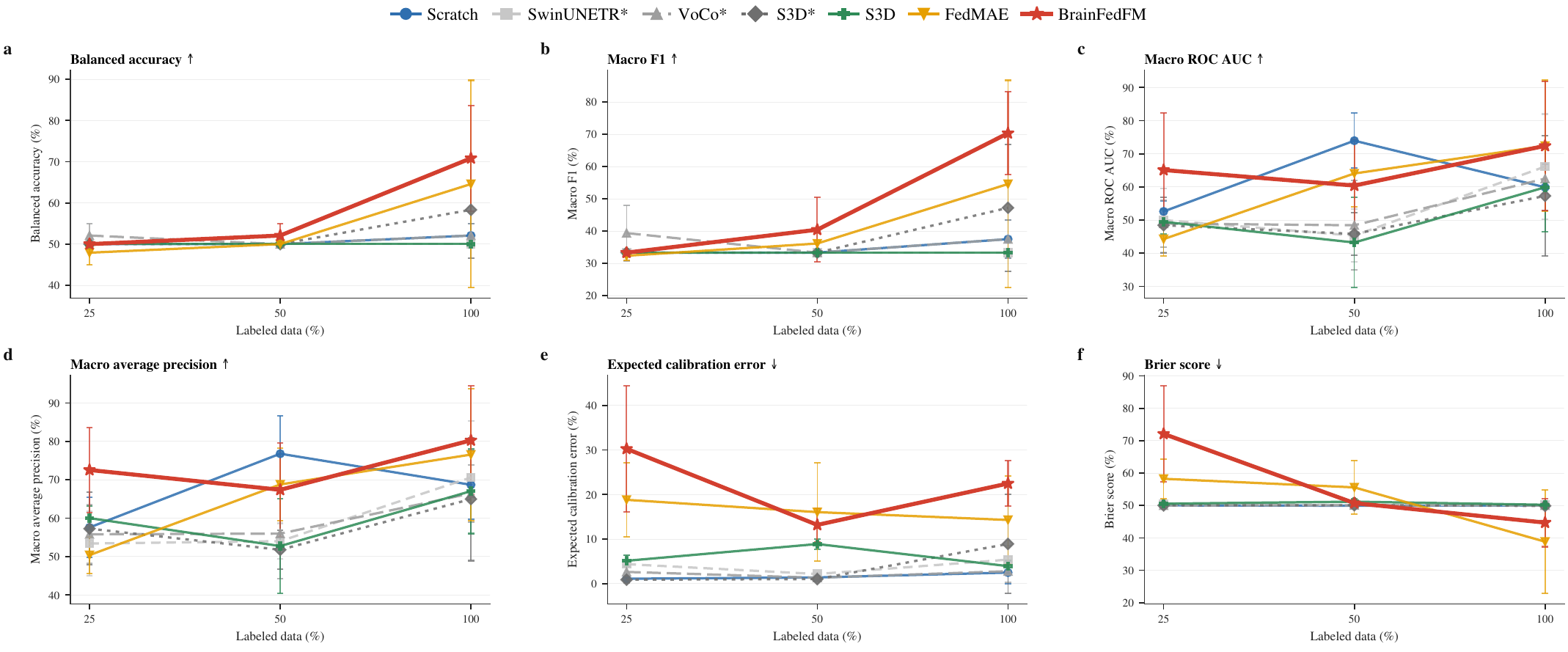} \par\vspace{3pt}
\includegraphics[width=1.0\textwidth,keepaspectratio]{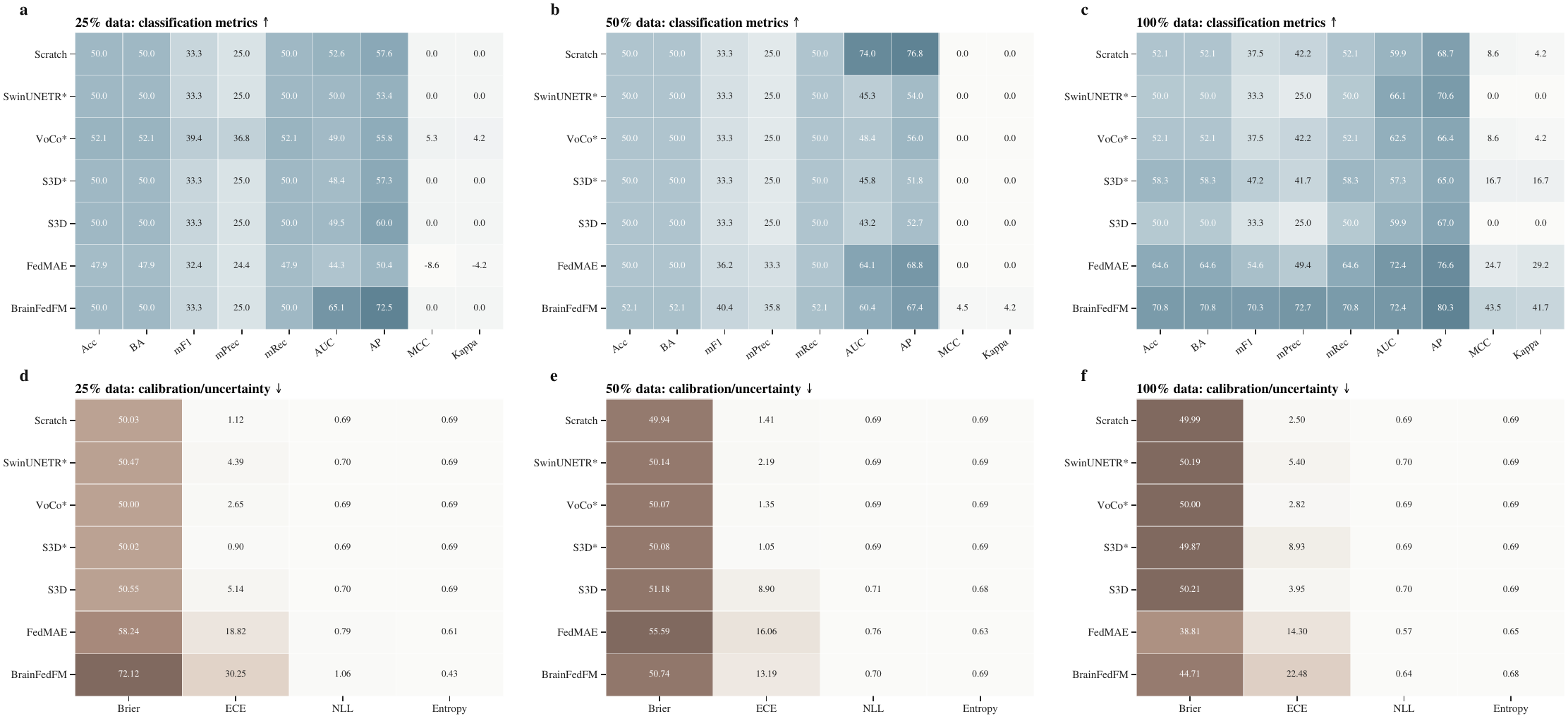} \par\vspace{5pt}
\includegraphics[width=0.935\textwidth,keepaspectratio]{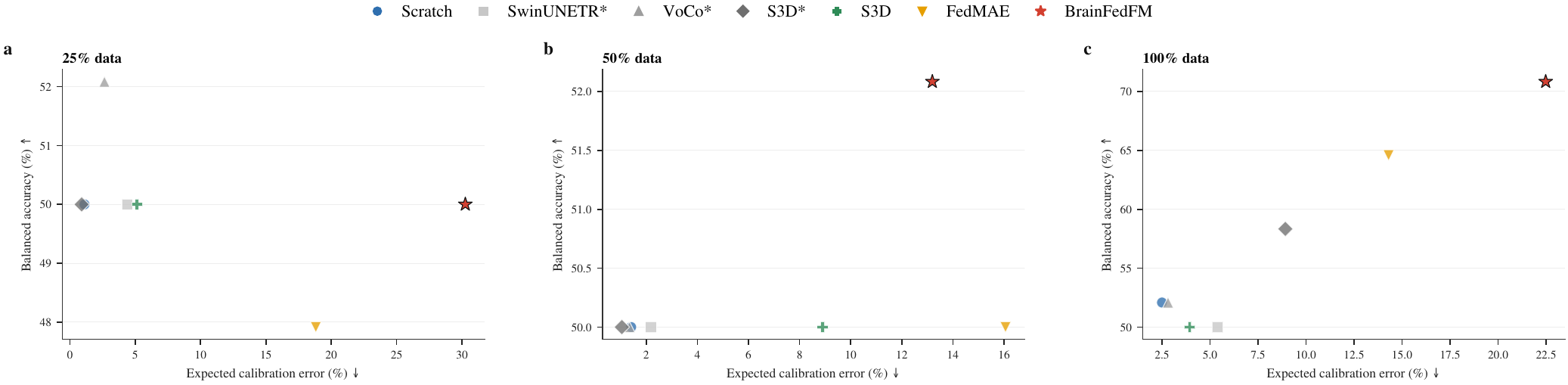}
\end{center}

\vspace{-2pt}

\noindent{\scriptsize\bfseries Panel C. Fold stability and full-data performance analyses.}
\vspace{2pt}
{\tiny\raggedright
Examines robustness and error modes beyond the aggregate scores. \textbf{C1}, fold-to-fold variability and average method ranking across balanced accuracy, macro F1, and macro ROC AUC; \textbf{C2}, probability-level analysis at 100\% labelled data, including micro-averaged ROC/precision--recall curves, reliability curves, and calibration scores; \textbf{C3}, row-normalized confusion matrices at 100\% labelled data, with rows denoting true cannabis-dependence classes and columns denoting predicted classes.\par}

\clearpage

\begin{center}
\includegraphics[width=0.838\textwidth,keepaspectratio]{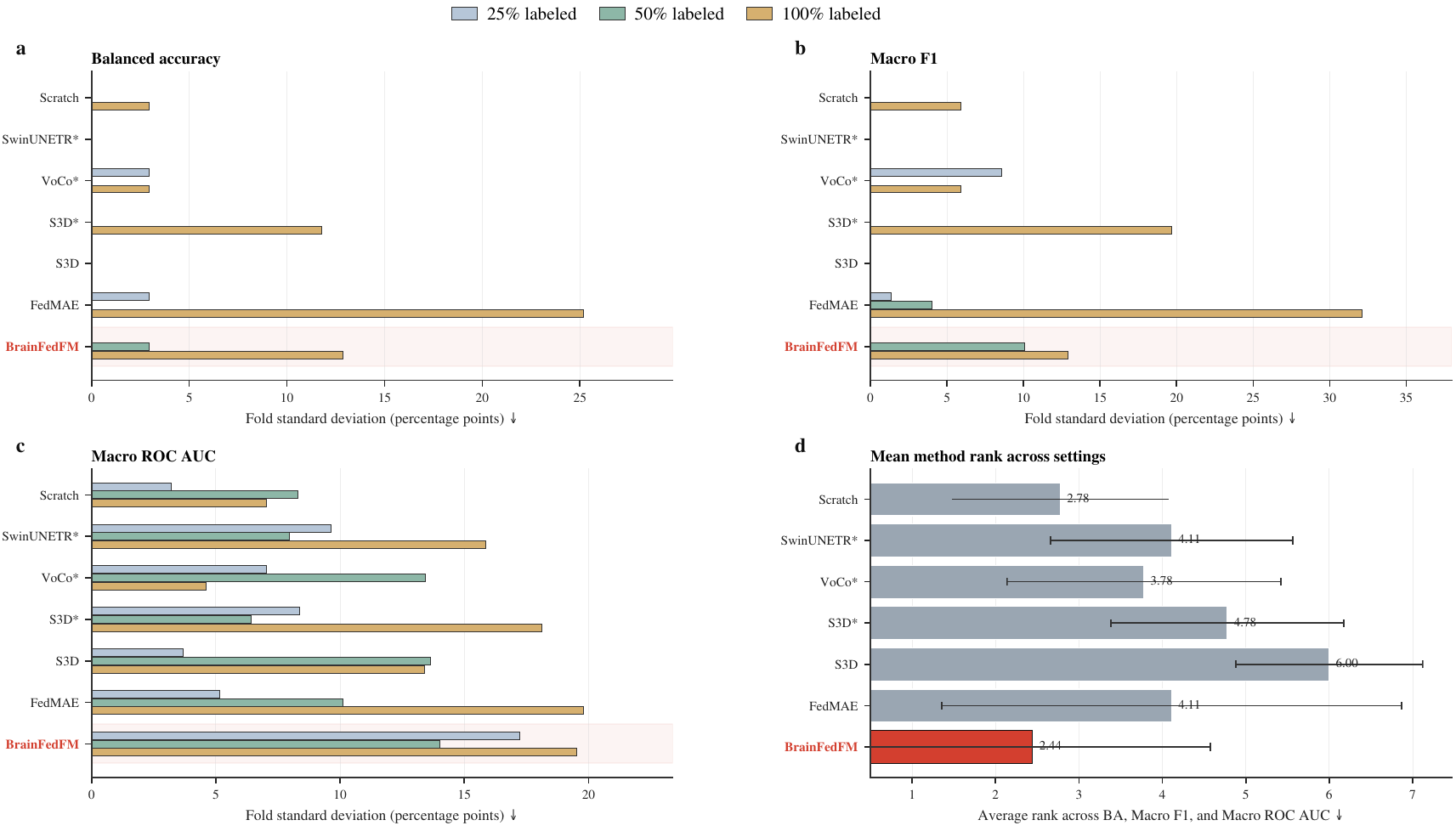}   \par\vspace{6pt}
\includegraphics[width=0.785\textwidth,keepaspectratio]{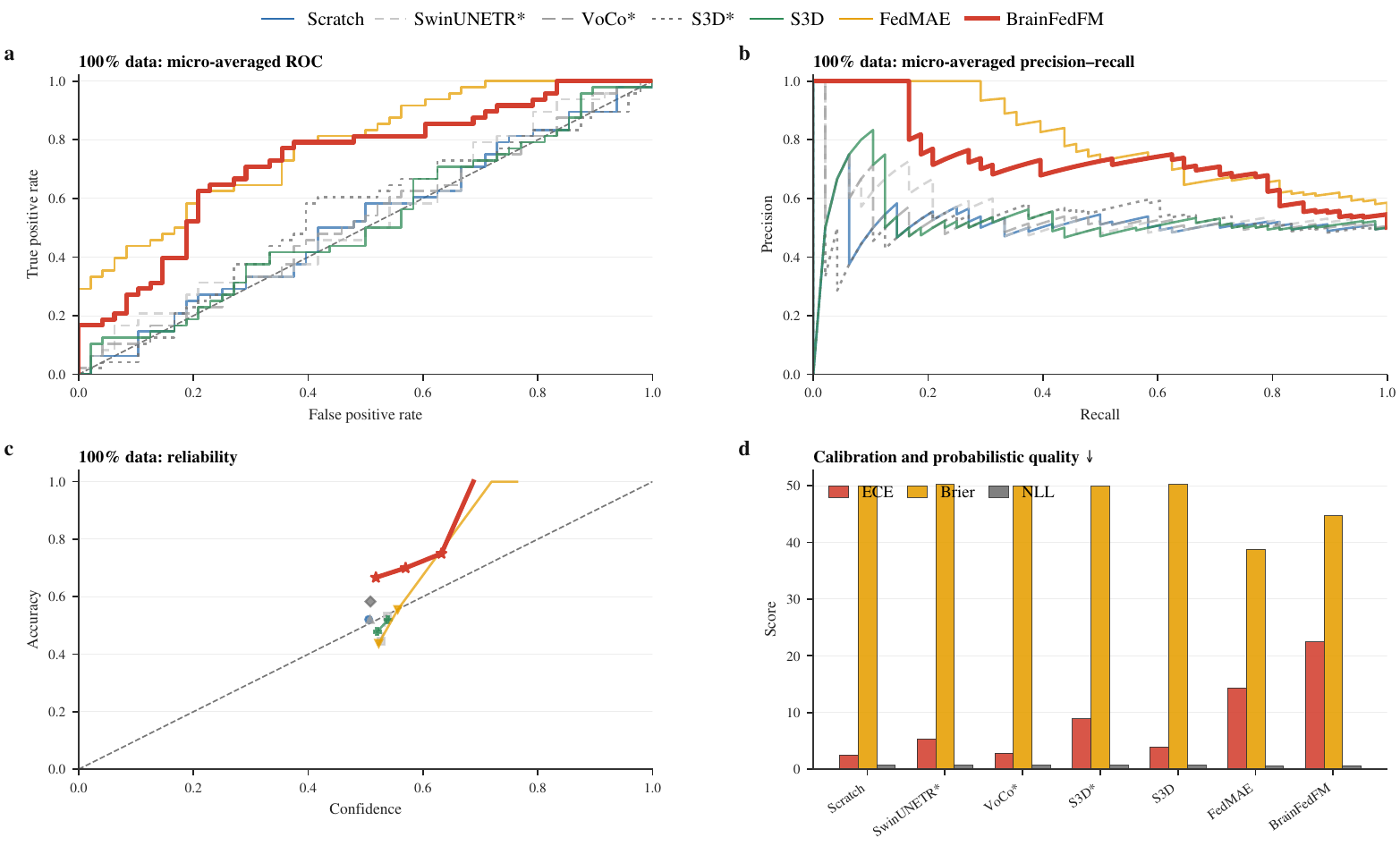}   \par\vspace{4pt}
\includegraphics[width=0.83\textwidth,keepaspectratio]{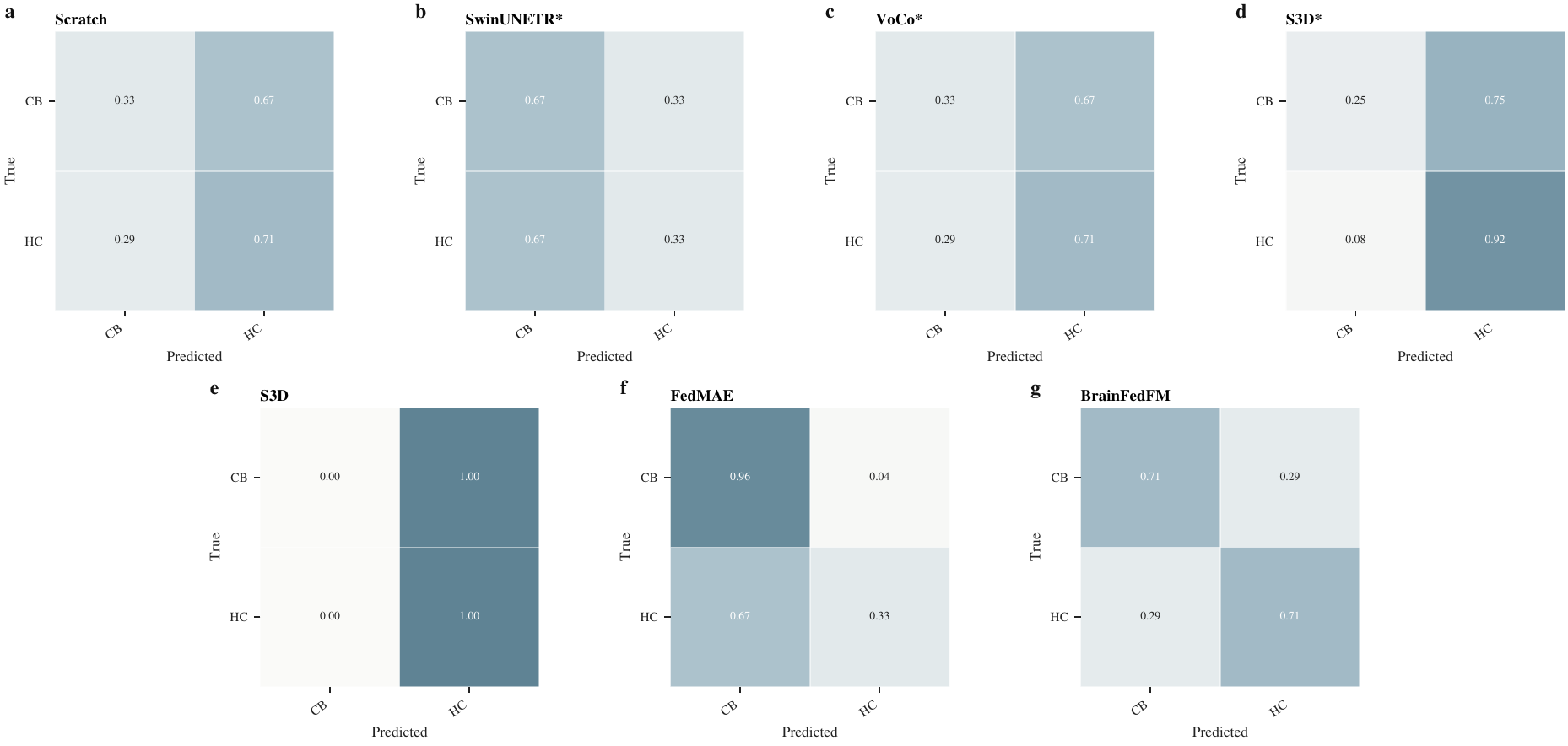}
\end{center}

\definecolor{pct100}{gray}{0.97}
\definecolor{pct75}{gray}{1.0}
\definecolor{pct50}{gray}{0.99}
\definecolor{pct25}{gray}{1.0}
\definecolor{overallgray}{gray}{0.97}
\definecolor{ourscol}{RGB}{244,248,252}

\begin{table*}[h]
\centering
\tiny
\caption[Pediatric demyelinating segmentation]{\textbf{Pediatric demyelinating lesion segmentation task.}
We evaluate foundation model performance on the PediDemi benchmark, where the goal is to delineate demyelinating lesions from pediatric brain MRI. Accurate segmentation enables quantitative assessment of lesion burden and supports longitudinal evaluation of disease evolution, treatment response, and imaging-based outcomes in demyelinating disorders. Panel~A reports Dice, intersection-over-union (IoU), and normalized surface Dice (NSD) with 100\% labelled training data, Panel~B reports the corresponding results with 25\% and 50\% labelled data. The labelled-data percentages denote the fraction of the training--validation pool used for downstream fine-tuning and model selection; the held-out test set remains unchanged across settings. We compare a randomly initialized scratch baseline, centralized pretrained foundation models, and federated pretrained foundation models. Higher Dice, IoU, and NSD indicate better performance.}
\label{tab:result_demy}

\renewcommand{\arraystretch}{3.6}
{\scriptsize\bfseries Panel A. Detailed results under 100\% labelled data.}
\vspace{5pt}

{\setlength{\tabcolsep}{3.38pt}

\resizebox{1.0\textwidth}{!}{
\begin{tabular}{
l c c c c c c
>
{\columncolor{ourscol}\centering\arraybackslash}p{1.62cm}
}

\toprule
\multicolumn{2}{c}{\textbf{Brain sMRI FMs }}
& \multicolumn{4}{c}{Centralized Pretraining}
& \multicolumn{2}{c}{Federated Pretraining} \\
\cmidrule(lr){1-2}\cmidrule(lr){3-6}\cmidrule(lr){7-8}
\makecell[c]{Metric/Methods}
& Scratch
& \makecell[c]{SwinUNETR$^{\ast}$} & VoCo$^{\ast}$ & S3D$^{\ast}$ & S3D
& \makecell[c]{FedMAE} & \makecell[c]{BrainFedFM} \\
\midrule
\rowcolor{overallgray}
\textbf{Overall Dice $\uparrow$}
& 44.20($\pm$10.46)
& 44.19($\pm$14.89)
& 51.56($\pm$16.26)
& 46.01($\pm$14.01)
& 40.40($\pm$8.590)
& 39.94($\pm$9.202)
& \cellcolor{ourscol}48.58($\pm$14.80) \\

\rowcolor{overallgray}
\textbf{Overall IoU $\uparrow$}
& 31.90($\pm$6.329)
& 31.27($\pm$11.04)
& 39.15($\pm$11.33)
& 33.60($\pm$9.579)
& 29.19($\pm$4.842)
& 28.89($\pm$5.837)
& \cellcolor{ourscol}36.79($\pm$10.57) \\

\rowcolor{overallgray}
\textbf{Overall NSD $\uparrow$}
& 54.81($\pm$14.78)
& 55.52($\pm$21.67)
& 63.06($\pm$22.17)
& 56.53($\pm$18.56)
& 48.41($\pm$12.01)
& 48.28($\pm$12.989)
& \cellcolor{ourscol}56.36($\pm$18.42) \\
\midrule

Demyelinating (Dice)
& 44.20($\pm$10.46)
& 44.19($\pm$14.89)
& 51.56($\pm$16.26)
& 46.01($\pm$14.01)
& 40.40($\pm$8.590)
& 39.94($\pm$9.202)
& 48.58($\pm$14.80) \\

Demyelinating (IoU)
& 31.90($\pm$6.329)
& 31.27($\pm$11.04)
& 39.15($\pm$11.33)
& 33.60($\pm$9.579)
& 29.19($\pm$4.842)
& 28.89($\pm$5.837)
& 36.79($\pm$10.57) \\

Demyelinating (NSD)
& 54.81($\pm$14.78)
& 55.52($\pm$21.67)
& 63.06($\pm$22.17)
& 56.53($\pm$18.56)
& 48.41($\pm$12.01)
& 48.28($\pm$12.989)
& 56.36($\pm$18.42) \\
\bottomrule
\end{tabular}
}}

\vspace{8pt}

{\scriptsize\bfseries Panel B. Summary under different labelled-data percentages.}
\vspace{5pt}

\renewcommand{\arraystretch}{2.3}
\resizebox{1.0\textwidth}{!}{%
\begin{tabular}{
l
ccc!{\hspace{2pt}\color{black!45}\vrule width 0.6pt\hspace{2pt}}
ccc
}
\toprule
\multicolumn{1}{c}{\textbf{Percentages}}
& \multicolumn{3}{c}{25\%}
& \multicolumn{3}{c}{50\%} \\
\cmidrule(lr){1-1}\cmidrule(lr){2-4}\cmidrule(lr){5-7}
\makecell[c]{Methods/Metric} & \multicolumn{1}{c}{Dice} & \multicolumn{1}{c}{IoU} & \multicolumn{1}{c}{NSD}
& \multicolumn{1}{c}{Dice} & \multicolumn{1}{c}{IoU} & \multicolumn{1}{c}{NSD} \\
\midrule
Scratch
& 13.63($\pm$9.510) & 8.886($\pm$6.329) & 15.69($\pm$10.82)
& 20.90($\pm$4.028) & 13.13($\pm$4.043) & 28.68($\pm$3.710) \\
SwinUNETR$^{\ast}$
& 19.92($\pm$6.751) & 12.83($\pm$4.862) & 25.77($\pm$7.252)
& 28.50($\pm$8.203) & 19.32($\pm$5.091) & 35.88($\pm$12.87) \\
VoCo$^{\ast}$
& 20.19($\pm$10.34) & 13.71($\pm$7.430) & 22.82($\pm$12.06)
& 39.15($\pm$7.503) & 28.02($\pm$4.219) & 47.42($\pm$9.982) \\
S3D$^{\ast}$
& 24.21($\pm$8.813) & 16.50($\pm$7.035) & 30.82($\pm$9.553)
& 34.37($\pm$7.844) & 23.48($\pm$4.853) & 42.38($\pm$8.799) \\
S3D
& 22.51($\pm$12.83) & 15.83($\pm$9.583) & 28.26($\pm$14.33)
& 33.22($\pm$10.11) & 24.44($\pm$5.376) & 38.54($\pm$14.38) \\
\hline
FedMAE
& 20.55($\pm$14.53) & 14.83($\pm$10.54) & 23.90($\pm$16.81)
& 28.97($\pm$8.810) & 20.39($\pm$6.542) & 33.42($\pm$10.75) \\
\rowcolor{ourscol}
BrainFedFM
& 25.55($\pm$5.725) & 17.86($\pm$5.268) & 33.06($\pm$4.564)
& 37.44($\pm$6.661) & 24.87($\pm$4.656) & 46.29($\pm$7.669) \\
\bottomrule
\end{tabular}
}
\vspace{3pt}

\begin{minipage}{1.0\textwidth}
\scriptsize
\textit{Notes. Dice, Dice similarity coefficient; IoU, intersection-over-union; NSD, normalized surface Dice. All models are fine-tuned for 150 epochs with 50 training iterations and 50 validation iterations per epoch using SGD with momentum 0.99, Nesterov acceleration, weight decay of $3\times10^{-5}$, and an initial learning rate of $1\times10^{-3}$. Training uses a 15-epoch linear warm-up followed by polynomial learning-rate decay, and test performance is reported from the checkpoint with the best validation result.}
\end{minipage}
\end{table*}

\clearpage

\noindent{\scriptsize\bfseries Panel C. Data efficiency, complementary profiles, and qualitative lesion delineation.}
\vspace{2pt}
{\tiny\raggedright
Examines annotation efficiency and complementary segmentation behavior beyond the tabulated overlap scores. \textbf{C1}, fold-averaged trajectories for Dice, NSD, voxel sensitivity, lesion F1, HD95, and absolute relative volume error; \textbf{C2}, direction-aligned profiles across overlap/surface, voxel/volume, and lesion-level metrics, together with a family-level summary; \textbf{C3}, representative axial MRI examples with reference lesions and three-fold ensemble error maps, where green, orange, and purple denote true-positive, false-positive, and false-negative lesion voxels, respectively.\par}
\par\vspace{-2pt}

\begin{center}
\includegraphics[width=0.975\textwidth,keepaspectratio]{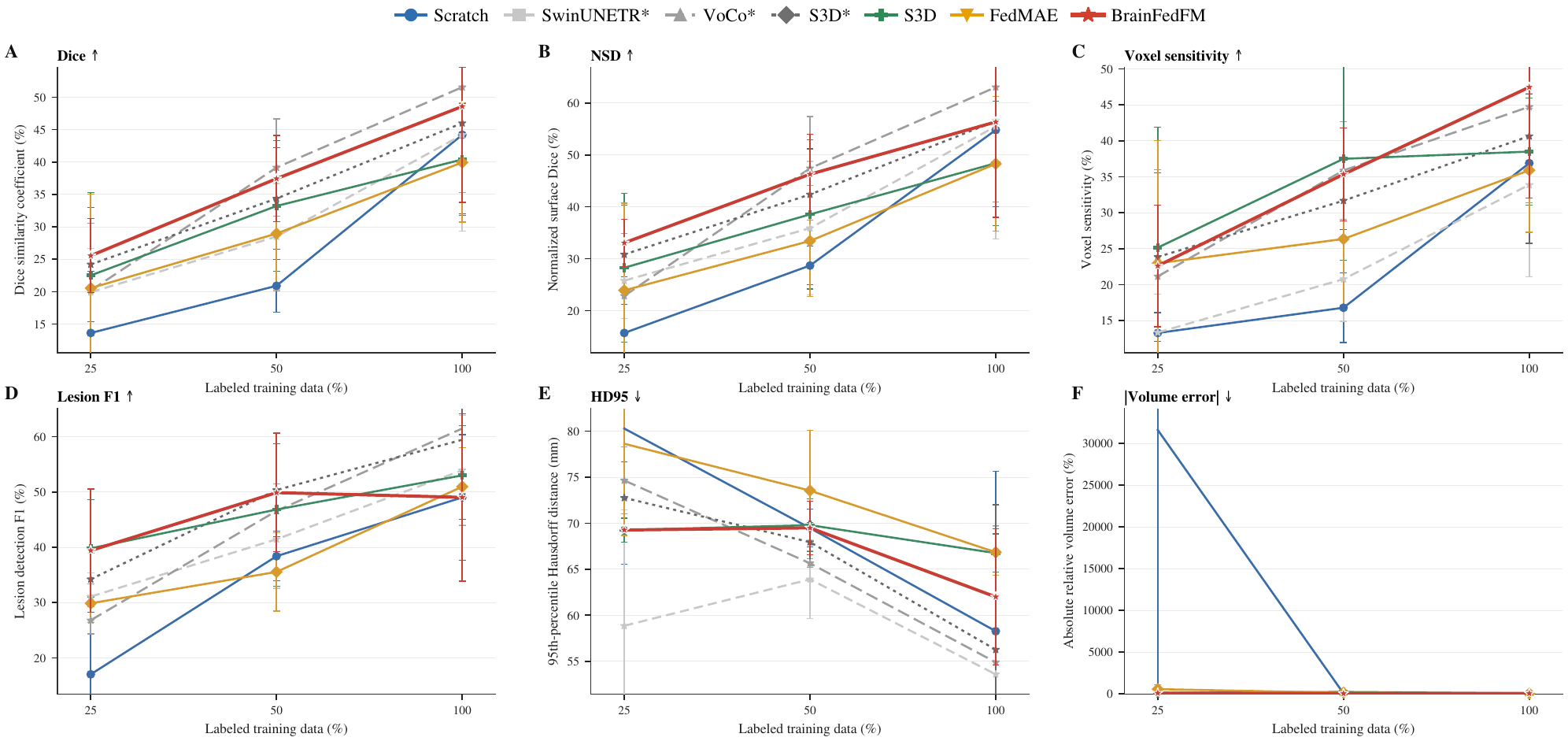}
\includegraphics[width=0.842\textwidth,keepaspectratio]{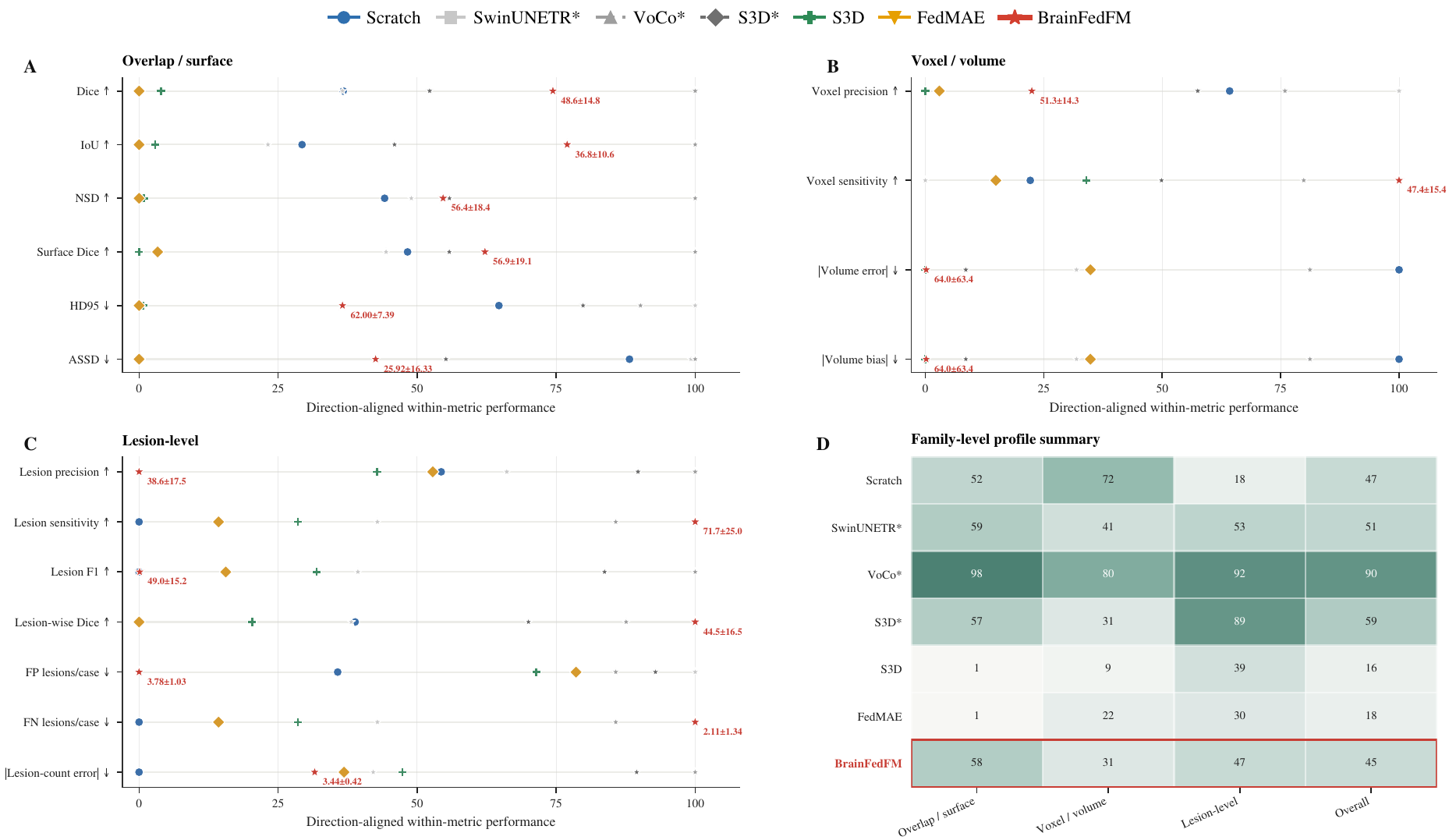}
\includegraphics[width=0.99\textwidth,keepaspectratio]{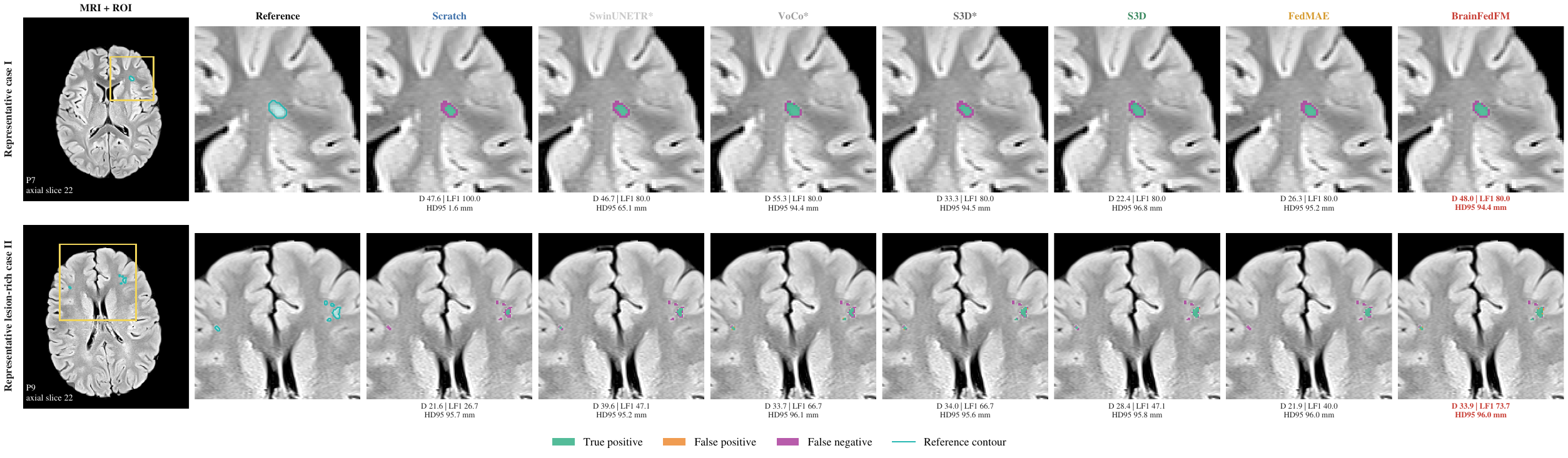}
\end{center}

\clearpage

\noindent{\scriptsize\bfseries Panel D. Error decomposition, case-level variability, and lesion-scale performance.}
\vspace{2pt}
{\tiny\raggedright
Characterizes segmentation failure modes and robustness beyond aggregate accuracy. \textbf{D1}, false-negative and false-positive voxel burden, signed volume bias, Dice--NSD performance, and voxel precision--sensitivity balance at 100\% labelled data; \textbf{D2}, matched case-by-fold distributions of HD95, average symmetric surface distance (ASSD), absolute relative volume error, and absolute lesion-count error; \textbf{D3}, lesion-detection sensitivity and lesion-wise Dice across reference-lesion volume quintiles, with dedicated summaries for the smallest lesions.\par}
\par\vspace{-2pt}

\begin{center}
\includegraphics[width=0.744\textwidth,keepaspectratio]{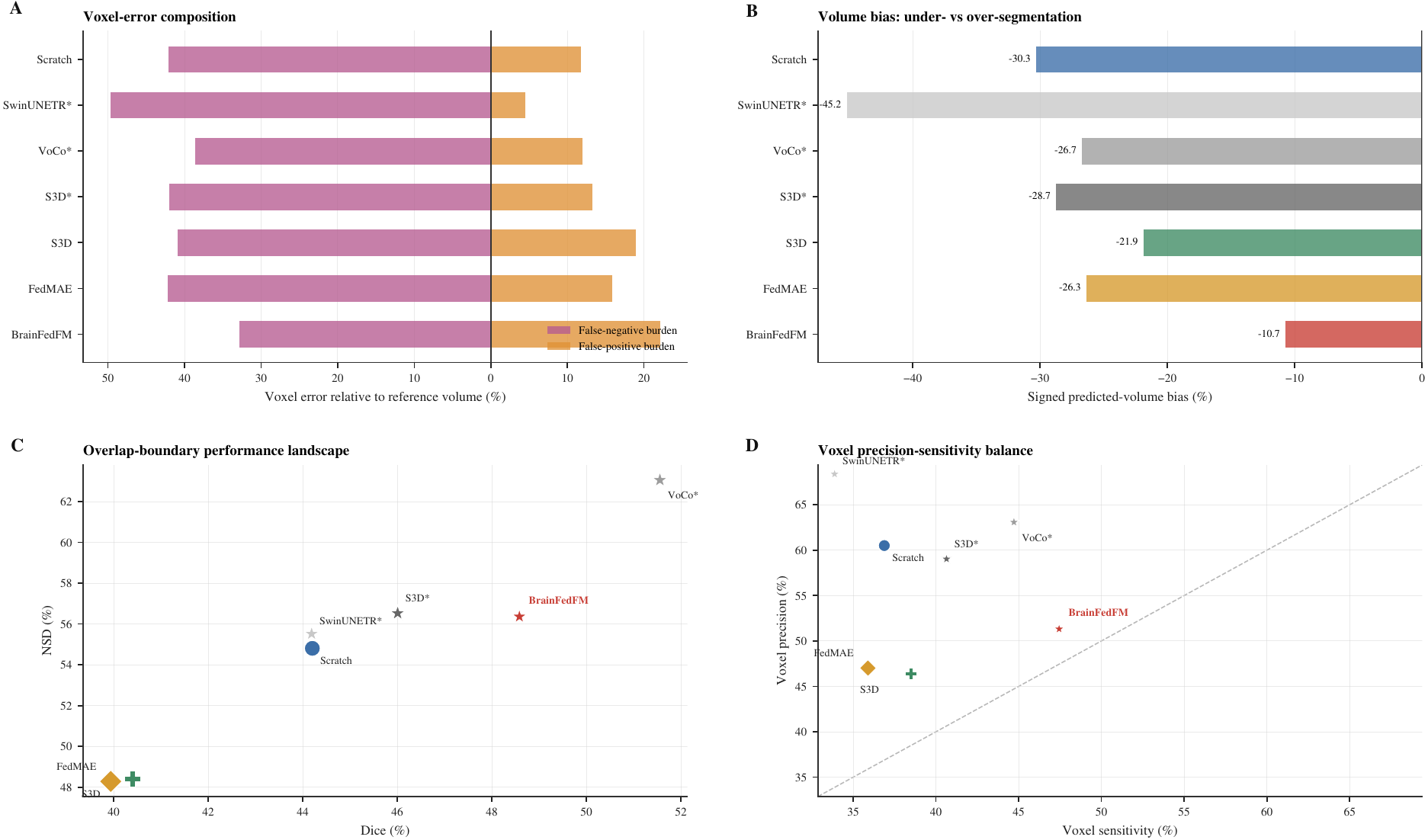}
\includegraphics[width=0.732\textwidth,keepaspectratio]{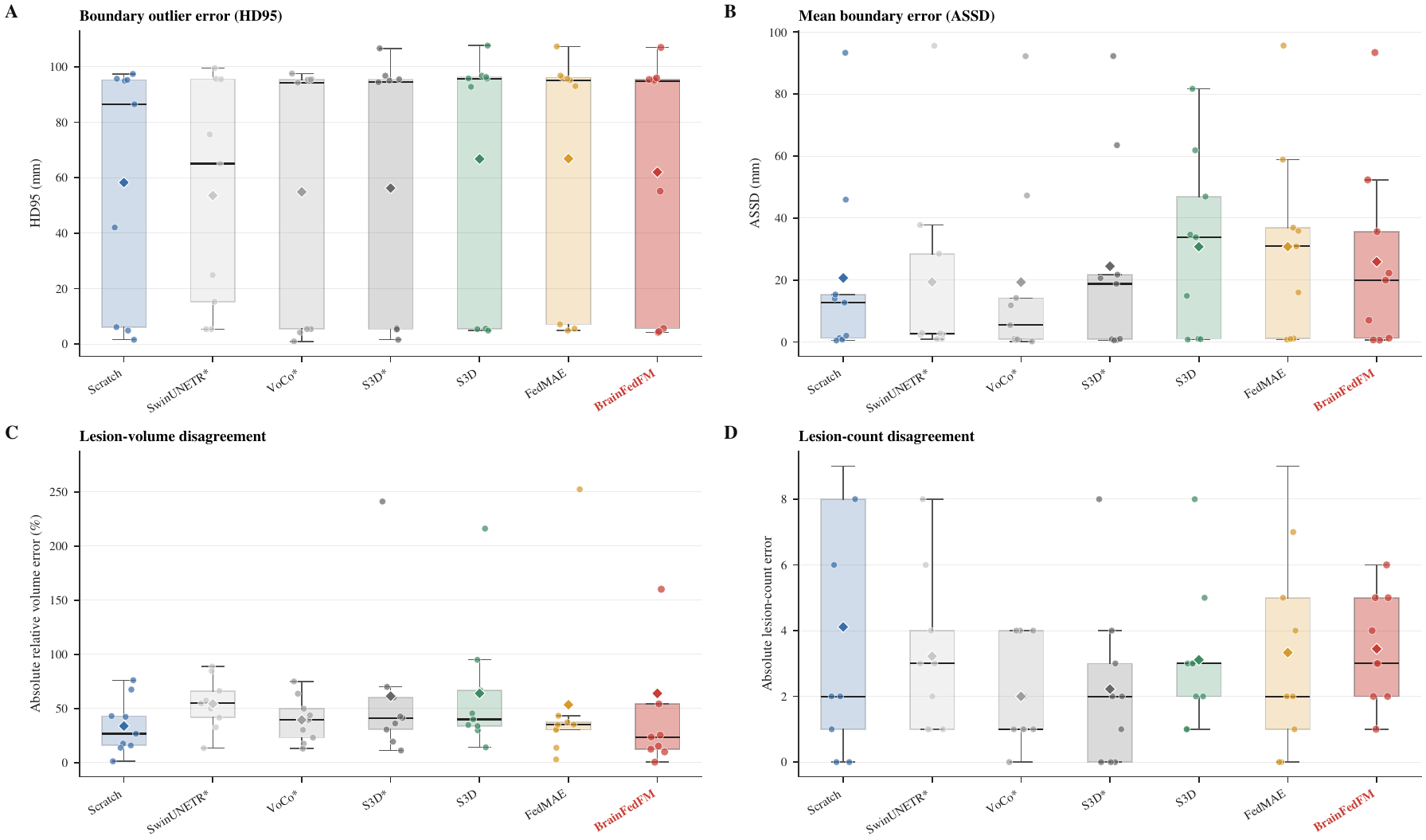}
\includegraphics[width=0.738\textwidth,keepaspectratio]{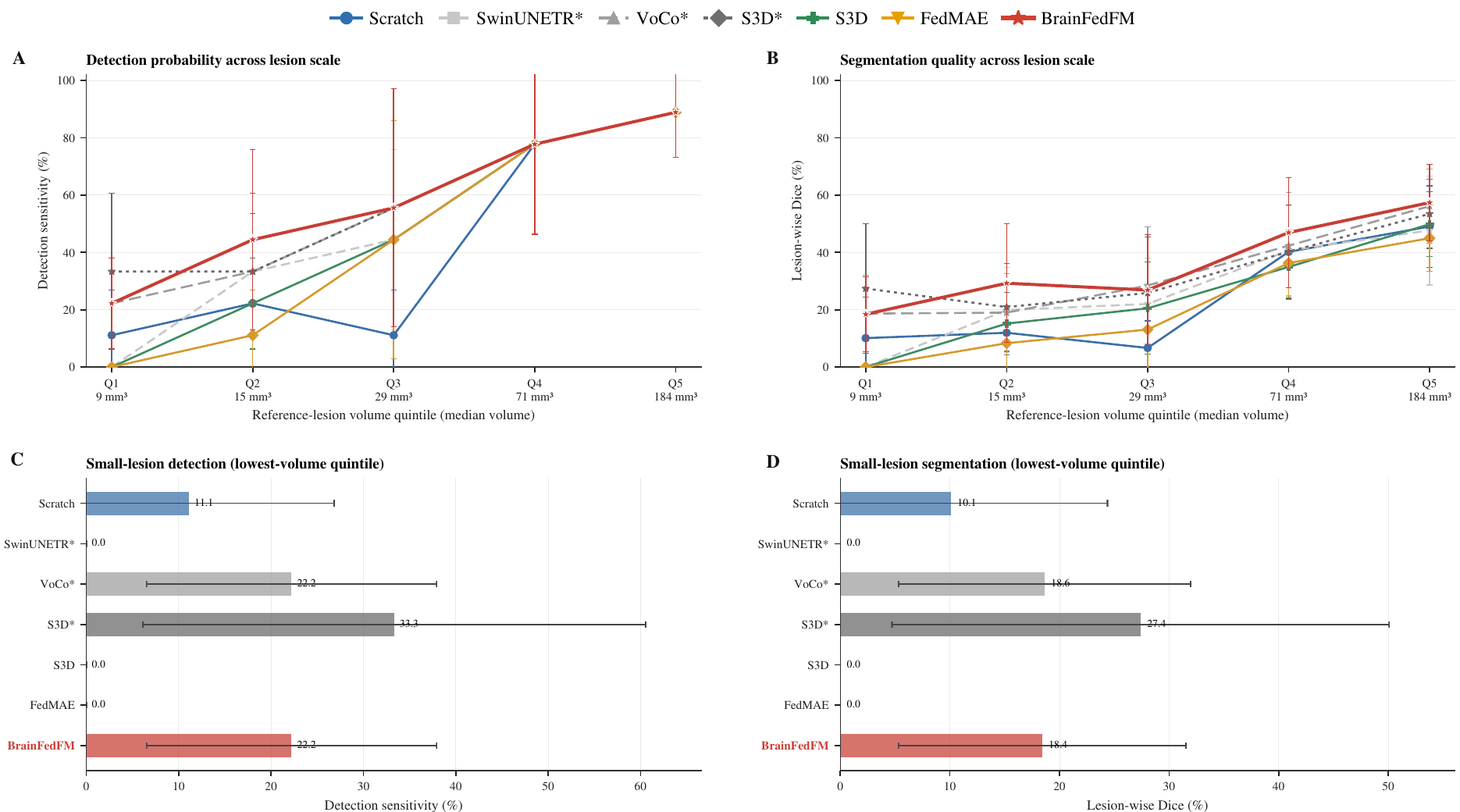}
\end{center}

\definecolor{pct100}{gray}{0.97}
\definecolor{pct75}{gray}{1.0}
\definecolor{pct50}{gray}{0.99}
\definecolor{pct25}{gray}{1.0}
\definecolor{overallgray}{gray}{1.0}
\definecolor{ourscol}{RGB}{244,248,252}

\begin{table*}[t]
\centering
\tiny
\caption[Multiple sclerosis diagnosis task]{\textbf{Multiple sclerosis diagnosis task.}
We evaluate foundation model performance on the SibBMS benchmark. This task aims to classify structural MRI scans into multiple sclerosis (MS) and healthy control (HC) categories.
Accurate MS diagnosis from brain MRI is important for demyelinating disease assessment, supporting consistent subject stratification, cohort curation, and downstream model evaluation.
The table summarizes performance under 25\%, 50\%, and 100\% labelled-data settings, which indicate the proportion of the labelled training--validation pool used for downstream fine-tuning and model selection; the held-out test set remains unchanged across all settings.
We compare a randomly initialized scratch baseline, centralized pretrained foundation models, and federated pretrained foundation models.}
\label{tab:result_mms}

\renewcommand{\arraystretch}{3.85}
{\scriptsize\bfseries Panel A. Summary under different labelled-data percentages.}
\vspace{5pt}

{\setlength{\tabcolsep}{3.38pt}

\resizebox{1.0\textwidth}{!}{
\begin{tabular}{
l c c c c c c
>{\columncolor{ourscol}\centering\arraybackslash}p{1.62cm}
}
\toprule
\multicolumn{2}{c}{\textbf{Brain sMRI FMs }}
& \multicolumn{4}{c}{Centralized Pretraining}
& \multicolumn{2}{c}{Federated Pretraining} \\
\cmidrule(lr){1-2}\cmidrule(lr){3-6}\cmidrule(lr){7-8}
\makecell[c]{Metric/Methods}
& Scratch
& \makecell[c]{SwinUNETR$^{\ast}$}  &\ VoCo$^{\ast}$ \ & \ S3D$^{\ast}$  & \ S3D 
& \makecell[c]{FedMAE} & \makecell[c]{BrainFedFM} \\
\midrule

\rowcolor{pct100}
\textbf{100\%  BA $\uparrow$}
& 75.63($\pm$2.076)& 67.15($\pm$0.447)  & 83.21($\pm$6.018) & 72.65($\pm$3.619) & 71.39($\pm$3.840) & 79.37($\pm$0.773)
& \cellcolor{ourscol} 84.02($\pm$1.501) \\
\rowcolor{pct100}
\textbf{100\%   mF1 $\uparrow$}
& 75.10($\pm$2.112) & 64.82($\pm$0.685)  & 83.09($\pm$6.432) & 71.24($\pm$4.786) & 69.90($\pm$4.881) & 79.32($\pm$0.847)
& \cellcolor{ourscol}  84.12($\pm$1.552) \\
\rowcolor{pct100}
\textbf{100\%   AUC $\uparrow$}
& 81.00($\pm$2.421) & 76.43($\pm$1.571) &  94.75($\pm$1.975) & 82.42($\pm$1.119) &  81.76($\pm$3.208) & 91.22($\pm$0.731)
& \cellcolor{ourscol} 94.39($\pm$1.320)  \\
\midrule

\rowcolor{pct50}
\textbf{50\%  BA $\uparrow$}
& 56.06($\pm$4.088) & 56.34($\pm$4.839)  & 76.54($\pm$12.35) & 65.81($\pm$11.55) & 71.82($\pm$2.533) & 74.65($\pm$3.910)
& \cellcolor{ourscol}76.69($\pm$3.843) \\
\rowcolor{pct50}
\textbf{50\%  mF1 $\uparrow$}
& 46.03($\pm$7.200) & 47.84($\pm$10.76)  & 74.18($\pm$15.87) & 59.84($\pm$18.37) &  70.76($\pm$3.485) & 74.37($\pm$4.054)
& \cellcolor{ourscol}75.90($\pm$4.444) \\
\rowcolor{pct50}
\textbf{50\% AUC $\uparrow$}
& 73.13($\pm$1.709) & 68.83($\pm$5.703)  & 88.70($\pm$6.078) &  79.87($\pm$6.473) & 81.90($\pm$4.832) &  83.12($\pm$3.596)
& \cellcolor{ourscol}87.88($\pm$3.892)  \\
\midrule

\rowcolor{pct25}
\textbf{25\%  BA $\uparrow$}
& 50.00($\pm$0.000) & 50.00($\pm$0.000)  &  68.59($\pm$4.279) & 56.21($\pm$8.785) & 65.63($\pm$1.911) & 62.73($\pm$0.866)
& \cellcolor{ourscol}67.02($\pm$1.299) \\
\rowcolor{pct25}
\textbf{25\%  mF1 $\uparrow$}
& 33.00($\pm$0.911) & 34.29($\pm$0.000)  & 66.63($\pm$5.717) & 45.74($\pm$16.19) & 64.10($\pm$1.532) & 62.70($\pm$0.855)
& \cellcolor{ourscol}65.39($\pm$2.705)\\
\rowcolor{pct25}
\textbf{25\% AUC $\uparrow$}
& 68.42($\pm$5.880) & 64.36($\pm$3.980)  & 79.74($\pm$3.431) & 72.18($\pm$4.330) & 72.52($\pm$2.217) & 71.70($\pm$1.807)
& \cellcolor{ourscol}74.30($\pm$2.390) \\
\bottomrule
\end{tabular}
}}

\vspace{5pt}

\begin{minipage}{1.0\textwidth}
\scriptsize
\textit{Notes. Values are mean($\pm$std) over three folds. BA, balanced accuracy; mF1, macro F1-score; AUC, macro one-vs-rest ROC AUC. Scratch uses random initialization; SwinUNETR$^{\ast}$, VoCo$^{\ast}$, and S3D$^{\ast}$ use publicly released centralized foundation model weights; S3D uses centralized pretraining on our pretraining corpus, FedMAE and BrainFedFM use federated pretraining.
Fine-tuning uses 8 epochs of AdamW, learning rate $1\times10^{-6}$, weight decay $1\times10^{-2}$, cosine scheduling with a 1-epoch warm-up, and label smoothing of 0.2. The best checkpoint is selected according to validation AUC.}
\end{minipage}
\end{table*}

\clearpage

\noindent{\scriptsize\bfseries Panel B. Labelled-data scaling and aggregate metric summaries.}
\vspace{2pt}
{\tiny\raggedright
Summarizes method behavior across labelled-data settings.
\textbf{B1}, fold-averaged trends for classification and calibration metrics; \textbf{B2}, extended metric heatmaps separating higher-is-better classification metrics from lower-is-better calibration/uncertainty metrics; \textbf{B3}, the accuracy--calibration trade-off, where points closer to the upper-left indicate stronger discrimination with better calibration.\par}
\vspace{-3pt}

\begin{center}
\includegraphics[width=1.0\textwidth,keepaspectratio]{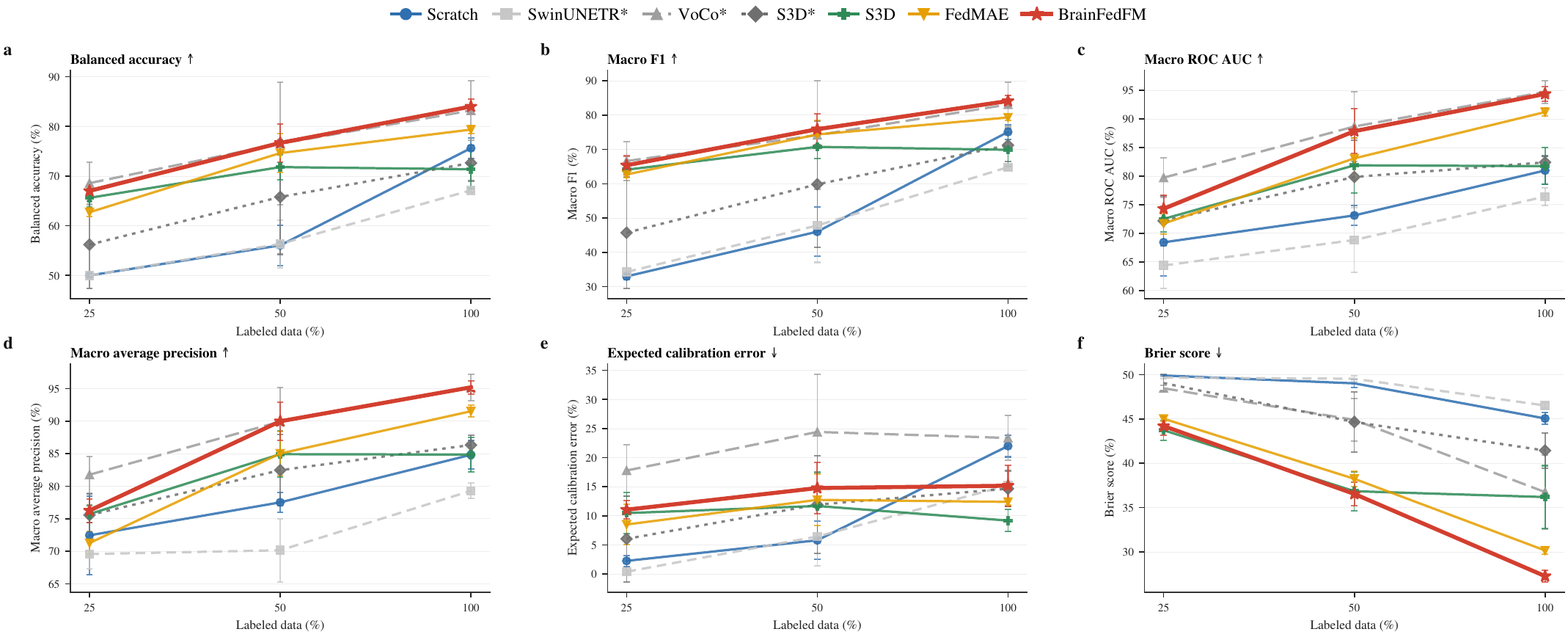} \par\vspace{3pt}
\includegraphics[width=1.0\textwidth,keepaspectratio]{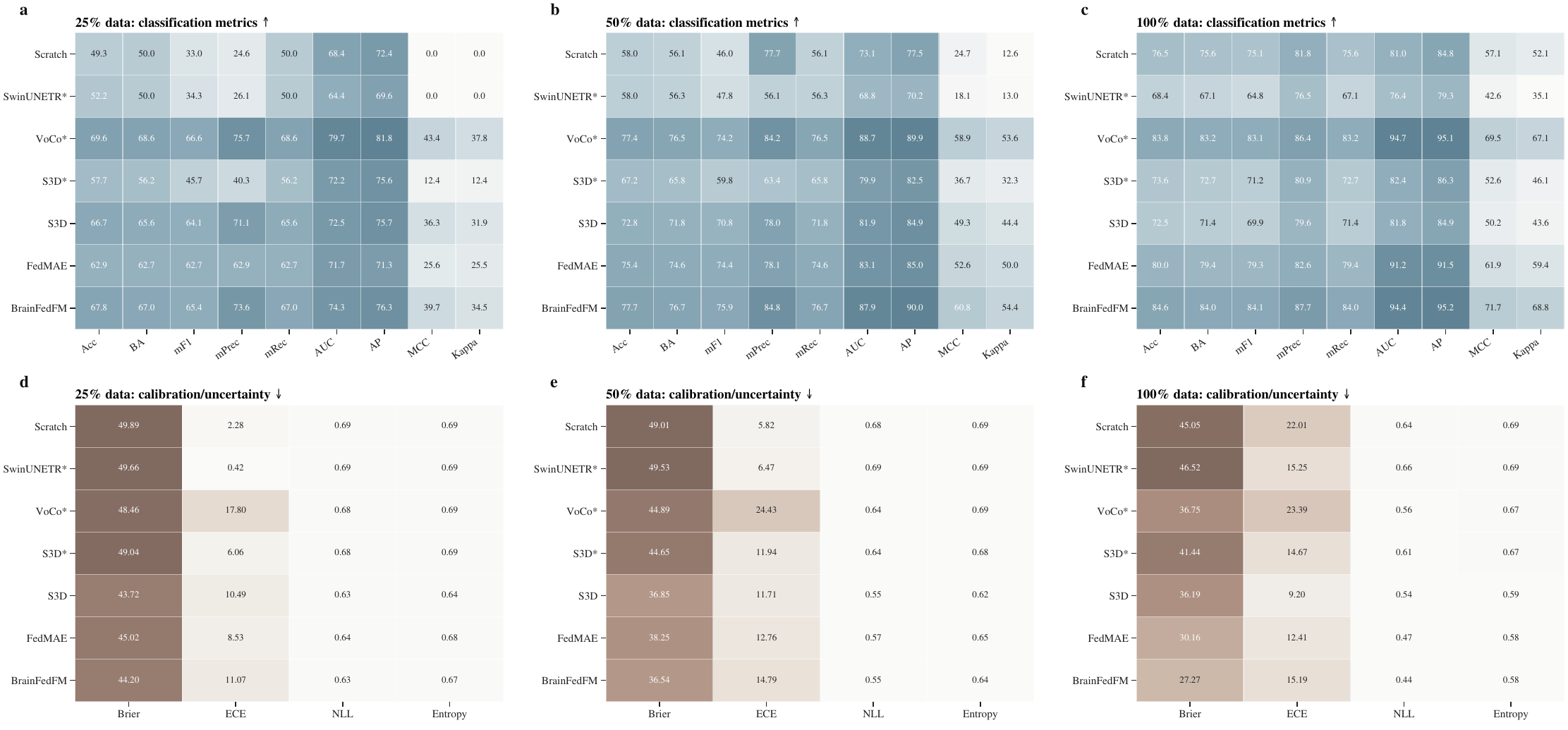} \par\vspace{5pt}
\includegraphics[width=0.935\textwidth,keepaspectratio]{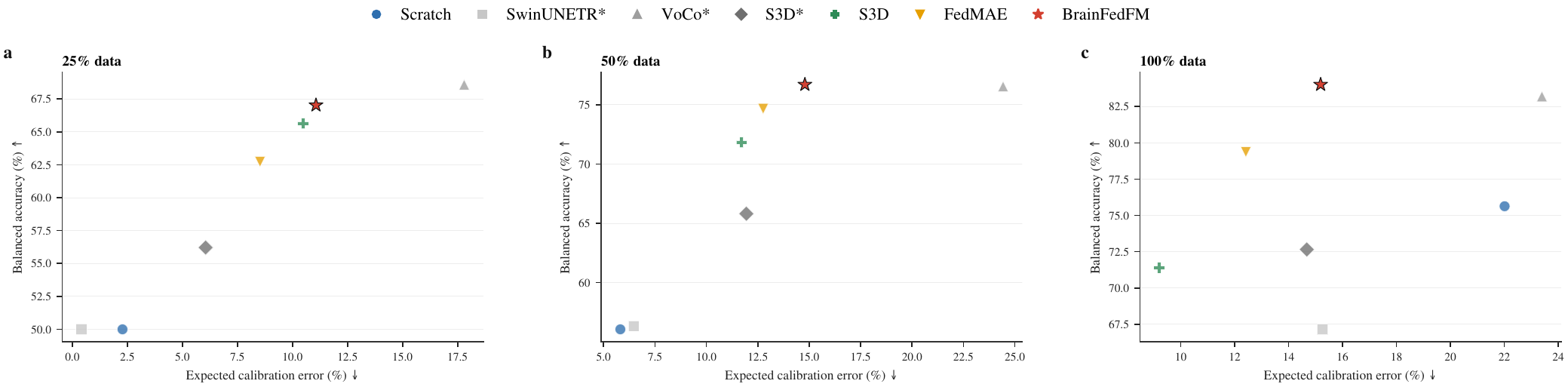}
\end{center}

\vspace{-2pt}

\noindent{\scriptsize\bfseries Panel C. Fold stability and full-data performance analyses.}
\vspace{2pt}
{\tiny\raggedright
Examines robustness and error modes beyond the aggregate scores. \textbf{C1}, fold-to-fold variability and average method ranking across balanced accuracy, macro F1, and macro ROC AUC; \textbf{C2}, probability-level analysis at 100\% labelled data, including micro-averaged ROC/precision--recall curves, reliability curves, and calibration scores; \textbf{C3}, row-normalized confusion matrices at 100\% labelled data, with rows denoting true MS/HC classes and columns denoting predicted classes.\par}

\clearpage

\begin{center}
\includegraphics[width=0.838\textwidth,keepaspectratio]{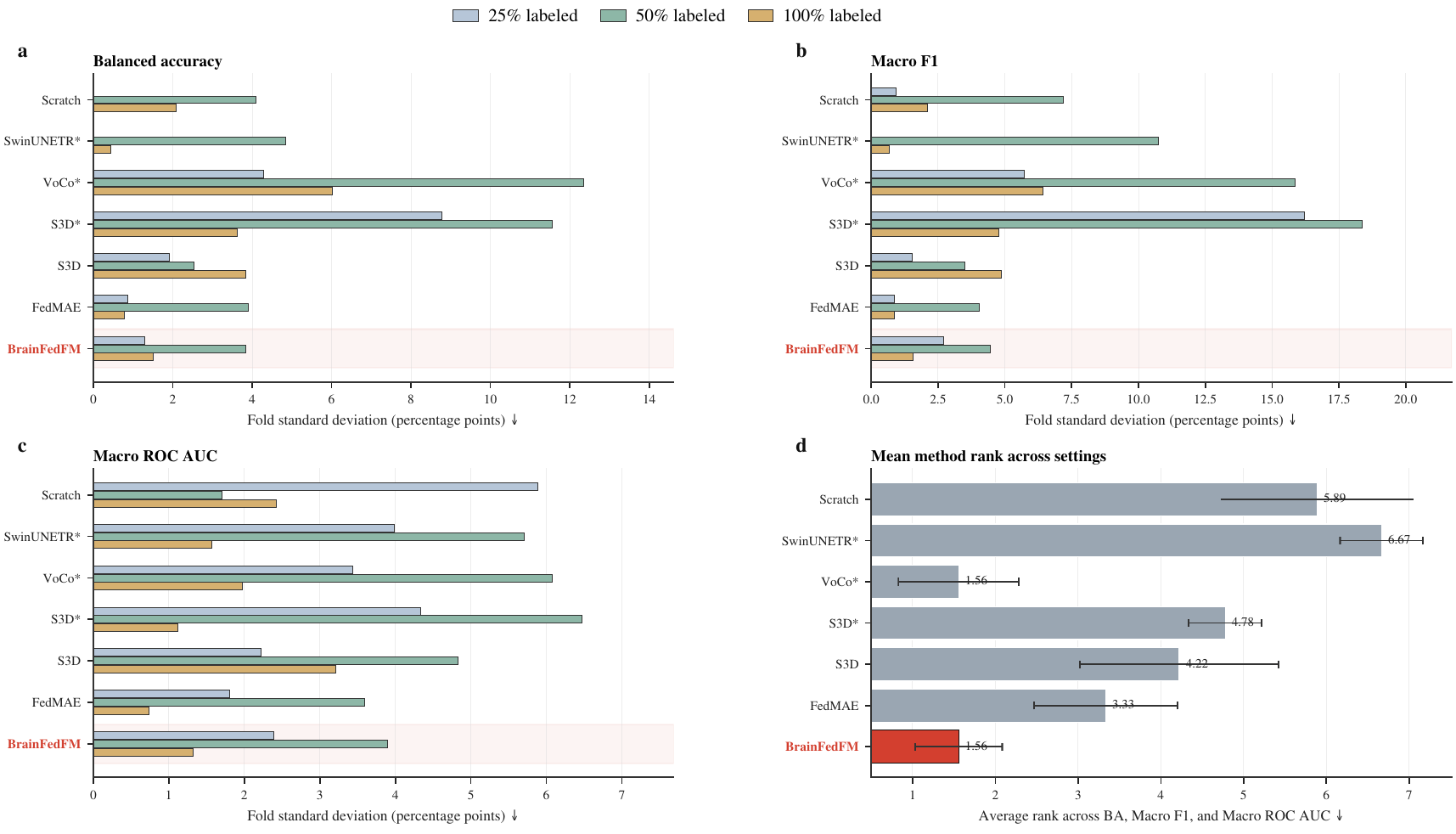}   \par\vspace{6pt}
\includegraphics[width=0.785\textwidth,keepaspectratio]{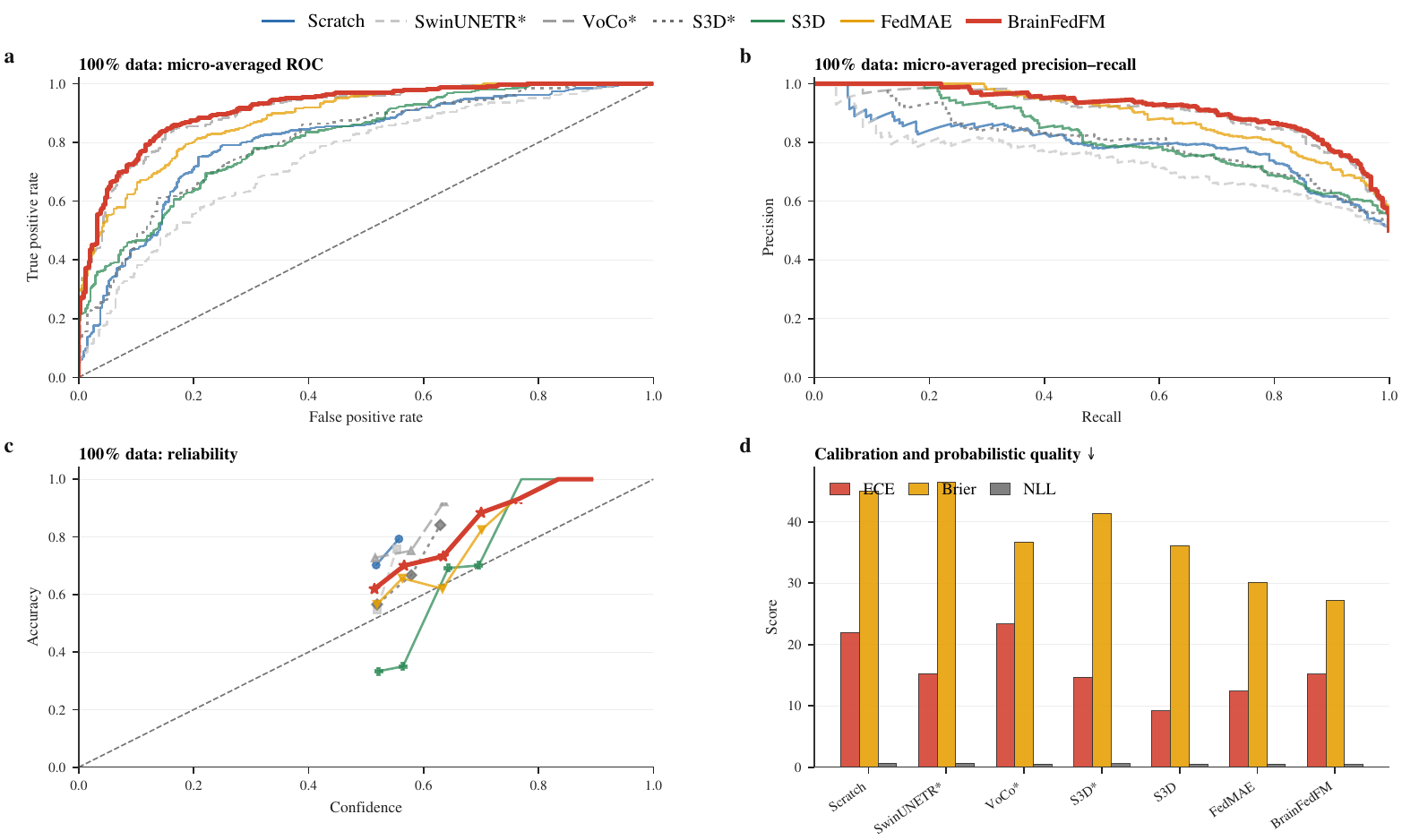}  \par\vspace{4pt}
\includegraphics[width=0.83\textwidth,keepaspectratio]{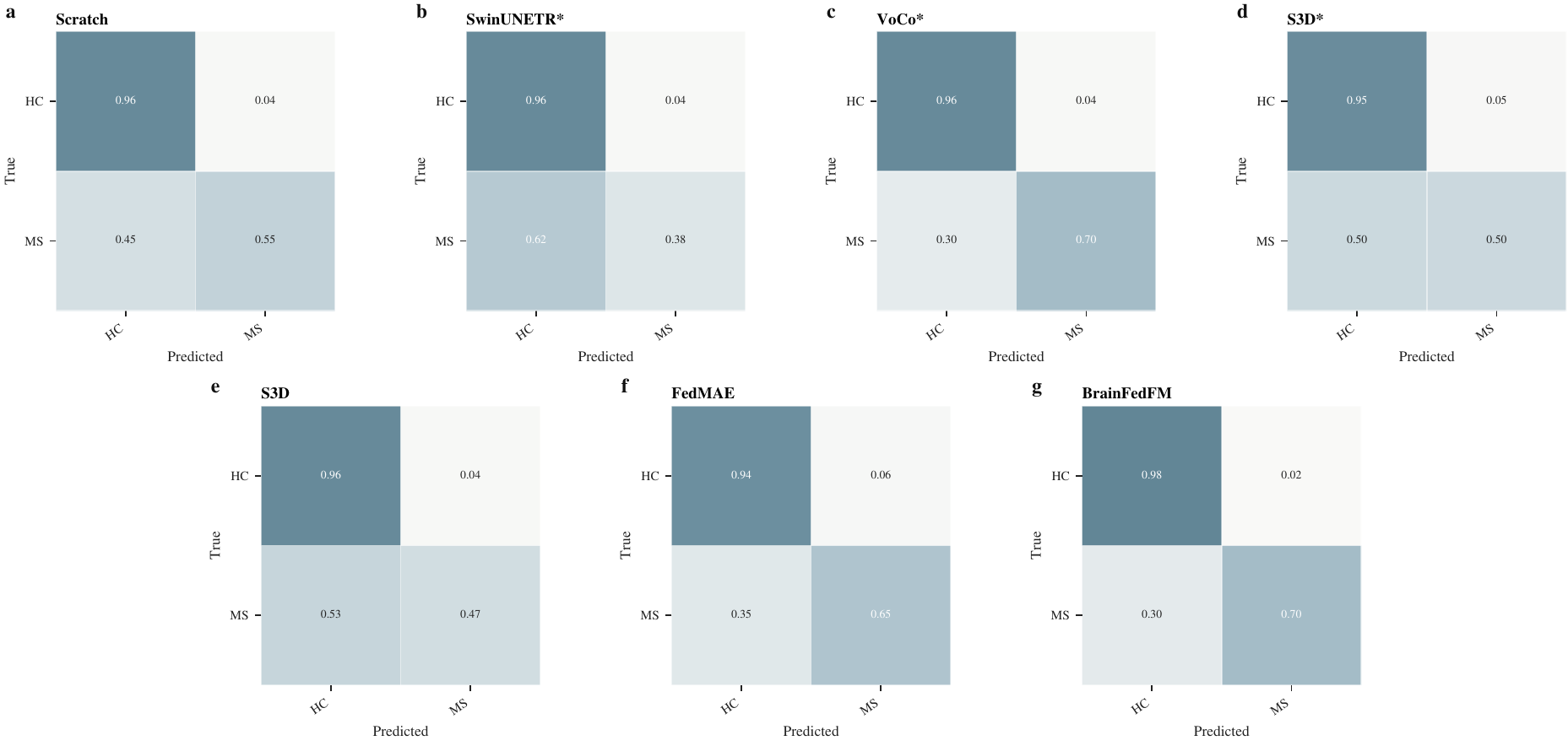}
\end{center}

\definecolor{pct100}{gray}{0.97}
\definecolor{pct75}{gray}{1.0}
\definecolor{pct50}{gray}{0.99}
\definecolor{pct25}{gray}{1.0}
\definecolor{overallgray}{gray}{1.0}
\definecolor{ourscol}{RGB}{244,248,252}

\begin{table*}[t]
\centering
\tiny
\caption[Alzheimer's disease prediction task]{\textbf{Alzheimer's disease prediction task.}
We evaluate foundation model performance on the ADNI-4 benchmark. This task aims to classify brain MRI scans into cognitively normal control (CN), mild cognitive impairment (MCI), and Alzheimer's disease (AD) categories. 
Accurate disease-stage prediction from brain MRI is important for neurodegenerative disease analysis, supporting consistent subject stratification, cohort curation, and downstream model evaluation in Alzheimer's disease studies.
The table summarizes performance under 25\%, 50\%, and 100\% labelled-data settings, which indicate the proportion of the labelled training--validation pool used for downstream fine-tuning and model selection; the held-out test set remains unchanged across all settings.
We compare a randomly initialized scratch baseline, centralized pretrained foundation models, and federated pretrained foundation models.}
\label{tab:result_adni}

\renewcommand{\arraystretch}{3.85}
{\scriptsize\bfseries Panel A. Summary under different labelled-data percentages.}
\vspace{5pt}

{\setlength{\tabcolsep}{3.38pt}

\resizebox{1.0\textwidth}{!}{
\begin{tabular}{
l c c c c c c
>{\columncolor{ourscol}\centering\arraybackslash}p{1.62cm}
}
\toprule
\multicolumn{2}{c}{\textbf{Brain sMRI FMs }}
& \multicolumn{4}{c}{Centralized Pretraining}
& \multicolumn{2}{c}{Federated Pretraining} \\
\cmidrule(lr){1-2}\cmidrule(lr){3-6}\cmidrule(lr){7-8}
\makecell[c]{Metric/Methods}
& Scratch
& \makecell[c]{SwinUNETR$^{\ast}$}  &\ VoCo$^{\ast}$ \ & \ S3D$^{\ast}$  & \ S3D 
& \makecell[c]{FedMAE} & \makecell[c]{BrainFedFM} \\
\midrule

\rowcolor{pct100}
\textbf{100\%  BA $\uparrow$}
& 56.74($\pm$6.951) & 54.13($\pm$4.198) & 91.48($\pm$5.768) & 98.32($\pm$1.794) & 98.94($\pm$0.851) & 96.53($\pm$2.598)
& \cellcolor{ourscol}98.46($\pm$1.464) \\
\rowcolor{pct100}
\textbf{100\%   mF1 $\uparrow$}
& 51.59($\pm$6.245) & 48.74($\pm$3.671) & 92.61($\pm$5.148) & 98.45($\pm$1.309) & 98.97($\pm$0.727) & 97.25($\pm$2.060)
& \cellcolor{ourscol}97.72($\pm$2.202) \\
\rowcolor{pct100}
\textbf{100\%   AUC $\uparrow$}
& 92.41($\pm$4.105) & 81.16($\pm$1.523) & 98.88($\pm$0.819) & 100.0($\pm$0.000) & 99.99($\pm$0.019) & 99.98($\pm$0.029)
& \cellcolor{ourscol}99.99($\pm$0.009) \\
\midrule

\rowcolor{pct50}
\textbf{50\%  BA $\uparrow$}
& 33.33($\pm$0.000) & 33.33($\pm$0.000) & 61.51($\pm$1.508) & 67.95($\pm$8.842) & 82.76($\pm$2.743) & 76.28($\pm$9.494)
& \cellcolor{ourscol}84.66($\pm$2.687) \\
\rowcolor{pct50}
\textbf{50\%  mF1 $\uparrow$}
& 20.77($\pm$0.000) & 19.69($\pm$1.518) & 56.85($\pm$1.507) & 65.06($\pm$12.56) & 83.47($\pm$2.831) & 74.35($\pm$7.445)
& \cellcolor{ourscol}81.14($\pm$2.545) \\
\rowcolor{pct50}
\textbf{50\% AUC $\uparrow$}
& 79.61($\pm$2.679) & 67.82($\pm$2.742) & 87.99($\pm$1.206) & 91.04($\pm$4.067) & 96.04($\pm$0.937) & 97.19($\pm$0.352)
& \cellcolor{ourscol}97.67($\pm$0.893) \\
\midrule

\rowcolor{pct25}
\textbf{25\%  BA $\uparrow$}
& 33.33($\pm$0.000) & 33.33($\pm$0.000) & 38.32($\pm$4.606) & 48.88($\pm$13.69) & 70.88($\pm$3.534) & 64.37($\pm$4.435)
& \cellcolor{ourscol}67.48($\pm$3.783) \\
\rowcolor{pct25}
\textbf{25\%  mF1 $\uparrow$}
& 19.69($\pm$1.518) & 19.69($\pm$1.518) & 30.63($\pm$7.395) & 44.12($\pm$19.95) & 72.03($\pm$4.640) & 65.04($\pm$6.648)
& \cellcolor{ourscol}67.79($\pm$5.329) \\
\rowcolor{pct25}
\textbf{25\% AUC $\uparrow$}
& 61.83($\pm$9.625) & 62.22($\pm$6.703) & 75.72($\pm$3.379) & 80.68($\pm$6.570) & 87.40($\pm$1.041) & 86.21($\pm$2.673)
& \cellcolor{ourscol}86.87($\pm$2.023) \\
\bottomrule
\end{tabular}
}}

\vspace{5pt}

\begin{minipage}{1.0\textwidth}
\scriptsize
\textit{Notes. Values are mean($\pm$std) over three folds. BA, balanced accuracy; mF1, macro F1-score; AUC, macro one-vs-rest ROC AUC. Scratch uses random initialization; SwinUNETR$^{\ast}$, VoCo$^{\ast}$, and S3D$^{\ast}$ use publicly released centralized foundation model weights; S3D uses centralized pretraining on our pretraining corpus, FedMAE and BrainFedFM use federated pretraining. 
Fine-tuning uses 25 epochs of AdamW, learning rate $1\times10^{-6}$, weight decay $1\times10^{-2}$, cosine scheduling with a 3-epoch warm-up, and label smoothing of 0.2. The best checkpoint is selected according to validation AUC.}
\end{minipage}
\end{table*}

\clearpage

\noindent{\scriptsize\bfseries Panel B. Labelled-data scaling and aggregate metric summaries.}
\vspace{2pt}
{\tiny\raggedright
Summarizes method behavior across labelled-data settings.
\textbf{B1}, fold-averaged trends for classification and calibration metrics; \textbf{B2}, extended metric heatmaps separating higher-is-better classification metrics from lower-is-better calibration/uncertainty metrics; \textbf{B3}, the accuracy--calibration trade-off, where points closer to the upper-left indicate stronger discrimination with better calibration.\par}
\vspace{-3pt}

\begin{center}
\includegraphics[width=1.0\textwidth,keepaspectratio]{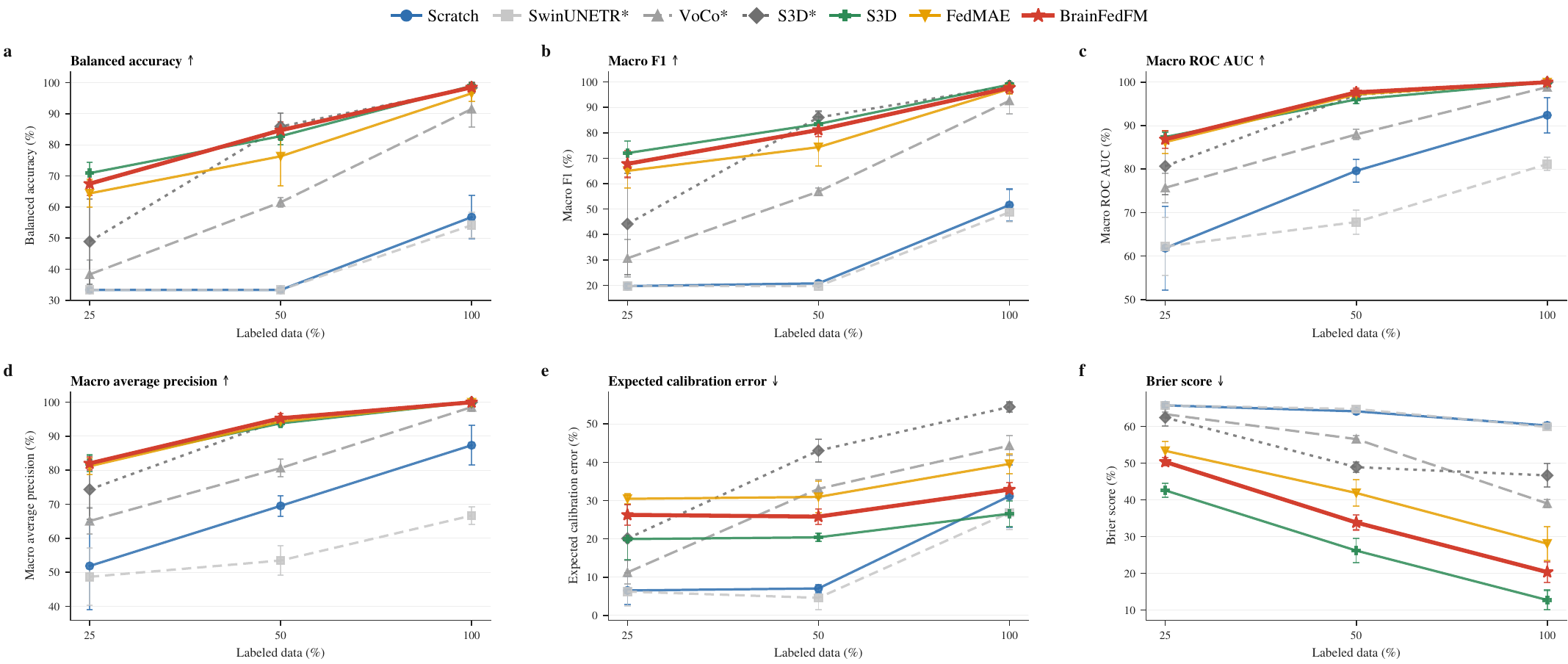} \par\vspace{3pt}
\includegraphics[width=1.0\textwidth,keepaspectratio]{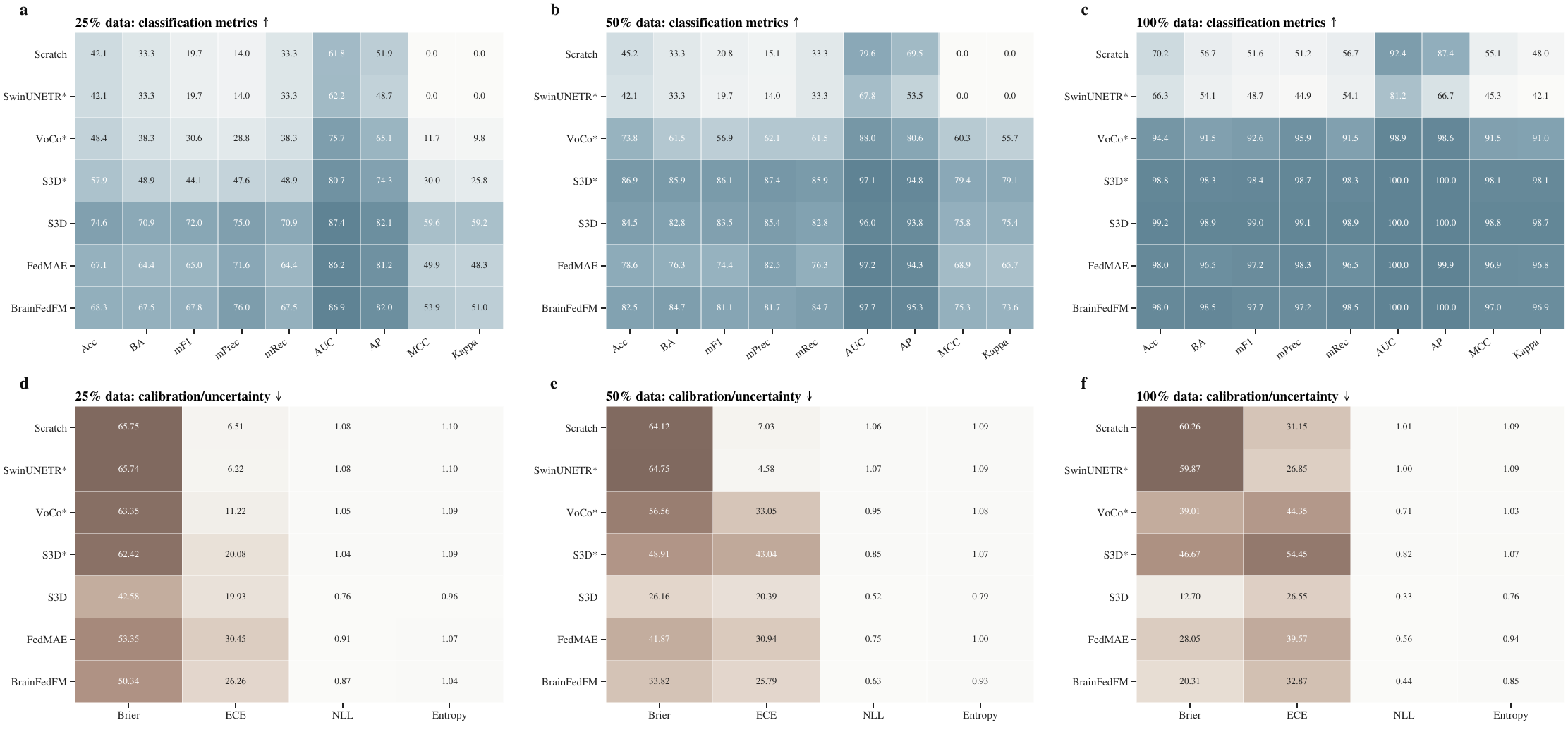} \par\vspace{5pt}
\includegraphics[width=0.935\textwidth,keepaspectratio]{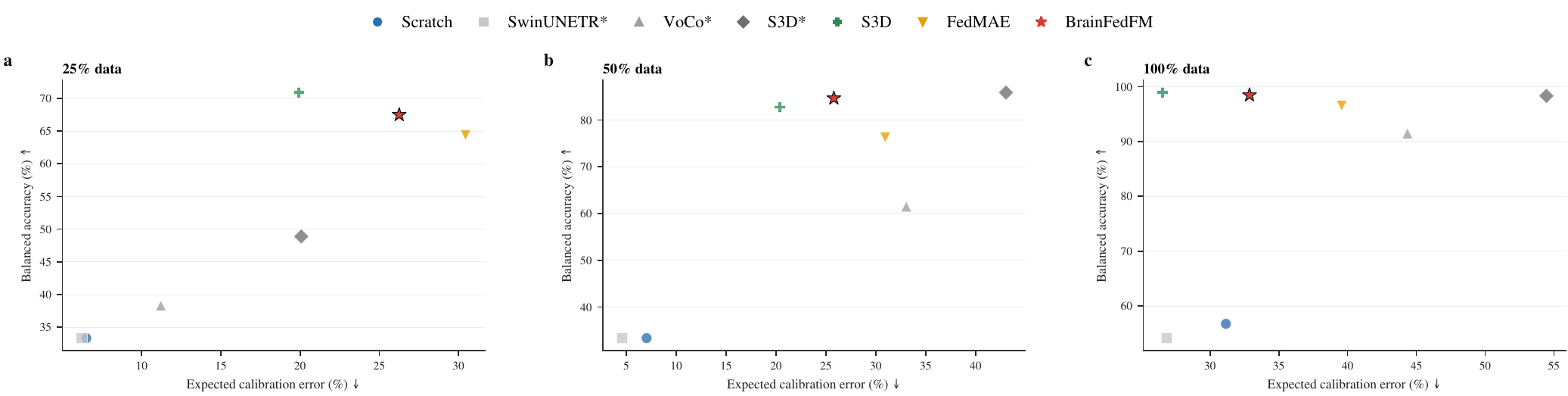}
\end{center}

\vspace{-2pt}

\noindent{\scriptsize\bfseries Panel C. Fold stability and full-data performance analyses.}
\vspace{2pt}
{\tiny\raggedright
Examines robustness and error modes beyond the aggregate scores. \textbf{C1}, fold-to-fold variability and average method ranking across balanced accuracy, macro F1, and macro ROC AUC; \textbf{C2}, probability-level analysis at 100\% labelled data, including micro-averaged ROC/precision--recall curves, reliability curves, and calibration scores; \textbf{C3}, row-normalized confusion matrices at 100\% labelled data, with rows denoting true disease classes and columns denoting predicted classes.\par}

\clearpage

\begin{center}
\includegraphics[width=0.838\textwidth,keepaspectratio]{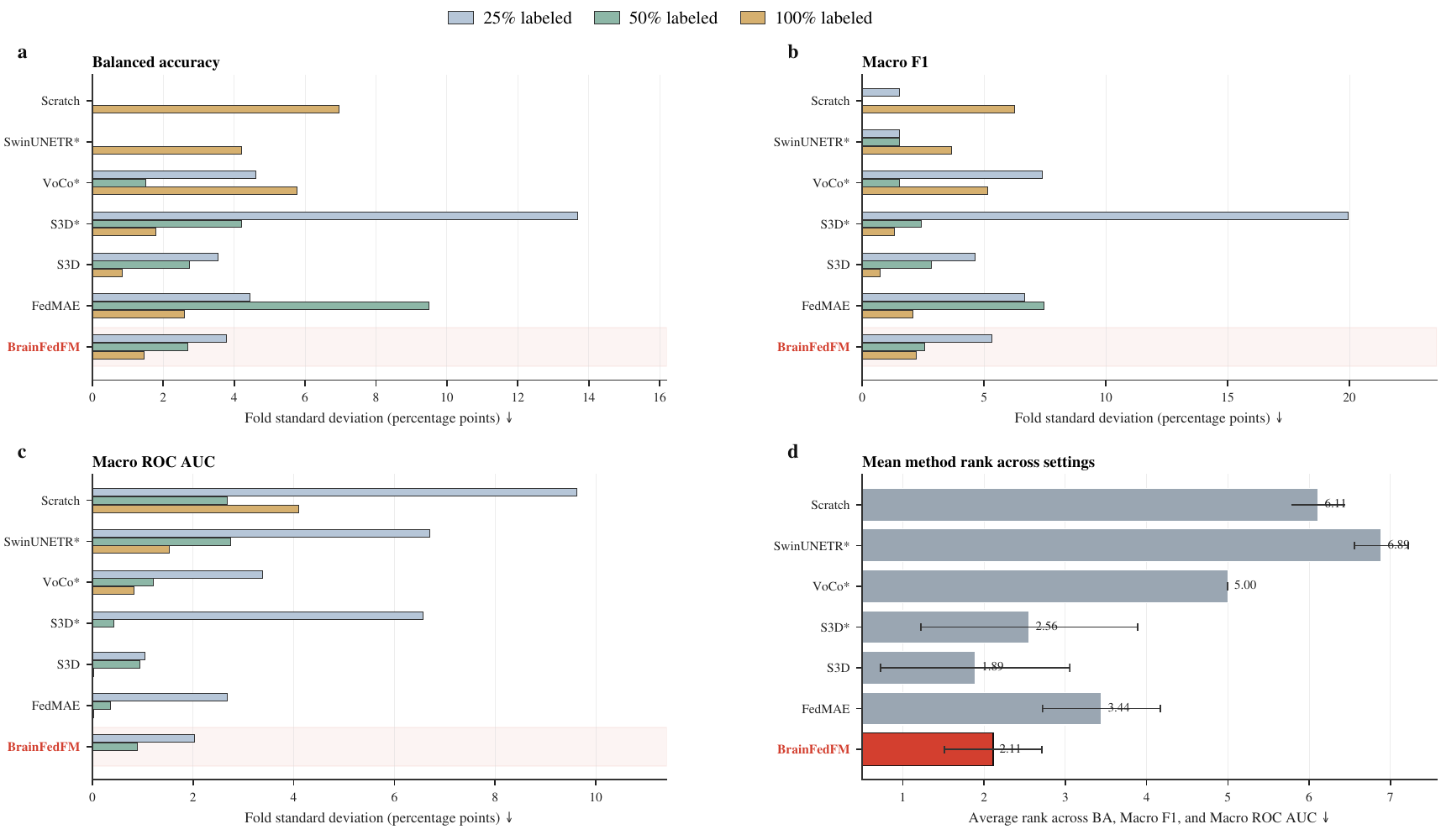}   \par\vspace{5pt}
\includegraphics[width=0.785\textwidth,keepaspectratio]{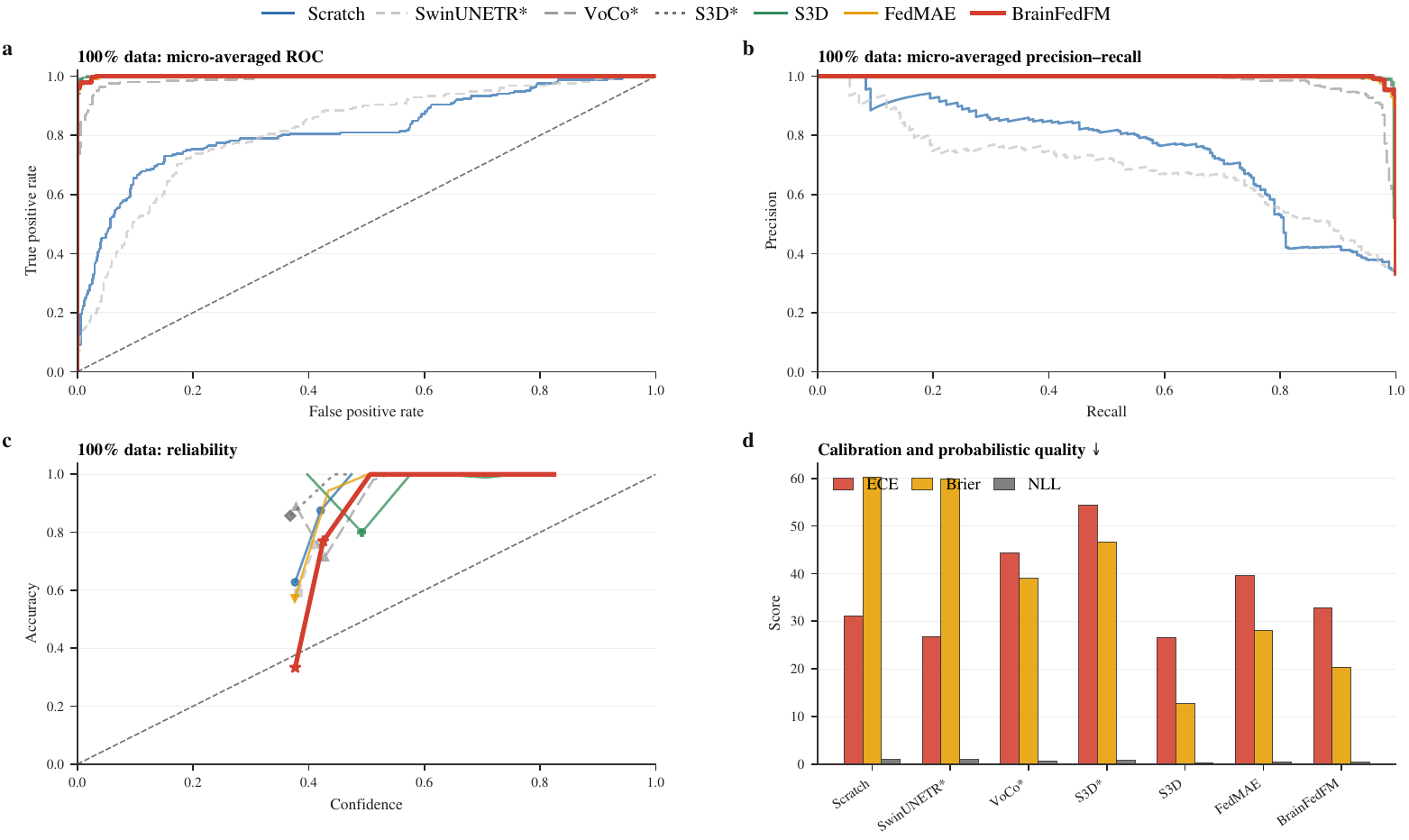}  \par\vspace{4pt}
\includegraphics[width=0.83\textwidth,keepaspectratio]{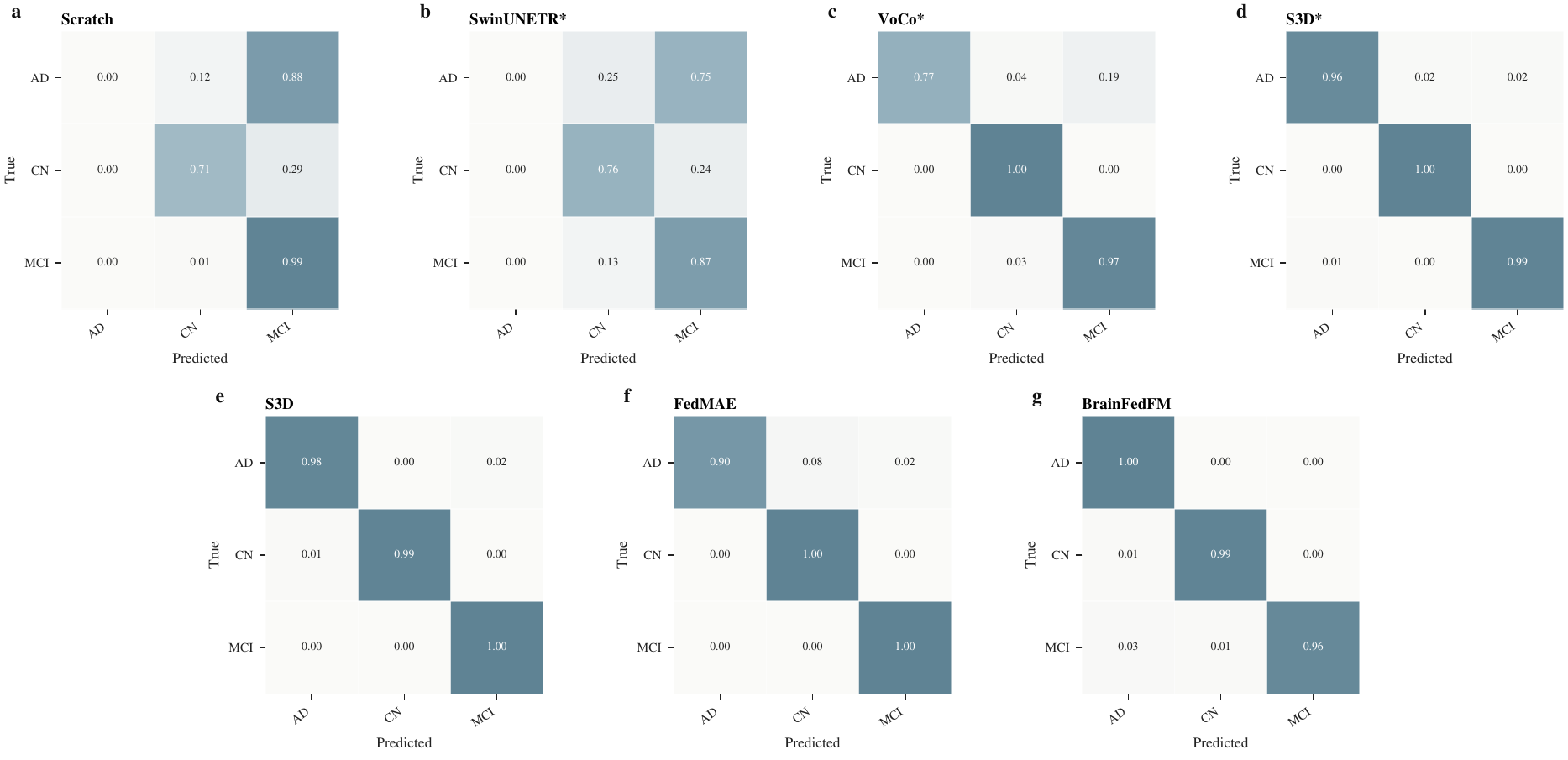}
\end{center}

\definecolor{pct100}{gray}{0.97}
\definecolor{pct75}{gray}{1.0}
\definecolor{pct50}{gray}{0.99}
\definecolor{pct25}{gray}{1.0}
\definecolor{overallgray}{gray}{1.0}
\definecolor{ourscol}{RGB}{244,248,252}

\begin{table*}[t]
\centering
\tiny
\caption[Focal epilepsy diagnosis task]{\textbf{Focal epilepsy diagnosis task.}
We evaluate foundation model performance on the IDEAS benchmark. This task aims to classify brain MRI scans into focal epilepsy and healthy control categories.
Accurate MRI-based classification of focal epilepsy is important for identifying disease-associated structural patterns, supporting clinical assessment, and enabling reliable patient stratification for downstream analysis.
The table summarizes performance under 25\%, 50\%, and 100\% labelled-data settings, which indicate the proportion of the labelled training--validation pool used for downstream fine-tuning and model selection; the held-out test set remains unchanged across all settings. We compare a randomly initialized scratch baseline, centralized pretrained foundation models, and federated pretrained foundation models.}
\label{tab:result_ideas}

\renewcommand{\arraystretch}{3.85}
{\scriptsize\bfseries Panel A. Summary under different labelled-data percentages.}
\vspace{5pt}

{\setlength{\tabcolsep}{3.38pt}

\resizebox{1.0\textwidth}{!}{
\begin{tabular}{
l c c c c c c
>{\columncolor{ourscol}\centering\arraybackslash}p{1.62cm}
}
\toprule
\multicolumn{2}{c}{\textbf{Brain sMRI FMs }}
& \multicolumn{4}{c}{Centralized Pretraining}
& \multicolumn{2}{c}{Federated Pretraining} \\
\cmidrule(lr){1-2}\cmidrule(lr){3-6}\cmidrule(lr){7-8}
\makecell[c]{Metric/Methods}
& Scratch
& \makecell[c]{SwinUNETR$^{\ast}$}  &\ VoCo$^{\ast}$ \ & \ S3D$^{\ast}$  & \ S3D
& \makecell[c]{FedMAE} & \makecell[c]{BrainFedFM} \\
\midrule

\rowcolor{pct100}
\textbf{100\%  BA $\uparrow$}
&  58.85($\pm$6.293)  &  50.00($\pm$0.000) & 61.46($\pm$0.737) & 61.46($\pm$8.495) &  68.75($\pm$1.276) &  66.15($\pm$0.737)
& \cellcolor{ourscol} 68.75($\pm$1.276) \\
\rowcolor{pct100}
\textbf{100\%  mF1 $\uparrow$}
& 53.26($\pm$14.10) & 34.20($\pm$1.219) & 60.24($\pm$1.323) & 55.76($\pm$16.09) & 68.73($\pm$1.276) & 66.03($\pm$0.766)
& \cellcolor{ourscol} 68.71($\pm$1.302) \\
\rowcolor{pct100}
\textbf{100\%  AUC $\uparrow$}
& 60.90($\pm$0.931) & 62.37($\pm$2.049) & 66.18($\pm$1.325) & 62.89($\pm$3.108) & 72.69($\pm$1.427) & 71.26($\pm$1.272)
& \cellcolor{ourscol} 71.52($\pm$0.924) \\
\midrule

\rowcolor{pct50}
\textbf{50\%  BA $\uparrow$}
&  51.56($\pm$2.210) & 50.00($\pm$0.000) &  52.08($\pm$4.101) & 65.10($\pm$1.949) & 66.15($\pm$1.949) & 63.02($\pm$3.211)
& \cellcolor{ourscol} 68.23($\pm$1.473) \\
\rowcolor{pct50}
\textbf{50\%  mF1 $\uparrow$}
& 38.63($\pm$7.489) & 33.33($\pm$0.000) & 40.00($\pm$8.203)
 & 64.78($\pm$2.161) &  65.86($\pm$2.076) & 62.30($\pm$3.561) & \cellcolor{ourscol} 68.18($\pm$1.504) \\
\rowcolor{pct50}
\textbf{50\%  AUC $\uparrow$}
& 62.43($\pm$1.563) & 54.92($\pm$5.814) & 63.28($\pm$3.317) & 65.14($\pm$2.417) & 66.99($\pm$1.150) & 68.72($\pm$4.392)
& \cellcolor{ourscol} 70.15($\pm$2.543) \\
\midrule

\rowcolor{pct25}
\textbf{25\%  BA $\uparrow$}
& 50.00($\pm$0.000) & 51.04($\pm$1.473) & 50.00($\pm$0.000) & 55.21($\pm$7.366)  & 55.73($\pm$8.102) & 56.25($\pm$4.600)
& \cellcolor{ourscol} 59.38($\pm$5.846) \\
\rowcolor{pct25}
\textbf{25\%  mF1 $\uparrow$}
& 33.33($\pm$0.000) & 39.92($\pm$9.308) & 33.33($\pm$0.000) & 45.61($\pm$14.08) & 44.59($\pm$15.93) & 47.09($\pm$10.64)
& \cellcolor{ourscol} 55.05($\pm$8.944) \\
\rowcolor{pct25}
\textbf{25\%  AUC $\uparrow$}
& 55.76($\pm$7.927) & 58.89($\pm$6.766) & 64.75($\pm$9.715) & 65.46($\pm$1.325) & 63.05($\pm$3.207) & 70.25($\pm$3.167)
& \cellcolor{ourscol} 68.98($\pm$3.910) \\
\bottomrule
\end{tabular}
}}

\vspace{5pt}

\begin{minipage}{1.0\textwidth}
\scriptsize
\textit{Notes. Values are mean($\pm$std) over three folds. BA, balanced accuracy; mF1, macro F1-score; AUC, macro one-vs-rest ROC AUC. Scratch uses random initialization; SwinUNETR$^{\ast}$, VoCo$^{\ast}$, and S3D$^{\ast}$ use publicly released centralized foundation model weights; S3D uses centralized pretraining on our pretraining corpus, FedMAE and BrainFedFM use federated pretraining.
Fine-tuning uses 8 epochs of AdamW, learning rate $1\times10^{-6}$, weight decay $1\times10^{-2}$, cosine scheduling with a 1-epoch warm-up, and label smoothing of 0.2. The best checkpoint is selected according to validation AUC.}
\end{minipage}
\end{table*}

\clearpage

\noindent{\scriptsize\bfseries Panel B. Labelled-data scaling and aggregate metric summaries.}
\vspace{2pt}
{\tiny\raggedright
Summarizes method behavior across labelled-data settings.
\textbf{B1}, fold-averaged trends for classification and calibration metrics; \textbf{B2}, extended metric heatmaps separating higher-is-better classification metrics from lower-is-better calibration/uncertainty metrics; \textbf{B3}, the accuracy--calibration trade-off, where points closer to the upper-left indicate stronger discrimination with better calibration.\par}
\vspace{-3pt}

\begin{center}
\includegraphics[width=1.0\textwidth,keepaspectratio]{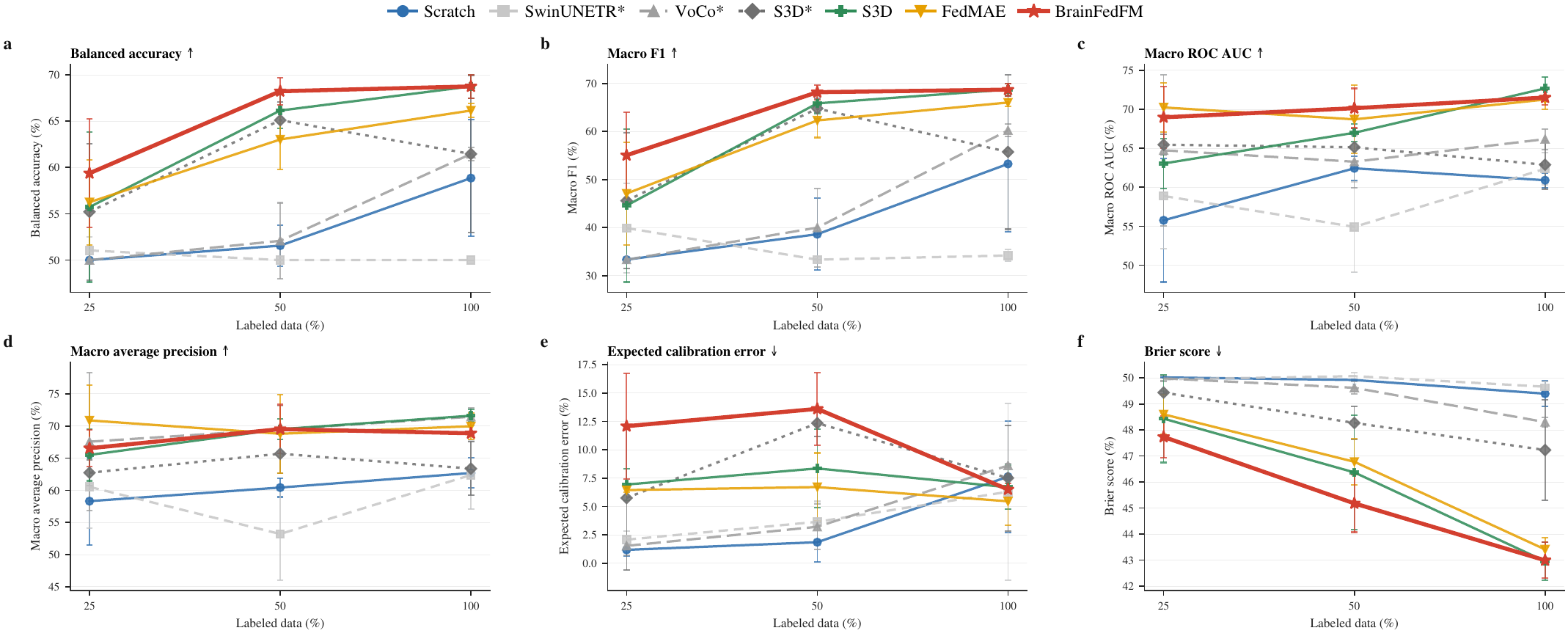} \par\vspace{3pt}
\includegraphics[width=1.0\textwidth,keepaspectratio]{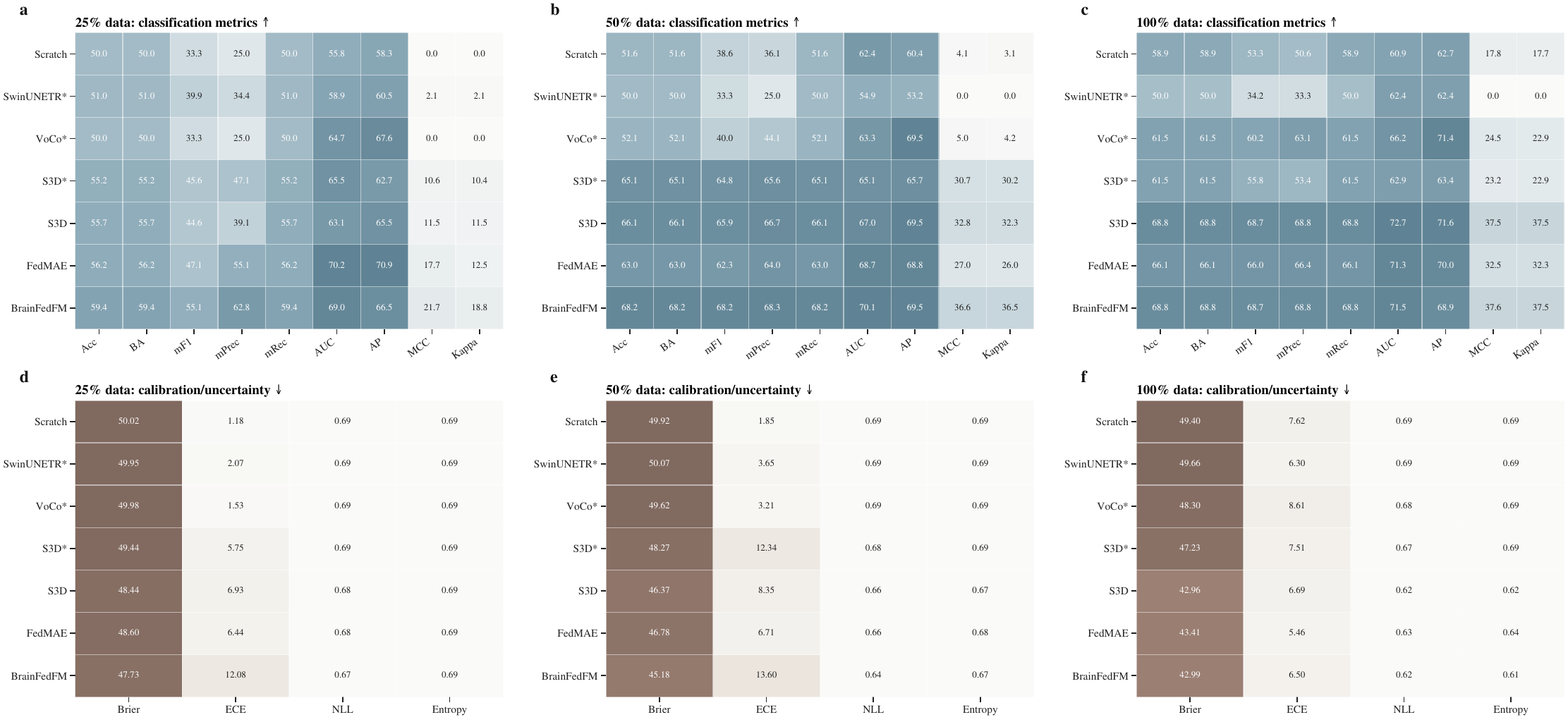} \par\vspace{5pt}
\includegraphics[width=0.935\textwidth,keepaspectratio]{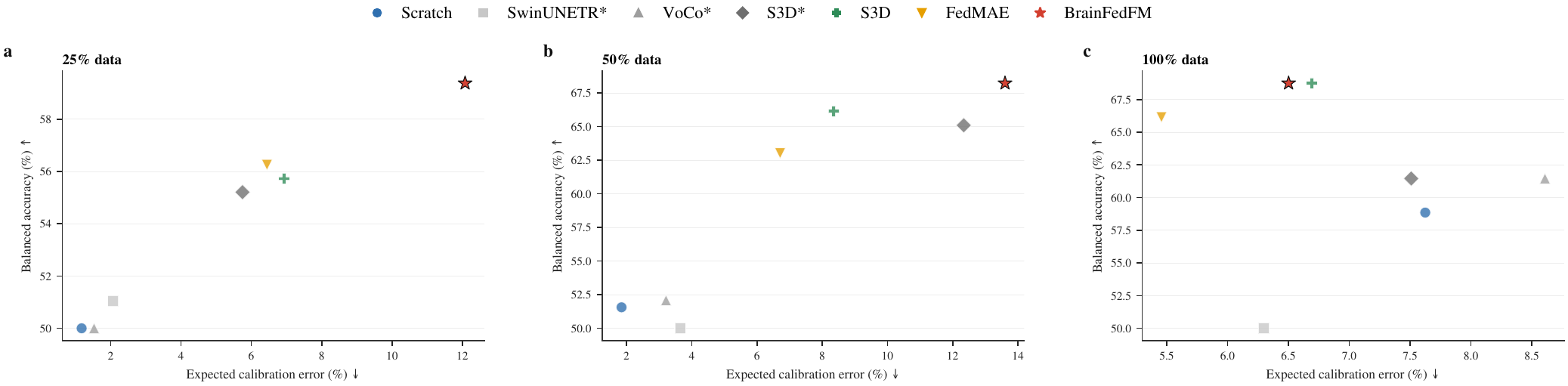}
\end{center}

\vspace{-2pt}

\noindent{\scriptsize\bfseries Panel C. Fold stability and full-data performance analyses.}
\vspace{2pt}
{\tiny\raggedright
Examines robustness and error modes beyond the aggregate scores. \textbf{C1}, fold-to-fold variability and average method ranking across balanced accuracy, macro F1, and macro ROC AUC; \textbf{C2}, probability-level analysis at 100\% labelled data, including micro-averaged ROC/precision--recall curves, reliability curves, and calibration scores; \textbf{C3}, row-normalized confusion matrices at 100\% labelled data, with rows denoting true focal epilepsy or healthy control classes and columns denoting predicted classes.\par}

\clearpage

\begin{center}
\includegraphics[width=0.84\textwidth,keepaspectratio]{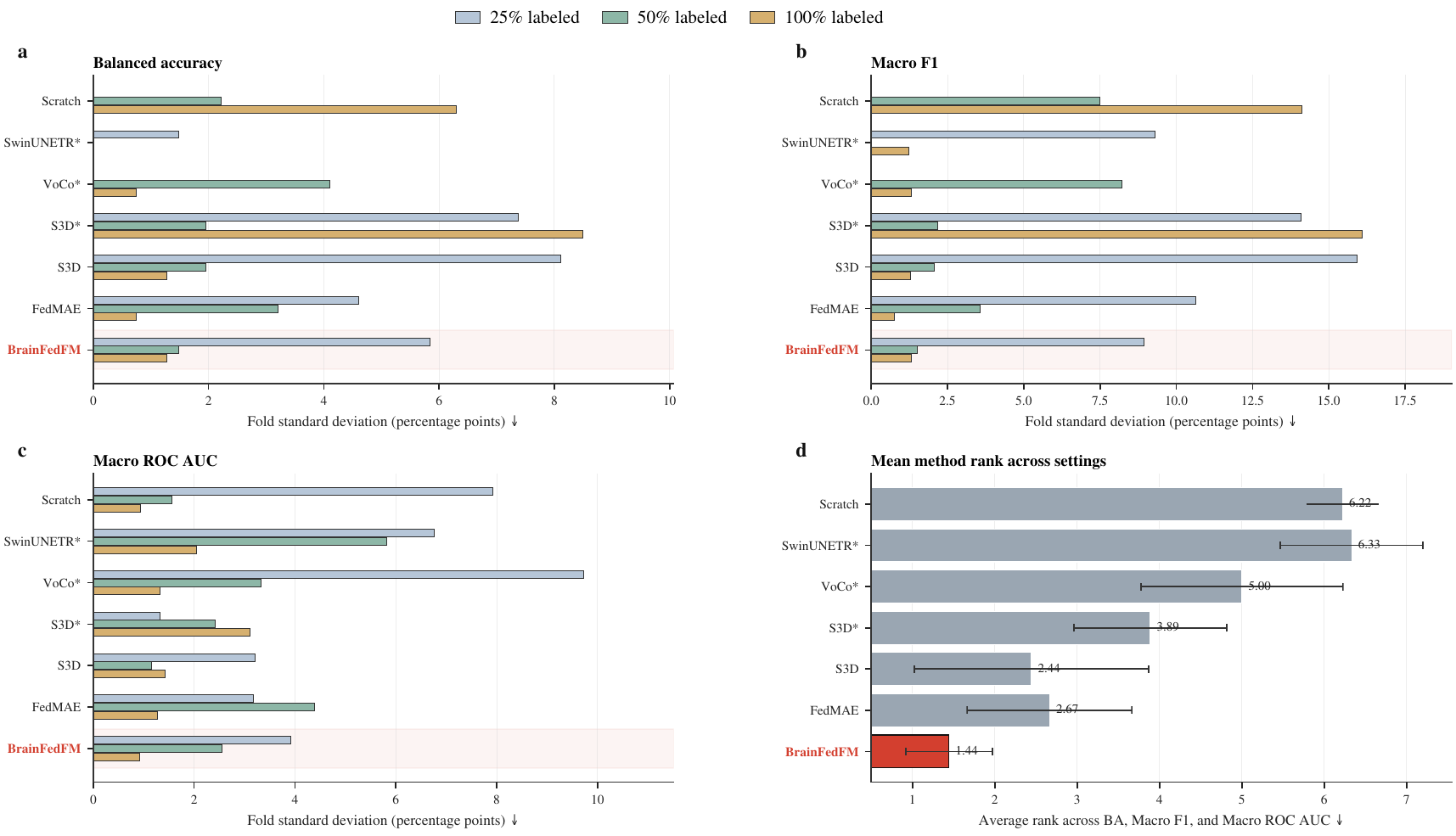} \par\vspace{5pt}
\includegraphics[width=0.7889\textwidth,keepaspectratio]{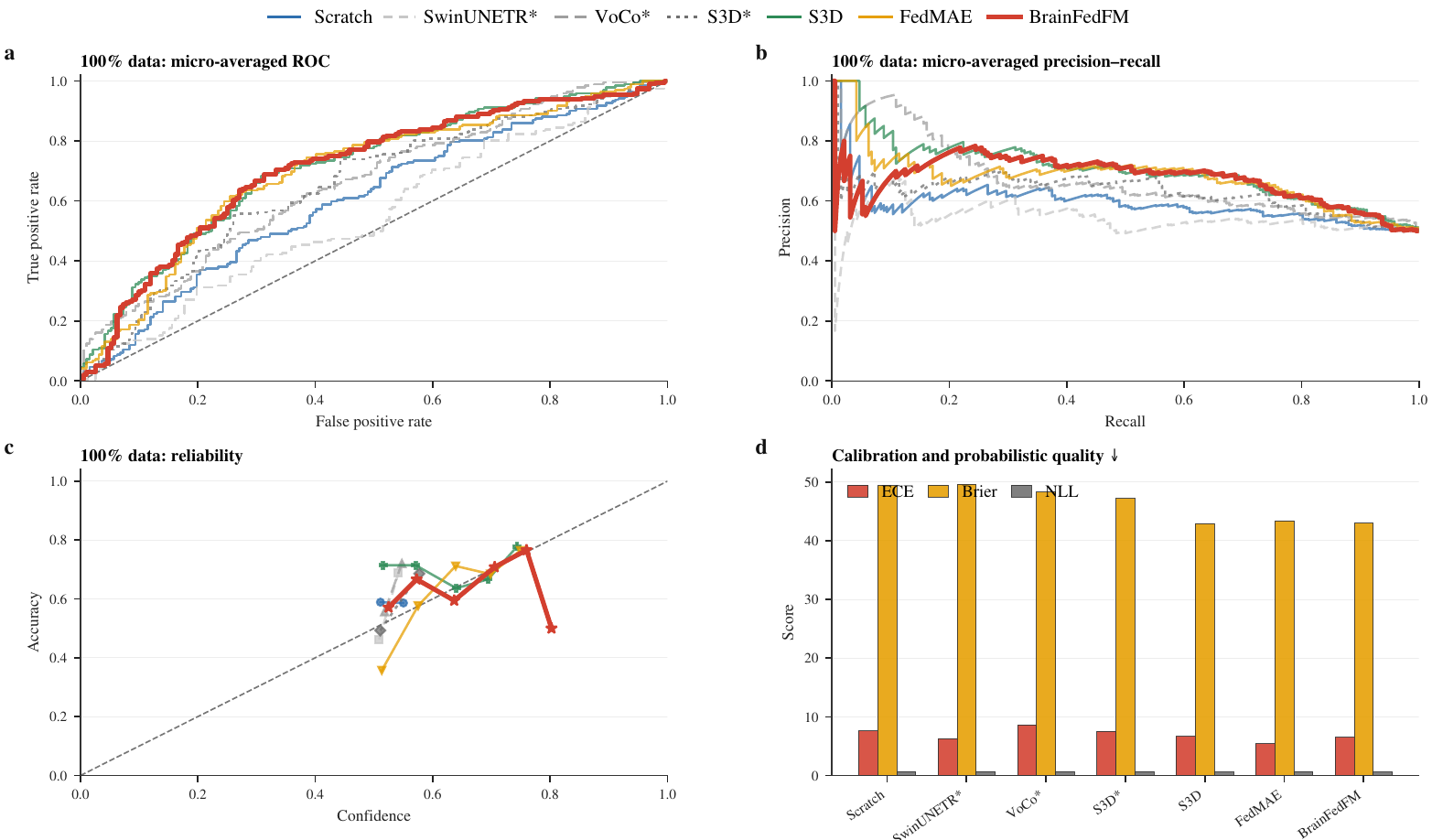} \par\vspace{4pt}
\includegraphics[width=0.845\textwidth,keepaspectratio]{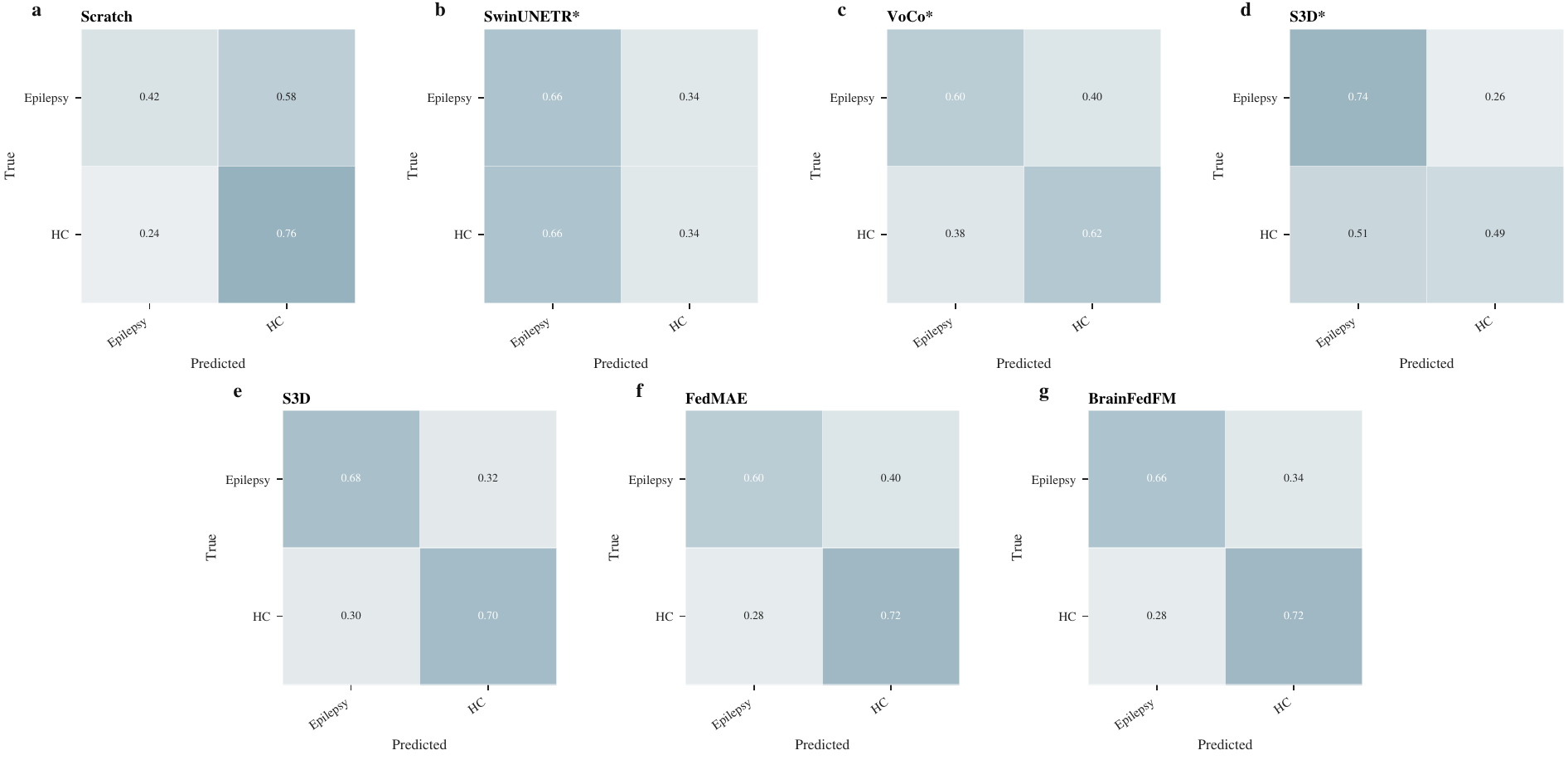}
\end{center}

\definecolor{pct100}{gray}{0.97}
\definecolor{pct75}{gray}{1.0}
\definecolor{pct50}{gray}{0.99}
\definecolor{pct25}{gray}{1.0}
\definecolor{overallgray}{gray}{0.97}
\definecolor{ourscol}{RGB}{244,248,252}

\begin{table*}[h]
\centering
\tiny
\caption[Brain tumor segmentation task]{\textbf{Brain tumor segmentation task.}
We evaluate foundation model performance on the MSD-Brain benchmark, where the goal is to delineate tumor core, edema, and enhancing tumor from multi-sequence brain MRI. Accurate tumor-subregion segmentation enables quantitative assessment of tumor burden and intra-tumoral heterogeneity, and supports treatment planning, longitudinal monitoring, and response evaluation. Panel~A reports overall and subregion-wise Dice, intersection-over-union (IoU), and normalized surface Dice (NSD) with 100\% labelled training data, whereas Panel~B reports the corresponding overall results with 25\% and 50\% labelled data. We compare a randomly initialized scratch baseline, centralized pretrained foundation models, and federated pretrained foundation models.}
\label{tab:result_msd}

\par\vspace{-2.5pt}

\renewcommand{\arraystretch}{2.53}
{\scriptsize\bfseries Panel A. Detailed results under 100\% labelled data.}
\vspace{5pt}

{\setlength{\tabcolsep}{3.38pt}

\resizebox{1.0\textwidth}{!}{
\begin{tabular}{
l c c c c c c
>
{\columncolor{ourscol}\centering\arraybackslash}p{1.58cm}
}
\toprule
\multicolumn{2}{c}{\textbf{Brain sMRI FMs }}
& \multicolumn{4}{c}{Centralized OpenMind}
& \multicolumn{2}{c}{Federated FMs} \\
\cmidrule(lr){1-2}\cmidrule(lr){3-6}\cmidrule(lr){7-8}
\makecell[c]{Metric/Methods}
& Scratch
& \makecell[c]{SwinUNETR$^{\ast}$}  &\ VoCo$^{\ast}$ \ & \ S3D$^{\ast}$ & \ S3D
& \makecell[c]{FedMAE} & \makecell[c]{BrainFedFM} \\
\midrule

\rowcolor{overallgray}
\textbf{Overall Dice $\uparrow$}
& 70.44($\pm$0.844) &  71.91($\pm$0.463)   &  71.80($\pm$0.207) & 71.62($\pm$0.525) &  71.04($\pm$0.279)   & 71.12($\pm$0.321) 
& \cellcolor{ourscol}71.17($\pm$0.235)  \\
\rowcolor{overallgray}
\textbf{Overall IoU $\uparrow$}
&  58.19($\pm$0.954) & 59.92($\pm$0.426)  & 59.88($\pm$0.219) & 59.68($\pm$0.511)  &  59.08($\pm$0.325)  & 59.12($\pm$0.331)
& \cellcolor{ourscol} 59.20($\pm$0.250)  \\
\rowcolor{overallgray}
\textbf{Overall NSD $\uparrow$}
&  75.26($\pm$1.153)  & 76.97($\pm$0.666) & 76.88($\pm$0.308) & 76.29($\pm$1.037) & 75.89($\pm$0.204)  & 76.06($\pm$0.499)
& \cellcolor{ourscol}76.10($\pm$0.223) \\
\midrule

Tumor Core (Dice) & 78.21($\pm$1.164) & 79.84($\pm$0.446)  & 80.07($\pm$0.287) & 80.07($\pm$0.263) & 79.84($\pm$0.150)  & 79.69($\pm$0.249)  & 79.47($\pm$0.375) \\
Tumor Core (IoU)  & 65.58($\pm$1.394) & 67.60($\pm$0.552)  & 67.92($\pm$0.337) & 67.90($\pm$0.337)  &  67.60($\pm$0.184)   &  67.43($\pm$0.313)  & 67.16($\pm$0.425)  \\
Tumor Core (NSD)  & 76.77($\pm$1.506) & 78.92($\pm$0.889)  & 79.62($\pm$0.203) & 79.42($\pm$0.707)  & 78.86($\pm$0.240)  & 78.77($\pm$0.260)  & 78.95($\pm$0.198)  \\

Edema & 56.41($\pm$0.750) & 58.40($\pm$1.217)  & 57.89($\pm$0.469) & 57.30($\pm$1.066) &  56.51($\pm$0.415) & 56.99($\pm$0.348)  & 57.32($\pm$0.334) \\
Edema  & 42.88($\pm$0.707) & 44.90($\pm$0.948)  & 44.63($\pm$0.417) & 44.01($\pm$0.900) & 43.24($\pm$0.313) & 43.65($\pm$0.269)  & 44.01($\pm$0.318)  \\
Edema  & 65.28($\pm$1.570) & 68.11($\pm$1.795)  & 66.92($\pm$1.337) & 65.43($\pm$1.905) & 65.68($\pm$0.745)  & 66.41($\pm$0.887)  & 66.27($\pm$0.690)  \\

Enhancing Tumor& 76.69($\pm$0.970) & 77.48($\pm$0.737)  & 77.43($\pm$0.605) & 77.48($\pm$0.346)  & 76.77($\pm$0.491) & 76.68($\pm$0.451)  & 76.71($\pm$0.086) \\
Enhancing Tumor & 66.12($\pm$1.032) & 67.25($\pm$0.754)  & 67.09($\pm$0.713) & 67.14($\pm$0.458) & 66.40($\pm$0.592)  & 66.30($\pm$0.488)  & 66.42($\pm$0.127)  \\
Enhancing Tumor  & 83.73($\pm$0.978) & 83.88($\pm$1.065)  & 84.10($\pm$0.887) & 84.02($\pm$0.630) & 83.14($\pm$0.455) & 83.01($\pm$0.569)  & 83.07($\pm$0.281)  \\
\bottomrule
\end{tabular}
}}

\vspace{8pt}

{\scriptsize\bfseries Panel B. Summary under different labelled-data percentages.}
\vspace{5pt}

\renewcommand{\arraystretch}{2.012}
\resizebox{1.0\textwidth}{!}{%
\begin{tabular}{
l
ccc!{\hspace{2pt}\color{black!45}\vrule width 0.6pt\hspace{2pt}}
ccc
}
\toprule
\multicolumn{1}{c}{\textbf{Percentages}}
& \multicolumn{3}{c}{25\%}
& \multicolumn{3}{c}{50\%} \\
\cmidrule(lr){1-1}\cmidrule(lr){2-4}\cmidrule(lr){5-7}
\makecell[c]{Methods/Metric} & \multicolumn{1}{c}{Dice} & \multicolumn{1}{c}{IoU} & \multicolumn{1}{c}{NSD}
& \multicolumn{1}{c}{Dice} & \multicolumn{1}{c}{IoU} & \multicolumn{1}{c}{NSD} \\
\midrule
Scratch
& 66.47($\pm$0.734)  & 54.12($\pm$0.692) & 70.12($\pm$0.913)
& 68.37($\pm$0.535)  & 56.29($\pm$0.537)  & 72.02($\pm$1.300)  \\
SwinUNETR$^{\ast}$
& 67.16($\pm$0.927) &  55.03($\pm$0.843) &  70.31($\pm$1.789) 
&  70.04($\pm$0.503)  & 58.14($\pm$0.399)  &  74.01($\pm$1.334)  \\
VoCo$^{\ast}$
& 67.99($\pm$0.474)  &  55.64($\pm$0.518)  & 71.87($\pm$0.769) 
& 70.50($\pm$0.510) & 58.44($\pm$0.440)  &  74.85($\pm$1.260) \\
S3D$^{\ast}$
& 66.03($\pm$1.072)  & 53.83($\pm$0.972)  & 68.62($\pm$2.289)
&  69.37($\pm$0.070)  & 57.35($\pm$0.100) & 73.27($\pm$0.200)  \\
S3D
& 66.38($\pm$0.636) & 54.04($\pm$0.503)  & 69.29($\pm$1.241)
& 69.42($\pm$0.088)  & 57.35($\pm$0.191)  & 73.32($\pm$0.161)  \\
\hline
FedMAE
&  66.57($\pm$0.872)  & 54.32($\pm$0.822)  & 69.72($\pm$1.672) 
& 68.95($\pm$0.154)  & 56.81($\pm$0.355) & 73.07($\pm$0.417)  \\
\rowcolor{ourscol}
BrainFedFM
& 66.70($\pm$0.518) & 54.55($\pm$0.434)  &  69.74($\pm$0.762)
& 69.60($\pm$0.264)  & 57.66($\pm$0.312) & 73.53($\pm$0.755)  \\
\bottomrule
\end{tabular}
}
\vspace{3pt}

\begin{minipage}{1.0\textwidth}
\scriptsize
\textit{Notes. Dice, Dice similarity coefficient; IoU, intersection-over-union; NSD, normalized surface Dice. All models are fine-tuned for 150 epochs with 100 training iterations and 50 validation iterations per epoch using SGD with momentum 0.99, Nesterov acceleration, weight decay of $3\times10^{-5}$, and an initial learning rate of $1\times10^{-3}$. Training uses a 15-epoch linear warm-up followed by polynomial learning-rate decay, and test performance is reported from the checkpoint with the best validation result.}
\end{minipage}
\end{table*}

\clearpage

\noindent{\scriptsize\bfseries Panel C. Data efficiency, complementary profiles, and qualitative lesion delineation.}
\vspace{2pt}
{\tiny\raggedright
Examines annotation efficiency and complementary segmentation behavior beyond the tabulated overlap scores. \textbf{C1}, fold-averaged trajectories for Dice, NSD, voxel sensitivity, lesion F1, HD95, and absolute relative volume error; \textbf{C2}, direction-aligned profiles across overlap/surface, voxel/volume, and lesion-level metrics, together with a family-level summary; \textbf{C3}, representative axial MRI examples with reference lesions and three-fold ensemble error maps, where green, orange, and purple denote true-positive, false-positive, and false-negative lesion voxels, respectively.\par}
\par\vspace{-2pt}

\begin{center}
\includegraphics[width=0.974\textwidth,keepaspectratio]{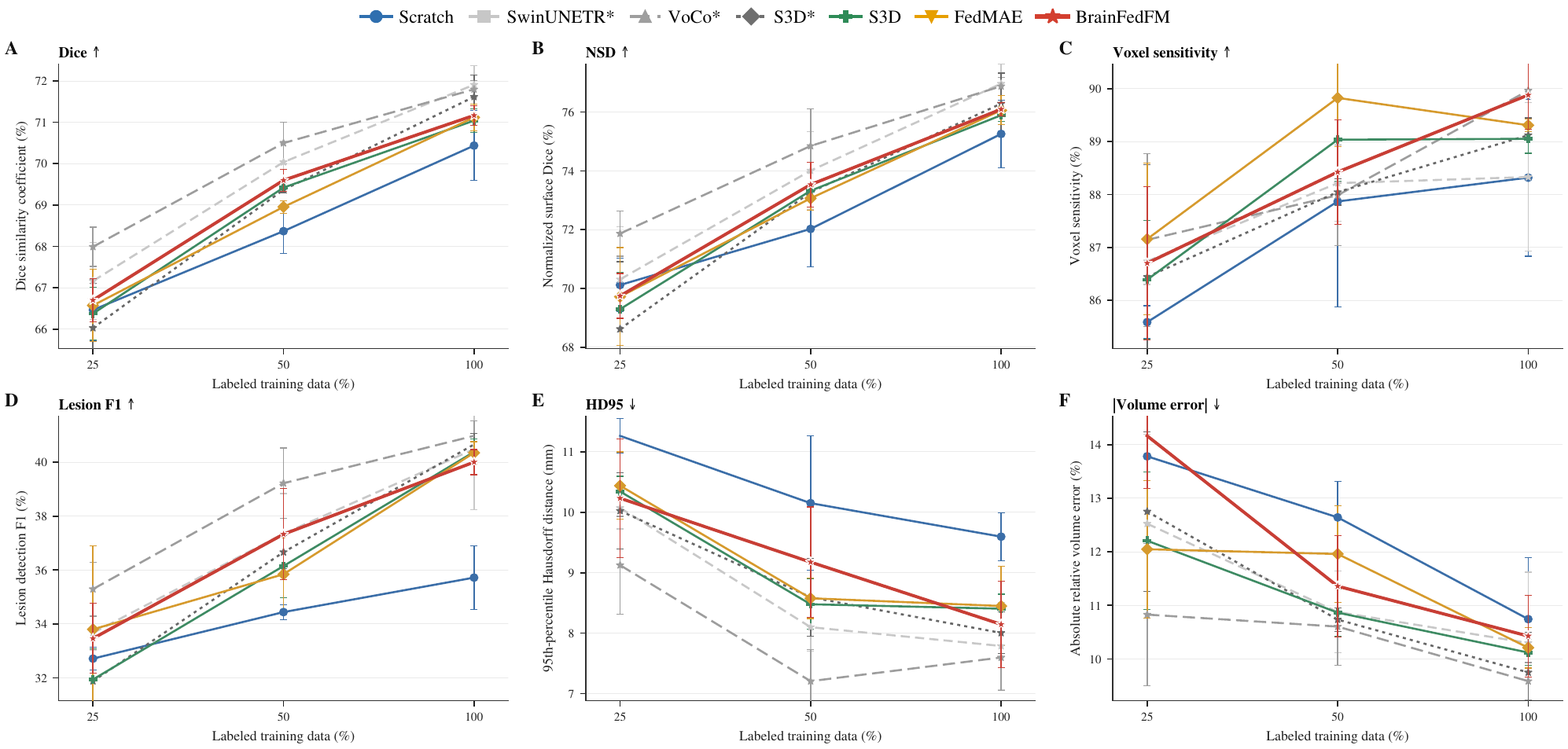}
\includegraphics[width=0.845\textwidth,keepaspectratio]{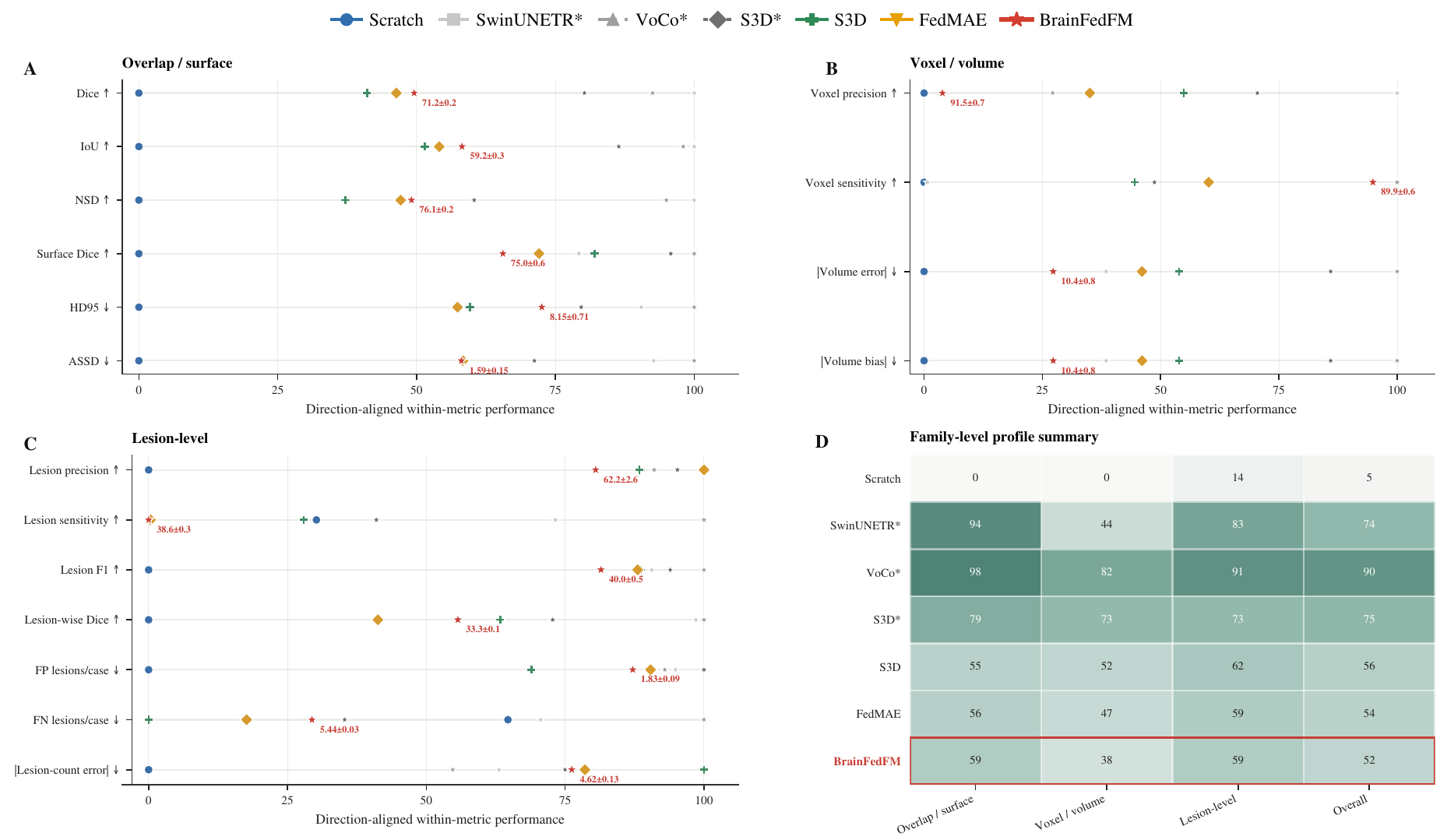}
\includegraphics[width=0.9888\textwidth,keepaspectratio]{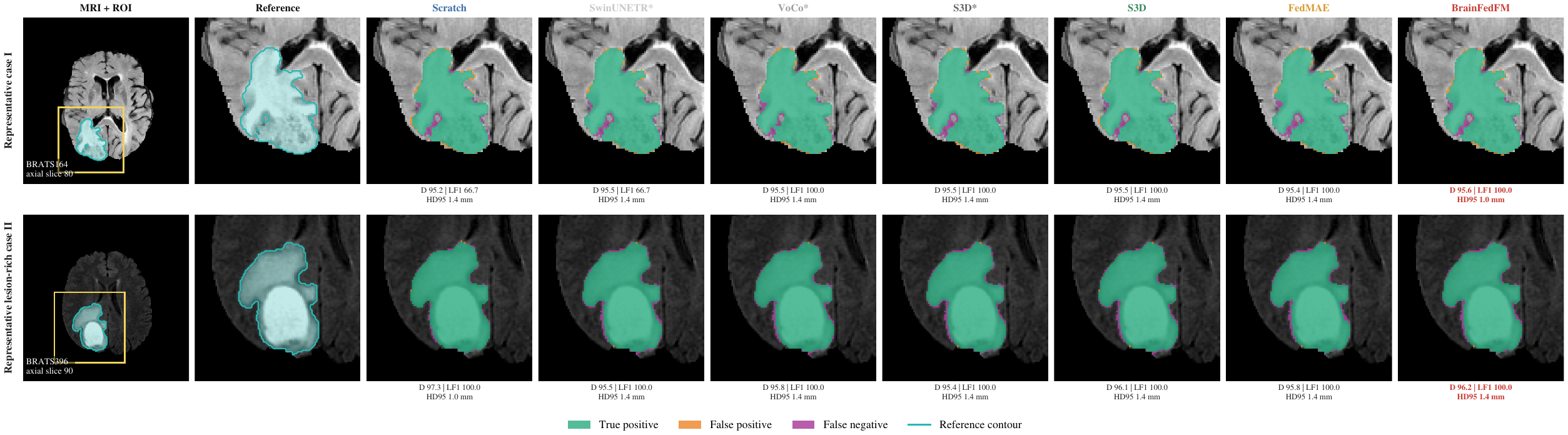}
\end{center}

\clearpage

\noindent{\scriptsize\bfseries Panel D. Error decomposition, case-level variability, and lesion-scale performance.}
\vspace{2pt}
{\tiny\raggedright
Characterizes segmentation failure modes and robustness beyond aggregate accuracy. \textbf{D1}, false-negative and false-positive voxel burden, signed volume bias, Dice--NSD performance, and voxel precision--sensitivity balance at 100\% labelled data; \textbf{D2}, matched case-by-fold distributions of HD95, average symmetric surface distance (ASSD), absolute relative volume error, and absolute lesion-count error; \textbf{D3}, lesion-detection sensitivity and lesion-wise Dice across reference-lesion volume quintiles, with dedicated summaries for the smallest lesions.\par}
\par\vspace{-3.5pt}

\begin{center}
\includegraphics[width=0.75\textwidth,keepaspectratio]{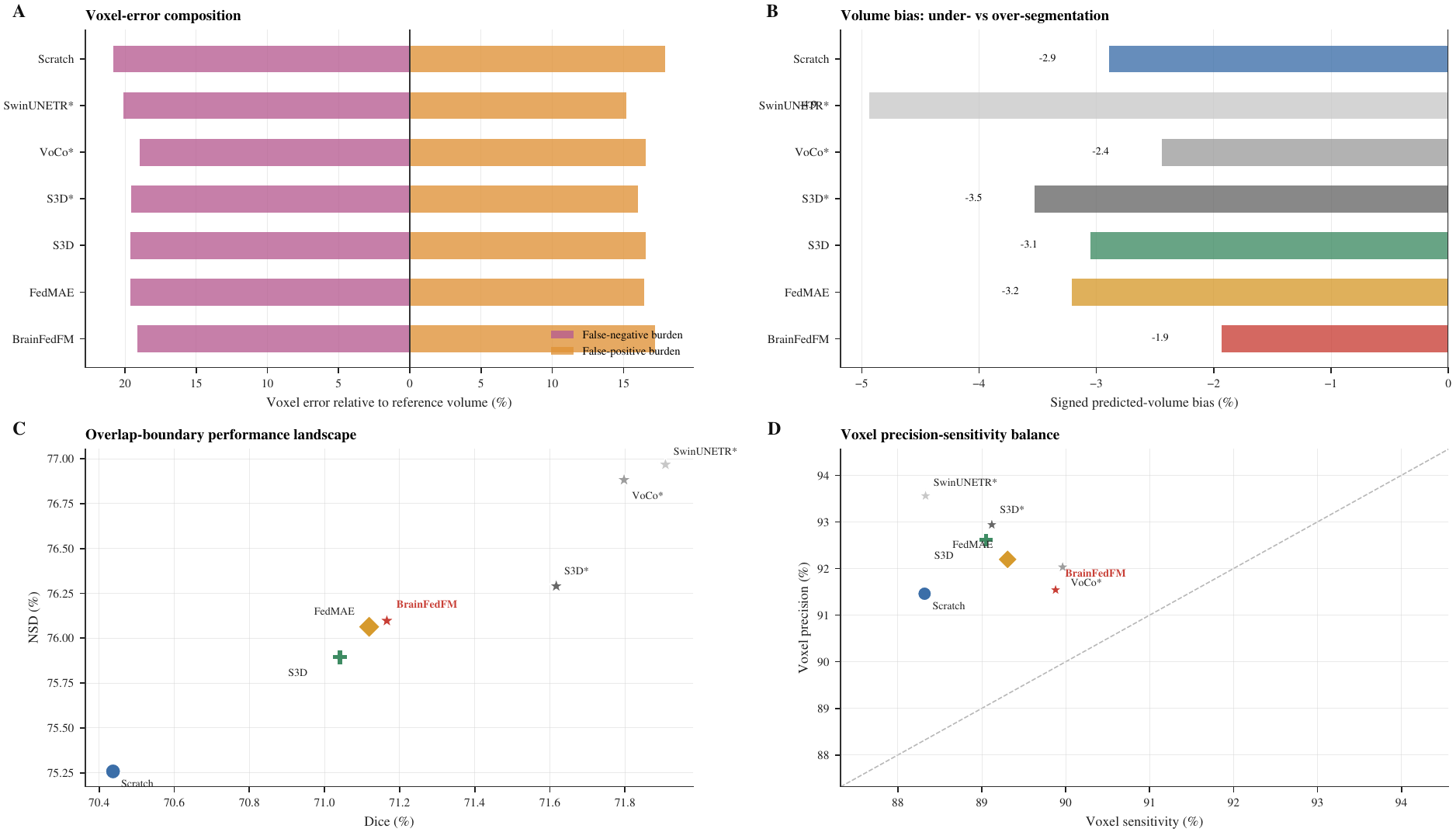}
\includegraphics[width=0.739\textwidth,keepaspectratio]{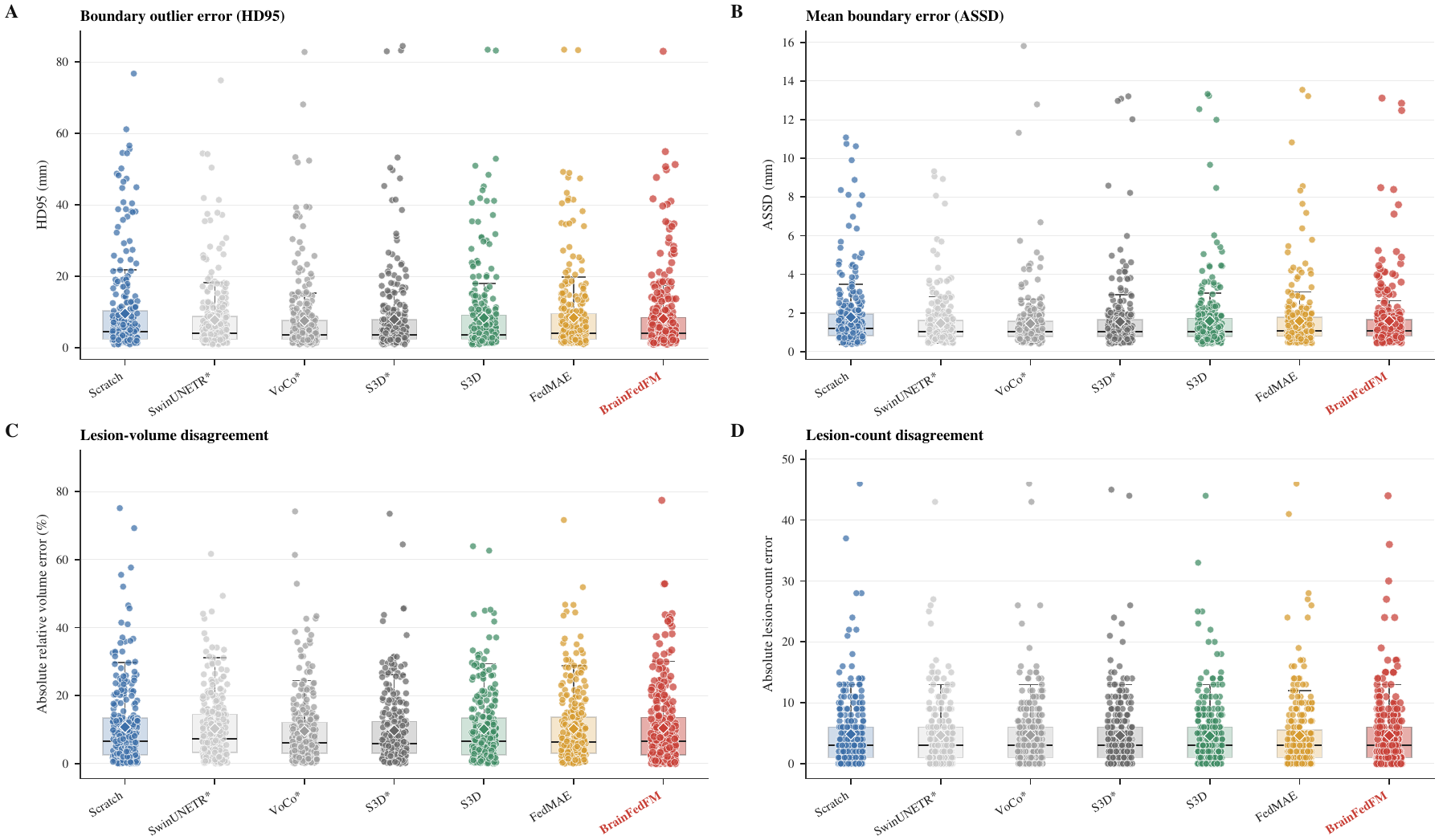}
\includegraphics[width=0.744\textwidth,keepaspectratio]{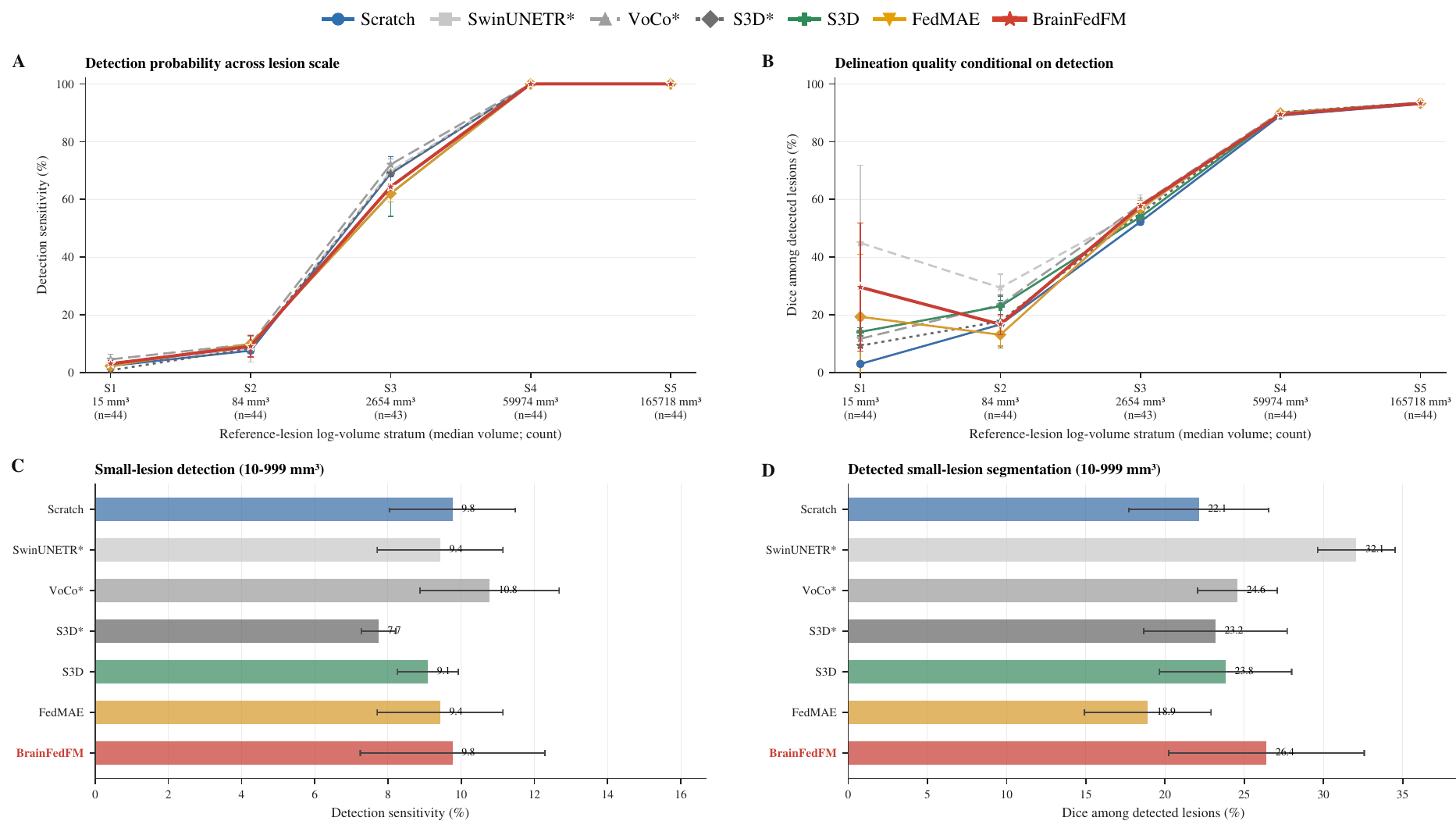}
\end{center}

\definecolor{pct100}{gray}{0.97}
\definecolor{pct75}{gray}{1.0}
\definecolor{pct50}{gray}{0.99}
\definecolor{pct25}{gray}{1.0}
\definecolor{overallgray}{gray}{1.0}
\definecolor{ourscol}{RGB}{244,248,252}

\begin{table*}[t]
\centering
\tiny
\caption[Disability severity estimation task]{\textbf{Disability severity estimation task.}
We evaluate foundation model performance on the MSLesSeg benchmark for estimating the Expanded Disability Status Scale (EDSS) score from structural brain MRI.
This task models MRI-visible disease burden associated with neurological disability in multiple sclerosis and provides a representative clinical regression setting. The table summarizes performance under 25\%, 50\%, and 100\% labelled-data settings, which indicate the proportion of the labelled training--validation pool used for downstream fine-tuning and model selection; the held-out test set remains unchanged across all settings.
We compare a randomly initialized scratch baseline, centralized pretrained foundation models, and federated pretrained foundation models.}
\label{tab:result_msles}

\renewcommand{\arraystretch}{3.8}
{\scriptsize\bfseries Panel A. Summary under different labelled-data percentages.}
\vspace{5pt}

{\setlength{\tabcolsep}{3.38pt}

\resizebox{1.0\textwidth}{!}{
\begin{tabular}{
l c c c c c c
>{\columncolor{ourscol}\centering\arraybackslash}p{1.62cm}
}
\toprule
\multicolumn{2}{c}{\textbf{Brain sMRI FMs }}
& \multicolumn{4}{c}{Centralized Pretraining}
& \multicolumn{2}{c}{Federated Pretraining} \\
\cmidrule(lr){1-2}\cmidrule(lr){3-6}\cmidrule(lr){7-8}
\makecell[c]{Metric/Methods}
& Scratch
& \makecell[c]{SwinUNETR$^{\ast}$}  &\ VoCo$^{\ast}$ \ & \ S3D$^{\ast}$  & \ S3D 
& \makecell[c]{FedMAE} & \makecell[c]{BrainFedFM} \\
\midrule

\rowcolor{pct100}
\textbf{100\%  RMSE $\downarrow$}
& 1.241($\pm$0.012) & 1.313($\pm$0.006) & 1.304($\pm$0.029) &   1.518($\pm$0.032) & 0.984($\pm$0.003) & 1.075($\pm$0.038)
& \cellcolor{ourscol}0.999($\pm$0.011) \\
\rowcolor{pct100}
\textbf{100\%   MAE $\downarrow$}
& 1.097($\pm$0.014) & 1.146($\pm$0.004)  & 1.140($\pm$0.016) &  1.291($\pm$0.020) & 0.816($\pm$0.008)  & 0.901($\pm$0.022)
& \cellcolor{ourscol}0.848($\pm$0.005) \\
\rowcolor{pct100}
\textbf{100\%   R$^2$ $\uparrow$}
&  -0.018($\pm$0.020) &  -0.140($\pm$0.010)  &  -0.125($\pm$0.050) & -0.524($\pm$0.064) &  0.360($\pm$0.004)  &  0.234($\pm$0.054)
& \cellcolor{ourscol}0.340($\pm$0.015) \\
\midrule

\rowcolor{pct50}
\textbf{50\%  RMSE $\downarrow$}
& 1.350($\pm$0.030) & 1.583($\pm$0.026)  & 1.443($\pm$0.009) & 1.513($\pm$0.039) &  1.234($\pm$0.101)  &  1.204($\pm$0.023)
& \cellcolor{ourscol} 1.167($\pm$0.034) \\
\rowcolor{pct50}
\textbf{50\%   MAE $\downarrow$} 
& 1.171($\pm$0.023) &  1.331($\pm$0.015) & 1.242($\pm$0.006) & 1.288($\pm$0.025) &  1.017($\pm$0.065)  & 0.996($\pm$0.035)
& \cellcolor{ourscol} 0.966($\pm$0.028) \\
\rowcolor{pct50}
\textbf{50\%  R$^2$ $\uparrow$}
& -0.206($\pm$0.053) &  -0.658($\pm$0.055)  & -0.377($\pm$0.017) &  -0.515($\pm$0.078) &  -0.014($\pm$0.168)  &  0.041($\pm$0.036)
& \cellcolor{ourscol}0.098($\pm$0.052) \\
\midrule

\rowcolor{pct25}
\textbf{25\%  RMSE $\downarrow$}
&  1.593($\pm$0.014) & 1.704($\pm$0.053)  & 1.632($\pm$0.035) & 1.608($\pm$0.014) & 1.350($\pm$0.129)  &  1.322($\pm$0.050)
& \cellcolor{ourscol}1.302($\pm$0.059) \\
\rowcolor{pct25}
\textbf{25\% MAE $\downarrow$}
& 1.337($\pm$0.008) & 1.399($\pm$0.028)  & 1.359($\pm$0.019) & 1.346($\pm$0.008) & 1.140($\pm$0.126) &  1.135($\pm$0.060)
& \cellcolor{ourscol} 1.105($\pm$0.085) \\
\rowcolor{pct25}
\textbf{25\% R$^2$ $\uparrow$}
& -0.678($\pm$0.028) & -0.921($\pm$0.120)  & -0.761($\pm$0.075) &  -0.709($\pm$0.029) &  -0.216($\pm$0.239)  &  -0.157($\pm$0.086)
& \cellcolor{ourscol} -0.124($\pm$0.102) \\
\bottomrule
\end{tabular}
}}

\vspace{5pt}

\begin{minipage}{1.0\textwidth}
\scriptsize
\textit{Notes. RMSE, root mean squared error; MAE, mean absolute error; R$^2$, coefficient of determination. RMSE and MAE are reported in EDSS scores; lower values indicate better performance, higher R$^2$ indicates better performance. Scratch uses random initialization; SwinUNETR$^{\ast}$, VoCo$^{\ast}$, and S3D$^{\ast}$ use publicly released centralized foundation models; S3D uses centralized pretraining on our pretraining corpus, FedMAE and BrainFedFM use federated pretraining. Fine-tuning uses 15 epochs of AdamW, a learning rate of $1\times10^{-6}$, weight decay of $1\times10^{-2}$, and cosine scheduling with a 2-epoch warm-up.}
\end{minipage}

\end{table*}

\clearpage

\begin{figure*}[p]
\noindent{\scriptsize\bfseries
Panel B. Labelled-data scaling and full-data regression analyses.}
{\tiny\raggedright
Characterizes regression performance from complementary perspectives.
\textbf{B1}, fold-averaged RMSE, MAE, R$^2$, Pearson correlation, Spearman correlation, and concordance correlation coefficient across the 25\%, 50\%, and 100\% labelled-data settings.
\textbf{B2}, cross-method analyses including error magnitude, association and agreement, signed-error distributions, and MAE across the observed EDSS range.
\textbf{B3}, the joint RMSE--CCC performance landscape across labelled-data settings, where lower RMSE and higher CCC indicate a more favorable error--agreement trade-off.\par}

\par\vspace{8pt}

\includegraphics[
    width=\textwidth,
    keepaspectratio
]{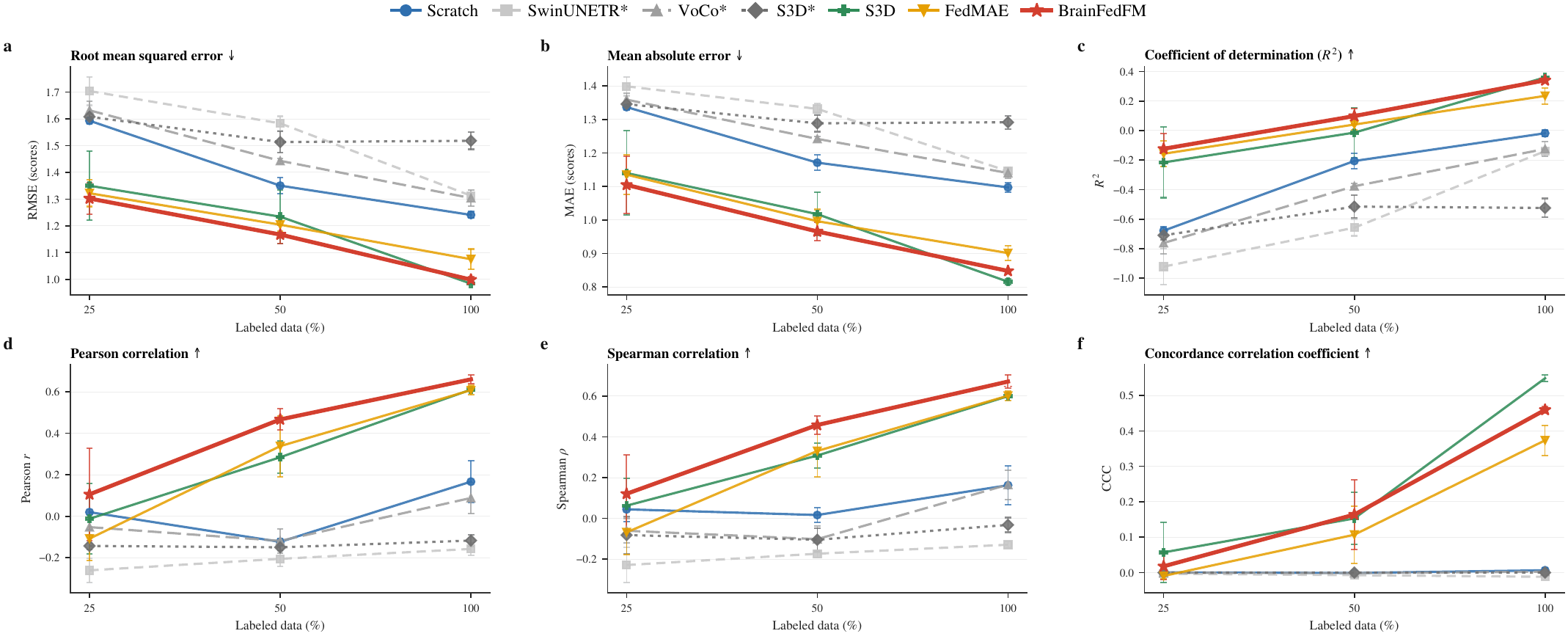}

\par\vspace{3pt}

\includegraphics[
    width=0.84\textwidth,
    keepaspectratio
]{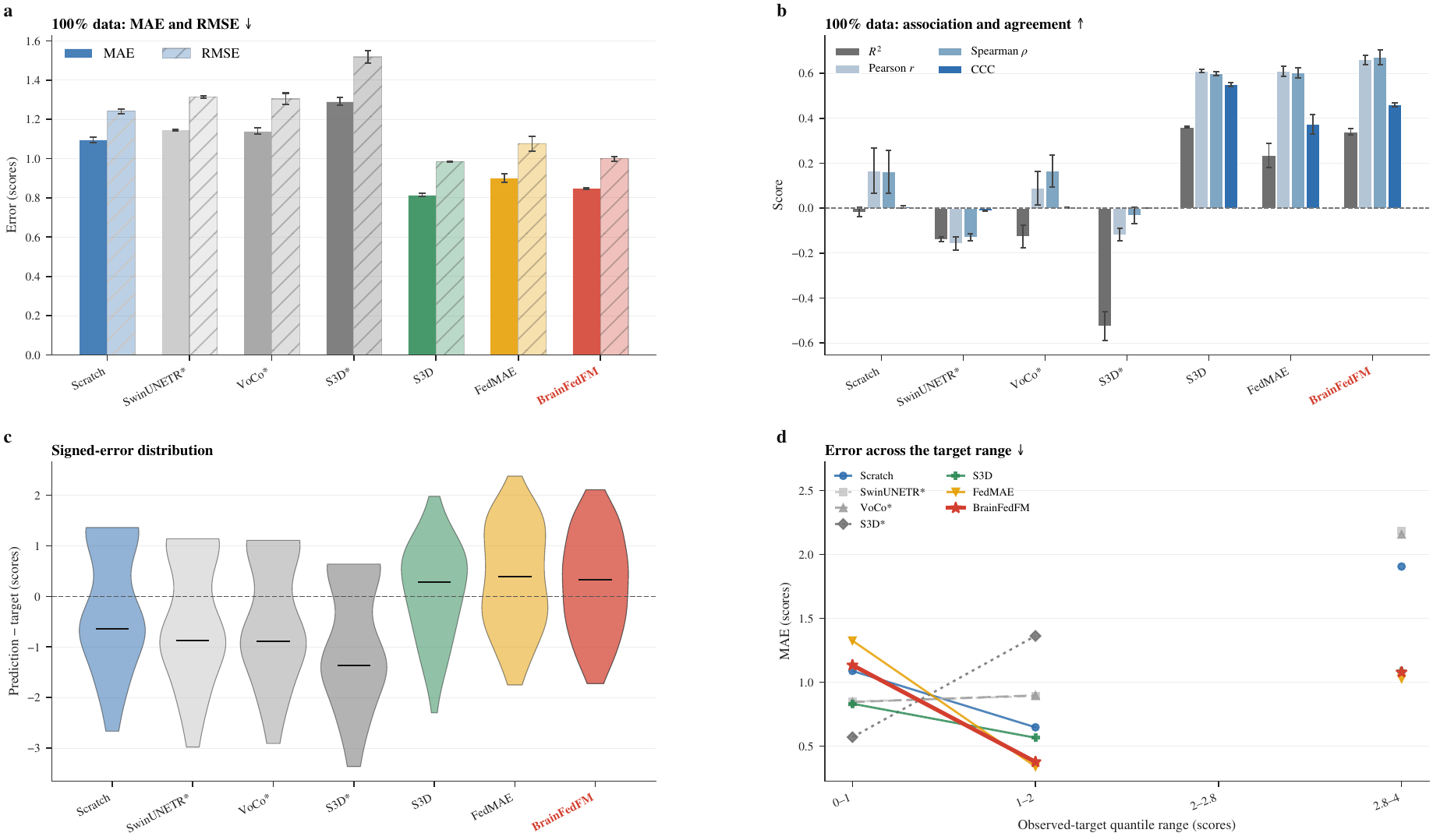}

\par\vspace{4pt}

\includegraphics[
    width=0.96\textwidth,
    keepaspectratio
]{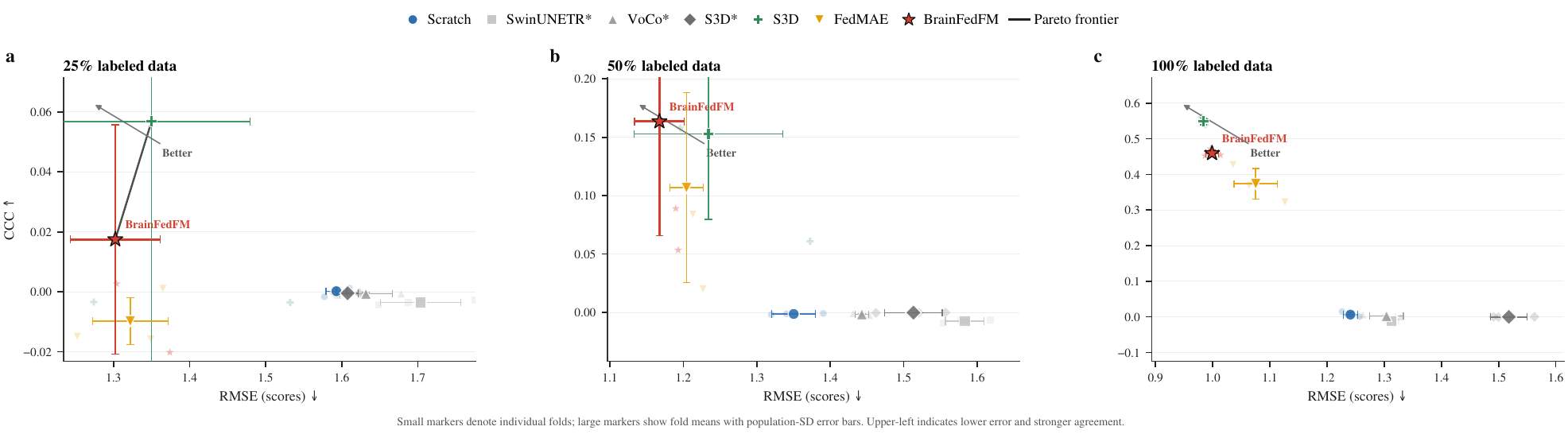}

\end{figure*}

\clearpage

\begin{figure*}[p]
\noindent{\scriptsize\bfseries
Panel C. Relative ranking, fold stability, and transfer-benefit analyses.}
{\tiny\raggedright
Summarizes comparative performance, robustness, and transfer benefit beyond the aggregate regression scores.
\textbf{C1}, relative method ranks across all labelled-data settings and six regression metrics, with rank 1 denoting the best result within each setting--metric combination.
\textbf{C2}, fold-to-fold variability in RMSE, MAE, and R$^2$, together with the average method rank across the core regression metrics.
\textbf{C3}, performance changes relative to the matched Scratch baseline across labelled-data settings, where positive values denote RMSE or MAE reduction and improvement in R$^2$.\par}

\par\vspace{5.5pt}

\includegraphics[
    width=\textwidth,
    keepaspectratio
]{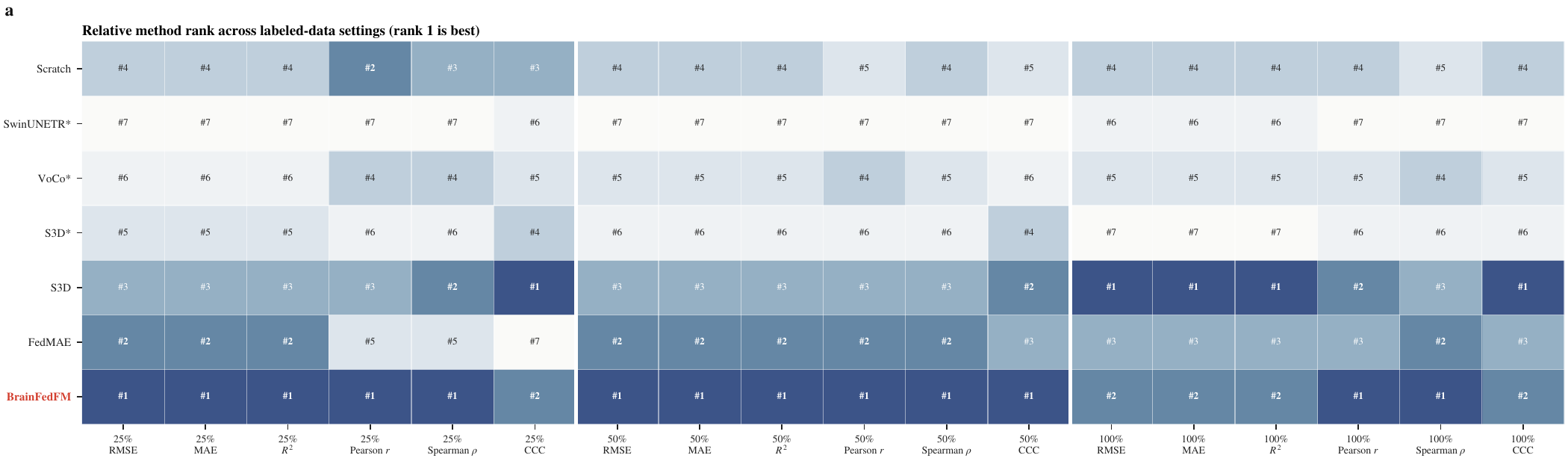}

\vspace{6pt}

\includegraphics[
    width=\textwidth,
    keepaspectratio
]{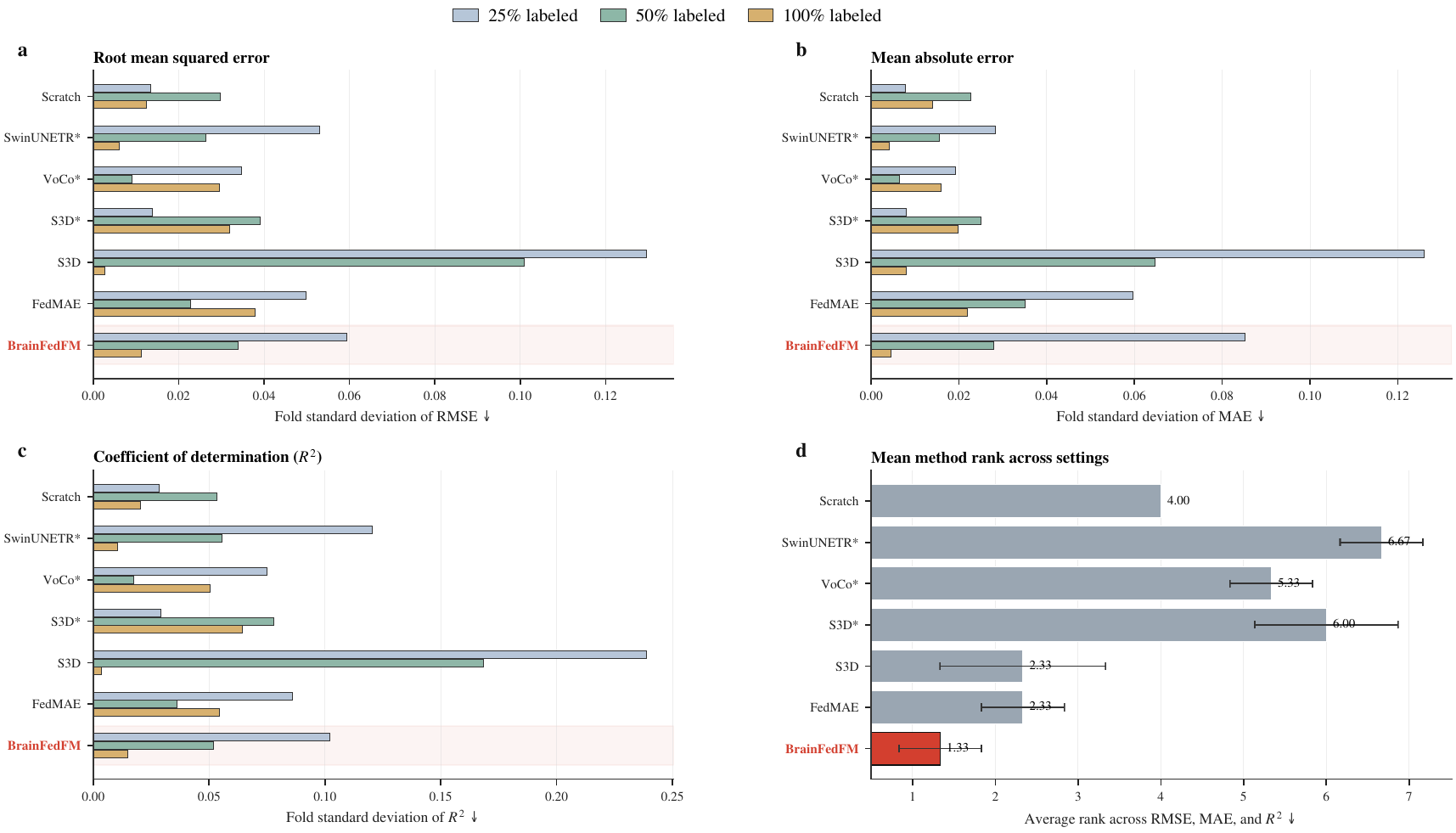}

\vspace{6pt}

\includegraphics[
    width=\textwidth,
    keepaspectratio
]{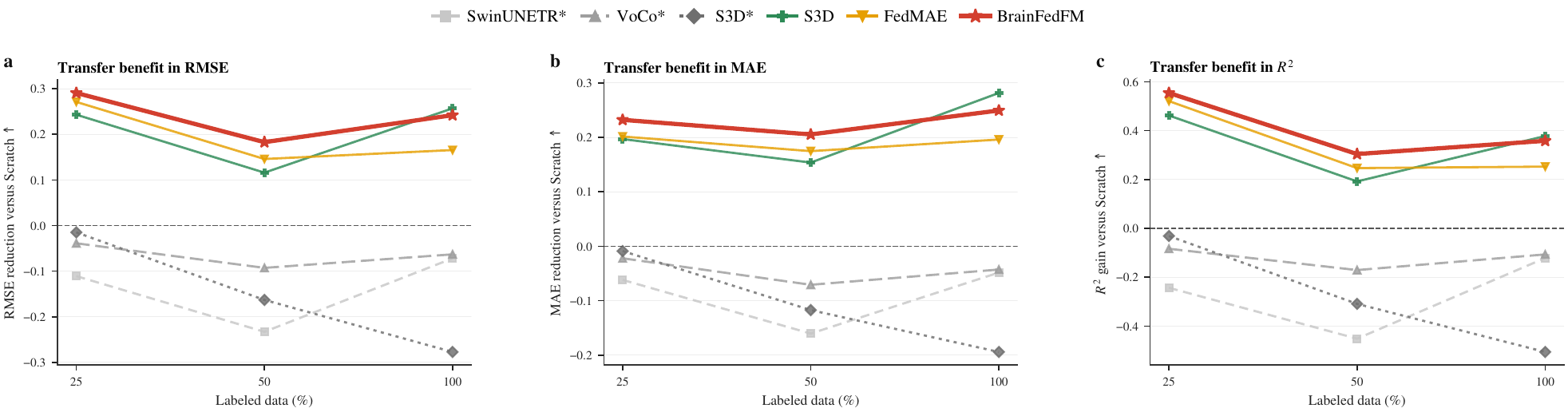}

\end{figure*}

\definecolor{pct100}{gray}{0.97}
\definecolor{pct75}{gray}{1.0}
\definecolor{pct50}{gray}{0.99}
\definecolor{pct25}{gray}{1.0}
\definecolor{overallgray}{gray}{0.97}
\definecolor{ourscol}{RGB}{244,248,252}

\begin{table*}[h]
\centering
\tiny
\caption[Infant brain tissue segmentation task]{\textbf{Infant brain tissue segmentation task.}
We evaluate foundation model performance on the iSeg benchmark, where the goal is to delineate cerebrospinal fluid (CSF), gray matter (GM), and white matter (WM) from infant brain MRI. Accurate tissue segmentation enables quantitative characterization of early brain development and supports morphometric and neurodevelopmental analysis during a period of rapid anatomical change. Panel~A reports macro-averaged and tissue-wise Dice, intersection-over-union (IoU), and normalized surface Dice (NSD) with 100\% labelled training data, whereas Panel~B reports the corresponding macro-averaged results with 25\% and 50\% labelled data. We compare a randomly initialized scratch baseline, centralized pretrained foundation models, and federated pretrained foundation models.}
\label{tab:result_iseg}

\renewcommand{\arraystretch}{2.43}
{\scriptsize\bfseries Panel A. Detailed results under 100\% labelled data.}
\vspace{5pt}

{\setlength{\tabcolsep}{3.38pt}

\resizebox{1.0\textwidth}{!}{
\begin{tabular}{
l c c c c c c
>
{\columncolor{ourscol}\centering\arraybackslash}p{1.62cm}
}

\toprule
\multicolumn{2}{c}{\textbf{Brain sMRI FMs}}
& \multicolumn{4}{c}{Centralized Pretraining}
& \multicolumn{2}{c}{Federated Pretraining} \\
\cmidrule(lr){1-2}\cmidrule(lr){3-6}\cmidrule(lr){7-8}
\makecell[c]{Metric/Methods}
& Scratch
& \makecell[c]{SwinUNETR$^{\ast}$} & VoCo$^{\ast}$ & S3D$^{\ast}$ & S3D
& \makecell[c]{FedMAE} & \makecell[c]{BrainFedFM} \\
\midrule

\rowcolor{overallgray}
\textbf{Overall Dice $\uparrow$}
& 89.16($\pm$0.197)
& 90.15($\pm$0.317)
& 90.98($\pm$0.294)
& 91.77($\pm$0.258)
& 91.83($\pm$0.506)
& 91.18($\pm$0.305)
& \cellcolor{ourscol}91.14($\pm$0.228) \\

\rowcolor{overallgray}
\textbf{Overall IoU $\uparrow$}
& 80.56($\pm$0.320)
& 82.18($\pm$0.526)
& 83.52($\pm$0.471)
& 84.87($\pm$0.420)
& 84.98($\pm$0.842)
& 83.89($\pm$0.515)
& \cellcolor{ourscol}83.81($\pm$0.383) \\

\rowcolor{overallgray}
\textbf{Overall NSD $\uparrow$}
& 97.16($\pm$0.322)
& 97.59($\pm$0.263)
& 98.24($\pm$0.238)
& 98.85($\pm$0.221)
& 98.65($\pm$0.303)
& 98.41($\pm$0.295)
& \cellcolor{ourscol}98.39($\pm$0.270) \\
\midrule

CSF (Dice)
& 92.49($\pm$0.439)
& 93.40($\pm$0.487)
& 93.50($\pm$0.463)
& 94.18($\pm$0.437)
& 94.38($\pm$0.404)
& 93.94($\pm$0.552)
& 93.84($\pm$0.299) \\

CSF (IoU)
& 86.07($\pm$0.748)
& 87.64($\pm$0.842)
& 87.83($\pm$0.806)
& 89.04($\pm$0.774)
& 89.39($\pm$0.707)
& 88.62($\pm$0.962)
& 88.44($\pm$0.528) \\

CSF (NSD)
& 99.20($\pm$0.061)
& 99.42($\pm$0.087)
& 99.46($\pm$0.092)
& 99.55($\pm$0.108)
& 99.57($\pm$0.078)
& 99.51($\pm$0.111)
& 99.48($\pm$0.087) \\

Gray matter
& 88.73($\pm$0.291)
& 89.99($\pm$0.186)
& 90.77($\pm$0.173)
& 91.33($\pm$0.259)
& 91.49($\pm$0.409)
& 90.82($\pm$0.047)
& 90.80($\pm$0.048) \\

Gray matter
& 79.75($\pm$0.467)
& 81.81($\pm$0.319)
& 83.10($\pm$0.287)
& 84.06($\pm$0.433)
& 84.32($\pm$0.683)
& 83.19($\pm$0.077)
& 83.15($\pm$0.078) \\

Gray matter
& 97.16($\pm$0.232)
& 97.63($\pm$0.214)
& 98.25($\pm$0.197)
& 98.87($\pm$0.192)
& 98.65($\pm$0.257)
& 98.43($\pm$0.251)
& 98.42($\pm$0.231) \\

White matter
& 86.26($\pm$0.380)
& 87.06($\pm$0.462)
& 88.66($\pm$0.489)
& 89.81($\pm$0.586)
& 89.64($\pm$0.743)
& 88.80($\pm$0.395)
& 88.79($\pm$0.347) \\

White matter
& 75.84($\pm$0.586)
& 77.09($\pm$0.751)
& 79.63($\pm$0.784)
& 81.51($\pm$0.959)
& 81.24($\pm$1.204)
& 79.85($\pm$0.638)
& 79.85($\pm$0.560) \\

White matter
& 95.12($\pm$0.676)
& 95.71($\pm$0.548)
& 97.02($\pm$0.512)
& 98.12($\pm$0.442)
& 97.74($\pm$0.583)
& 97.28($\pm$0.531)
& 97.27($\pm$0.501) \\
\bottomrule
\end{tabular}
}}

\vspace{6pt}

{\scriptsize\bfseries Panel B. Summary under different labelled-data percentages.}
\vspace{5pt}

\renewcommand{\arraystretch}{2.1}
\resizebox{1.0\textwidth}{!}{%
\begin{tabular}{
l
ccc!{\hspace{2pt}\color{black!45}\vrule width 0.6pt\hspace{2pt}}
ccc
}
\toprule
\multicolumn{1}{c}{\textbf{Percentages}}
& \multicolumn{3}{c}{25\%}
& \multicolumn{3}{c}{50\%} \\
\cmidrule(lr){1-1}\cmidrule(lr){2-4}\cmidrule(lr){5-7}
\makecell[c]{Methods/Metric}
& \multicolumn{1}{c}{Dice}
& \multicolumn{1}{c}{IoU}
& \multicolumn{1}{c}{NSD}
& \multicolumn{1}{c}{Dice}
& \multicolumn{1}{c}{IoU}
& \multicolumn{1}{c}{NSD} \\
\midrule

Scratch
& 81.12($\pm$2.418)
& 68.42($\pm$3.315)
& 91.19($\pm$2.029)
& 87.05($\pm$0.477)
& 77.15($\pm$0.747)
& 96.24($\pm$0.474) \\

SwinUNETR$^{\ast}$
& 81.54($\pm$3.188)
& 69.11($\pm$4.334)
& 91.40($\pm$3.162)
& 87.78($\pm$0.390)
& 78.31($\pm$0.609)
& 96.66($\pm$0.335) \\

VoCo$^{\ast}$
& 83.35($\pm$1.806)
& 71.61($\pm$2.590)
& 92.61($\pm$1.734)
& 88.47($\pm$0.222)
& 79.41($\pm$0.351)
& 97.10($\pm$0.134) \\

S3D$^{\ast}$
& 85.63($\pm$2.755)
& 75.14($\pm$4.047)
& 95.16($\pm$2.302)
& 90.27($\pm$0.139)
& 82.35($\pm$0.214)
& 98.20($\pm$0.054) \\

S3D
& 85.09($\pm$2.906)
& 74.35($\pm$4.212)
& 94.41($\pm$2.774)
& 90.17($\pm$0.605)
& 82.20($\pm$0.979)
& 98.12($\pm$0.306) \\
\hline

FedMAE
& 84.83($\pm$2.023)
& 73.89($\pm$2.956)
& 93.98($\pm$1.971)
& 89.63($\pm$0.221)
& 81.31($\pm$0.348)
& 97.84($\pm$0.152) \\

\rowcolor{ourscol}
BrainFedFM
& 84.75($\pm$2.939)
& 73.84($\pm$4.244)
& 94.14($\pm$2.412)
& 89.51($\pm$0.506)
& 81.11($\pm$0.803)
& 97.76($\pm$0.367) \\
\bottomrule
\end{tabular}
}
\vspace{3pt}

\begin{minipage}{1.0\textwidth}
\scriptsize
\textit{Notes. Dice, Dice similarity coefficient; IoU, intersection-over-union; NSD, normalized surface Dice. All models are fine-tuned for 150 epochs with 50 training iterations and 50 validation iterations per epoch using SGD with momentum 0.99, Nesterov acceleration, weight decay of $3\times10^{-5}$, and an initial learning rate of $1\times10^{-3}$. Training uses a 15-epoch linear warm-up followed by polynomial learning-rate decay, and test performance is reported from the checkpoint with the best validation result.}
\end{minipage}
\end{table*}

\clearpage

\noindent{\scriptsize\bfseries Panel C. Data efficiency, complementary profiles, and qualitative tissue segmentation.}
\vspace{2pt}
{\tiny\raggedright
Visualizes method behavior across labelled-data regimes and complementary tissue-segmentation criteria. \textbf{C1}, fold-averaged trajectories for macro Dice, macro NSD, macro sensitivity, macro precision, HD95, and absolute tissue-volume error across 25\%, 50\%, and 100\% labelled data; \textbf{C2}, direction-aligned profiles across overlap/surface, voxel/volume, and tissue-wise metric families, together with a family-level summary; \textbf{C3}, representative axial MRI examples with reference tissue maps and three-fold ensemble predictions with macro and tissue-wise Dice scores.\par}
\par\vspace{-4pt}

\begin{center}
\includegraphics[width=0.968\textwidth,keepaspectratio]{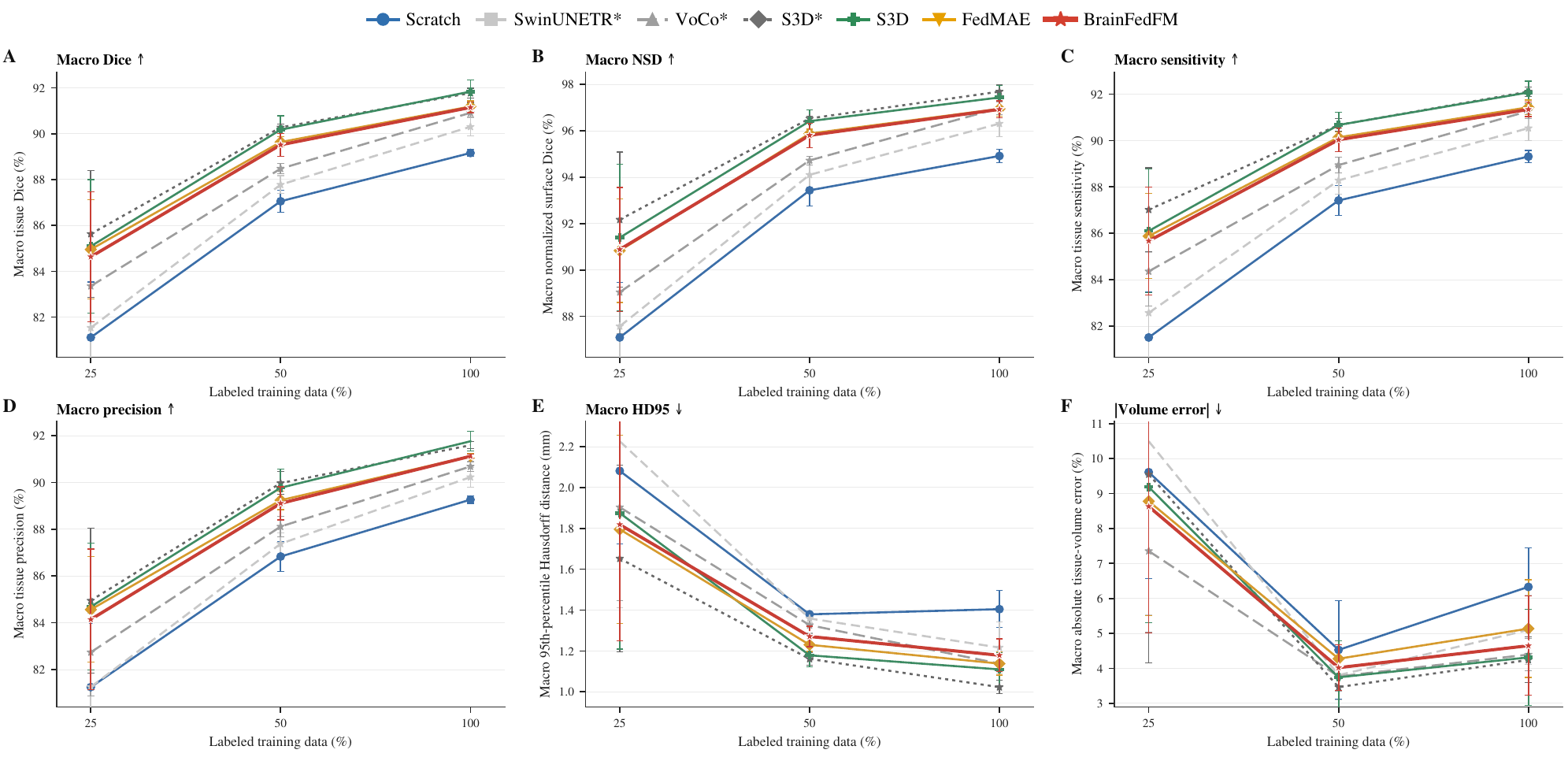}
\includegraphics[width=0.84\textwidth,keepaspectratio]{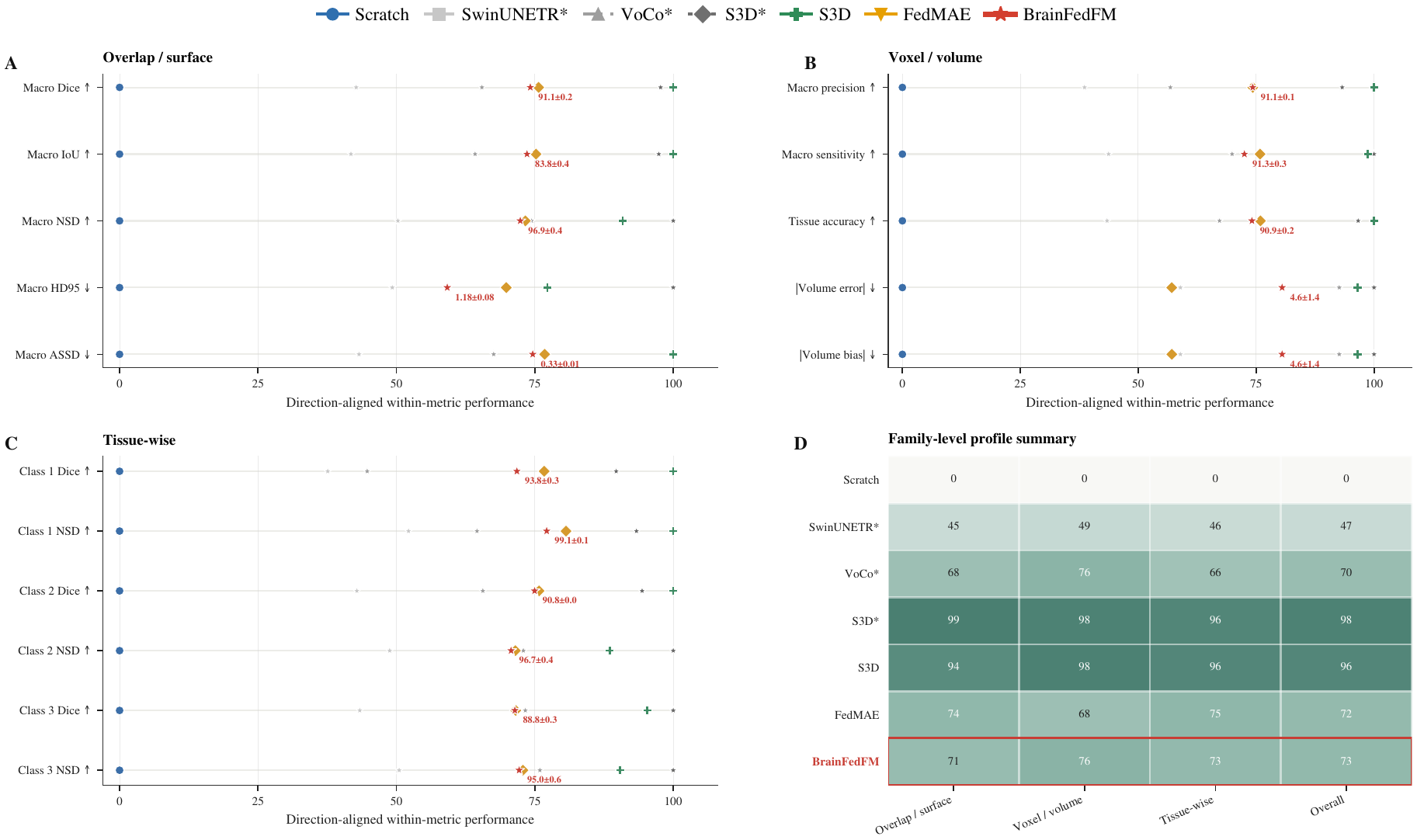}
\includegraphics[width=0.97\textwidth,keepaspectratio]{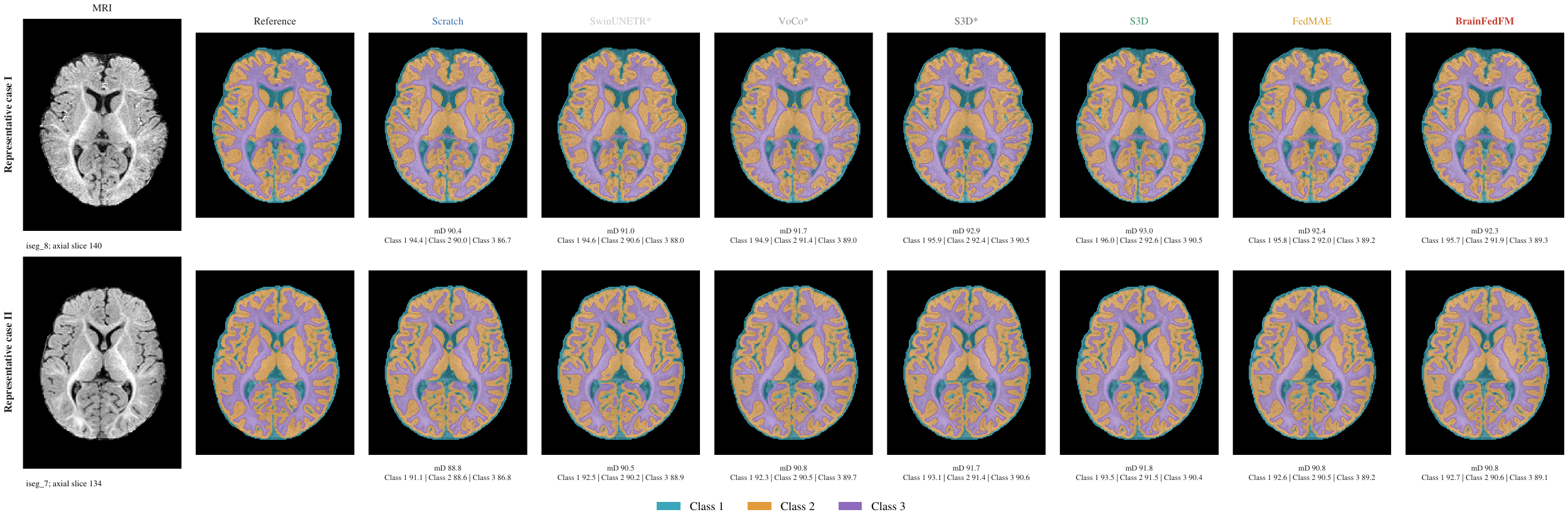}
\end{center}

\clearpage

\noindent{\scriptsize\bfseries Panel D. Tissue-wise decomposition, case-level variability, and cross-tissue confusion.}
\vspace{2pt}
{\tiny\raggedright
Characterizes tissue-specific performance and robustness beyond the aggregate overlap scores. \textbf{D1}, tissue-wise Dice and NSD, the macro Dice--NSD performance landscape, and the macro precision--sensitivity balance at 100\% labelled data; \textbf{D2}, matched case-by-fold distributions of macro Dice, HD95, average symmetric surface distance (ASSD), and absolute tissue-volume error; \textbf{D3}, row-normalized confusion matrices for WM, GM, and CSF, including predictions assigned to background, across all seven methods at 100\% labelled data.\par}
\par\vspace{-3pt}

\begin{center}
\includegraphics[width=0.76\textwidth,keepaspectratio]{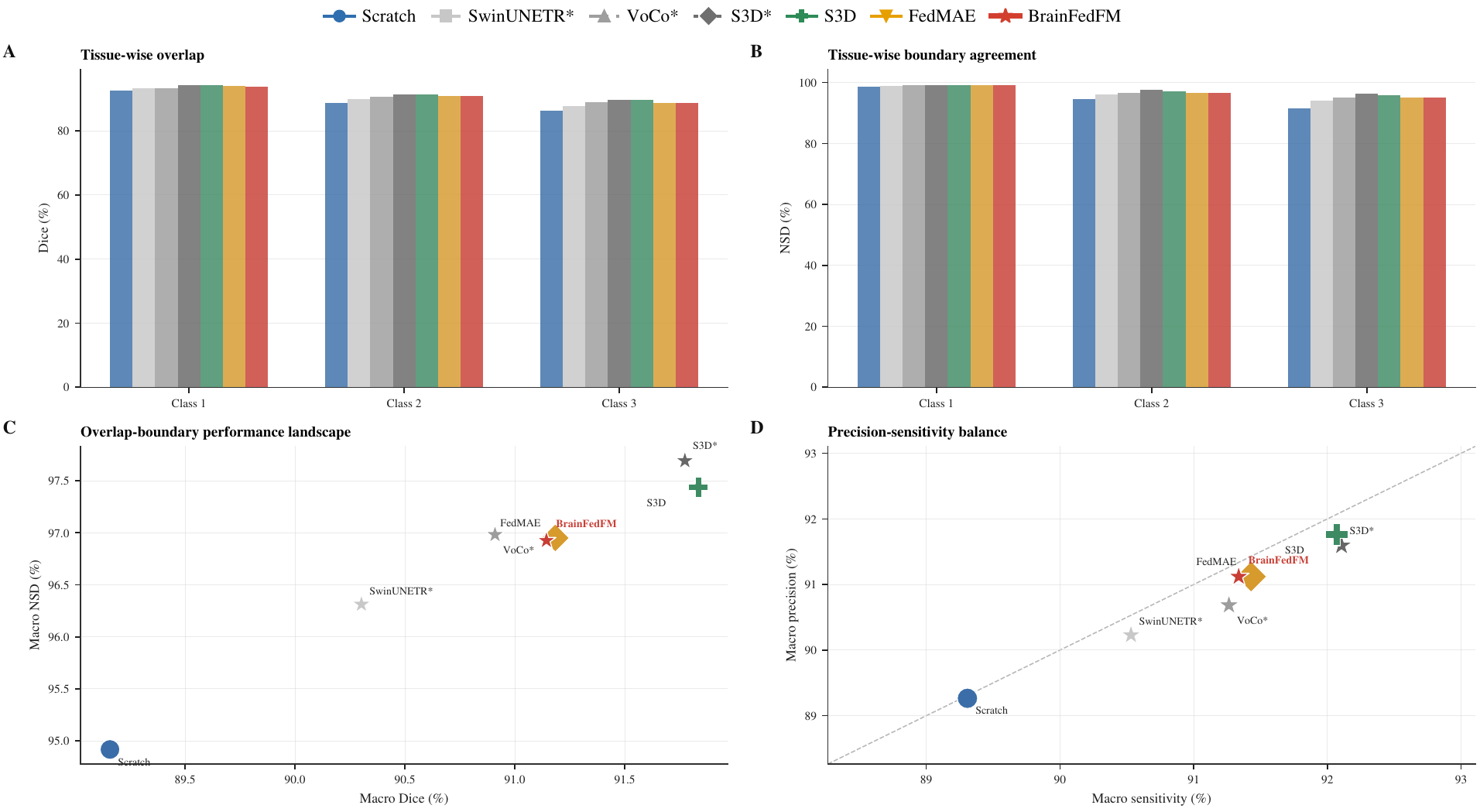}\par\vspace{1pt}
\includegraphics[width=0.745\textwidth,keepaspectratio]{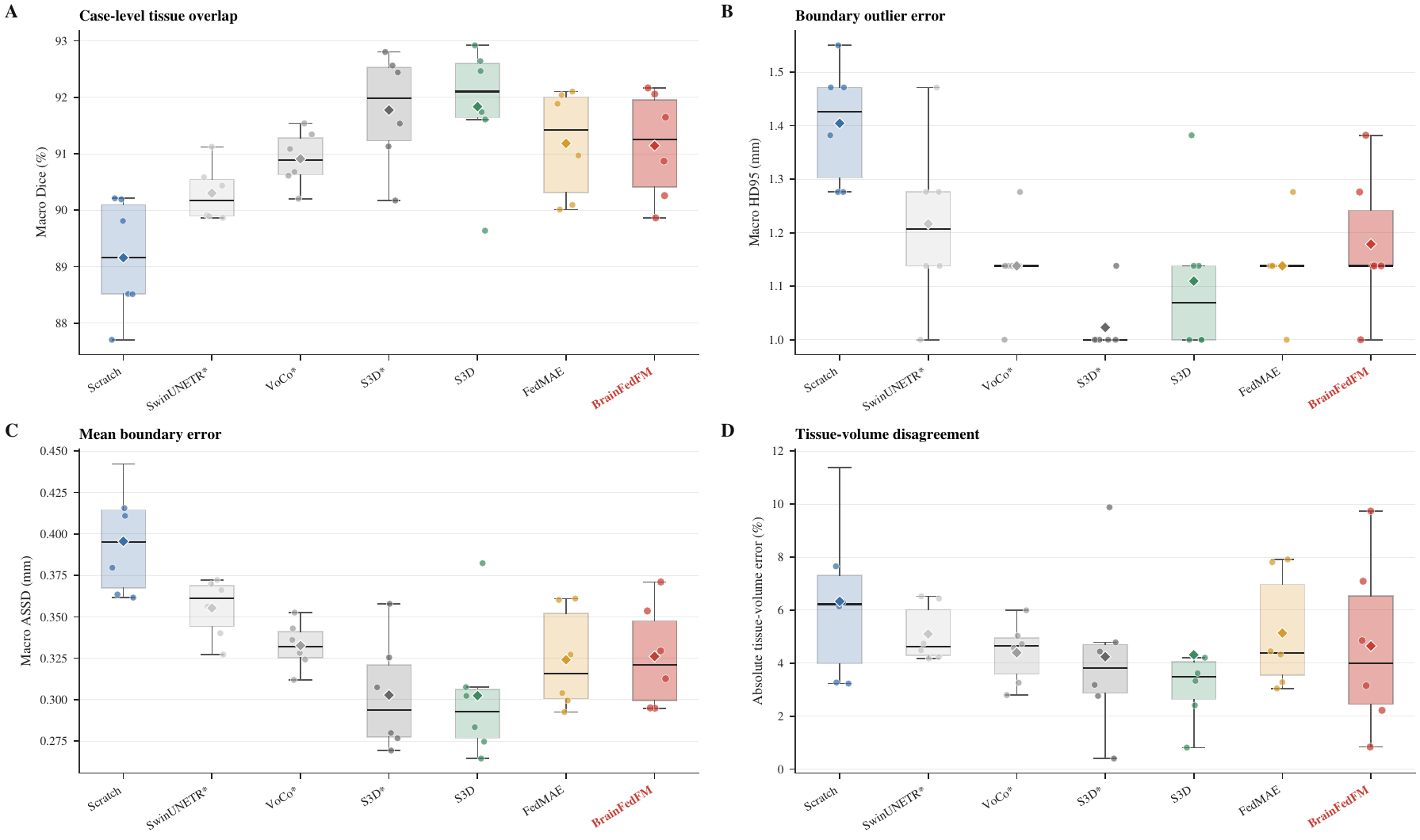}\par\vspace{1pt}
\includegraphics[width=0.875\textwidth,keepaspectratio]{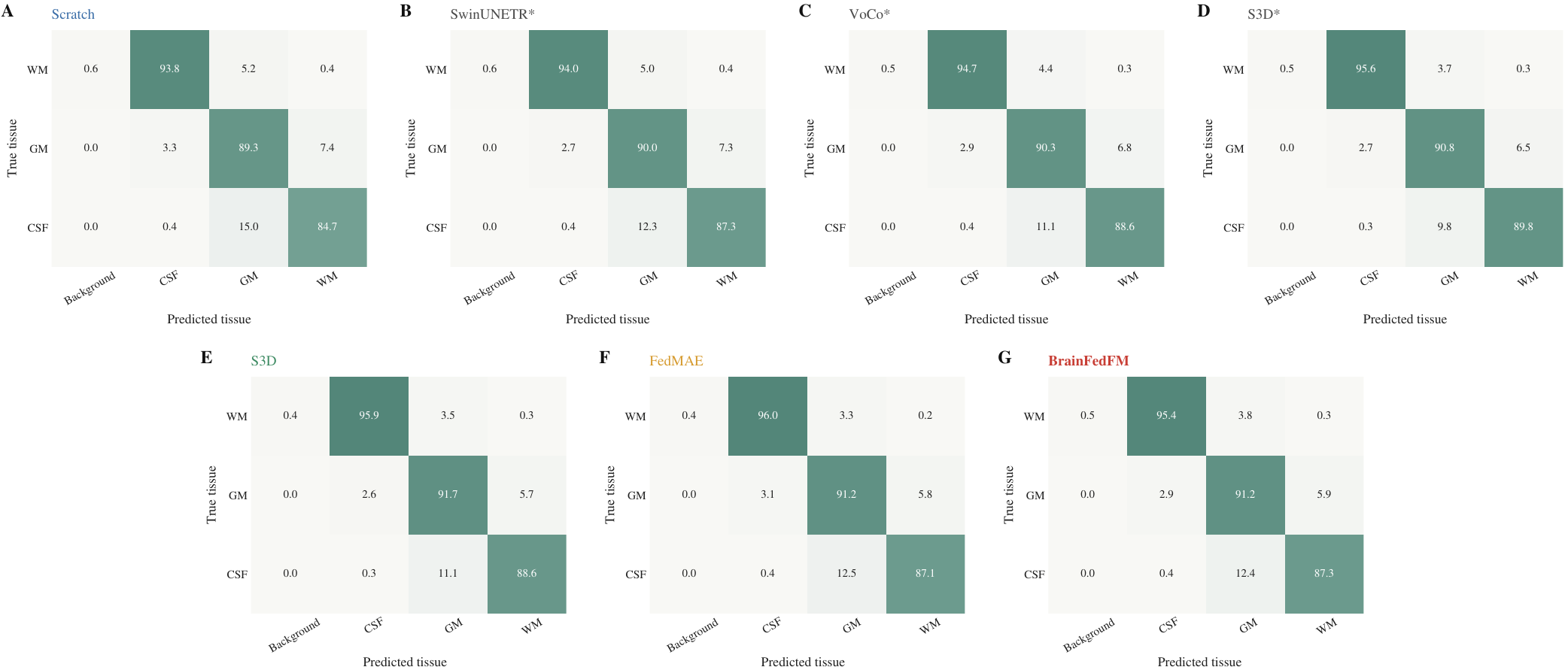}
\end{center}

\definecolor{pct100}{gray}{0.97}
\definecolor{pct75}{gray}{1.0}
\definecolor{pct50}{gray}{0.99}
\definecolor{pct25}{gray}{1.0}
\definecolor{overallgray}{gray}{1.0}
\definecolor{ourscol}{RGB}{244,248,252}

\begin{table*}[t]
\centering
\tiny
\caption[MRI sequence classification task]{\textbf{MRI sequence classification task.}
We evaluate foundation model performance on the UPENN-GBM benchmark. This task aims to identify the sequence modality of individual brain MRI scans, including T1w, contrast-enhanced T1w (T1ce), T2w, and FLAIR images. 
Accurate sequence recognition is essential for multi-sequence MRI analysis, as it ensures correct modality labeling, input-channel assignment, data harmonization, and quality control before model training and evaluation.
The table summarizes performance under 25\%, 50\%, and 100\% labelled-data settings, which indicate the proportion of the labelled training--validation pool used for downstream fine-tuning and model selection; the held-out test set remains unchanged across all settings.
We compare a randomly initialized scratch baseline, centralized pretrained foundation models, and federated pretrained foundation models.}
\label{tab:result_upenn}

\renewcommand{\arraystretch}{3.85}
{\scriptsize\bfseries Panel A. Summary under different labelled-data percentages.}
\vspace{5pt}

{\setlength{\tabcolsep}{3.38pt}

\resizebox{1.0\textwidth}{!}{
\begin{tabular}{
l c c c c c c
>{\columncolor{ourscol}\centering\arraybackslash}p{1.62cm}
}
\toprule
\multicolumn{2}{c}{\textbf{Brain sMRI FMs }}
& \multicolumn{4}{c}{Centralized Pretraining}
& \multicolumn{2}{c}{Federated Pretraining} \\
\cmidrule(lr){1-2}\cmidrule(lr){3-6}\cmidrule(lr){7-8}
\makecell[c]{Metric/Methods}
& Scratch
& \makecell[c]{SwinUNETR$^{\ast}$}  &\ VoCo$^{\ast}$ \ & \ S3D$^{\ast}$  & \ S3D 
& \makecell[c]{FedMAE} & \makecell[c]{BrainFedFM} \\
\midrule

\rowcolor{pct100}
\textbf{100\%  BA $\uparrow$}
& 94.58($\pm$3.444)  & 95.35($\pm$4.662) & 99.74($\pm$0.182) & 99.68($\pm$0.183) & 98.97($\pm$0.556)  & 99.10($\pm$0.594)
& \cellcolor{ourscol}99.48($\pm$0.091) \\
\rowcolor{pct100}
\textbf{100\%   mF1 $\uparrow$}
& 94.53($\pm$3.506) & 95.40($\pm$4.588) & 99.74($\pm$0.182) & 99.68($\pm$0.183) & 98.97($\pm$0.556)  & 99.10($\pm$0.597)
& \cellcolor{ourscol}99.48($\pm$0.092) \\
\rowcolor{pct100}
\textbf{100\%   AUC $\uparrow$}
& 99.42($\pm$0.299) & 99.86($\pm$0.171) & 100.0($\pm$0.001) & 100.0($\pm$0.000) & 99.98($\pm$0.026)  & 99.99($\pm$0.003)
& \cellcolor{ourscol}100.0($\pm$0.001)\\
\midrule

\rowcolor{pct50}
\textbf{50\%  BA $\uparrow$}
& 79.87($\pm$10.31) & 69.11($\pm$25.54) & 96.71($\pm$2.476) & 99.35($\pm$0.329)  & 99.16($\pm$0.240) & 98.39($\pm$0.183)
& \cellcolor{ourscol}98.26($\pm$0.989) \\
\rowcolor{pct50}
\textbf{50\%  mF1 $\uparrow$}
& 75.20($\pm$13.52) & 64.87($\pm$29.78) & 96.67($\pm$2.521) & 99.35($\pm$0.329) & 99.16($\pm$0.239) & 98.39($\pm$0.182)
& \cellcolor{ourscol}98.26($\pm$0.999) \\
\rowcolor{pct50}
\textbf{50\% AUC $\uparrow$}
& 98.34($\pm$0.801) & 99.33($\pm$0.132) & 99.77($\pm$0.272) & 99.98($\pm$0.020) & 99.99($\pm$0.003)  & 99.97($\pm$0.012)
& \cellcolor{ourscol}100.0($\pm$0.003)  \\
\midrule

\rowcolor{pct25}
\textbf{25\%  BA $\uparrow$}
& 57.40($\pm$8.585) & 48.91($\pm$10.86) & 75.58($\pm$13.76) & 89.47($\pm$10.38) & 97.81($\pm$0.241)  & 89.44($\pm$7.402)
& \cellcolor{ourscol} 98.78($\pm$0.396) \\
\rowcolor{pct25}
\textbf{25\%  mF1 $\uparrow$}
& 49.34($\pm$11.01) & 37.53($\pm$11.75)  & 70.57($\pm$16.88) & 86.59($\pm$14.45) & 97.81($\pm$0.244)  &89.18($\pm$7.823)
& \cellcolor{ourscol}98.78($\pm$0.394) \\
\rowcolor{pct25}
\textbf{25\% AUC $\uparrow$}
& 95.48($\pm$1.687) & 94.64($\pm$3.776) & 97.33($\pm$2.409) & 99.90($\pm$0.051) & 99.80($\pm$0.191)  & 99.68($\pm$0.159)
& \cellcolor{ourscol}99.97($\pm$0.005) \\
\bottomrule
\end{tabular}
}}

\vspace{5pt}

\begin{minipage}{1.0\textwidth}
\scriptsize
\textit{Notes. Values are mean($\pm$std) over three folds. BA, balanced accuracy; mF1, macro F1-score; AUC, macro one-vs-rest ROC AUC. Scratch uses random initialization; SwinUNETR$^{\ast}$, VoCo$^{\ast}$, and S3D$^{\ast}$ use publicly released centralized foundation model weights; S3D uses centralized pretraining on our pretraining corpus, FedMAE and BrainFedFM use federated pretraining. 
Fine-tuning uses 2 epochs of AdamW, learning rate $1\times10^{-6}$, weight decay $1\times10^{-2}$, cosine scheduling with a 1-epoch warm-up, and label smoothing of 0.2. The best checkpoint is selected according to validation AUC.}
\end{minipage}
\end{table*}

\clearpage

\noindent{\scriptsize\bfseries Panel B. Labelled-data scaling and aggregate metric summaries.}
\vspace{2pt}
{\tiny\raggedright
Summarizes method behavior across labelled-data settings.
\textbf{B1}, fold-averaged trends for classification and calibration metrics; \textbf{B2}, extended metric heatmaps separating higher-is-better classification metrics from lower-is-better calibration/uncertainty metrics; \textbf{B3}, the accuracy--calibration trade-off, where points closer to the upper-left indicate stronger discrimination with better calibration.\par}
\vspace{-3pt}

\begin{center}
\includegraphics[width=1.0\textwidth,keepaspectratio]{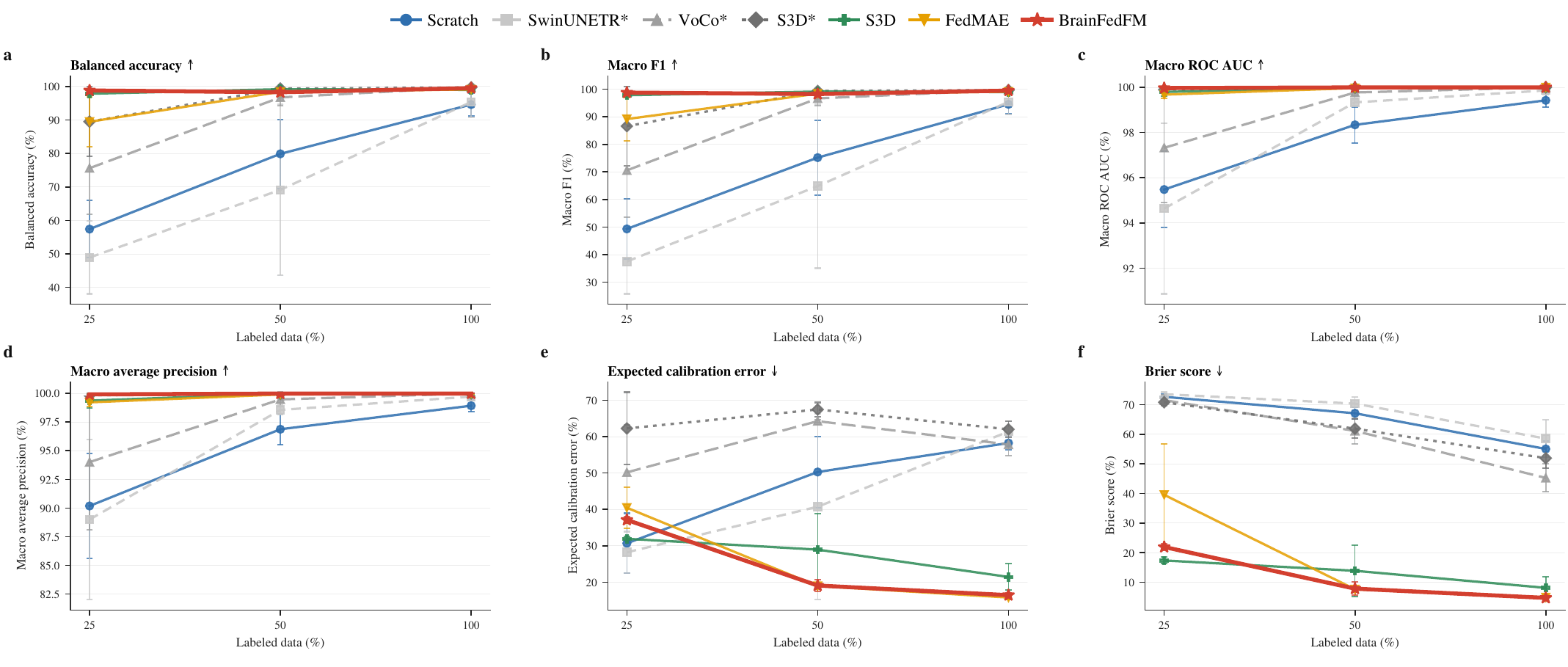} \par\vspace{3pt}
\includegraphics[width=1.0\textwidth,keepaspectratio]{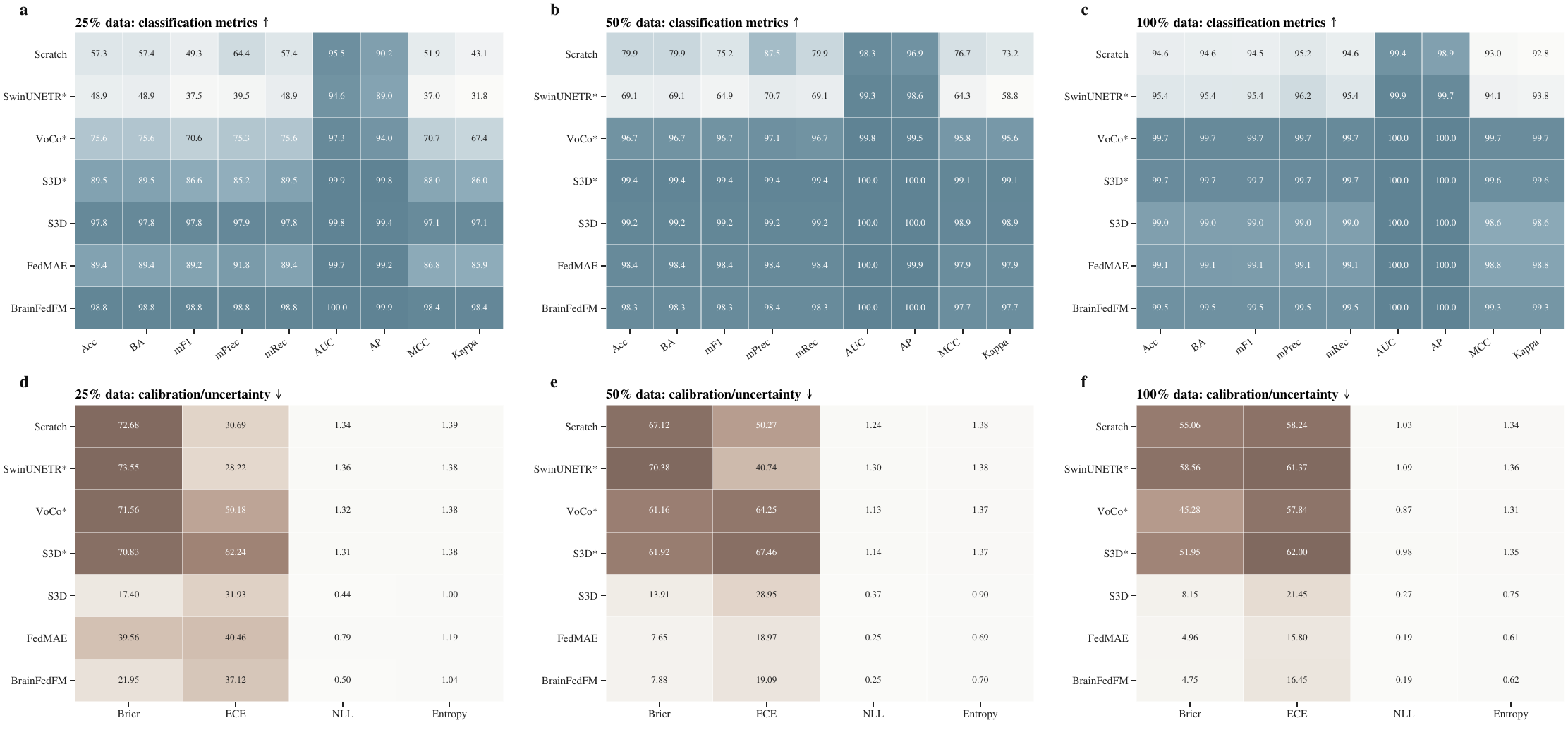} \par\vspace{5pt}
\includegraphics[width=0.935\textwidth,keepaspectratio]{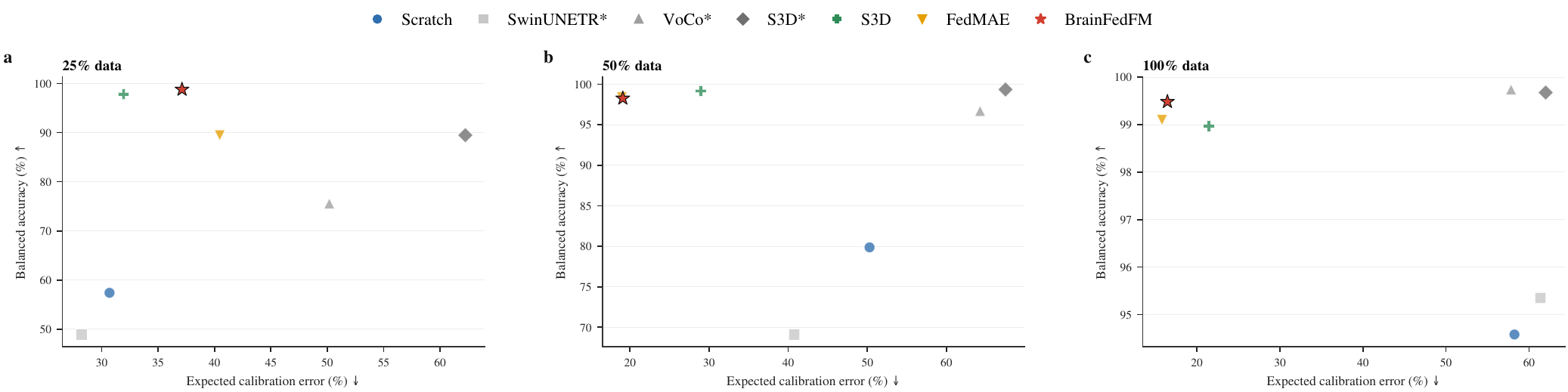}
\end{center}

\vspace{-2pt}

\noindent{\scriptsize\bfseries Panel C. Fold stability and full-data performance analyses.}
{\tiny\raggedright
Examines robustness and error modes beyond the aggregate scores. \textbf{C1}, fold-to-fold variability and average method ranking across balanced accuracy, macro F1, and macro ROC AUC; \textbf{C2}, probability-level analysis at 100\% labelled data, including micro-averaged ROC/precision--recall curves, reliability curves, and calibration scores; \textbf{C3}, row-normalized confusion matrices at 100\% labelled data, with rows denoting true sequence classes and columns denoting predicted classes.\par}

\clearpage

\begin{center}
\includegraphics[width=0.838\textwidth,keepaspectratio]{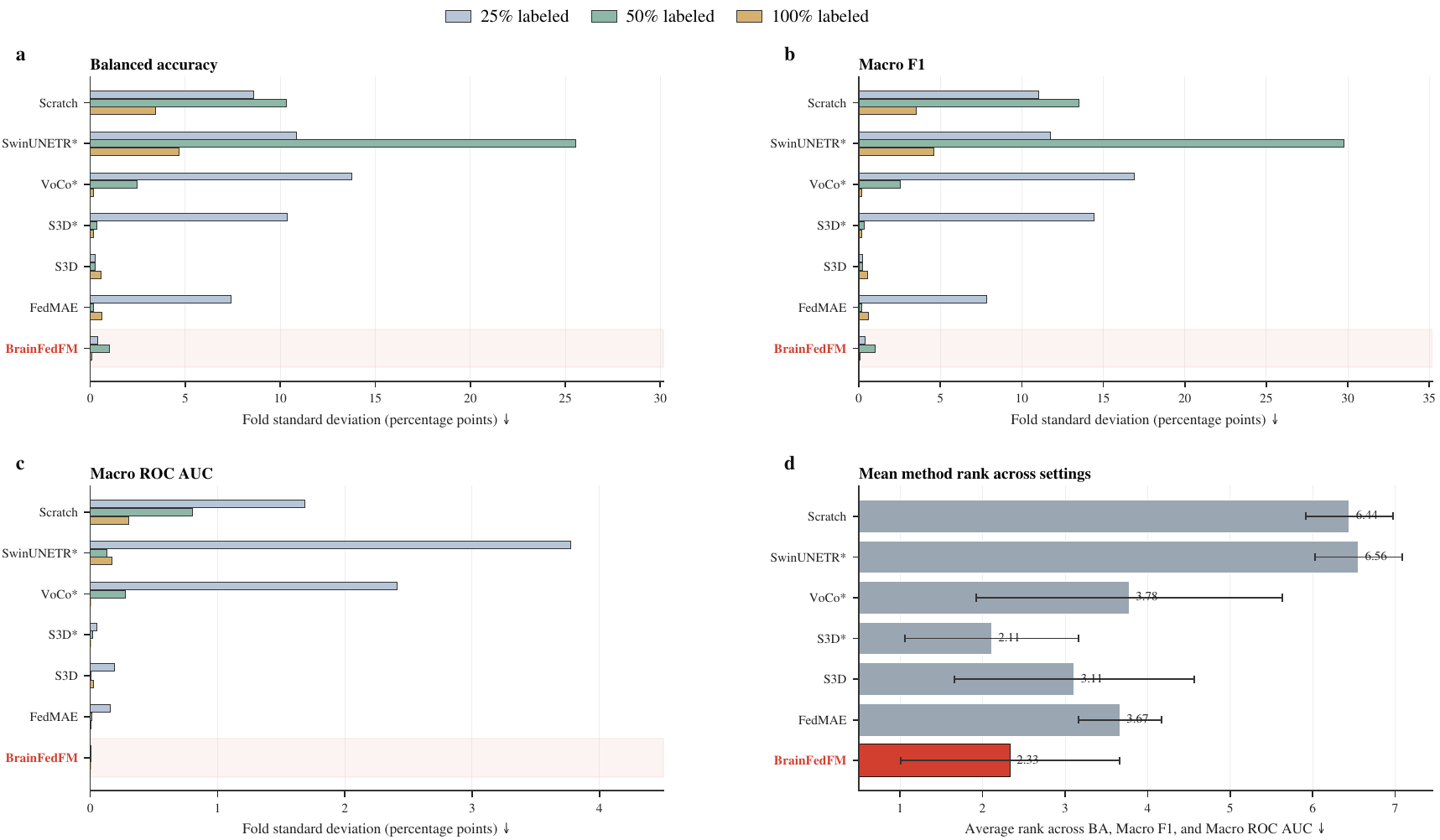}  \par\vspace{6pt}
\includegraphics[width=0.785\textwidth,keepaspectratio]{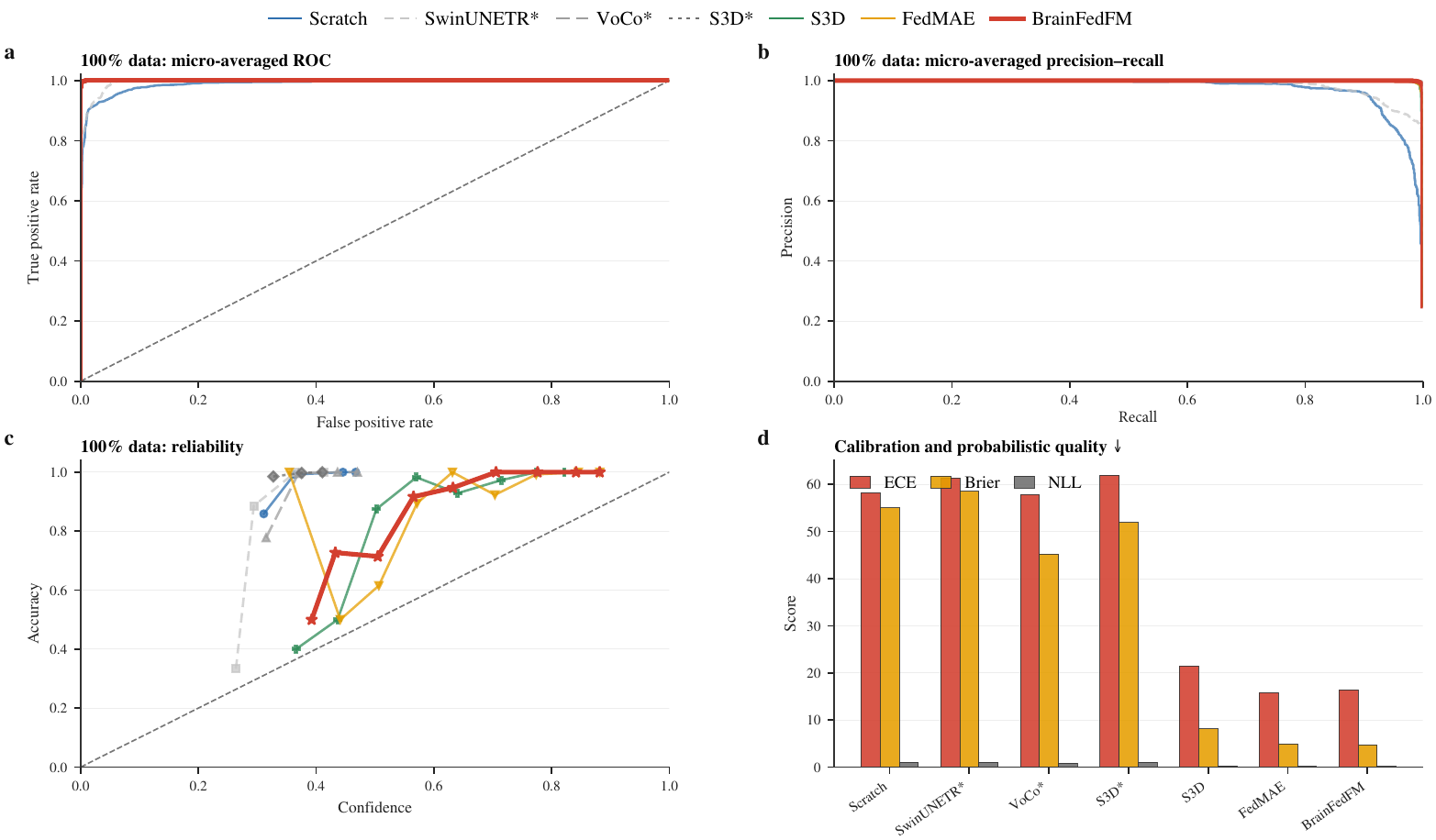}   \par\vspace{4pt}
\includegraphics[width=0.83\textwidth,keepaspectratio]{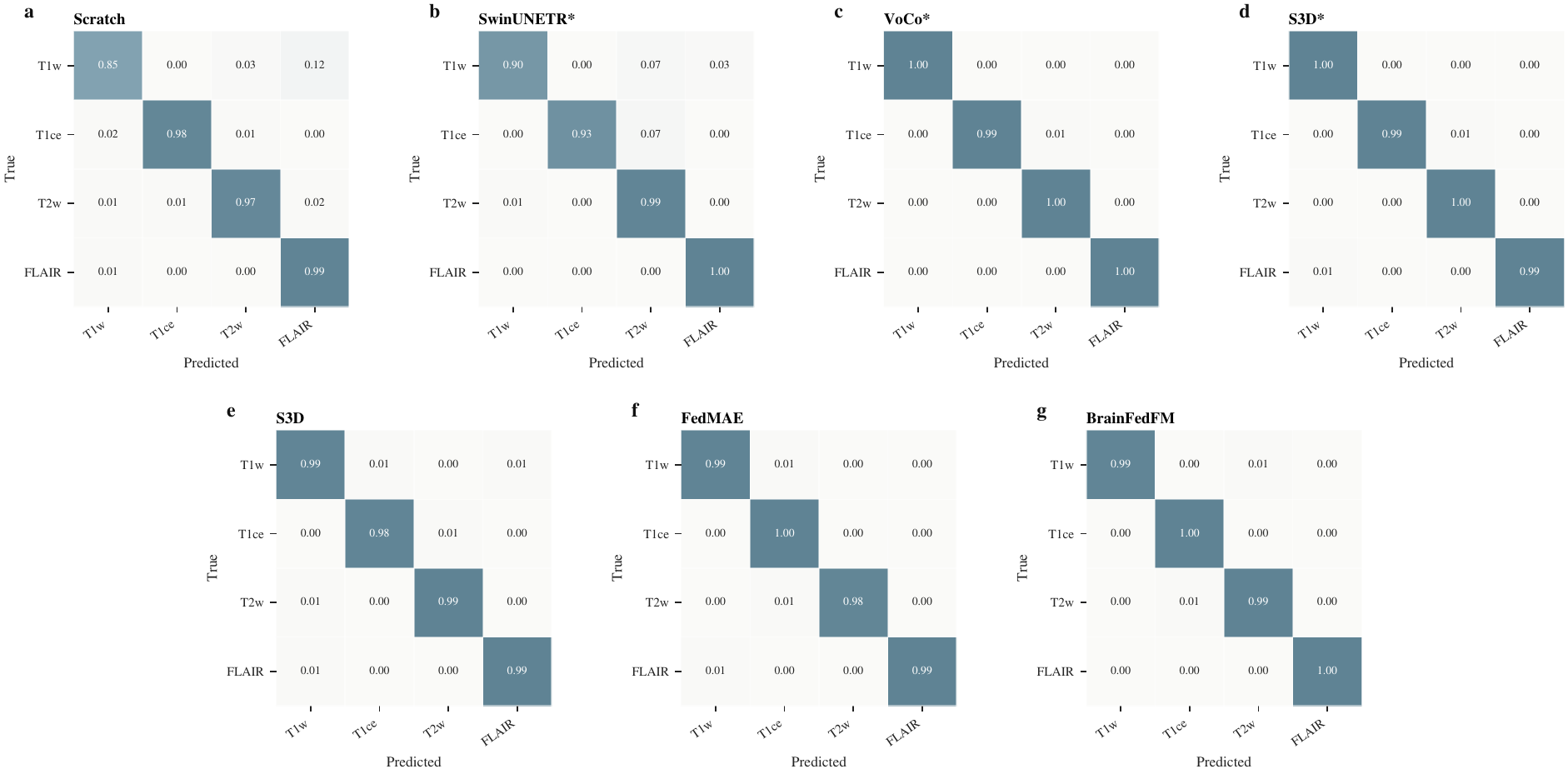}
\end{center}
\definecolor{pct100}{gray}{0.97}
\definecolor{pct75}{gray}{1.0}
\definecolor{pct50}{gray}{0.99}
\definecolor{pct25}{gray}{1.0}
\definecolor{overallgray}{gray}{1.0}
\definecolor{ourscol}{RGB}{244,248,252}

\begin{table*}[t]
\centering
\tiny
\caption[Parkinsonism subtype classification task]{\textbf{Parkinsonism subtype classification task.}
We evaluate foundation model performance on the 4RTNI benchmark. This task aims to classify structural brain MRI scans into corticobasal syndrome (CBS) and progressive supranuclear palsy (PSP) categories.
Accurate subtype classification from brain MRI is important for atypical parkinsonism analysis, supporting consistent subject stratification, subtype-specific neuroanatomical characterization, and downstream model evaluation.
The table summarizes performance under 25\%, 50\%, and 100\% labelled-data settings, which indicate the proportion of the labelled training--validation pool used for downstream fine-tuning and model selection; the held-out test set remains unchanged across all settings.
We compare a randomly initialized scratch baseline, centralized pretrained foundation models, and federated pretrained foundation models.}
\label{tab:result_4rnti}

\renewcommand{\arraystretch}{3.85}
{\scriptsize\bfseries Panel A. Summary under different labelled-data percentages.}
\vspace{5pt}

{\setlength{\tabcolsep}{3.38pt}

\resizebox{1.0\textwidth}{!}{
\begin{tabular}{
l c c c c c c
>{\columncolor{ourscol}\centering\arraybackslash}p{1.62cm}
}
\toprule
\multicolumn{2}{c}{\textbf{Brain sMRI FMs }}
& \multicolumn{4}{c}{Centralized Pretraining}
& \multicolumn{2}{c}{Federated Pretraining} \\
\cmidrule(lr){1-2}\cmidrule(lr){3-6}\cmidrule(lr){7-8}
\makecell[c]{Metric/Methods}
& Scratch
& \makecell[c]{SwinUNETR$^{\ast}$}  &\ VoCo$^{\ast}$ \ & \ S3D$^{\ast}$  & \ S3D 
& \makecell[c]{FedMAE} & \makecell[c]{BrainFedFM} \\
\midrule
\rowcolor{pct100}
\textbf{100\%  BA $\uparrow$}
& 78.21($\pm$5.688) & 75.96($\pm$5.701) & 89.27($\pm$2.800) & 86.82($\pm$1.892) & 89.57($\pm$1.886) & 90.30($\pm$2.288)
& \cellcolor{ourscol}91.68($\pm$2.990) \\
\rowcolor{pct100}
\textbf{100\%   mF1 $\uparrow$}
& 78.41($\pm$6.047) & 76.64($\pm$5.757) & 89.25($\pm$2.202) & 86.95($\pm$2.183) & 88.86($\pm$2.406) & 90.68($\pm$1.745)
& \cellcolor{ourscol}92.23($\pm$2.581) \\
\rowcolor{pct100}
\textbf{100\%   AUC $\uparrow$}
& 91.87($\pm$0.606) & 87.64($\pm$2.063) & 96.56($\pm$1.158) & 95.66($\pm$1.688) & 95.56($\pm$0.765) & 96.89($\pm$1.617)
& \cellcolor{ourscol}97.33($\pm$1.861) \\
\midrule

\rowcolor{pct50}
\textbf{50\%  BA $\uparrow$}
& 59.85($\pm$12.62) & 55.05($\pm$3.806) & 77.05($\pm$6.534) & 78.80($\pm$3.067) & 86.36($\pm$3.997) & 88.66($\pm$1.500)
& \cellcolor{ourscol}86.74($\pm$5.975) \\
\rowcolor{pct50}
\textbf{50\%  mF1 $\uparrow$}
& 52.81($\pm$18.24) & 46.24($\pm$6.523) & 77.23($\pm$7.074) & 78.30($\pm$3.988) & 86.32($\pm$4.418) & 88.26($\pm$2.449)
& \cellcolor{ourscol}86.65($\pm$6.997) \\
\rowcolor{pct50}
\textbf{50\% AUC $\uparrow$}
& 85.83($\pm$6.436) & 74.41($\pm$2.648) & 90.25($\pm$1.914) & 88.90($\pm$2.552) & 93.85($\pm$3.508) & 96.65($\pm$1.275)
& \cellcolor{ourscol}95.46($\pm$3.106) \\
\midrule

\rowcolor{pct25}
\textbf{25\%  BA $\uparrow$}
& 50.00($\pm$0.000) & 50.00($\pm$0.000) & 61.71($\pm$8.370) & 56.85($\pm$4.967) & 63.89($\pm$6.326) & 62.40($\pm$1.743)
& \cellcolor{ourscol}65.79($\pm$2.882) \\
\rowcolor{pct25}
\textbf{25\%  mF1 $\uparrow$}
& 37.46($\pm$1.110) & 38.07($\pm$1.251) & 58.15($\pm$13.62) & 52.19($\pm$11.27) & 61.88($\pm$9.361) & 62.04($\pm$1.587)
& \cellcolor{ourscol}65.90($\pm$3.074) \\
\rowcolor{pct25}
\textbf{25\% AUC $\uparrow$}
& 59.80($\pm$7.225) & 66.58($\pm$6.896) & 69.76($\pm$12.67) & 68.39($\pm$1.623) & 75.14($\pm$3.625) & 75.87($\pm$5.688)
& \cellcolor{ourscol}75.04($\pm$4.284) \\
\bottomrule
\end{tabular}
}}

\vspace{5pt}

\begin{minipage}{1.0\textwidth}
\scriptsize
\textit{Notes. Values are mean($\pm$std) over three folds. BA, balanced accuracy; mF1, macro F1-score; AUC, macro one-vs-rest ROC AUC. Scratch uses random initialization; SwinUNETR$^{\ast}$, VoCo$^{\ast}$, and S3D$^{\ast}$ use publicly released centralized foundation model weights; S3D uses centralized pretraining on our pretraining corpus, FedMAE and BrainFedFM use federated pretraining.
Fine-tuning uses 8 epochs of AdamW, learning rate $1\times10^{-6}$, weight decay $1\times10^{-2}$, cosine scheduling with a 1-epoch warm-up, and label smoothing of 0.2. The best checkpoint is selected according to validation AUC.}
\end{minipage}
\end{table*}

\clearpage

\noindent{\scriptsize\bfseries Panel B. Labelled-data scaling and aggregate metric summaries.}
\vspace{2pt}
{\tiny\raggedright
Summarizes method behavior across labelled-data settings.
\textbf{B1}, fold-averaged trends for classification and calibration metrics; \textbf{B2}, extended metric heatmaps separating higher-is-better classification metrics from lower-is-better calibration/uncertainty metrics; \textbf{B3}, the accuracy--calibration trade-off, where points closer to the upper-left indicate stronger discrimination with better calibration.\par}
\vspace{-3pt}

\begin{center}
\includegraphics[width=1.0\textwidth,keepaspectratio]{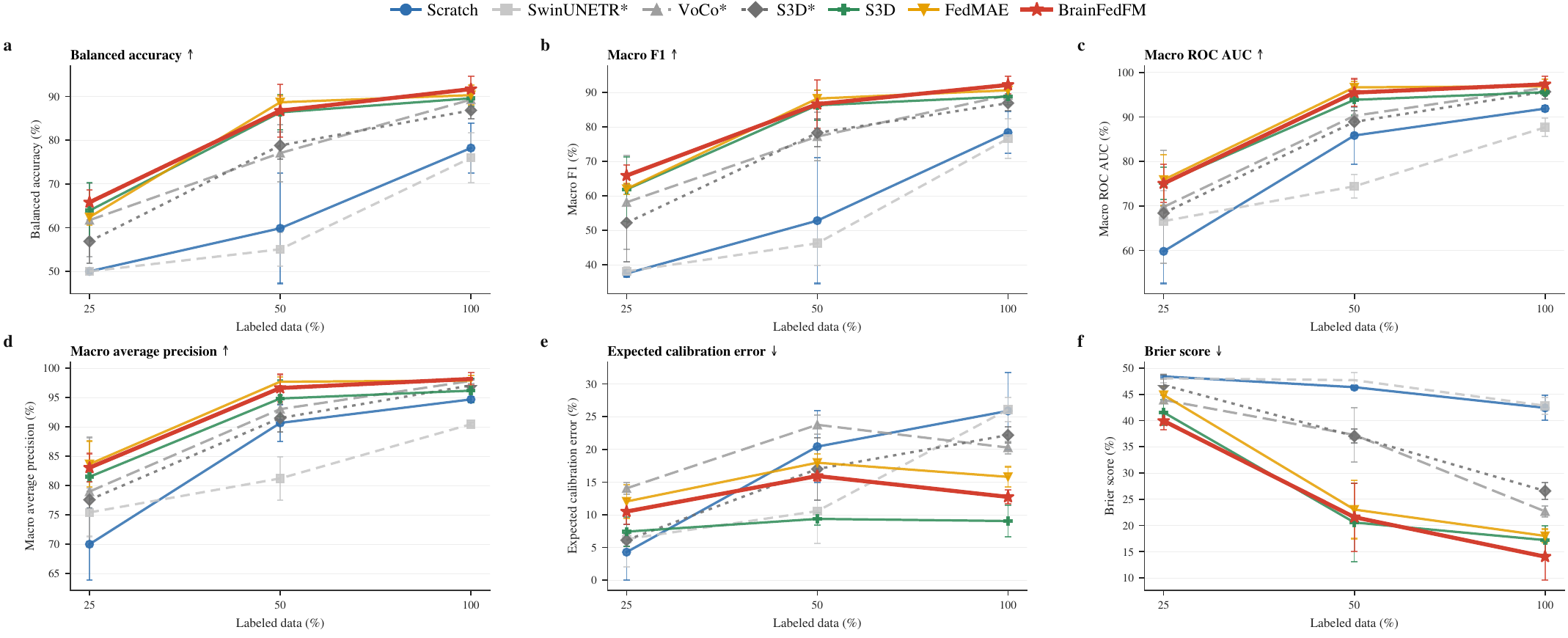} \par\vspace{3pt}
\includegraphics[width=1.0\textwidth,keepaspectratio]{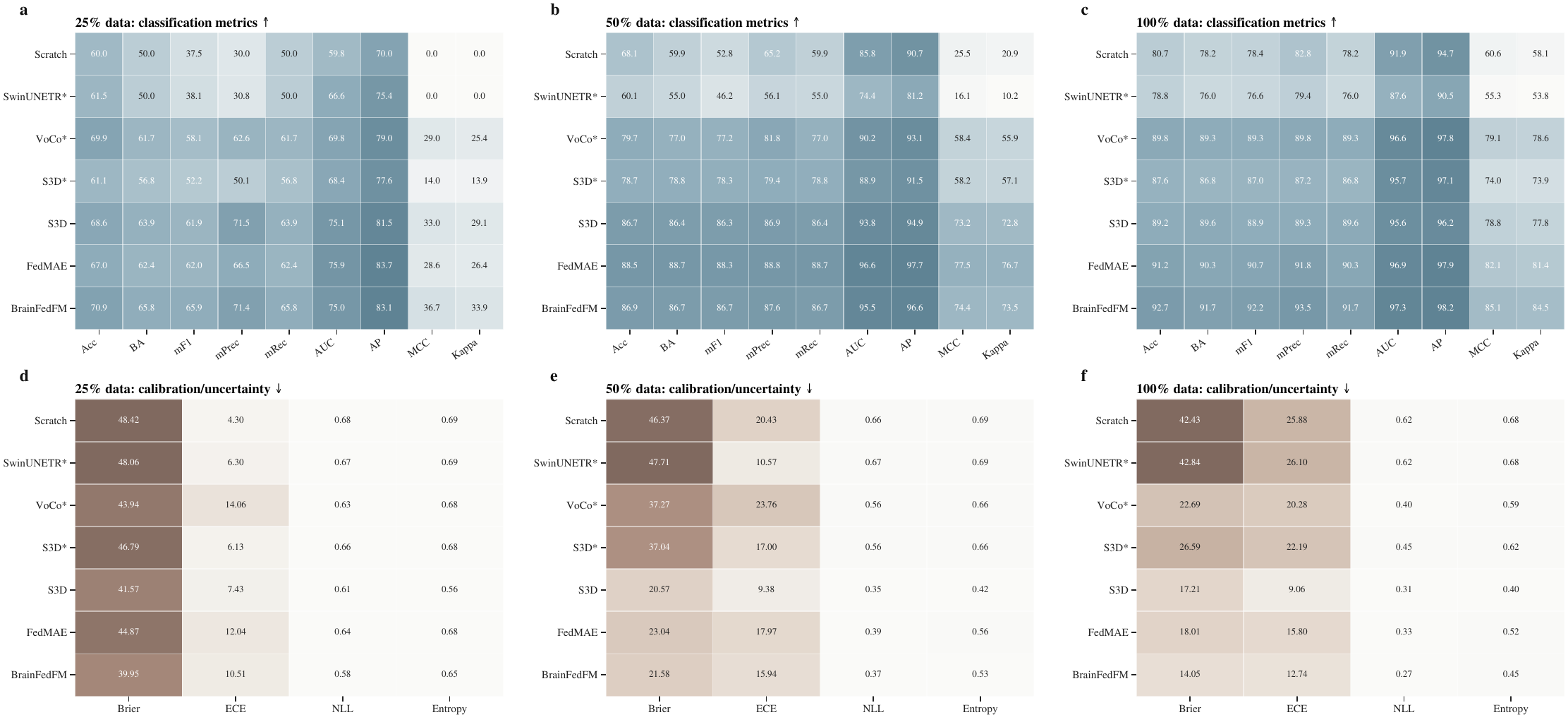} \par\vspace{5pt}
\includegraphics[width=0.935\textwidth,keepaspectratio]{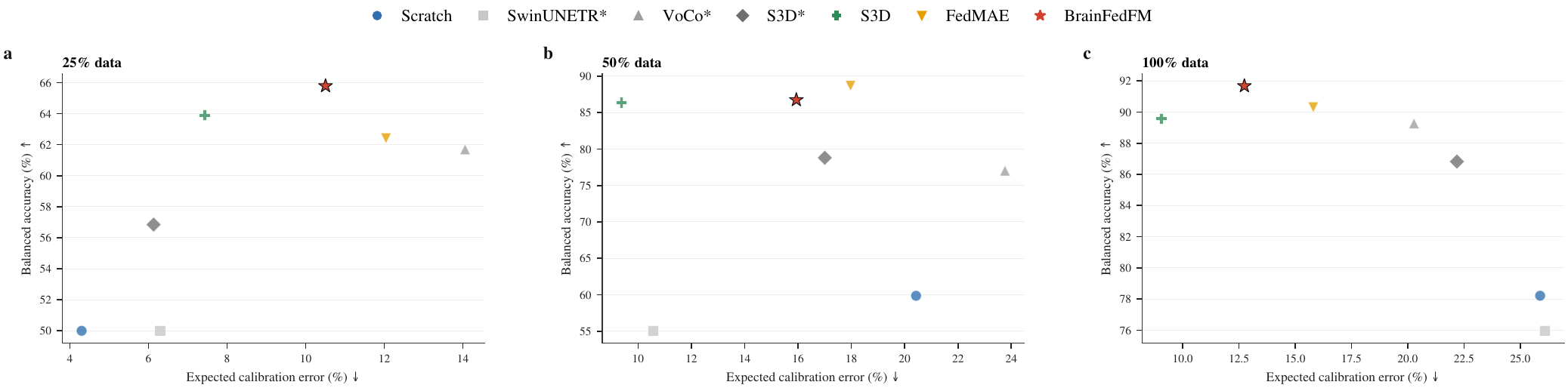}
\end{center}

\vspace{-2pt}

\noindent{\scriptsize\bfseries Panel C. Fold stability and full-data performance analyses.}
\vspace{2pt}
{\tiny\raggedright
Examines robustness and error modes beyond the aggregate scores. \textbf{C1}, fold-to-fold variability and average method ranking across balanced accuracy, macro F1, and macro ROC AUC; \textbf{C2}, probability-level analysis at 100\% labelled data, including micro-averaged ROC/precision--recall curves, reliability curves, and calibration scores; \textbf{C3}, row-normalized confusion matrices at 100\% labelled data, with rows denoting true parkinsonism subtypes and columns denoting predicted subtypes.\par}

\clearpage

\begin{center}
\includegraphics[width=0.838\textwidth,keepaspectratio]{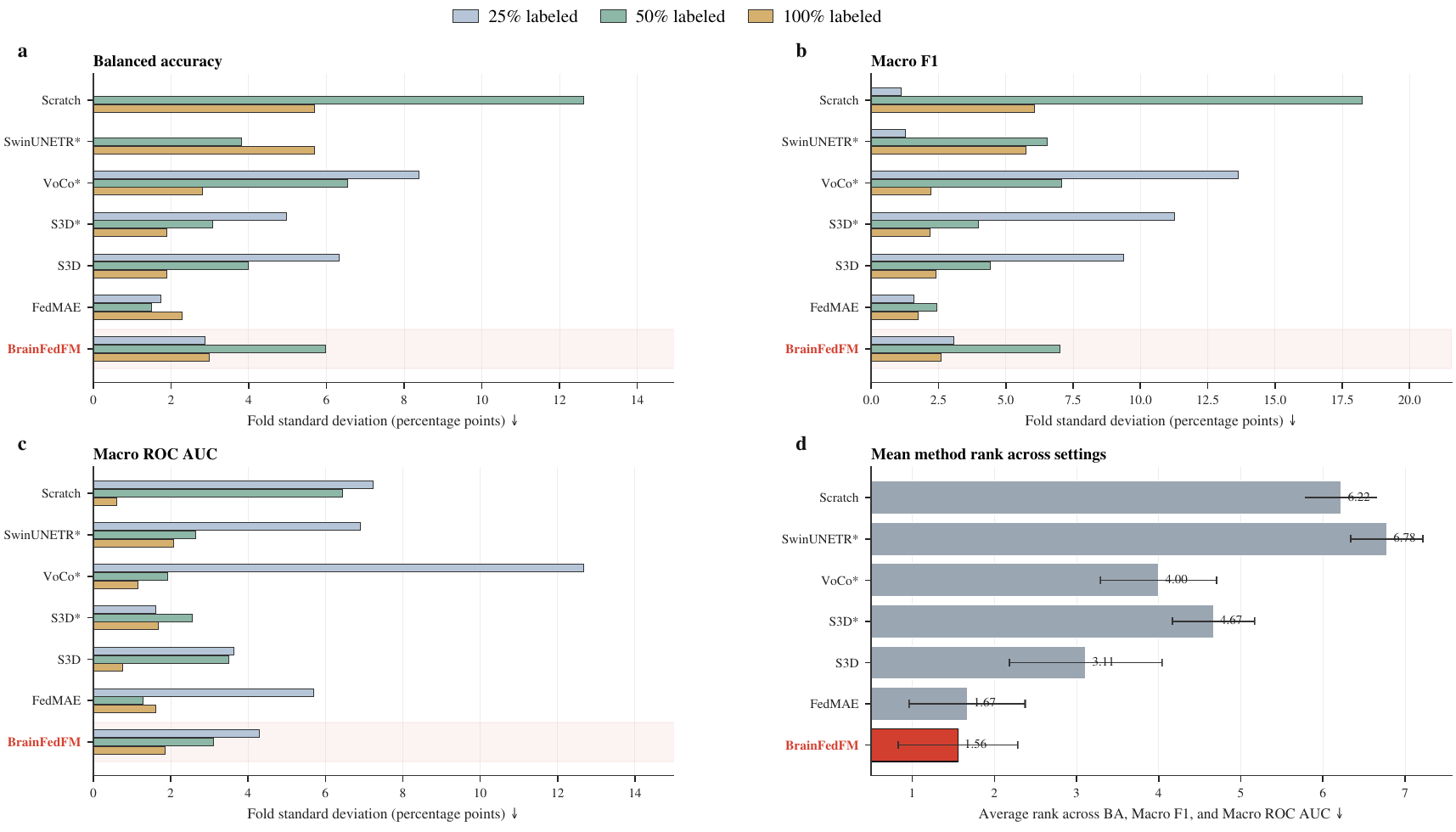}   \par\vspace{3pt}
\includegraphics[width=0.785\textwidth,keepaspectratio]{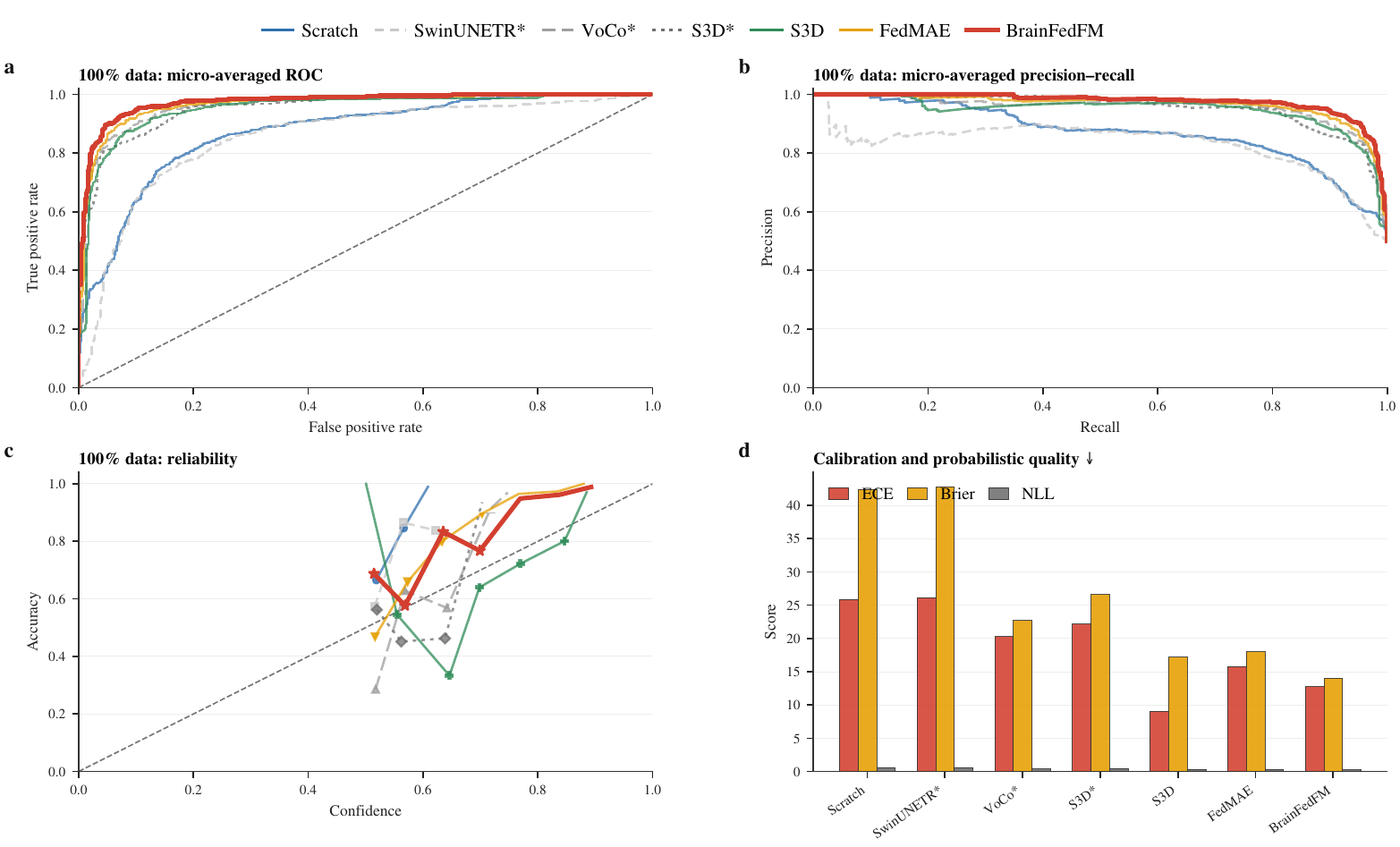}  \par\vspace{4pt}
\includegraphics[width=0.83\textwidth,keepaspectratio]{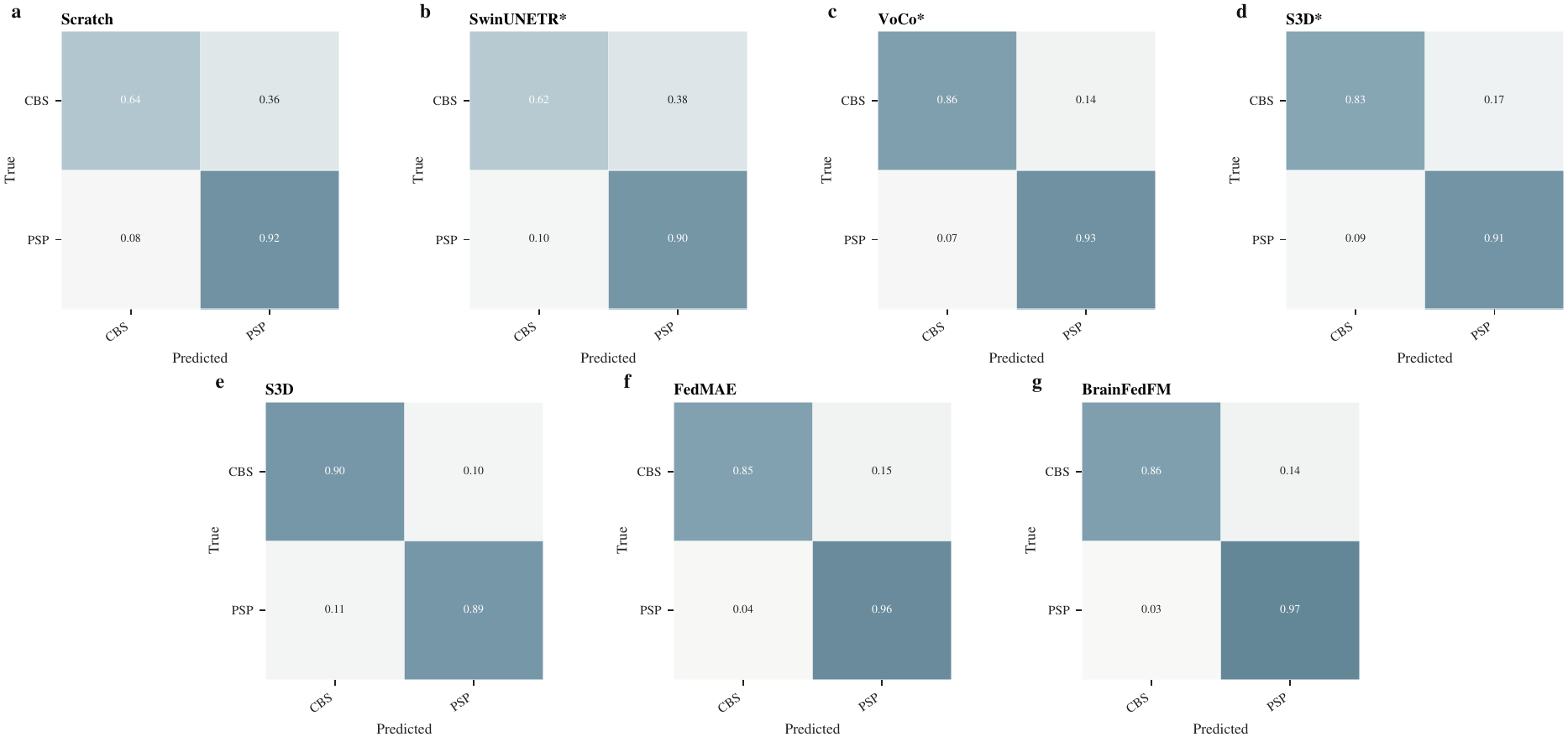}
\end{center}

\definecolor{pct100}{gray}{0.97}
\definecolor{pct75}{gray}{1.0}
\definecolor{pct50}{gray}{0.99}
\definecolor{pct25}{gray}{1.0}
\definecolor{overallgray}{gray}{1.0}
\definecolor{ourscol}{RGB}{244,248,252}

\begin{table*}[t]
\centering
\tiny
\caption[Tumor radiogenomic prediction task]{\textbf{Tumor radiogenomic prediction task.}
We evaluate foundation model performance on the RSNA-MICCAI benchmark. This task aims to classify brain MRI scans into MGMT promoter methylated and unmethylated categories.
Accurate tumor radiogenomic prediction from brain MRI is important for neuro-oncology analysis, supporting non-invasive molecular characterization, and patient stratification in brain tumor studies.
The table summarizes performance under 25\%, 50\%, and 100\% labelled-data settings, which indicate the proportion of the labelled training--validation pool used for downstream fine-tuning and model selection; the held-out test set remains unchanged across all settings. We compare a randomly initialized scratch baseline, centralized pretrained foundation models, and federated pretrained foundation models.}
\label{tab:result_rsna}

\renewcommand{\arraystretch}{3.85}
{\scriptsize\bfseries Panel A. Summary under different labelled-data percentages.}
\vspace{5pt}

{\setlength{\tabcolsep}{3.38pt}

\resizebox{1.0\textwidth}{!}{
\begin{tabular}{
l c c c c c c
>{\columncolor{ourscol}\centering\arraybackslash}p{1.62cm}
}
\toprule
\multicolumn{2}{c}{\textbf{Brain sMRI FMs }}
& \multicolumn{4}{c}{Centralized Pretraining}
& \multicolumn{2}{c}{Federated Pretraining} \\
\cmidrule(lr){1-2}\cmidrule(lr){3-6}\cmidrule(lr){7-8}
\makecell[c]{Metric/Methods}
& Scratch
& \makecell[c]{SwinUNETR$^{\ast}$}  &\ VoCo$^{\ast}$ \ & \ S3D$^{\ast}$  & \ S3D 
& \makecell[c]{FedMAE} & \makecell[c]{BrainFedFM} \\
\midrule

\rowcolor{pct100}
\textbf{100\%  BA $\uparrow$}
& 55.92($\pm$2.353) &  51.17($\pm$1.661) & 57.84($\pm$2.176) & 50.87($\pm$1.272) & 54.22($\pm$3.036) & 53.74($\pm$2.973)
& \cellcolor{ourscol} 56.10($\pm$2.346) \\
\rowcolor{pct100}
\textbf{100\%   mF1 $\uparrow$}
& 54.87($\pm$3.038) &  38.19($\pm$5.525)  & 57.25($\pm$2.118) & 41.26($\pm$7.900) & 48.60($\pm$10.12) & 47.81($\pm$11.01)
& \cellcolor{ourscol} 55.17($\pm$2.419) \\
\rowcolor{pct100}
\textbf{100\%   AUC $\uparrow$}
& 60.50($\pm$2.176) &  61.10($\pm$1.468)  &  61.98($\pm$2.609) & 56.58($\pm$2.316) & 58.01($\pm$2.061) & 59.26($\pm$0.872)
& \cellcolor{ourscol} 60.67($\pm$2.200) \\
\midrule

\rowcolor{pct50}
\textbf{50\%  BA $\uparrow$}
& 50.25($\pm$0.348) & 49.12($\pm$0.962)  & 55.52($\pm$0.504) & 51.09($\pm$1.166) & 55.18($\pm$1.315) & 55.83($\pm$2.413)
& \cellcolor{ourscol} 56.19($\pm$2.239)\\
\rowcolor{pct50}
\textbf{50\%  mF1 $\uparrow$}
& 36.33($\pm$2.896)& 33.72($\pm$1.000)  & 55.23($\pm$0.604) & 45.09($\pm$7.689) & 50.74($\pm$5.830) & 55.32($\pm$1.942)
& \cellcolor{ourscol} 51.78($\pm$5.024) \\
\rowcolor{pct50}
\textbf{50\% AUC $\uparrow$}
&  52.85($\pm$2.006) & 55.88($\pm$1.281)  & 59.16($\pm$0.822) & 54.91($\pm$2.058) & 58.42($\pm$1.751) &  57.83($\pm$2.299)
& \cellcolor{ourscol} 61.33($\pm$0.975) \\
\midrule

\rowcolor{pct25}
\textbf{25\%  BA $\uparrow$}
& 50.03($\pm$0.036) &  50.13($\pm$0.587)  & 51.58($\pm$1.394) & 52.53($\pm$3.788) & 52.99($\pm$4.905) & 51.24($\pm$3.229)
& \cellcolor{ourscol}   54.19($\pm$1.554)\\
\rowcolor{pct25}
\textbf{25\%  mF1 $\uparrow$}
& 34.81($\pm$0.741) & 39.01($\pm$6.824)  & 44.71($\pm$7.436) & 46.61($\pm$9.191) & 52.86($\pm$4.800) & 50.70($\pm$2.921)
& \cellcolor{ourscol} 53.85($\pm$1.784) \\
\rowcolor{pct25}
\textbf{25\% AUC $\uparrow$}
& 50.47($\pm$1.433) & 51.78($\pm$2.312)  &  57.35($\pm$5.571) &  56.14($\pm$5.359) & 54.42($\pm$6.591) &  54.63($\pm$1.973)
& \cellcolor{ourscol} 54.53($\pm$1.353) \\
\bottomrule
\end{tabular}
}}

\vspace{5pt}

\begin{minipage}{1.0\textwidth}
\scriptsize
\textit{Notes. Values are mean($\pm$std) over three folds. BA, balanced accuracy; mF1, macro F1-score; AUC, macro one-vs-rest ROC AUC. Scratch uses random initialization; SwinUNETR$^{\ast}$, VoCo$^{\ast}$, and S3D$^{\ast}$ use publicly released centralized foundation model weights; S3D uses centralized pretraining on our pretraining corpus, FedMAE and BrainFedFM use federated pretraining.
Fine-tuning uses 8 epochs of AdamW, learning rate $1\times10^{-6}$, weight decay $1\times10^{-2}$, cosine scheduling with a 1-epoch warm-up, and label smoothing of 0.2. The best checkpoint is selected according to validation AUC.}
\end{minipage}
\end{table*}

\clearpage

\noindent{\scriptsize\bfseries Panel B. Labelled-data scaling and aggregate metric summaries.}
\vspace{2pt}
{\tiny\raggedright
Summarizes method behavior across labelled-data settings.
\textbf{B1}, fold-averaged trends for classification and calibration metrics; \textbf{B2}, extended metric heatmaps separating higher-is-better classification metrics from lower-is-better calibration/uncertainty metrics; \textbf{B3}, the accuracy--calibration trade-off, where points closer to the upper-left indicate stronger discrimination with better calibration.\par}
\vspace{-3pt}

\begin{center}
\includegraphics[width=1.0\textwidth,keepaspectratio]{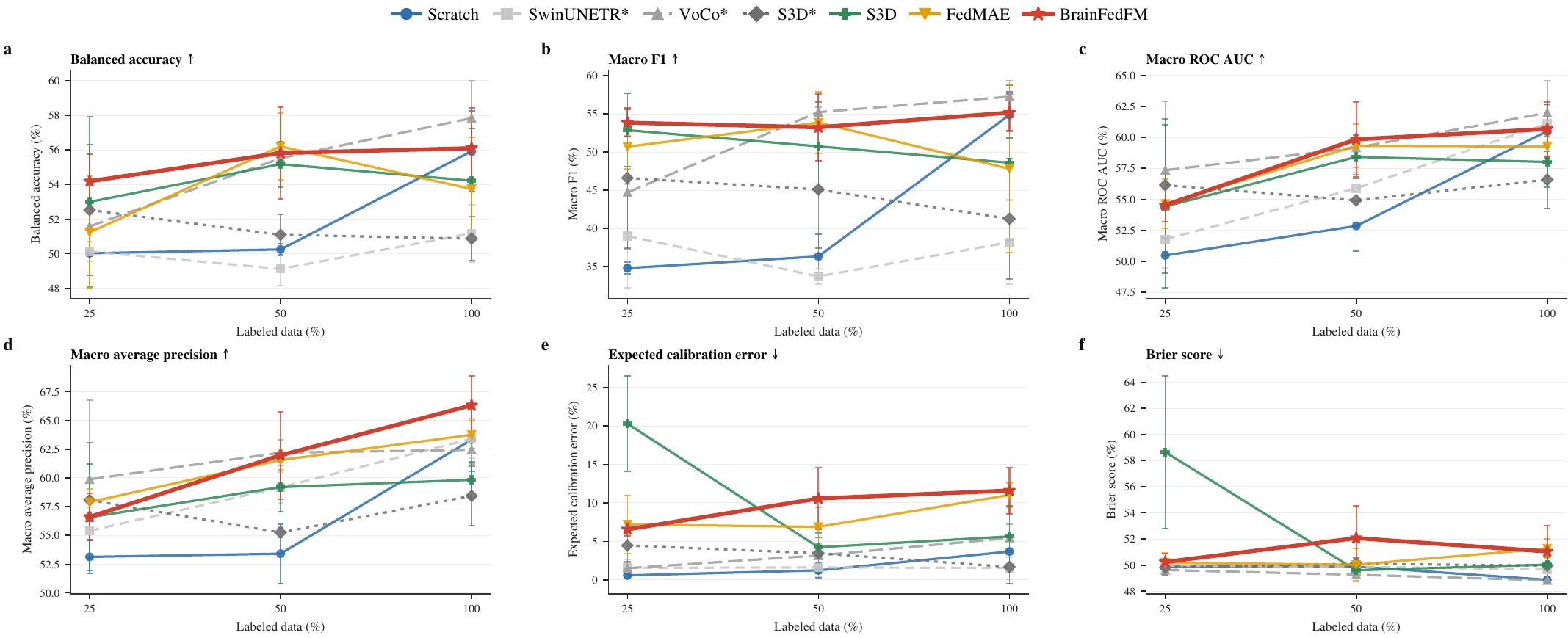} \par\vspace{3pt}
\includegraphics[width=1.0\textwidth,keepaspectratio]{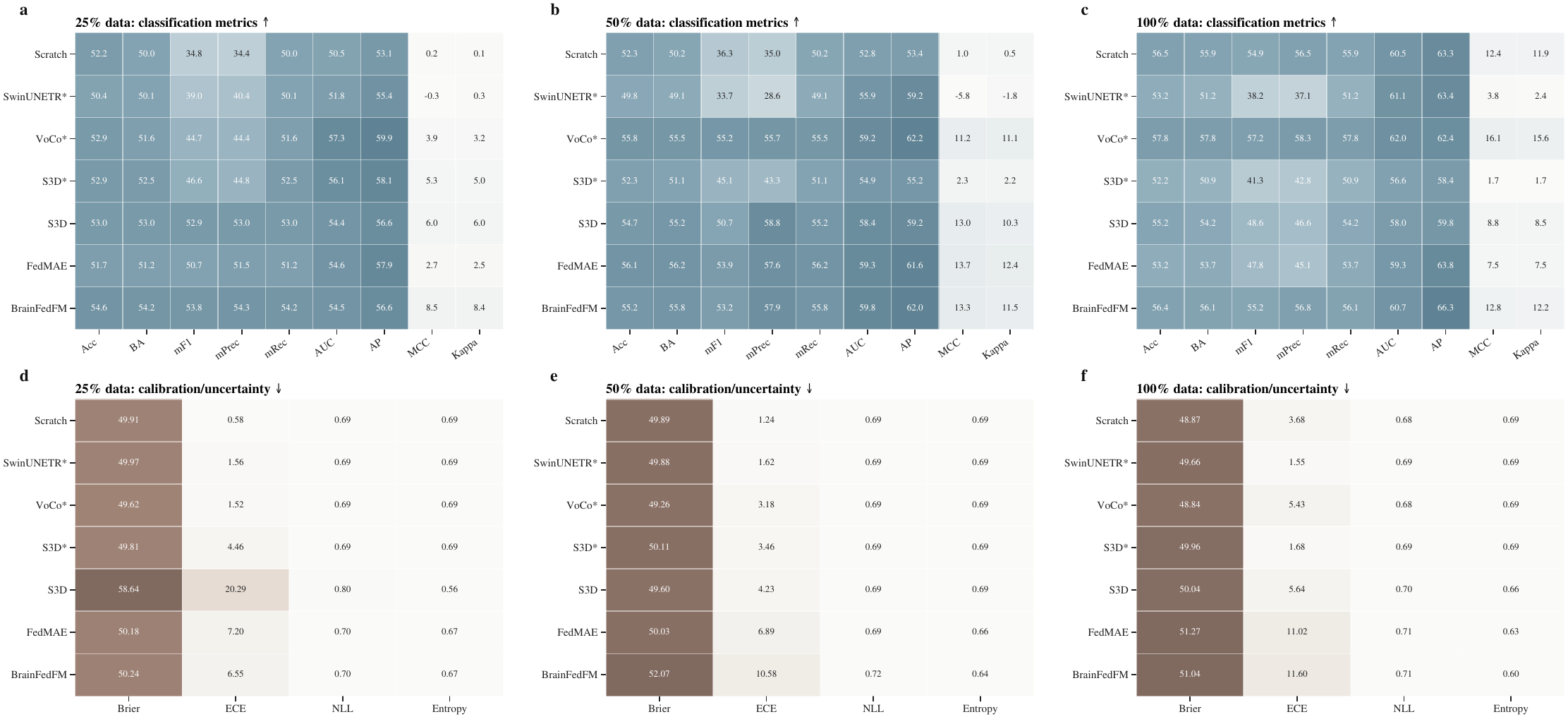} \par\vspace{5pt}
\includegraphics[width=0.935\textwidth,keepaspectratio]{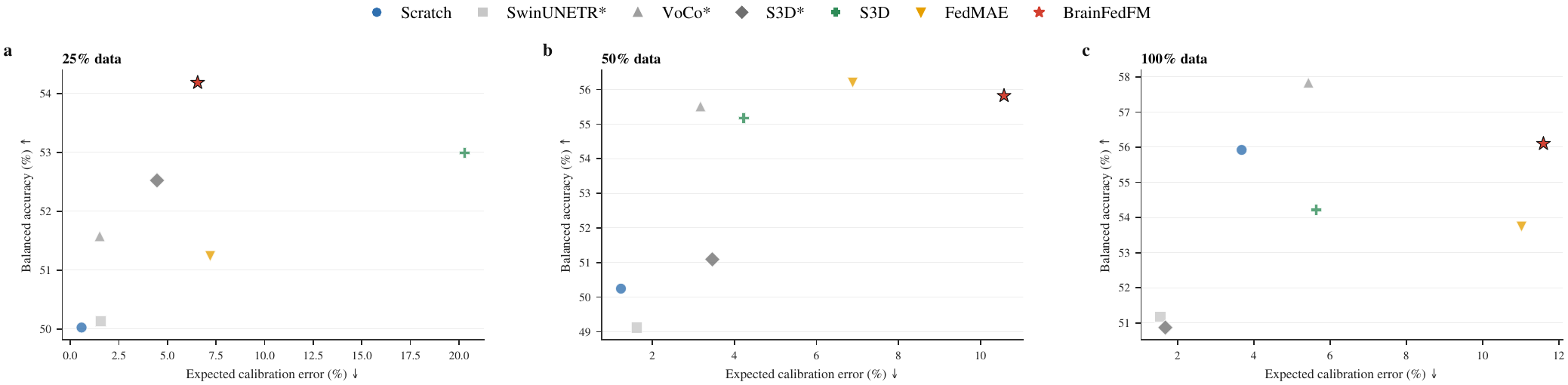}
\end{center}

\vspace{-2pt}

\noindent{\scriptsize\bfseries Panel C. Fold stability and full-data performance analyses.}
\vspace{2pt}
{\tiny\raggedright
Examines robustness and error modes beyond the aggregate scores. \textbf{C1}, fold-to-fold variability and average method ranking across balanced accuracy, macro F1, and macro ROC AUC; \textbf{C2}, probability-level analysis at 100\% labelled data, including micro-averaged ROC/precision--recall curves, reliability curves, and calibration scores; \textbf{C3}, row-normalized confusion matrices at 100\% labelled data, with rows denoting true methylation-status classes and columns denoting predicted classes.\par}

\clearpage

\begin{center}
\includegraphics[width=0.838\textwidth,keepaspectratio]{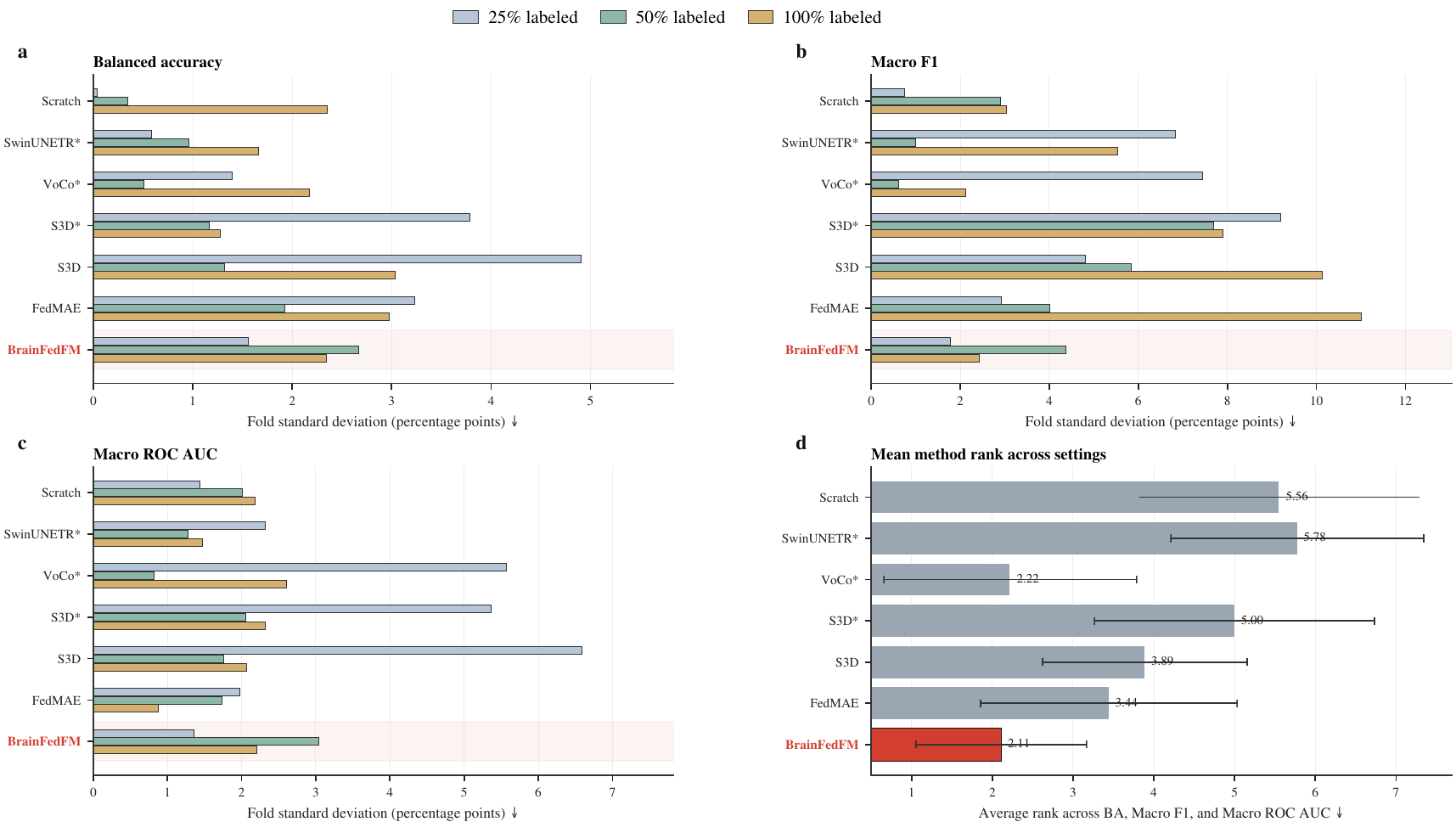}   \par\vspace{6pt}
\includegraphics[width=0.785\textwidth,keepaspectratio]{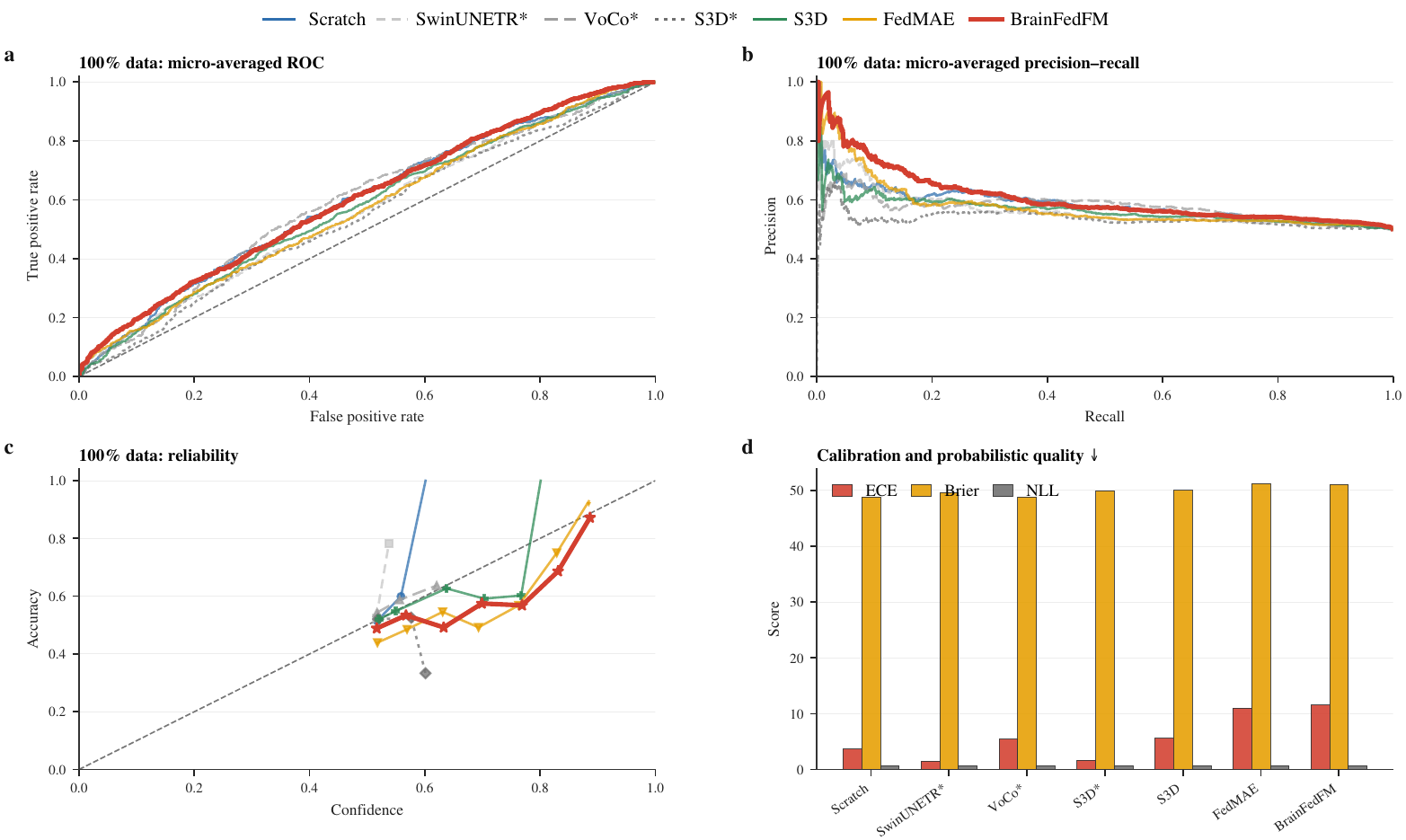}  \par\vspace{4pt}
\includegraphics[width=0.83\textwidth,keepaspectratio]{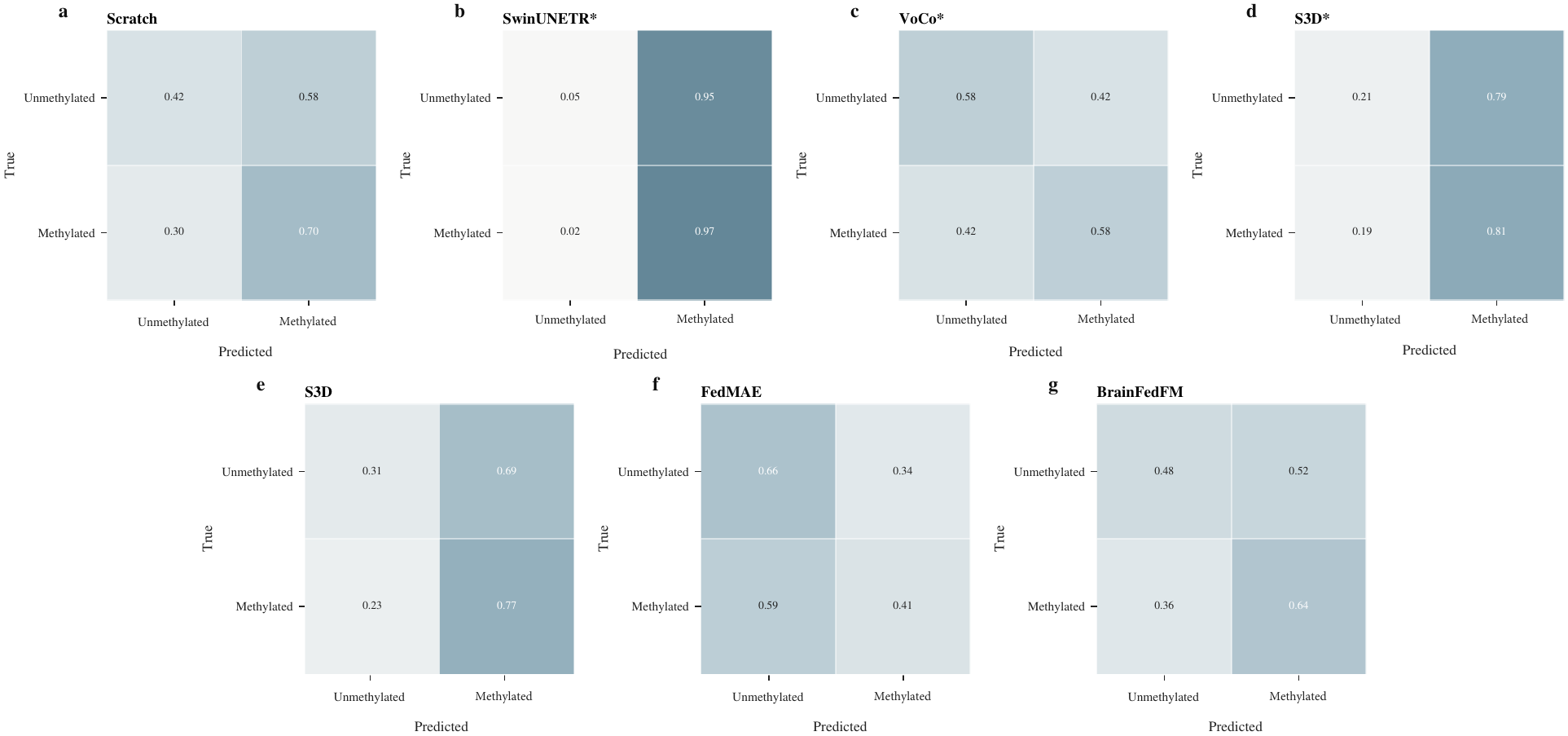}
\end{center}

\definecolor{pct100}{gray}{0.97}
\definecolor{pct75}{gray}{1.0}
\definecolor{pct50}{gray}{0.99}
\definecolor{pct25}{gray}{1.0}
\definecolor{overallgray}{gray}{1.0}
\definecolor{ourscol}{RGB}{244,248,252}

\begin{table*}[t]
\centering
\tiny
\caption[Adult brain age prediction task]{\textbf{Adult brain age prediction task.}
We evaluate foundation model performance on the SALD benchmark for estimating chronological age from structural brain MRI of healthy people.
This task captures age-related neuroanatomical variation for assessing the transferability of learned structural representations, and it evaluates whether foundation models capture subtle, spatially distributed aging patterns and transfer them effectively to a biologically meaningful continuous prediction task.
The table summarizes performance under 25\%, 50\%, and 100\% labelled-data settings, which indicate the proportion of the labelled training--validation pool used for downstream fine-tuning and model selection; the held-out test set remains unchanged across all settings.
We compare a randomly initialized scratch baseline, centralized pretrained foundation models, and federated pretrained foundation models.}
\label{tab:result_sald}

\renewcommand{\arraystretch}{3.8}
{\scriptsize\bfseries Panel A. Summary under different labelled-data percentages.}
\vspace{5pt}

{\setlength{\tabcolsep}{3.38pt}

\resizebox{1.0\textwidth}{!}{
\begin{tabular}{
l c c c c c c
>{\columncolor{ourscol}\centering\arraybackslash}p{1.62cm}
}
\toprule
\multicolumn{2}{c}{\textbf{Brain sMRI FMs }}
& \multicolumn{4}{c}{Centralized Pretraining}
& \multicolumn{2}{c}{Federated Pretraining} \\
\cmidrule(lr){1-2}\cmidrule(lr){3-6}\cmidrule(lr){7-8}
\makecell[c]{Metric/Methods}
& Scratch
& \makecell[c]{SwinUNETR$^{\ast}$}  &\ VoCo$^{\ast}$ \ & \ S3D$^{\ast}$  & \ S3D 
& \makecell[c]{FedMAE} & \makecell[c]{BrainFedFM} \\
\midrule

\rowcolor{pct100}
\textbf{100\%  RMSE $\downarrow$}
& 9.213($\pm$0.413) & 10.32($\pm$0.191)  & 11.31($\pm$0.843) & 14.24($\pm$0.339) & 13.22($\pm$0.672) & 8.779($\pm$0.423)
& \cellcolor{ourscol} 8.649($\pm$0.314) \\
\rowcolor{pct100}
\textbf{100\%   MAE $\downarrow$}
& 6.717($\pm$0.329) & 8.034($\pm$0.167)  & 8.720($\pm$0.531) & 12.60($\pm$0.387) & 11.22($\pm$0.979) & 6.405($\pm$0.304)
& \cellcolor{ourscol} 6.310($\pm$0.039) \\
\rowcolor{pct100}
\textbf{100\%   R$^2$ $\uparrow$}
& 0.742($\pm$0.023) & 0.677($\pm$0.012)  & 0.609($\pm$0.059) &  0.384($\pm$0.029) & 0.468($\pm$0.055) & 0.765($\pm$0.022)
& \cellcolor{ourscol} 0.773($\pm$0.017) \\
\midrule

\rowcolor{pct50}
\textbf{50\%  RMSE $\downarrow$}
& 13.08($\pm$0.921) & 21.63($\pm$0.434)  & 21.94($\pm$0.463)  & 19.43($\pm$0.059) & 19.01($\pm$0.211) & 13.09($\pm$0.498)
& \cellcolor{ourscol}  12.48($\pm$0.757) \\
\rowcolor{pct50}
\textbf{50\%   MAE $\downarrow$}
& 10.89($\pm$0.976) & 18.35($\pm$0.202)  & 18.50($\pm$0.222) & 17.16($\pm$0.036)  & 16.89($\pm$0.124) & 10.41($\pm$0.829)
& \cellcolor{ourscol}  9.925($\pm$0.649) \\
\rowcolor{pct50}
\textbf{50\%  R$^2$ $\uparrow$}
& 0.478($\pm$0.072) & -0.422($\pm$0.057)   &  -0.463($\pm$0.061) &  -0.147($\pm$0.007)  & -0.097($\pm$0.024) & 0.479($\pm$0.040)
& \cellcolor{ourscol}  0.525($\pm$0.058) \\
\midrule

\rowcolor{pct25}
\textbf{25\%  RMSE $\downarrow$}
&  17.90($\pm$1.249) & 35.35($\pm$0.192) & 35.94($\pm$0.681) & 34.38($\pm$0.517) & 29.44($\pm$0.112) & 18.39($\pm$0.315) 
& \cellcolor{ourscol} 18.21($\pm$0.032) \\
\rowcolor{pct25}
\textbf{25\% MAE $\downarrow$}
& 15.95($\pm$1.145) & 30.30($\pm$0.223) & 30.98($\pm$0.790) & 29.20($\pm$0.608)& 23.31($\pm$0.124) & 16.53($\pm$0.191)
& \cellcolor{ourscol} 16.42($\pm$0.039) \\
\rowcolor{pct25}
\textbf{25\% R$^2$ $\uparrow$}
& 0.022($\pm$0.134) &  -2.794($\pm$0.041)  &  -2.925($\pm$0.149) & -2.590($\pm$0.108) &  -1.632($\pm$0.020) &  -0.028($\pm$0.035)
& \cellcolor{ourscol}  -0.007($\pm$0.004) \\
\bottomrule
\end{tabular}
}}

\vspace{5pt}

\begin{minipage}{1.0\textwidth}
\scriptsize
\textit{Notes. RMSE, root mean squared error; MAE, mean absolute error; R$^2$, coefficient of determination. RMSE and MAE are reported in years; lower values indicate better performance, higher R$^2$ indicates better performance. Scratch uses random initialization; SwinUNETR$^{\ast}$, VoCo$^{\ast}$, and S3D$^{\ast}$ use publicly released centralized foundation models; S3D uses centralized pretraining on our pretraining corpus, FedMAE and BrainFedFM use federated pretraining. Fine-tuning uses 25 epochs of AdamW, a learning rate of $1\times10^{-5}$, weight decay of $1\times10^{-2}$, and cosine scheduling with a 3-epoch warm-up.}
\end{minipage}

\end{table*}

\clearpage

\begin{figure*}[p]
\noindent{\scriptsize\bfseries
Panel B. Labelled-data scaling and full-data regression analyses.}
{\tiny\raggedright
Characterizes brain age prediction performance from complementary perspectives.
\textbf{B1}, fold-averaged RMSE, MAE, R$^2$, Pearson correlation, Spearman correlation, and concordance correlation coefficient.
\textbf{B2}, cross-method analyses including error magnitude, association and agreement, signed age-error distributions, and MAE across the observed chronological-age range.
\textbf{B3}, the joint RMSE--CCC performance landscape across labelled-data settings, where lower RMSE and higher CCC indicate a more favorable error--agreement trade-off.\par}

\par\vspace{8pt}

\includegraphics[
    width=\textwidth,
    keepaspectratio
]{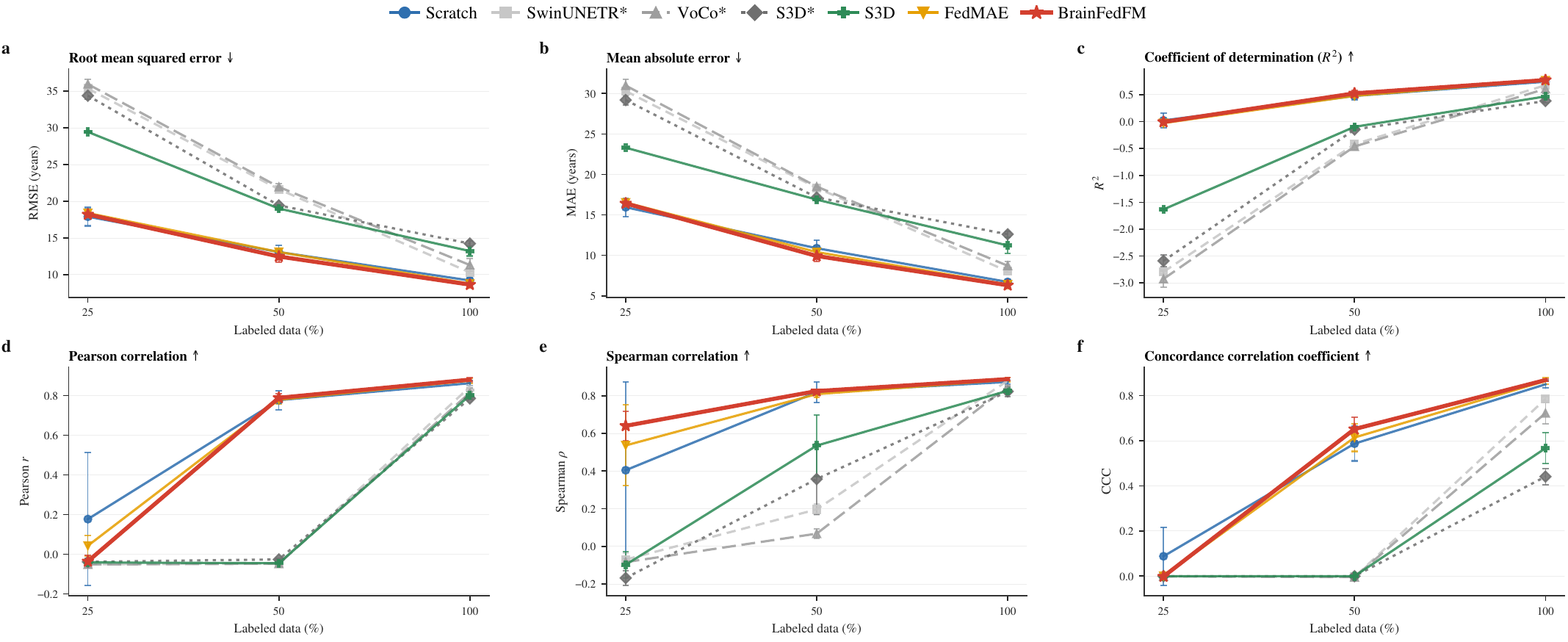}

\par\vspace{3pt}

\includegraphics[
    width=0.84\textwidth,
    keepaspectratio
]{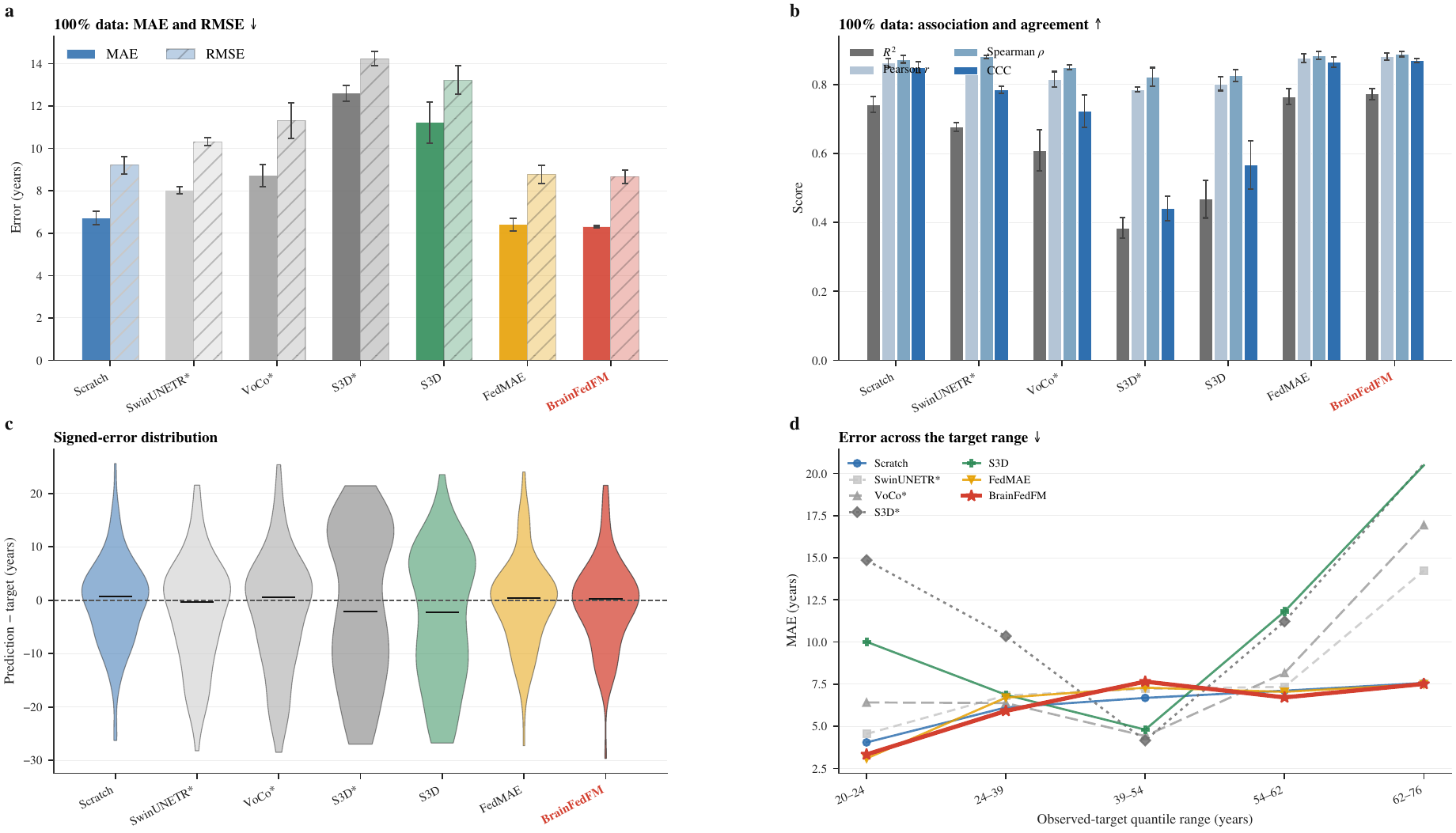}

\par\vspace{4pt}

\includegraphics[
    width=0.96\textwidth,
    keepaspectratio
]{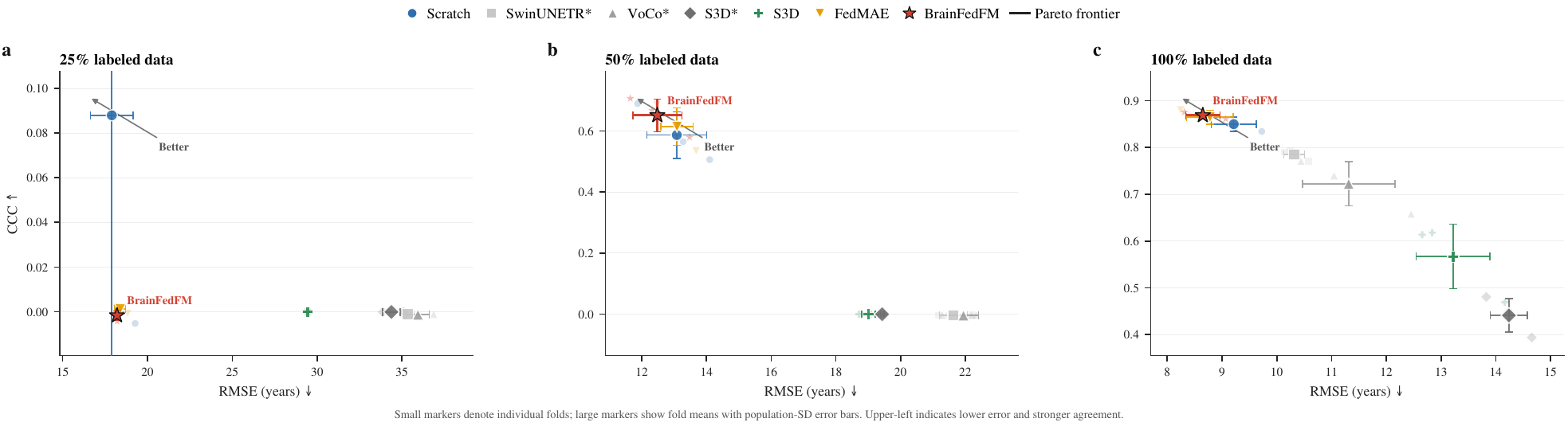}

\end{figure*}

\clearpage

\begin{figure*}[p]
\noindent{\scriptsize\bfseries
Panel C. Relative ranking, fold stability, and transfer-benefit analyses.}
{\tiny\raggedright
Summarizes comparative performance, robustness, and transfer benefit beyond the aggregate brain age regression scores.
\textbf{C1}, relative method ranks across all labelled-data settings and six regression metrics, with rank 1 denoting the best result within each setting--metric combination.
\textbf{C2}, fold-to-fold variability in RMSE, MAE, and R$^2$, together with the average method rank across the core regression metrics.
\textbf{C3}, performance changes relative to the matched Scratch baseline, where positive values denote RMSE or MAE reduction and improvement in R$^2$.\par}

\par\vspace{5.5pt}

\includegraphics[
    width=\textwidth,
    keepaspectratio
]{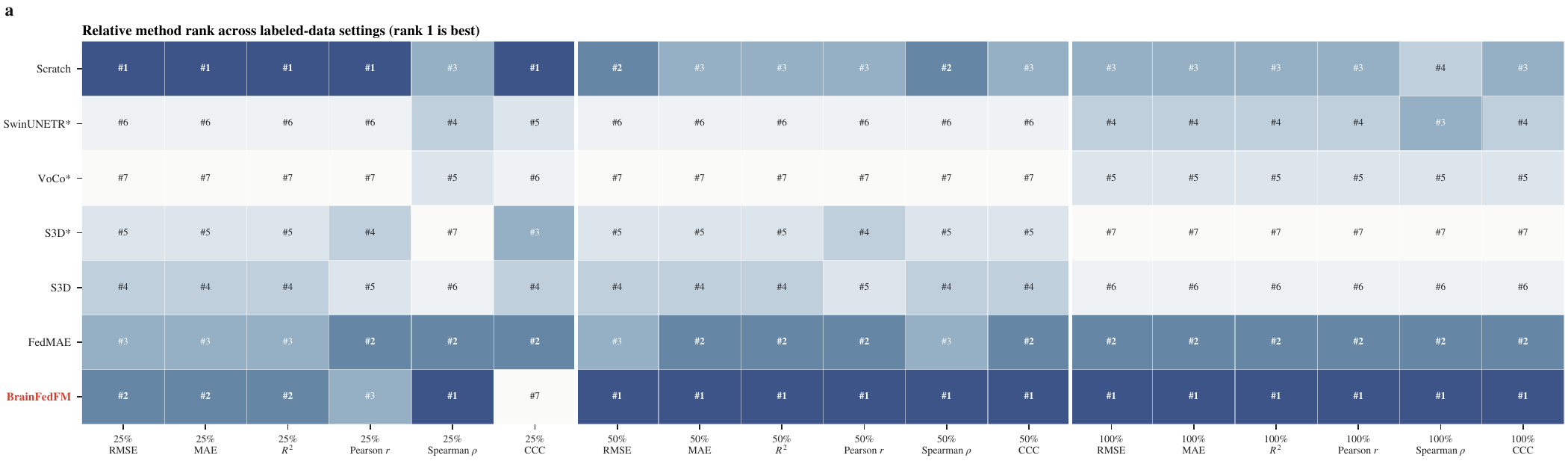}

\vspace{6pt}

\includegraphics[
    width=\textwidth,
    keepaspectratio
]{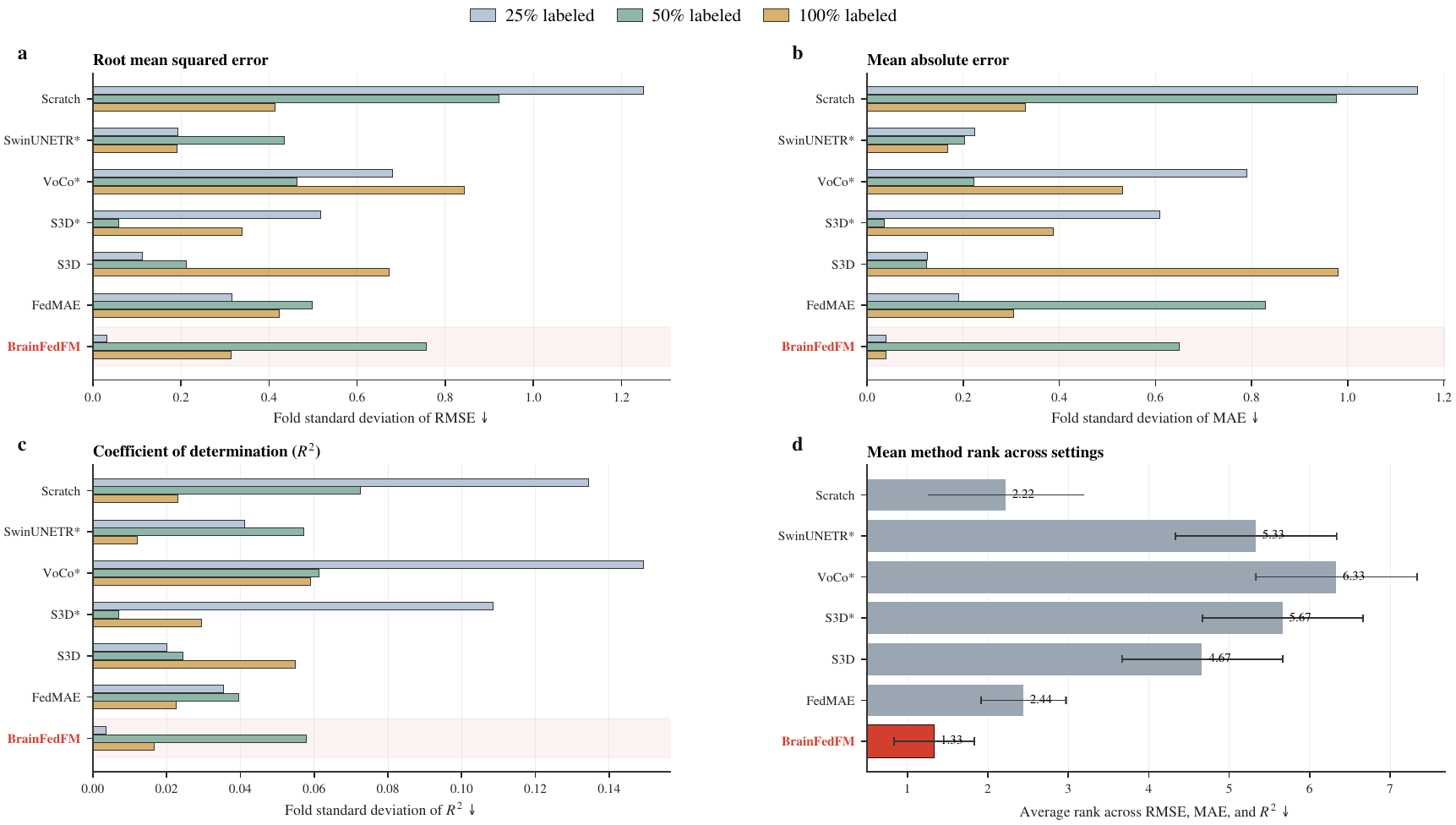}

\vspace{6pt}

\includegraphics[
    width=\textwidth,
    keepaspectratio
]{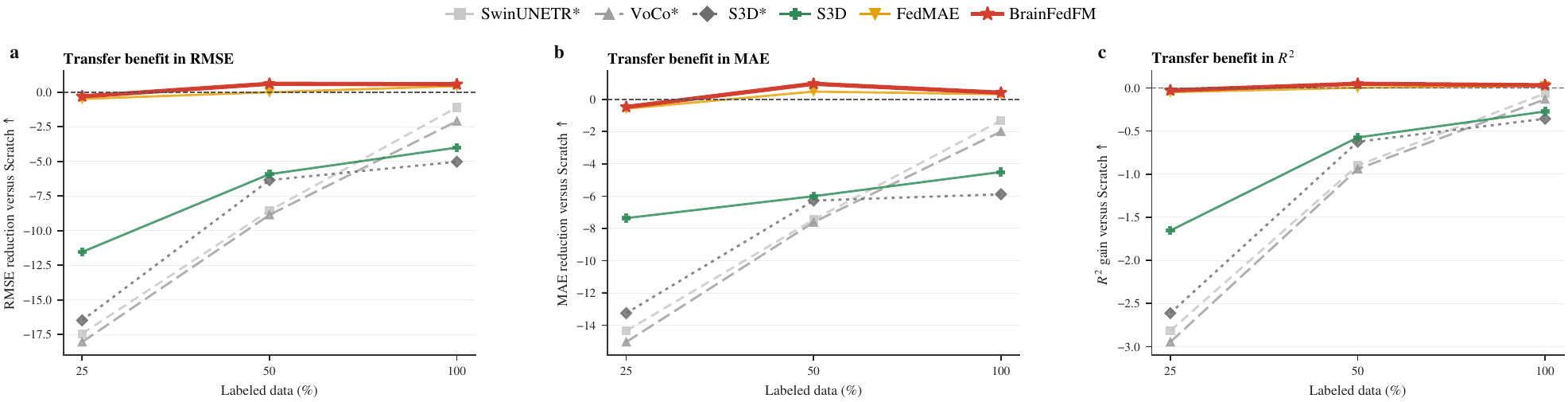}

\end{figure*}
\definecolor{pct100}{gray}{0.97}
\definecolor{pct75}{gray}{1.0}
\definecolor{pct50}{gray}{0.99}
\definecolor{pct25}{gray}{1.0}
\definecolor{overallgray}{gray}{1.0}
\definecolor{ourscol}{RGB}{244,248,252}

\begin{table*}[t]
\centering
\tiny
\caption[Lifespan brain age prediction task]{\textbf{Lifespan brain age prediction task.}
We evaluate foundation model performance on the NKI-RS benchmark for estimating chronological age from structural brain MRI of healthy people.
This task captures age-related neuroanatomical variation for assessing the transferability of learned structural representations, and it evaluates whether foundation models capture subtle, spatially distributed aging patterns and transfer them effectively to a biologically meaningful continuous prediction task.
The table summarizes performance under 25\%, 50\%, and 100\% labelled-data settings, which indicate the proportion of the labelled training--validation pool used for downstream fine-tuning and model selection; the held-out test set remains unchanged across all settings.
We compare a randomly initialized scratch baseline, centralized pretrained foundation models, and federated pretrained foundation models.}
\label{tab:result_nki}

\renewcommand{\arraystretch}{3.8}
{\scriptsize\bfseries Panel A. Summary under different labelled-data percentages.}
\vspace{5pt}

{\setlength{\tabcolsep}{3.38pt}

\resizebox{1.0\textwidth}{!}{
\begin{tabular}{
l c c c c c c
>{\columncolor{ourscol}\centering\arraybackslash}p{1.62cm}
}
\toprule
\multicolumn{2}{c}{\textbf{Brain sMRI FMs }}
& \multicolumn{4}{c}{Centralized Pretraining}
& \multicolumn{2}{c}{Federated Pretraining} \\
\cmidrule(lr){1-2}\cmidrule(lr){3-6}\cmidrule(lr){7-8}
\makecell[c]{Metric/Methods}
& Scratch
& \makecell[c]{SwinUNETR$^{\ast}$}  &\ VoCo$^{\ast}$ \ & \ S3D$^{\ast}$  & \ S3D 
& \makecell[c]{FedMAE} & \makecell[c]{BrainFedFM} \\
\midrule

\rowcolor{pct100}
\textbf{100\%  RMSE $\downarrow$}
& 7.670($\pm$0.043) &  7.575($\pm$0.334) & 7.615($\pm$0.644) & 11.96($\pm$0.703)  & 9.222($\pm$0.516) & 6.847($\pm$0.044)
& \cellcolor{ourscol} 6.795($\pm$0.234)\\
\rowcolor{pct100}
\textbf{100\%   MAE $\downarrow$}
& 5.668($\pm$0.120) &  5.793($\pm$0.312) & 5.800($\pm$0.618) & 9.858($\pm$0.614)  & 7.410($\pm$0.439) & 4.960($\pm$0.086)
& \cellcolor{ourscol}  4.881($\pm$0.089) \\
\rowcolor{pct100}
\textbf{100\%   R$^2$ $\uparrow$}
&  0.878($\pm$0.001)  &  0.880($\pm$0.011) &  0.878($\pm$0.021) &  0.701($\pm$0.034)  & 0.823($\pm$0.020) & 0.902($\pm$0.001)
& \cellcolor{ourscol} 0.904($\pm$0.007)\\
\midrule

\rowcolor{pct50}
\textbf{50\%  RMSE $\downarrow$}
& 9.770($\pm$0.862) &  12.96($\pm$0.409) &  12.38($\pm$0.255) & 18.88($\pm$1.488)  & 14.65($\pm$0.351) & 8.180($\pm$0.304)
& \cellcolor{ourscol}  7.915($\pm$0.370) \\
\rowcolor{pct50}
\textbf{50\%   MAE $\downarrow$}
& 7.635($\pm$0.622) &   9.951($\pm$0.269) &  9.365($\pm$0.212) & 16.45($\pm$1.503)  & 11.66($\pm$0.281) & 6.112($\pm$0.211)
& \cellcolor{ourscol}  5.849($\pm$0.302) \\
\rowcolor{pct50}
\textbf{50\%  R$^2$ $\uparrow$}
& 0.800($\pm$0.035) &  0.650($\pm$0.022) & 0.681($\pm$0.013) & 0.254($\pm$0.118)  &  0.553($\pm$0.022) & 0.861($\pm$0.010)
& \cellcolor{ourscol}   0.869($\pm$0.012) \\
\midrule

\rowcolor{pct25}
\textbf{25\%  RMSE $\downarrow$}
& 13.59($\pm$0.417) &  26.68($\pm$0.421) &  26.94($\pm$0.485) & 24.85($\pm$0.389)  & 23.42($\pm$0.797) & 12.93($\pm$0.866)
& \cellcolor{ourscol} 11.10($\pm$0.943)\\
\rowcolor{pct25}
\textbf{25\% MAE $\downarrow$}
& 10.64($\pm$0.433) &  20.52($\pm$0.470) & 20.86($\pm$0.817) & 20.56($\pm$0.448)  & 18.41($\pm$0.894) & 9.688($\pm$0.850)
& \cellcolor{ourscol} 8.249($\pm$0.698) \\
\rowcolor{pct25}
\textbf{25\% R$^2$ $\uparrow$}
&  0.616($\pm$0.024) & -0.481($\pm$0.047) & -0.510($\pm$0.054) &  -0.285($\pm$0.040)  & -0.143($\pm$0.077) & 0.651($\pm$0.046)
& \cellcolor{ourscol} 0.742($\pm$0.043) \\
\bottomrule
\end{tabular}
}}

\vspace{5pt}

\begin{minipage}{1.0\textwidth}
\scriptsize
\textit{Notes. RMSE, root mean squared error; MAE, mean absolute error; R$^2$, coefficient of determination. RMSE and MAE are reported in years; lower values indicate better performance, higher R$^2$ indicates better performance. Scratch uses random initialization; SwinUNETR$^{\ast}$, VoCo$^{\ast}$, and S3D$^{\ast}$ use publicly released centralized foundation models; S3D uses centralized pretraining on our pretraining corpus, FedMAE and BrainFedFM use federated pretraining. Fine-tuning uses 15 epochs of AdamW, a learning rate of $1\times10^{-5}$, weight decay of $1\times10^{-2}$, and cosine scheduling with a 2-epoch warm-up.}
\end{minipage}
\end{table*}

\clearpage

\begin{figure*}[p]
\noindent{\scriptsize\bfseries
Panel B. Labelled-data scaling and full-data brain age regression analyses.}
{\tiny\raggedright
Characterizes brain age prediction performance from complementary perspectives.
\textbf{B1}, fold-averaged RMSE, MAE, R$^2$, Pearson correlation, Spearman correlation, and concordance correlation coefficient.
\textbf{B2}, cross-method analyses including error magnitude, association and agreement, signed age-error distributions, and MAE across the observed chronological-age range.
\textbf{B3}, the joint RMSE--CCC performance landscape across labelled-data settings, where lower RMSE and higher CCC indicate a more favorable error--agreement trade-off.\par}

\par\vspace{8pt}

\includegraphics[
    width=\textwidth,
    keepaspectratio
]{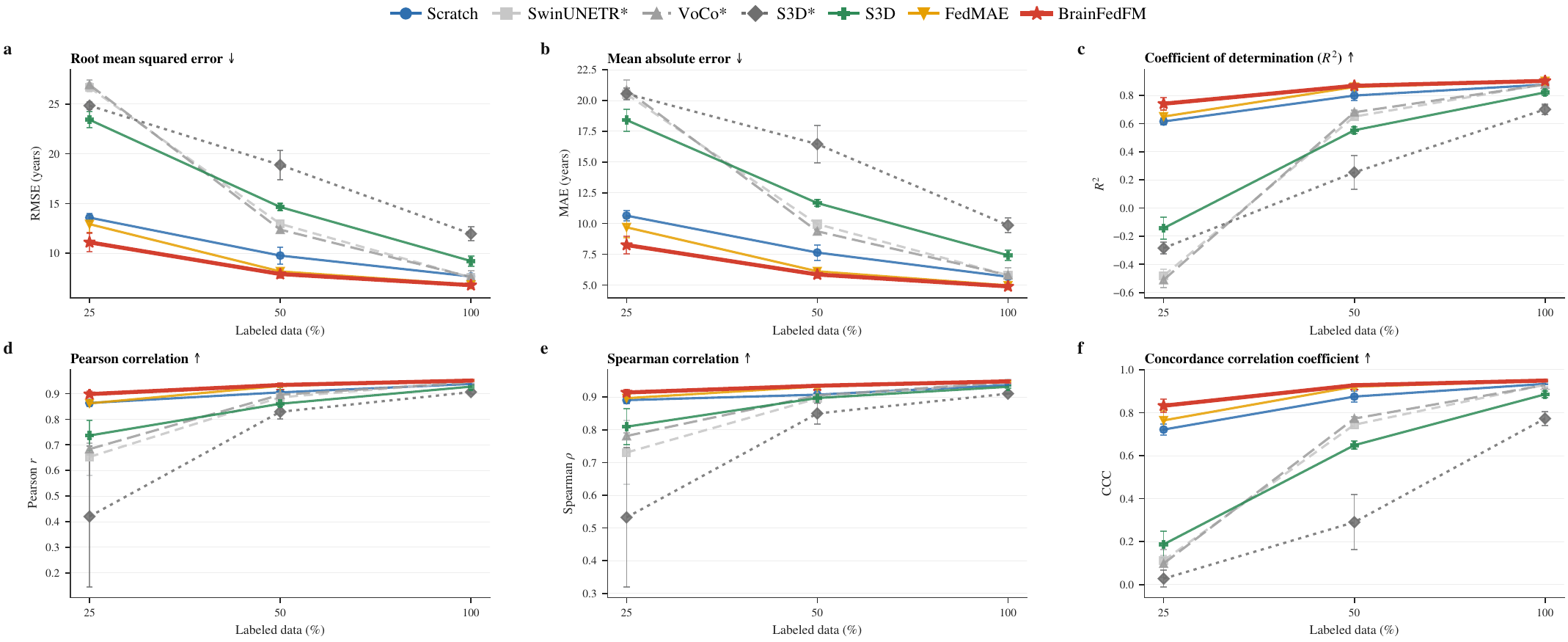}

\par\vspace{3pt}

\includegraphics[
    width=0.84\textwidth,
    keepaspectratio
]{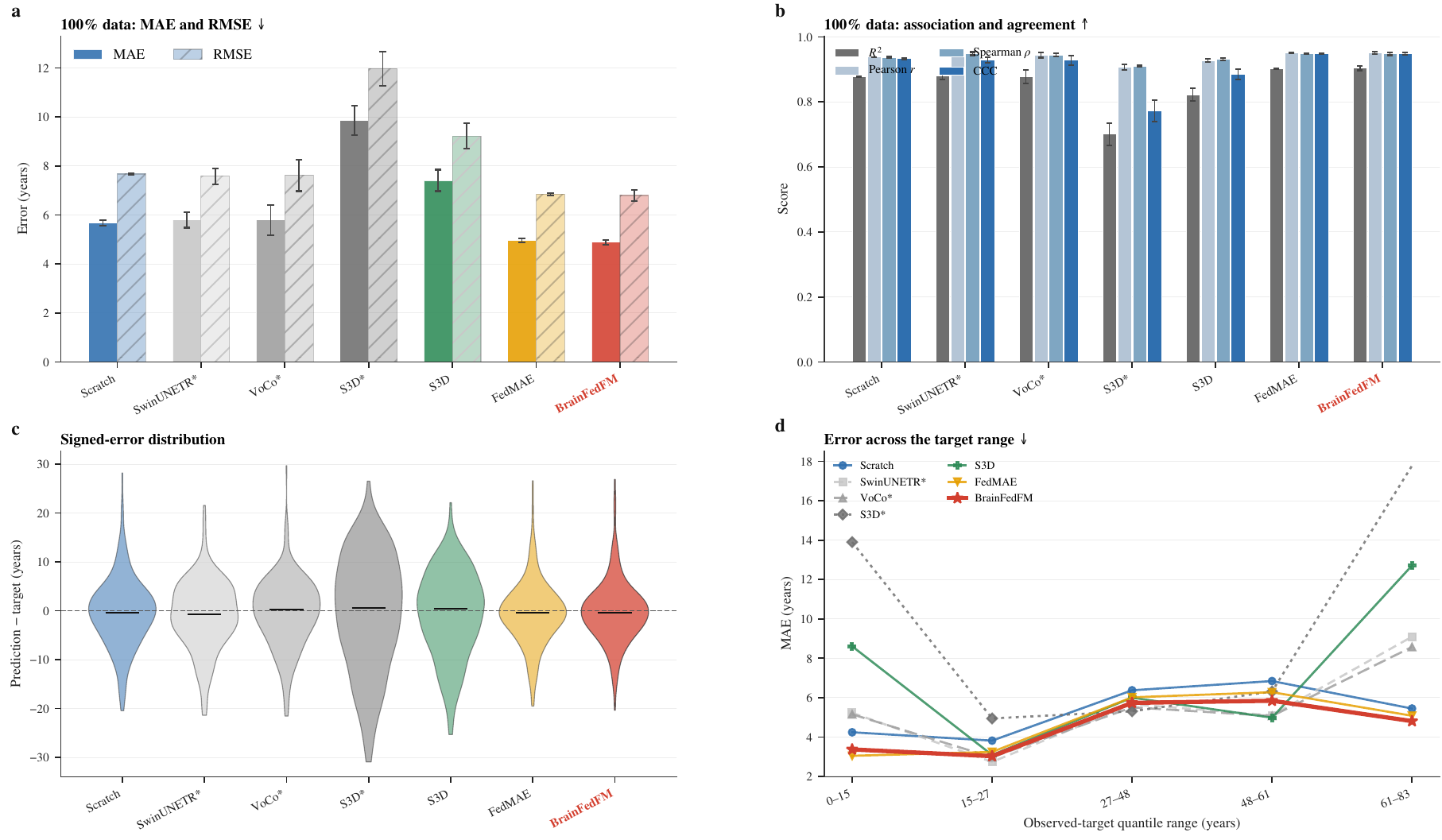}

\par\vspace{4pt}

\includegraphics[
    width=0.96\textwidth,
    keepaspectratio
]{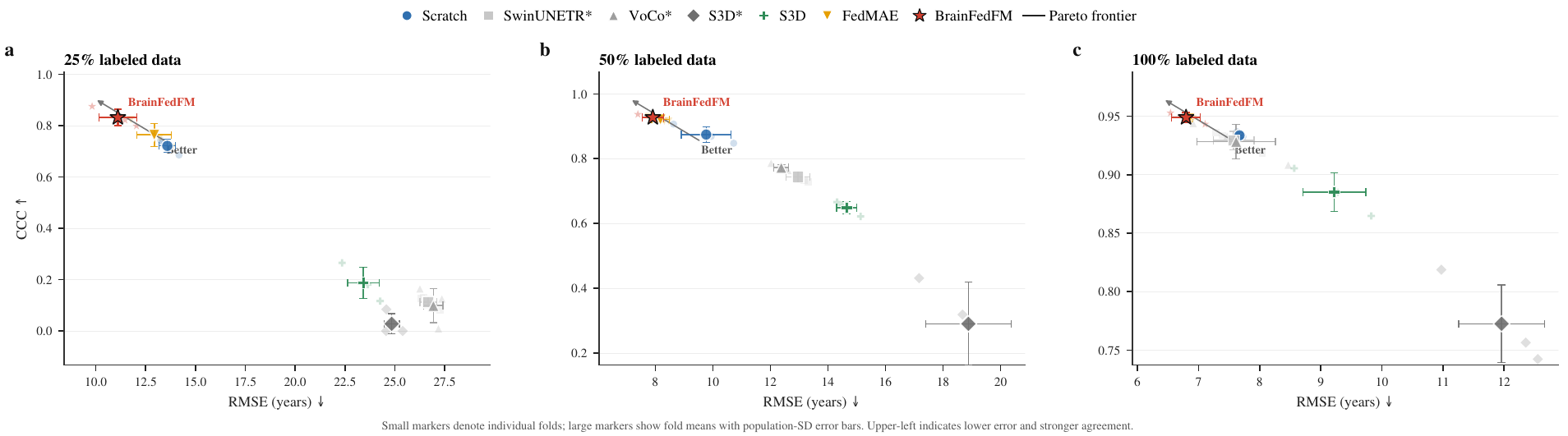}

\end{figure*}

\clearpage

\begin{figure*}[p]
\noindent{\scriptsize\bfseries
Panel C. Relative ranking, fold stability, and transfer-benefit analyses.}
{\tiny\raggedright
Summarizes comparative performance, robustness, and transfer benefit beyond the aggregate brain age regression scores.
\textbf{C1}, relative method ranks across all labelled-data settings and six regression metrics, with rank 1 denoting the best result within each setting--metric combination.
\textbf{C2}, fold-to-fold variability in RMSE, MAE, and R$^2$, together with the average method rank across the core regression metrics.
\textbf{C3}, performance changes relative to the matched Scratch baseline, where positive values denote RMSE or MAE reduction and improvement in R$^2$.\par}

\par\vspace{4pt}

\includegraphics[
    width=\textwidth,
    keepaspectratio
]{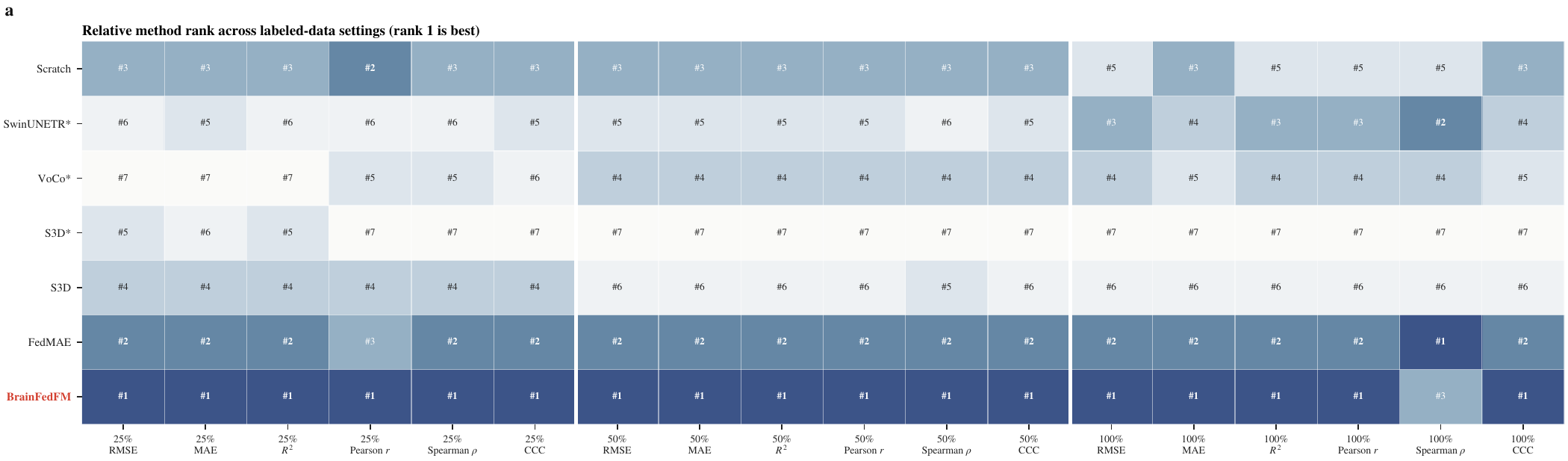}

\vspace{6pt}

\includegraphics[
    width=\textwidth,
    keepaspectratio
]{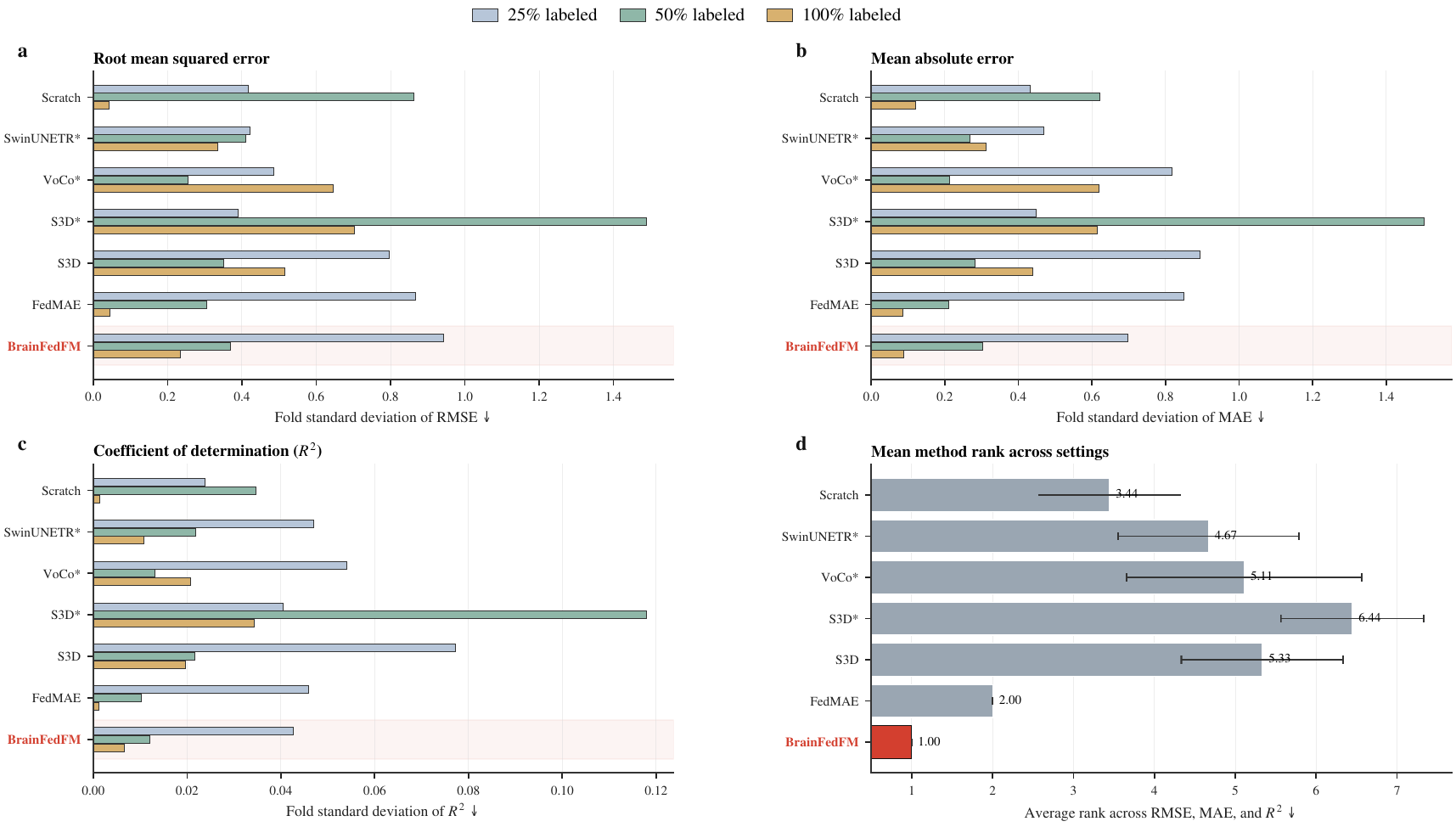}

\vspace{6pt}

\includegraphics[
    width=0.98\textwidth,
    keepaspectratio
]{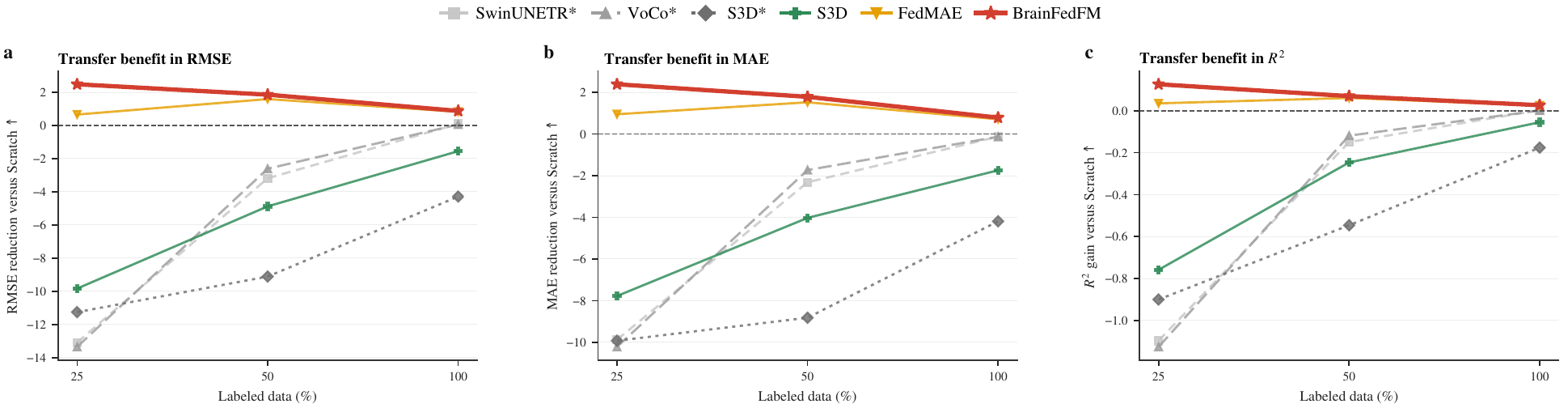}

\end{figure*}

\definecolor{pct100}{gray}{0.97}
\definecolor{pct75}{gray}{1.0}
\definecolor{pct50}{gray}{0.99}
\definecolor{pct25}{gray}{1.0}
\definecolor{overallgray}{gray}{0.97}
\definecolor{ourscol}{RGB}{244,248,252}

\begin{table*}[h]
\centering
\tiny
\caption[Adult MS lesion segmentation task]{\textbf{Adult MS lesion segmentation task.}
We evaluate foundation model performance on the ISBI-2015 multiple-sclerosis benchmark, where the goal is to delineate MS lesions from brain MRI. Accurate lesion segmentation enables quantitative assessment of lesion burden and supports longitudinal monitoring of disease progression, treatment response, and imaging-based outcome analysis. Panel~A reports Dice, intersection-over-union (IoU), and normalized surface Dice (NSD) with 100\% labelled training data, whereas Panel~B reports the corresponding results with 25\% and 50\% labelled data. The labelled-data percentages denote the fraction of the training--validation pool used for downstream fine-tuning and model selection; the held-out test set remains unchanged across settings. We compare a randomly initialized scratch baseline, centralized pretrained foundation models, and federated pretrained foundation models. Higher Dice, IoU, and NSD indicate better performance.}
\label{tab:result_isbi}

\renewcommand{\arraystretch}{3.5}
{\scriptsize\bfseries Panel A. Detailed results under 100\% labelled data.}
\vspace{5pt}

{\setlength{\tabcolsep}{3.38pt}

\resizebox{1.0\textwidth}{!}{
\begin{tabular}{
l c c c c c c
>
{\columncolor{ourscol}\centering\arraybackslash}p{1.62cm}
}

\toprule
\multicolumn{2}{c}{\textbf{Brain sMRI FMs }}
& \multicolumn{4}{c}{Centralized Pretraining}
& \multicolumn{2}{c}{Federated Pretraining} \\
\cmidrule(lr){1-2}\cmidrule(lr){3-6}\cmidrule(lr){7-8}
\makecell[c]{Metric/Methods}
& Scratch
& \makecell[c]{SwinUNETR$^{\ast}$}  & VoCo$^{\ast}$  &  S3D$^{\ast}$ &  S3D
& \makecell[c]{FedMAE} & \makecell[c]{BrainFedFM} \\
\midrule
\rowcolor{overallgray}
\textbf{Overall Dice $\uparrow$}
& 77.55($\pm$1.735) & 77.33($\pm$2.819)  & 77.65($\pm$1.032) & 78.83($\pm$0.759)  & 78.70($\pm$2.491) & 78.37($\pm$1.894) 
& \cellcolor{ourscol}78.97($\pm$1.062) \\
\rowcolor{overallgray}
\textbf{Overall IoU $\uparrow$}
& 64.18($\pm$2.160) & 63.95($\pm$3.429) & 64.39($\pm$1.225) & 65.89($\pm$1.129) & 65.79($\pm$2.864)  & 65.25($\pm$2.229) 
& \cellcolor{ourscol}65.97($\pm$1.219) \\
\rowcolor{overallgray}
\textbf{Overall NSD $\uparrow$}
& 86.85($\pm$1.624) & 87.06($\pm$1.885)  & 86.85($\pm$0.470) & 88.26($\pm$0.444) & 88.09($\pm$2.253)  & 87.97($\pm$1.210)
& \cellcolor{ourscol}88.52($\pm$0.698) \\
\midrule

MS Lesion (Dice)  & 77.55($\pm$1.735) & 77.33($\pm$2.819)    & 77.65($\pm$1.032)  & 78.83($\pm$0.759) & 78.70($\pm$2.491)  & 78.37($\pm$1.894)  & 78.97($\pm$1.062)   \\ 
MS Lesion (IoU) & 64.18($\pm$2.160) & 63.95($\pm$3.429)  & 64.39($\pm$1.225) & 65.89($\pm$1.129) & 65.79($\pm$2.864)  & 65.25($\pm$2.229)   & 65.97($\pm$1.219)  \\ 
MS Lesion (NSD)  & 86.85($\pm$1.624) & 87.06($\pm$1.885) & 86.85($\pm$0.470) & 88.26($\pm$0.444))  & 88.09($\pm$2.253) & 87.97($\pm$1.210)  & 88.52($\pm$0.698) \\ 
\bottomrule
\end{tabular}
}}

\vspace{10pt}

{\scriptsize \bfseries Panel B. Summary under different labelled-data percentages.}
\vspace{5pt}

\renewcommand{\arraystretch}{2.2}
\resizebox{1.0\textwidth}{!}{%
\begin{tabular}{
l
ccc!{\hspace{2pt}\color{black!45}\vrule width 0.6pt\hspace{2pt}}
ccc
}
\toprule
\multicolumn{1}{c}{\textbf{Percentages}}
& \multicolumn{3}{c}{25\%}
& \multicolumn{3}{c}{50\%} \\
\cmidrule(lr){1-1}\cmidrule(lr){2-4}\cmidrule(lr){5-7}
\makecell[c]{Methods/Metric} & \multicolumn{1}{c}{Dice} & \multicolumn{1}{c}{IoU} & \multicolumn{1}{c}{NSD}
& \multicolumn{1}{c}{Dice} & \multicolumn{1}{c}{IoU} & \multicolumn{1}{c}{NSD} \\
\midrule
Scratch
& 39.13($\pm$27.70) & 29.04($\pm$20.56) & 44.41($\pm$31.45)
& 70.29($\pm$3.294) & 55.45($\pm$3.532)  & 80.50($\pm$2.924)  \\
SwinUNETR$^{\ast}$
& 64.27($\pm$2.298) & 48.95($\pm$2.022)  & 73.74($\pm$2.897) 
& 72.04($\pm$3.779)  & 57.94($\pm$3.603)  & 82.54($\pm$3.775) \\
VoCo$^{\ast}$
& 36.67($\pm$26.32) & 26.50($\pm$19.23)  & 41.86($\pm$29.96) 
& 70.22($\pm$0.712)  & 55.55($\pm$1.134)  & 80.78($\pm$0.538)  \\
S3D$^{\ast}$
& 67.45($\pm$2.643) & 52.19($\pm$3.624)  & 78.13($\pm$2.170)
& 75.34($\pm$2.132) & 61.62($\pm$2.319)  &  85.31($\pm$2.046)  \\
S3D
& 61.22($\pm$10.38)  & 46.53($\pm$10.49)  & 71.87($\pm$8.519)
& 75.37($\pm$2.884) & 61.76($\pm$3.262)  & 85.78($\pm$2.550)  \\
\hline
FedMAE
& 66.13($\pm$1.335)  & 51.00($\pm$1.781)  & 76.87($\pm$1.215) 
& 73.28($\pm$2.709)  & 59.12($\pm$2.927)  & 83.91($\pm$2.763)  \\
\rowcolor{ourscol}
BrainFedFM
& 66.22($\pm$1.079) & 50.93($\pm$1.793) & 76.57($\pm$0.418) 
& 74.72($\pm$2.167)  & 60.76($\pm$2.370)  & 84.78($\pm$2.486)  \\
\bottomrule
\end{tabular}
}
\vspace{3pt}

\begin{minipage}{1.0\textwidth}
\scriptsize
\textit{Notes. Dice, Dice similarity coefficient; IoU, intersection-over-union; NSD, normalized surface Dice. All models are fine-tuned for 150 epochs with 20 training iterations and 50 validation iterations per epoch using SGD with momentum 0.99, Nesterov acceleration, weight decay of $3\times10^{-5}$, and an initial learning rate of $1\times10^{-3}$. Training uses a 15-epoch linear warm-up followed by polynomial learning-rate decay, and test performance is reported from the checkpoint with the best validation result.}
\end{minipage}
\end{table*}

\clearpage

\noindent{\scriptsize\bfseries Panel C. Data efficiency, complementary profiles, and qualitative lesion delineation.}
\vspace{2pt}
{\tiny\raggedright
Examines annotation efficiency and complementary segmentation behavior beyond the tabulated overlap scores. \textbf{C1}, fold-averaged trajectories for Dice, NSD, voxel sensitivity, lesion F1, HD95, and absolute relative volume error; \textbf{C2}, direction-aligned profiles across overlap/surface, voxel/volume, and lesion-level metrics, together with a family-level summary; \textbf{C3}, representative axial MRI examples with reference lesions and three-fold ensemble error maps, where green, orange, and purple denote true-positive, false-positive, and false-negative lesion voxels, respectively.\par}
\par\vspace{-2pt}

\begin{center}
\includegraphics[width=0.975\textwidth,keepaspectratio]{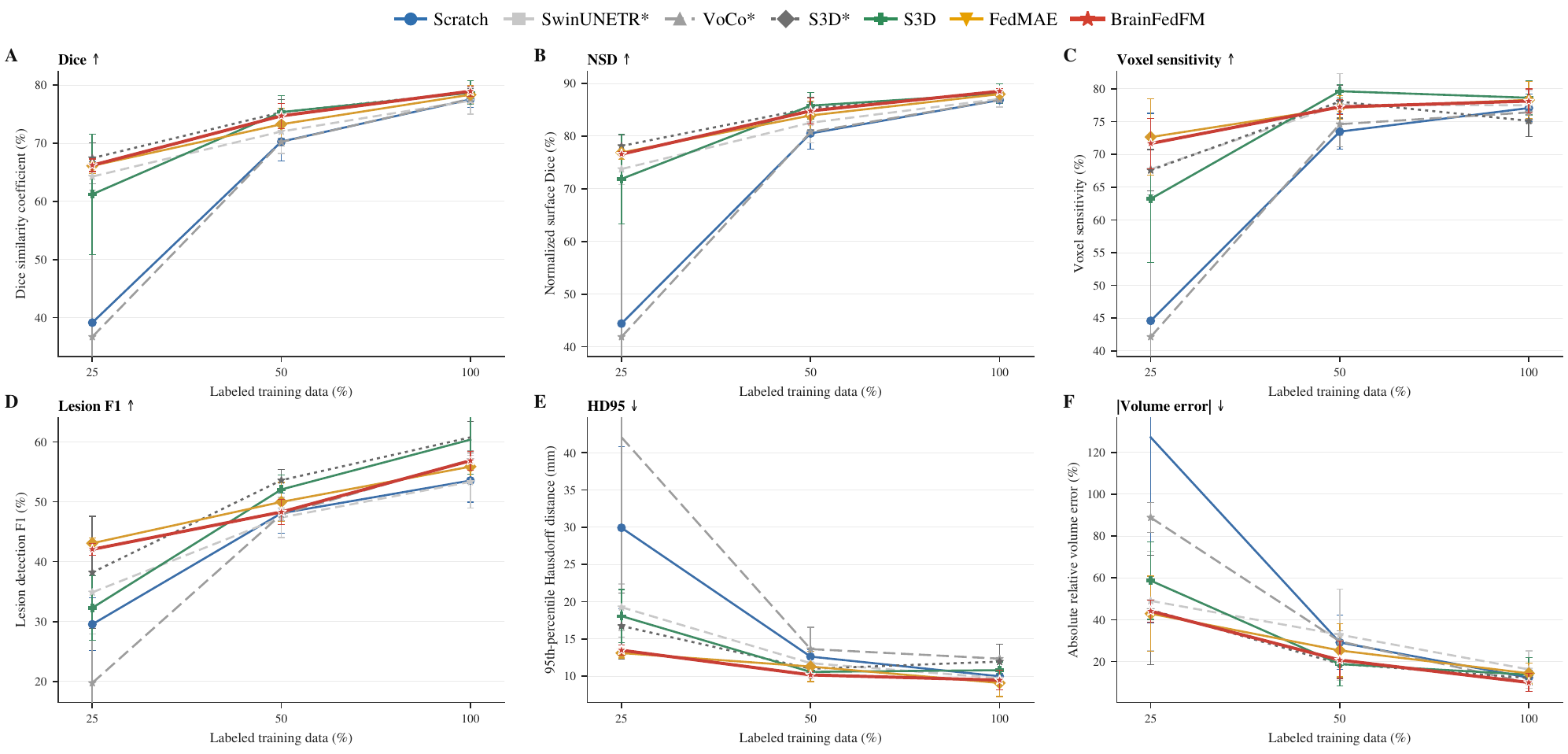}
\includegraphics[width=0.845\textwidth,keepaspectratio]{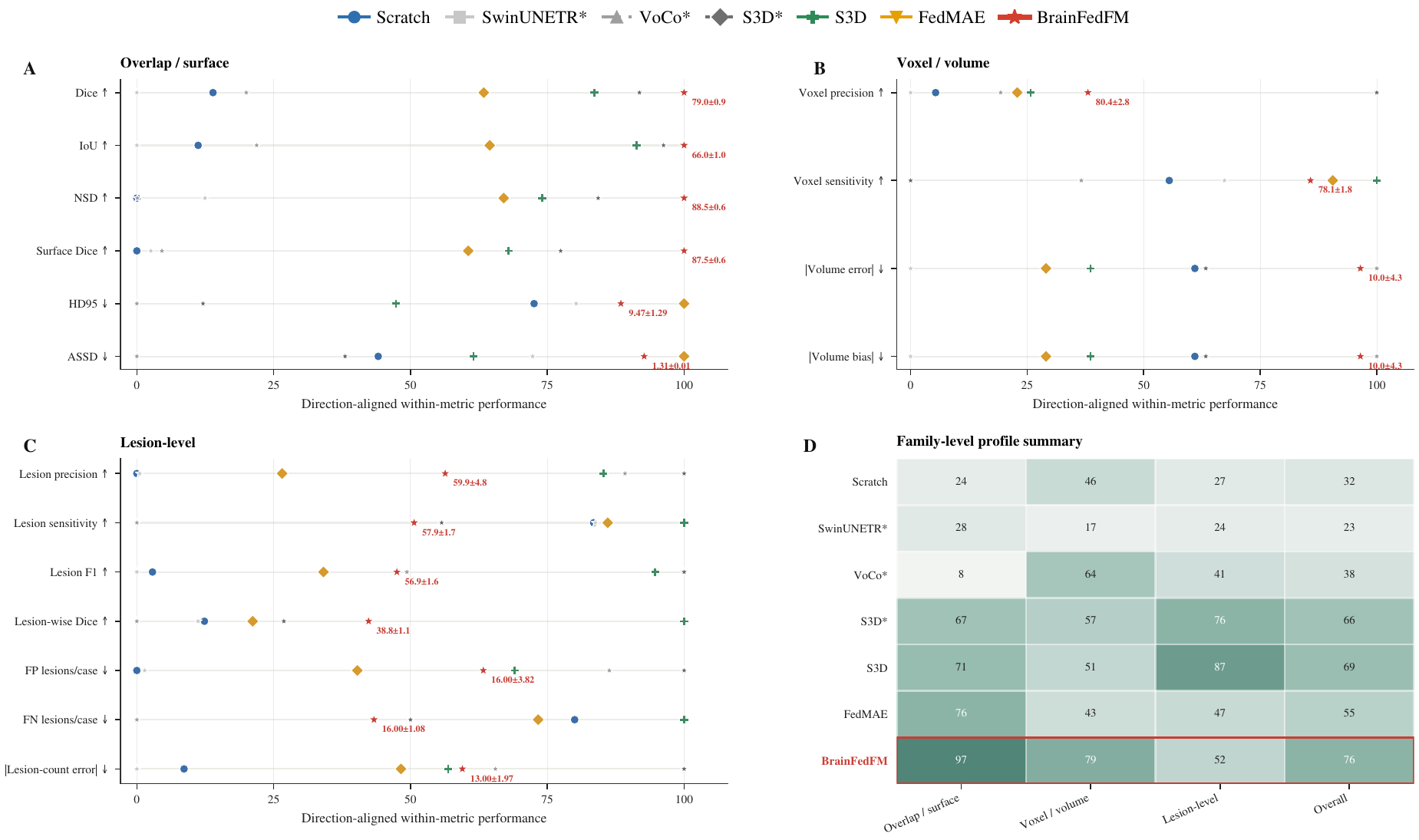}
\includegraphics[width=0.984\textwidth,keepaspectratio]{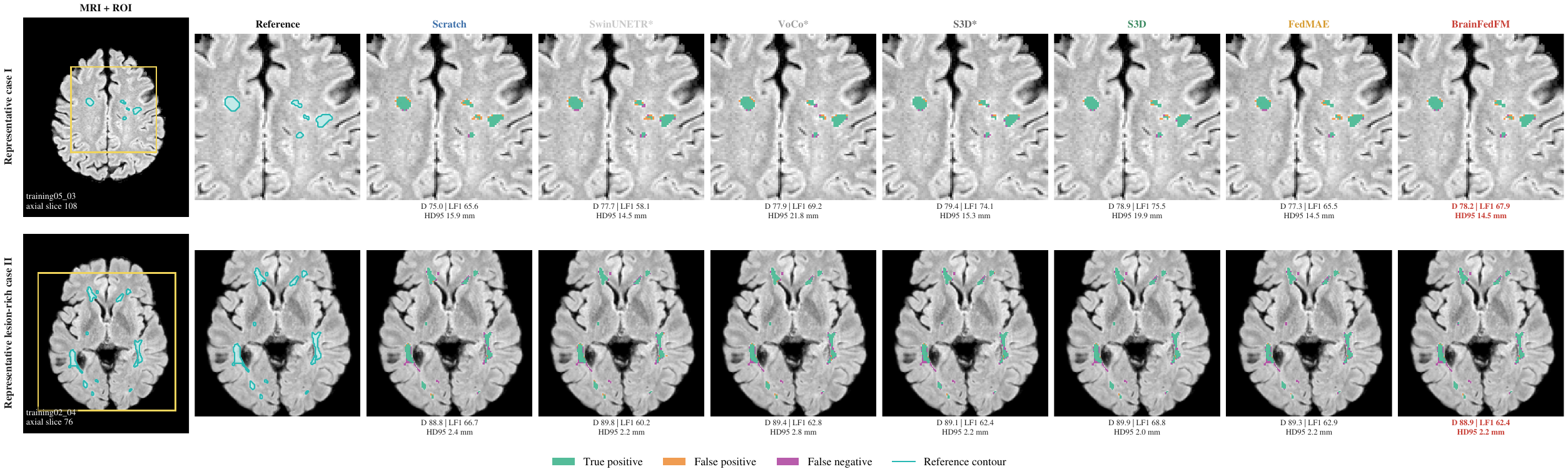}
\end{center}

\clearpage

\noindent{\scriptsize\bfseries Panel D. Error decomposition, case-level variability, and lesion-scale performance.}
\vspace{2pt}
{\tiny\raggedright
Characterizes segmentation failure modes and robustness beyond aggregate accuracy. \textbf{D1}, false-negative and false-positive voxel burden, signed volume bias, Dice--NSD performance, and voxel precision--sensitivity balance at 100\% labelled data; \textbf{D2}, matched case-by-fold distributions of HD95, average symmetric surface distance (ASSD), absolute relative volume error, and absolute lesion-count error; \textbf{D3}, lesion-detection sensitivity and lesion-wise Dice across reference-lesion volume quintiles, with dedicated summaries for the smallest lesions.\par}
\par\vspace{-2pt}

\begin{center}
\includegraphics[width=0.76\textwidth,keepaspectratio]{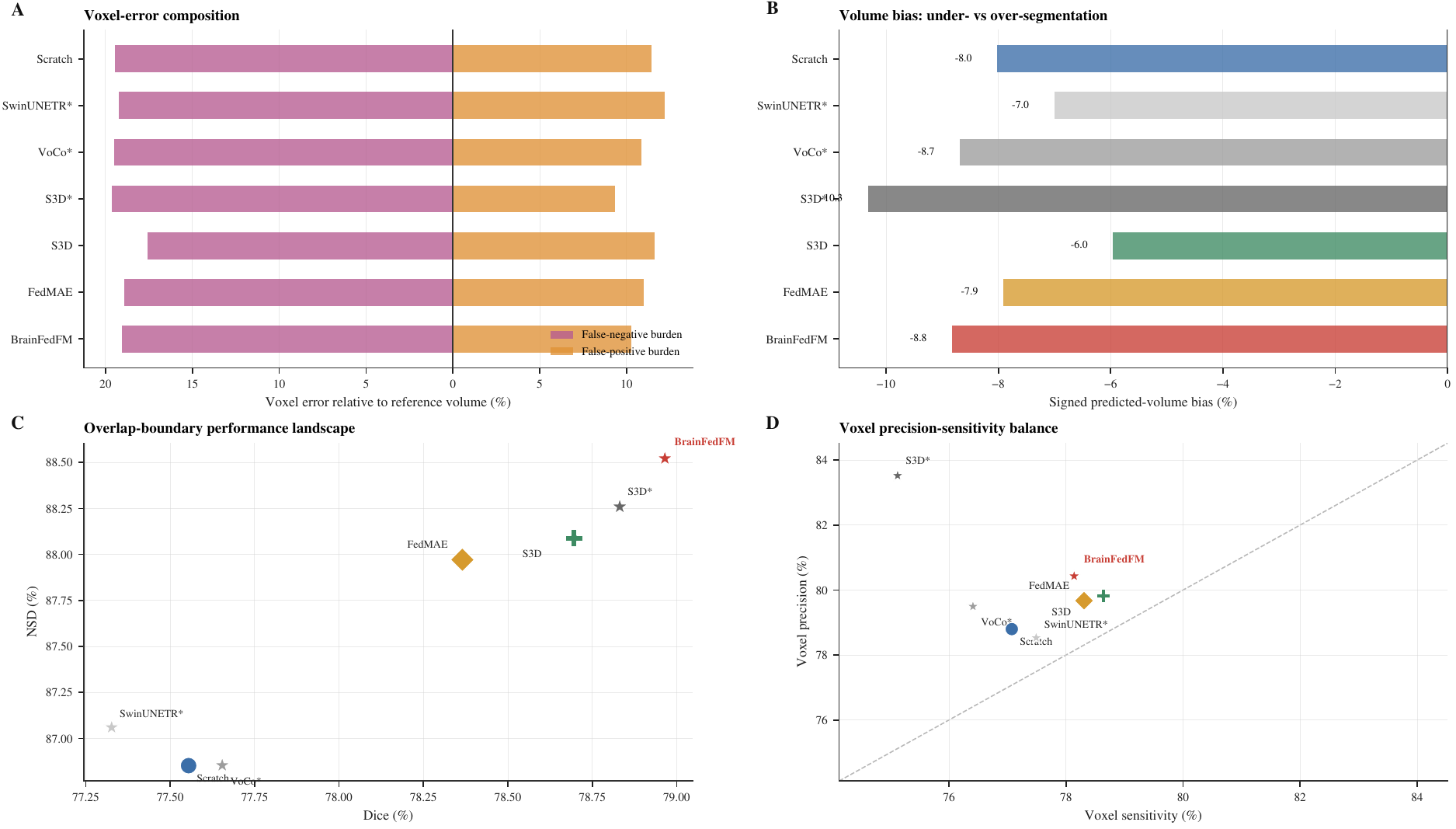}
\includegraphics[width=0.743\textwidth,keepaspectratio]{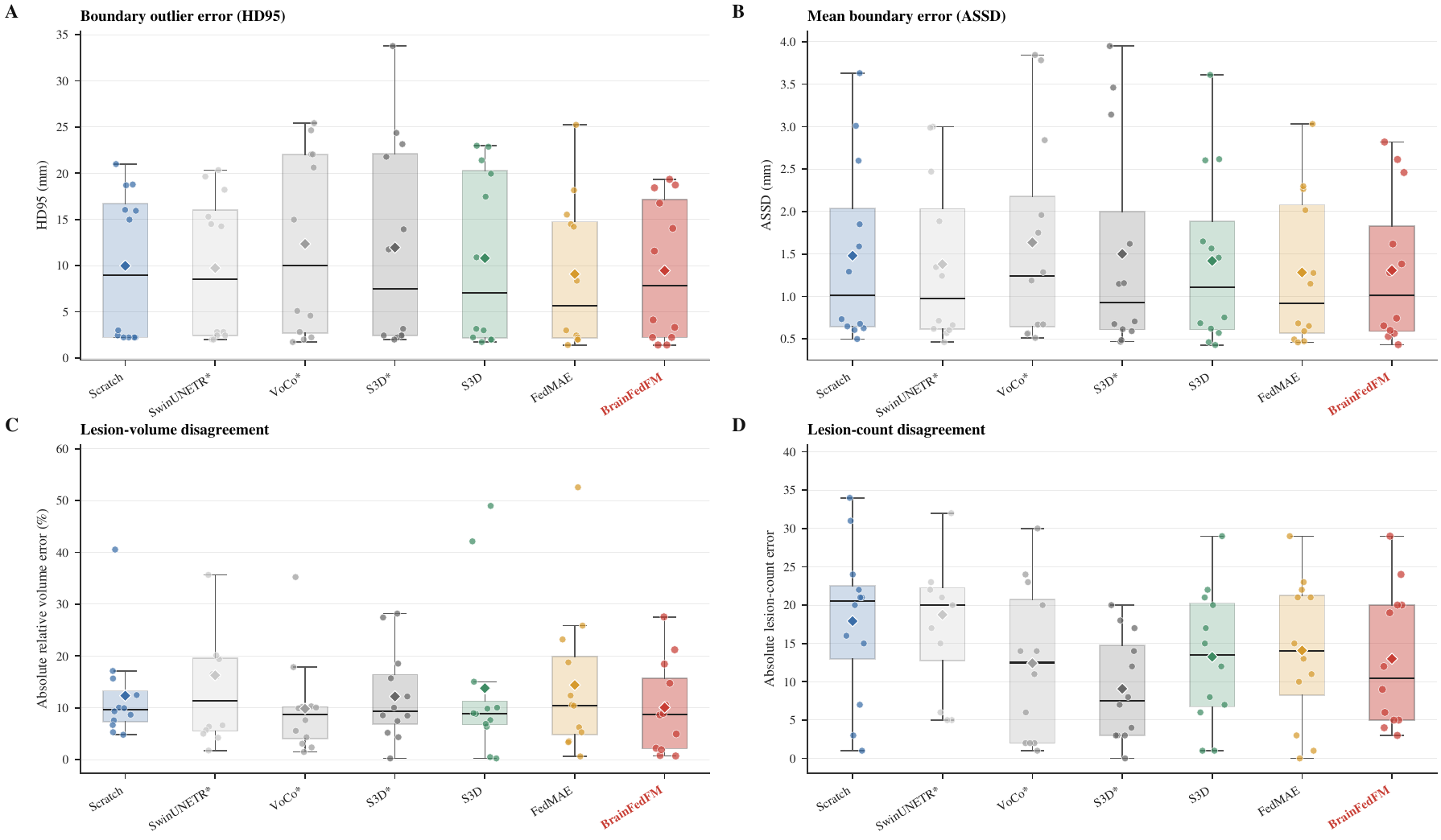}
\includegraphics[width=0.753\textwidth,keepaspectratio]{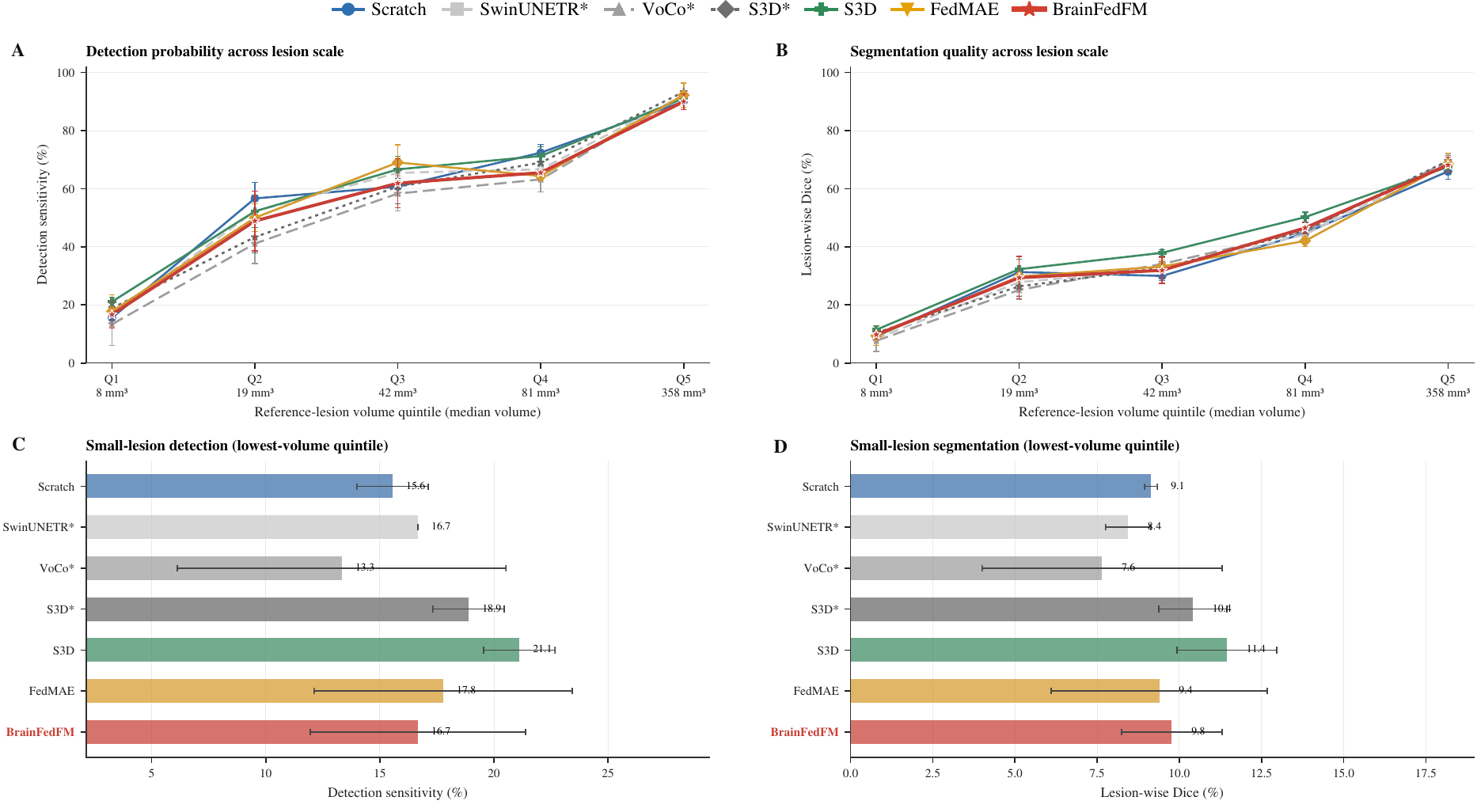}
\end{center}

\definecolor{pct100}{gray}{0.97}
\definecolor{pct75}{gray}{1.0}
\definecolor{pct50}{gray}{0.99}
\definecolor{pct25}{gray}{1.0}
\definecolor{overallgray}{gray}{0.97}
\definecolor{ourscol}{RGB}{244,248,252}

\begin{table*}[h]
\centering
\tiny
\caption[Pediatric MS lesion segmentation task]{\textbf{Pediatric MS lesion segmentation task.}
We evaluate foundation model performance on the PediMS benchmark, where the goal is to delineate pediatric multiple-sclerosis lesions from brain MRI. Accurate lesion segmentation enables quantitative assessment of lesion burden and supports longitudinal monitoring of disease progression, treatment response, and imaging-based outcome analysis in pediatric MS. Panel~A reports Dice, intersection-over-union (IoU), and normalized surface Dice (NSD) with 100\% labelled training data, whereas Panel~B reports the corresponding results with 25\% and 50\% labelled data. The labelled-data percentages denote the fraction of the training--validation pool used for downstream fine-tuning and model selection; the held-out test set remains unchanged across settings. We compare a randomly initialized scratch baseline, centralized pretrained foundation models, and federated pretrained foundation models. Higher Dice, IoU, and NSD indicate better performance.}
\label{tab:result_pediMS}

\renewcommand{\arraystretch}{3.65}
{\scriptsize\bfseries Panel A. Detailed results under 100\% labelled data.}
\vspace{5pt}

{\setlength{\tabcolsep}{3.38pt}

\resizebox{1.0\textwidth}{!}{
\begin{tabular}{
l c c c c c c
>
{\columncolor{ourscol}\centering\arraybackslash}p{1.62cm}
}

\toprule
\multicolumn{2}{c}{\textbf{Brain sMRI FMs }}
& \multicolumn{4}{c}{Centralized Pretraining}
& \multicolumn{2}{c}{Federated Pretraining} \\
\cmidrule(lr){1-2}\cmidrule(lr){3-6}\cmidrule(lr){7-8}
\makecell[c]{Metric/Methods}
& Scratch
& \makecell[c]{SwinUNETR$^{\ast}$} & VoCo$^{\ast}$ & S3D$^{\ast}$ & S3D
& \makecell[c]{FedMAE} & \makecell[c]{BrainFedFM} \\
\midrule
\rowcolor{overallgray}
\textbf{Overall Dice $\uparrow$}
& 72.16($\pm$0.104)
& 72.22($\pm$0.586)
& 73.29($\pm$1.134)
& 73.74($\pm$0.170)
& 73.28($\pm$1.100)
& 73.04($\pm$1.556)
& \cellcolor{ourscol}73.93($\pm$1.220) \\

\rowcolor{overallgray}
\textbf{Overall IoU $\uparrow$}
& 56.75($\pm$0.157)
& 56.80($\pm$0.799)
& 58.13($\pm$1.406)
& 58.62($\pm$0.227)
& 58.02($\pm$1.409)
& 57.80($\pm$1.873)
& \cellcolor{ourscol}58.81($\pm$1.560) \\

\rowcolor{overallgray}
\textbf{Overall NSD $\uparrow$}
& 82.12($\pm$0.480)
& 81.67($\pm$0.170)
& 82.13($\pm$1.238)
& 82.71($\pm$0.313)
& 82.46($\pm$0.796)
& 82.62($\pm$1.401)
& \cellcolor{ourscol}83.62($\pm$0.873) \\
\midrule

Pediatric MS (Dice)
& 72.16($\pm$0.104)
& 72.22($\pm$0.586)
& 73.29($\pm$1.134)
& 73.74($\pm$0.170)
& 73.28($\pm$1.100)
& 73.04($\pm$1.556)
& 73.93($\pm$1.220) \\

Pediatric MS (IoU)
& 56.75($\pm$0.157)
& 56.80($\pm$0.799)
& 58.13($\pm$1.406)
& 58.62($\pm$0.227)
& 58.02($\pm$1.409)
& 57.80($\pm$1.873)
& 58.81($\pm$1.560) \\

Pediatric MS (NSD)
& 82.12($\pm$0.480)
& 81.67($\pm$0.170)
& 82.13($\pm$1.238)
& 82.71($\pm$0.313)
& 82.46($\pm$0.796)
& 82.62($\pm$1.401)
& 83.62($\pm$0.873) \\
\bottomrule
\end{tabular}
}}

\vspace{8pt}

{\scriptsize\bfseries Panel B. Summary under different labelled-data percentages.}
\vspace{5pt}

\renewcommand{\arraystretch}{2.3}
\resizebox{1.0\textwidth}{!}{%
\begin{tabular}{
l
ccc!{\hspace{2pt}\color{black!45}\vrule width 0.6pt\hspace{2pt}}
ccc
}
\toprule
\multicolumn{1}{c}{\textbf{Percentages}}
& \multicolumn{3}{c}{25\%}
& \multicolumn{3}{c}{50\%} \\
\cmidrule(lr){1-1}\cmidrule(lr){2-4}\cmidrule(lr){5-7}
\makecell[c]{Methods/Metric} & \multicolumn{1}{c}{Dice} & \multicolumn{1}{c}{IoU} & \multicolumn{1}{c}{NSD}
& \multicolumn{1}{c}{Dice} & \multicolumn{1}{c}{IoU} & \multicolumn{1}{c}{NSD} \\
\midrule
Scratch
& 57.35($\pm$8.360) & 40.99($\pm$8.037) & 63.91($\pm$9.968)
& 66.94($\pm$2.571) & 50.77($\pm$3.033) & 75.71($\pm$2.468) \\
SwinUNETR$^{\ast}$
& 62.15($\pm$3.768) & 45.59($\pm$4.111) & 69.67($\pm$3.892)
& 65.63($\pm$3.680) & 49.43($\pm$4.207) & 74.18($\pm$3.683) \\
VoCo$^{\ast}$
& 64.29($\pm$3.856) & 47.77($\pm$4.227) & 71.71($\pm$4.647)
& 68.47($\pm$2.205) & 52.34($\pm$2.674) & 77.39($\pm$2.337) \\
S3D$^{\ast}$
& 64.65($\pm$2.512) & 48.08($\pm$2.800) & 72.78($\pm$2.526)
& 68.84($\pm$2.656) & 52.73($\pm$3.191) & 77.86($\pm$3.032) \\
S3D
& 64.17($\pm$3.238) & 47.50($\pm$3.606) & 72.31($\pm$3.734)
& 70.02($\pm$1.363) & 54.14($\pm$1.757) & 79.03($\pm$1.131) \\
\hline
FedMAE
& 64.32($\pm$1.919) & 47.64($\pm$2.108) & 72.15($\pm$2.993)
& 70.61($\pm$2.547) & 54.82($\pm$3.210) & 79.76($\pm$2.024) \\
\rowcolor{ourscol}
BrainFedFM
& 64.24($\pm$3.063) & 47.53($\pm$3.469) & 71.98($\pm$3.487)
& 70.20($\pm$1.726) & 54.29($\pm$2.266) & 78.51($\pm$1.656) \\
\bottomrule
\end{tabular}
}
\vspace{3pt}

\begin{minipage}{1.0\textwidth}
\scriptsize
\textit{Notes. Dice, Dice similarity coefficient; IoU, intersection-over-union; NSD, normalized surface Dice. All models are fine-tuned for 150 epochs with 50 training iterations and 50 validation iterations per epoch using SGD with momentum 0.99, Nesterov acceleration, weight decay of $3\times10^{-5}$, and an initial learning rate of $1\times10^{-3}$. Training uses a 15-epoch linear warm-up followed by polynomial learning-rate decay, and test performance is reported from the checkpoint with the best validation result.}
\end{minipage}
\end{table*}

\clearpage

\noindent{\scriptsize\bfseries Panel C. Data efficiency, complementary profiles, and qualitative lesion delineation.}
\vspace{2pt}
{\tiny\raggedright
Examines annotation efficiency and complementary segmentation behavior beyond the tabulated overlap scores. \textbf{C1}, fold-averaged trajectories for Dice, NSD, voxel sensitivity, lesion F1, HD95, and absolute relative volume error; \textbf{C2}, direction-aligned profiles across overlap/surface, voxel/volume, and lesion-level metrics, together with a family-level summary; \textbf{C3}, representative axial MRI examples with reference lesions and three-fold ensemble error maps, where green, orange, and purple denote true-positive, false-positive, and false-negative lesion voxels, respectively.\par}
\par\vspace{-2pt}

\begin{center}
\includegraphics[width=0.975\textwidth,keepaspectratio]{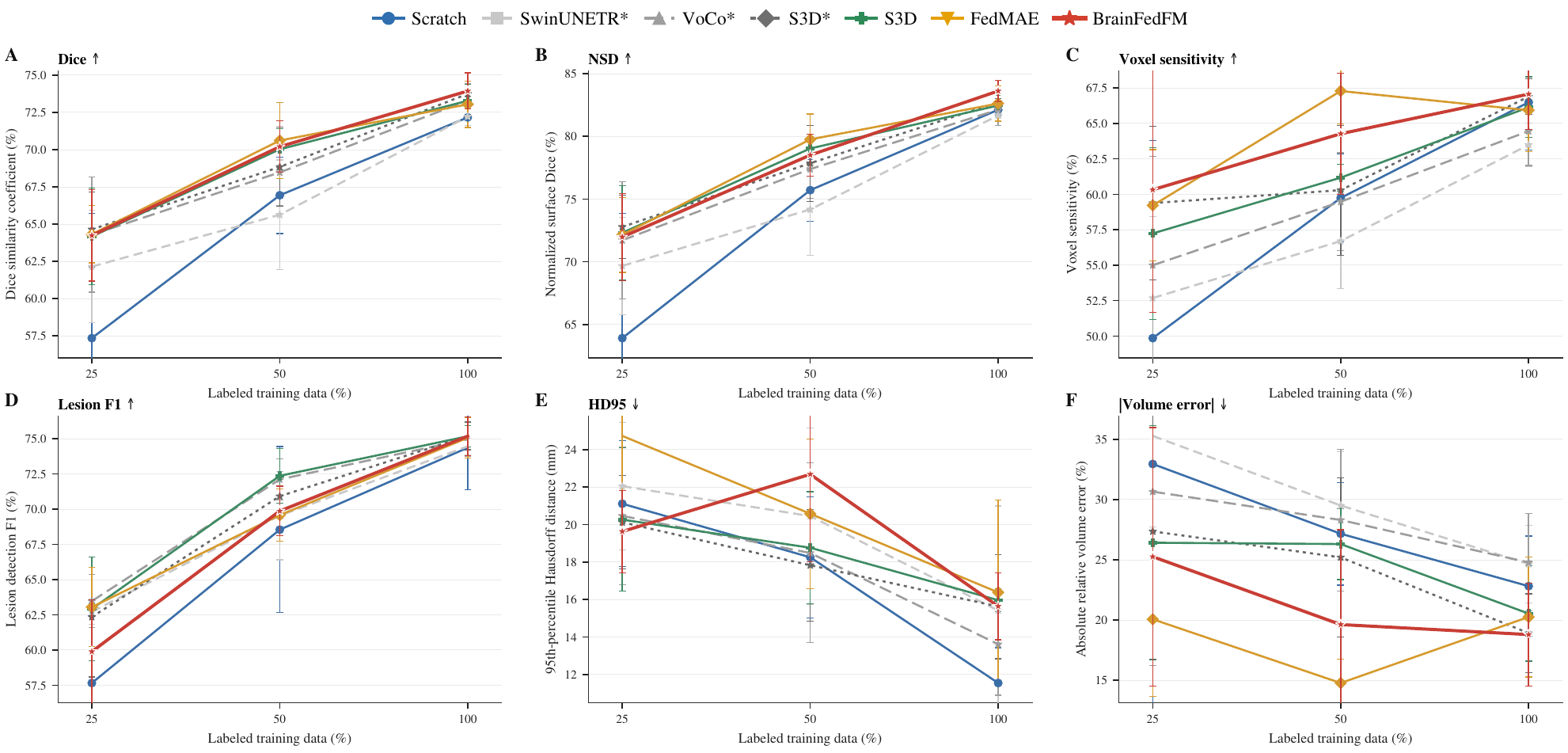}
\includegraphics[width=0.845\textwidth,keepaspectratio]{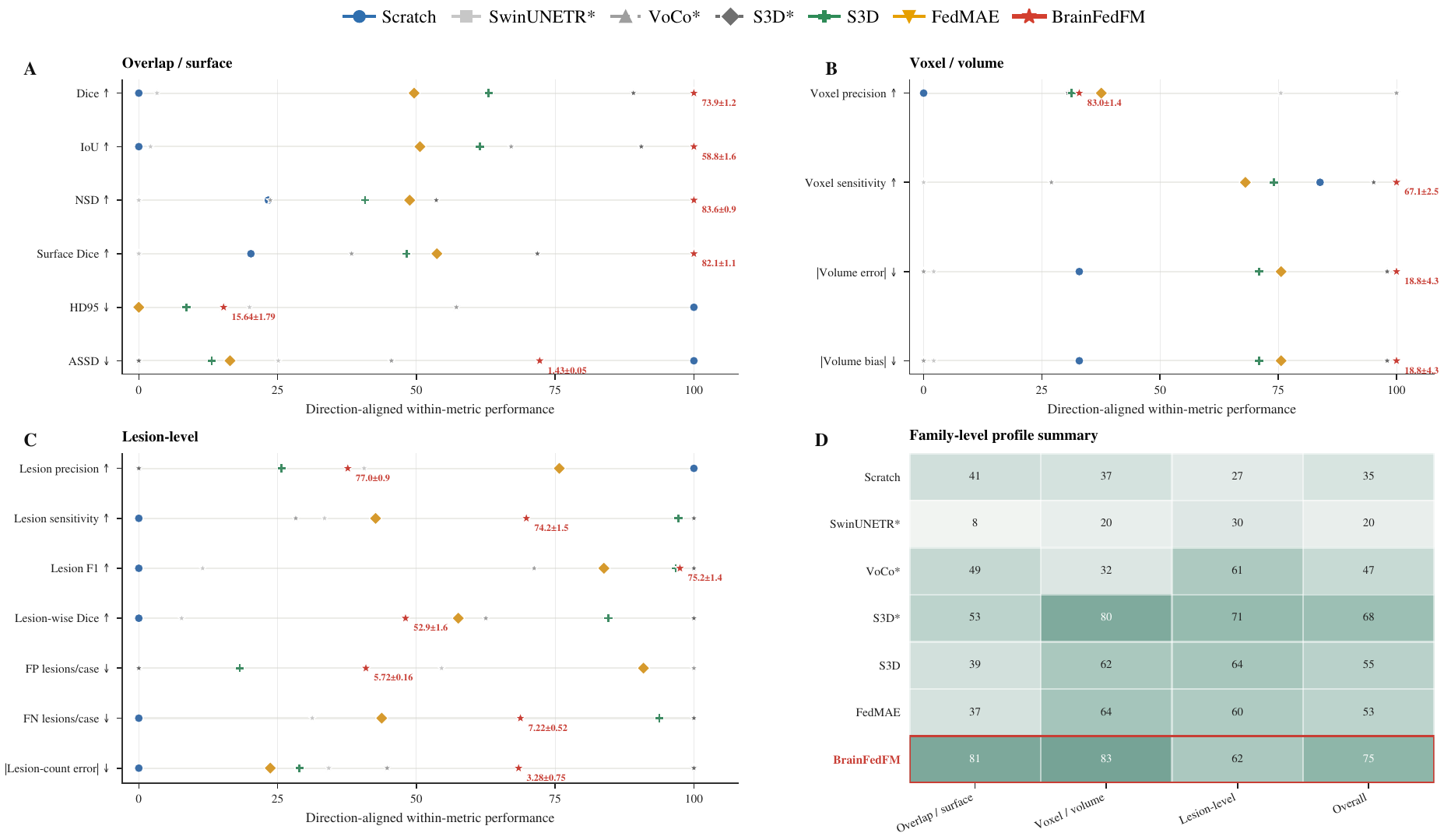}
\includegraphics[width=0.9888\textwidth,keepaspectratio]{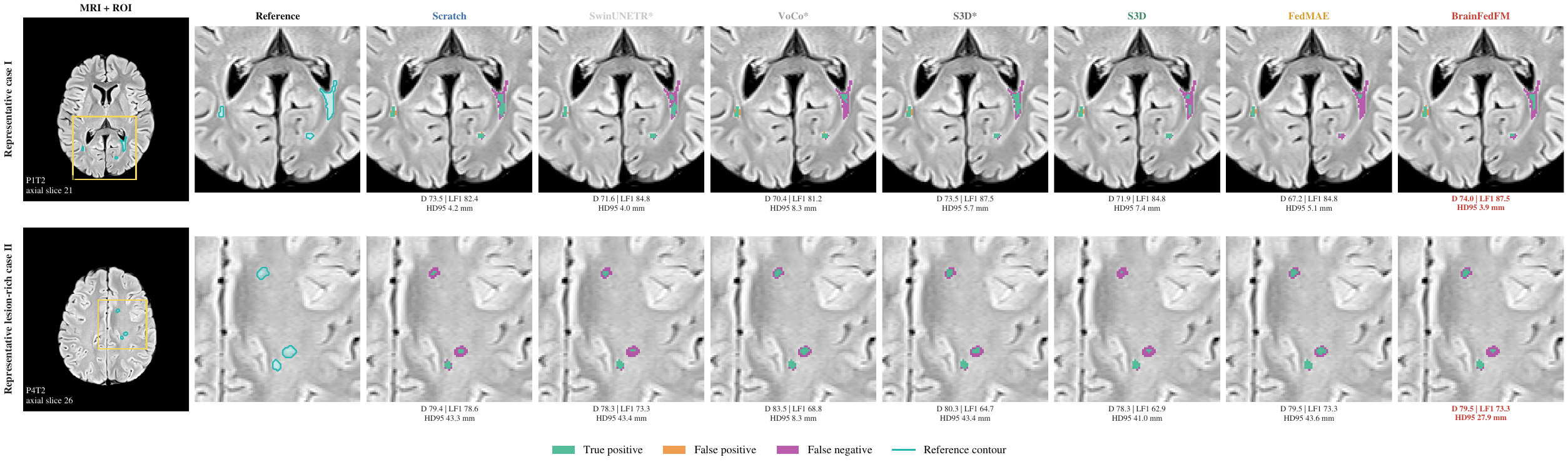}
\end{center}

\clearpage

\noindent{\scriptsize\bfseries Panel D. Error decomposition, case-level variability, and lesion-scale performance.}
\vspace{2pt}
{\tiny\raggedright
Characterizes segmentation failure modes and robustness beyond aggregate accuracy. \textbf{D1}, false-negative and false-positive voxel burden, signed volume bias, Dice--NSD performance, and voxel precision--sensitivity balance at 100\% labelled data; \textbf{D2}, matched case-by-fold distributions of HD95, average symmetric surface distance (ASSD), absolute relative volume error, and absolute lesion-count error; \textbf{D3}, lesion-detection sensitivity and lesion-wise Dice across reference-lesion volume quintiles, with dedicated summaries for the smallest lesions.\par}
\par\vspace{-2pt}

\begin{center}
\includegraphics[width=0.76\textwidth,keepaspectratio]{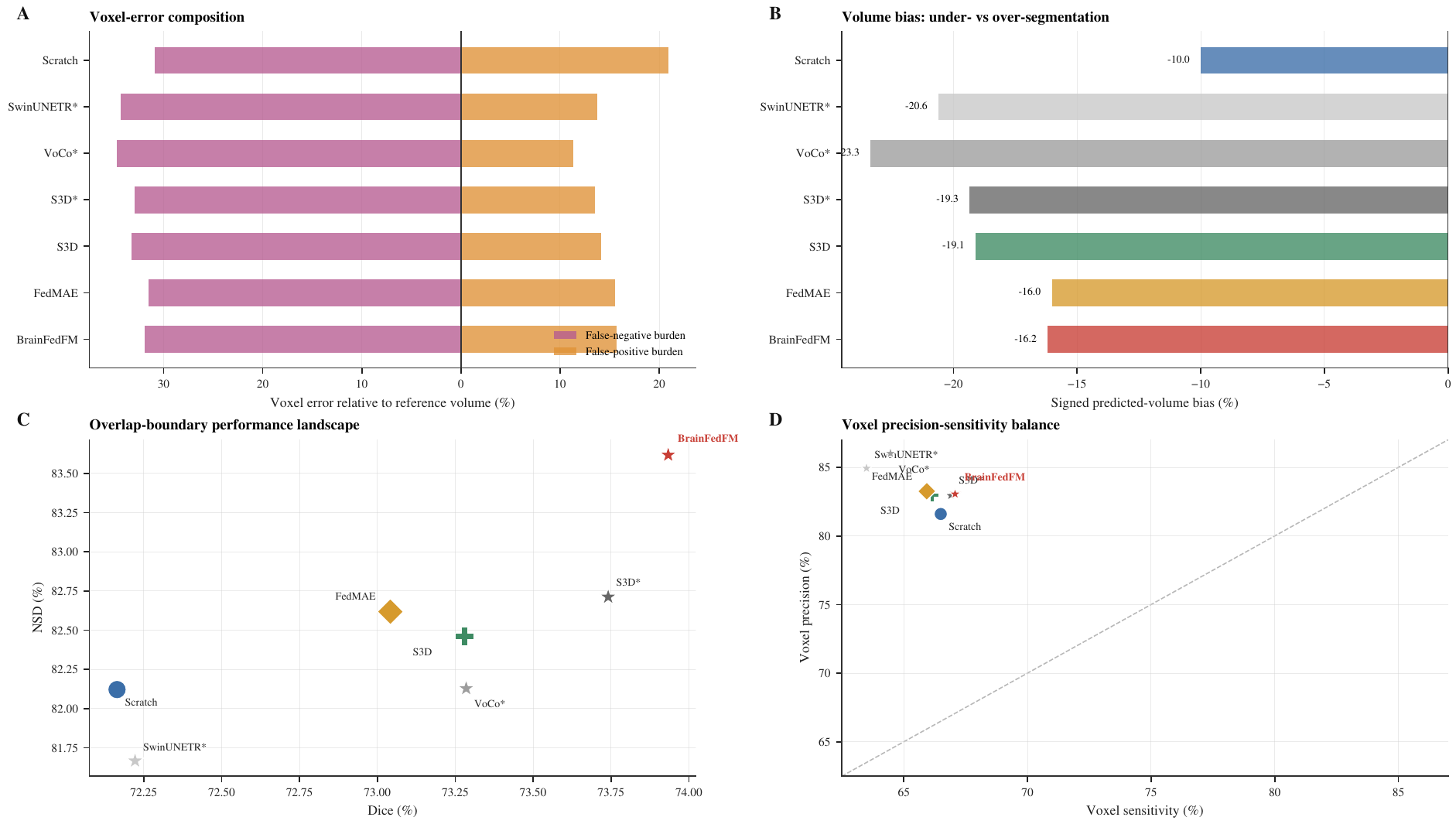}
\includegraphics[width=0.745\textwidth,keepaspectratio]{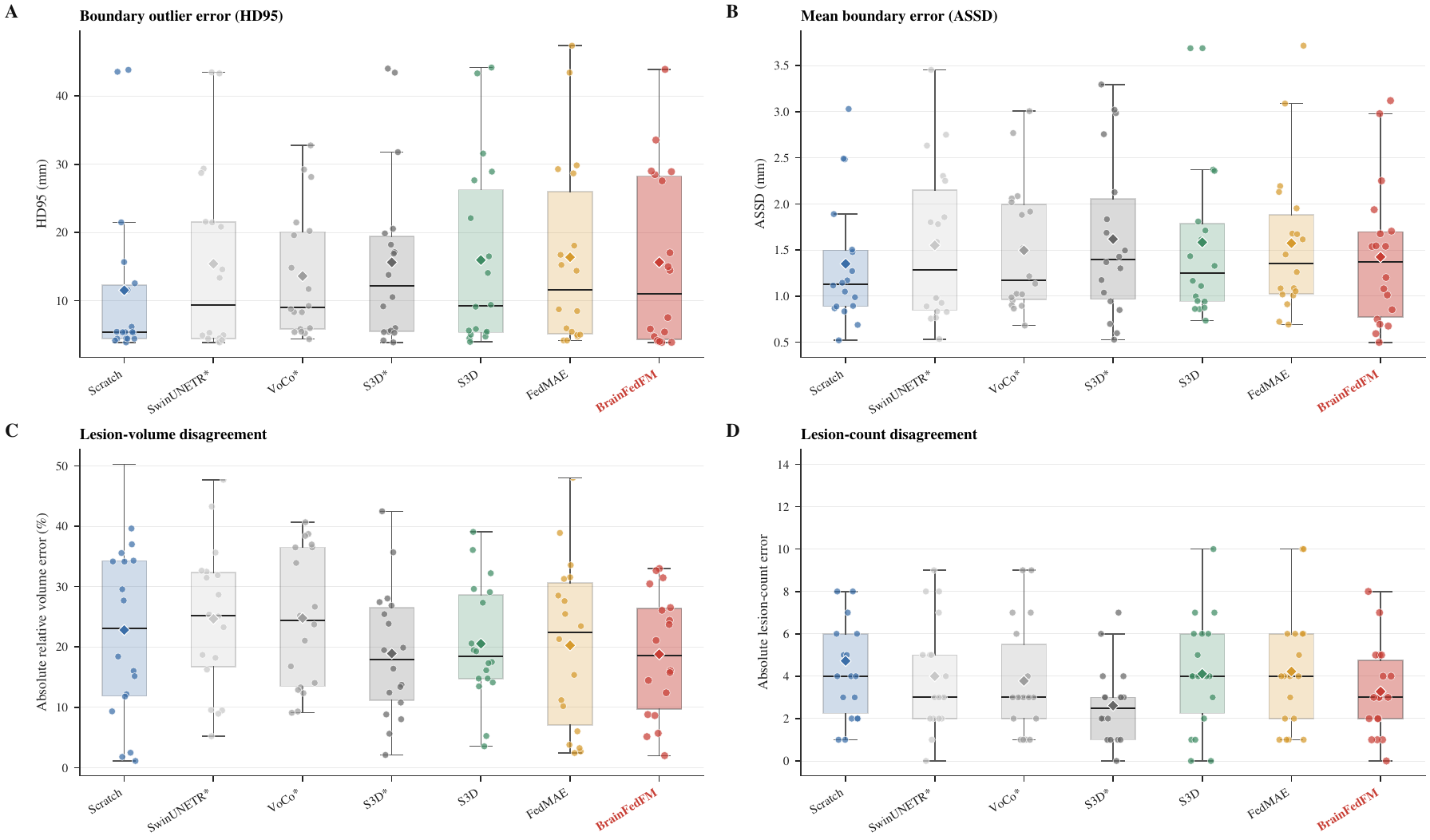}
\includegraphics[width=0.754\textwidth,keepaspectratio]{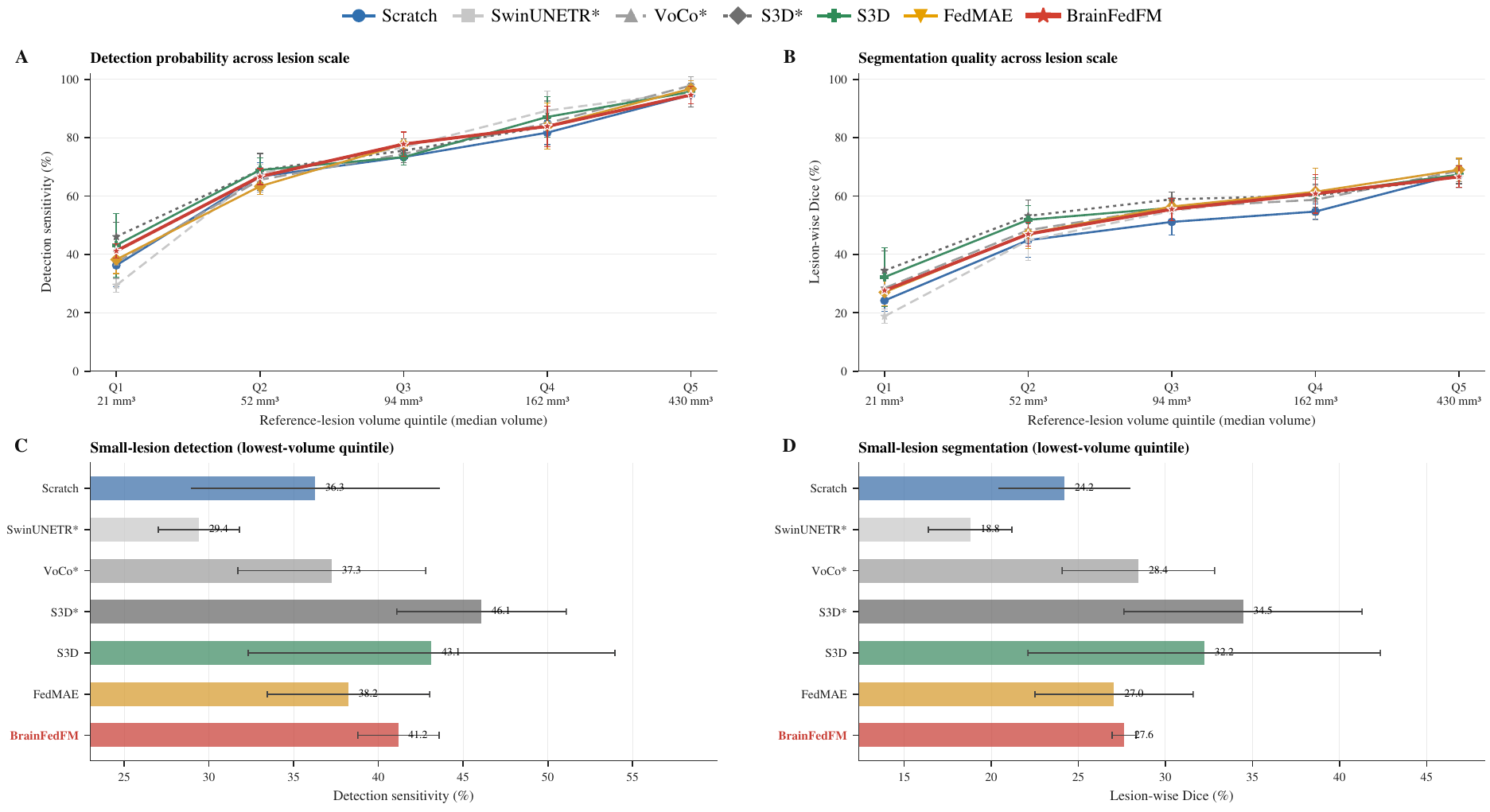}
\end{center}

\definecolor{pct100}{gray}{0.97}
\definecolor{pct75}{gray}{1.0}
\definecolor{pct50}{gray}{0.99}
\definecolor{pct25}{gray}{1.0}
\definecolor{overallgray}{gray}{1.0}
\definecolor{ourscol}{RGB}{244,248,252}

\begin{table*}[t]
\centering
\tiny
\caption[Psychiatric differential diagnosis task]{\textbf{Psychiatric differential diagnosis task.}
We evaluate foundation model performance on the UCLA-CNP benchmark. This task aims to classify brain MRI scans into schizophrenia (SCHZ), bipolar disorder (BIPOLAR), and attention-deficit/hyperactivity disorder (ADHD) categories.
Accurate cross-diagnostic psychiatric disorder classification is important for neuropsychiatric analysis, supporting consistent cohort stratification, dataset curation, and downstream model evaluation across heterogeneous psychiatric conditions.
The table summarizes performance under 25\%, 50\%, and 100\% labelled-data settings, which indicate the proportion of the labelled training--validation pool used for downstream fine-tuning and model selection; the held-out test set remains unchanged across all settings. We compare a randomly initialized scratch baseline, centralized pretrained foundation models, and federated pretrained foundation models.}
\label{tab:result_ucla}

\renewcommand{\arraystretch}{3.85}
{\scriptsize\bfseries Panel A. Summary under different labelled-data percentages.}
\vspace{5pt}

{\setlength{\tabcolsep}{3.38pt}

\resizebox{1.0\textwidth}{!}{
\begin{tabular}{
l c c c c c c
>{\columncolor{ourscol}\centering\arraybackslash}p{1.62cm}
}
\toprule
\multicolumn{2}{c}{\textbf{Brain sMRI FMs }}
& \multicolumn{4}{c}{Centralized Pretraining}
& \multicolumn{2}{c}{Federated Pretraining} \\
\cmidrule(lr){1-2}\cmidrule(lr){3-6}\cmidrule(lr){7-8}
\makecell[c]{Metric/Methods}
& Scratch
& \makecell[c]{SwinUNETR$^{\ast}$}  &\ VoCo$^{\ast}$ \ & \ S3D$^{\ast}$  & \ S3D 
& \makecell[c]{FedMAE} & \makecell[c]{BrainFedFM} \\
\midrule

\rowcolor{pct100}
\textbf{100\%  BA $\uparrow$}
&  33.33($\pm$0.000)  & 38.55($\pm$7.381)  & 33.33($\pm$0.000) & 40.40($\pm$9.999) & 33.33($\pm$0.000) & 46.47($\pm$5.151)
& \cellcolor{ourscol}   52.53($\pm$1.428) \\
\rowcolor{pct100}
\textbf{100\%   mF1 $\uparrow$}
& 16.46($\pm$3.318) & 19.78($\pm$11.33)  & 14.11($\pm$3.318) & 21.01($\pm$13.08) &  16.46($\pm$3.318) & 31.57($\pm$6.706)
& \cellcolor{ourscol}  41.23($\pm$1.230) \\
\rowcolor{pct100}
\textbf{100\%   AUC $\uparrow$}
& 54.34($\pm$8.329) & 54.94($\pm$14.94)  & 53.60($\pm$12.07) & 64.17($\pm$8.231)&  67.99($\pm$2.663)   & 70.80($\pm$3.682)
& \cellcolor{ourscol} 70.50($\pm$1.787) \\
\midrule

\rowcolor{pct50}
\textbf{50\%  BA $\uparrow$}
& 33.33($\pm$0.000) & 33.33($\pm$0.000)  & 33.33($\pm$0.000) & 33.33($\pm$0.000) & 33.33($\pm$0.000) & 34.34($\pm$1.428)
& \cellcolor{ourscol}34.34($\pm$1.428) \\
\rowcolor{pct50}
\textbf{50\%  mF1 $\uparrow$}
& 14.11($\pm$3.318) & 14.11($\pm$3.318)  &  14.11($\pm$3.318) & 14.11($\pm$3.318) & 11.77($\pm$0.000)& 13.73($\pm$2.781)
& \cellcolor{ourscol}14.00($\pm$3.157)  \\
\rowcolor{pct50}
\textbf{50\% AUC $\uparrow$}
& 48.85($\pm$1.271) & 45.68($\pm$7.914)  & 54.52($\pm$5.339) & 54.52($\pm$5.339) & 59.70($\pm$4.333) & 69.59($\pm$3.902)
& \cellcolor{ourscol} 59.00($\pm$2.240)  \\
\midrule

\rowcolor{pct25}
\textbf{25\%  BA $\uparrow$}
& 33.33($\pm$0.000)  & 33.33($\pm$0.000)  & 33.33($\pm$0.000) & 32.16($\pm$1.667) &  33.33($\pm$0.000) & 33.33($\pm$0.000)
& \cellcolor{ourscol} 33.33($\pm$0.000) \\
\rowcolor{pct25}
\textbf{25\%  mF1 $\uparrow$}
& 14.11($\pm$3.318) & 18.80($\pm$0.000)  & 14.11($\pm$3.318) & 15.11($\pm$4.728) & 16.46($\pm$3.318) & 14.11($\pm$3.318)
& \cellcolor{ourscol}  18.80($\pm$0.000) \\
\rowcolor{pct25}
\textbf{25\% AUC $\uparrow$}
& 46.45($\pm$5.373) & 49.15($\pm$3.446)  & 51.85($\pm$2.733) & 53.03($\pm$2.207) & 53.66($\pm$5.460) & 62.29($\pm$8.744)
& \cellcolor{ourscol} 46.07($\pm$0.912) \\
\bottomrule
\end{tabular}
}}

\vspace{5pt}

\begin{minipage}{1.0\textwidth}
\scriptsize
\textit{Notes. Values are mean($\pm$std) over three folds. BA, balanced accuracy; mF1, macro F1-score; AUC, macro one-vs-rest ROC AUC. Scratch uses random initialization; SwinUNETR$^{\ast}$, VoCo$^{\ast}$, and S3D$^{\ast}$ use publicly released centralized foundation model weights; S3D uses centralized pretraining on our pretraining corpus, FedMAE and BrainFedFM use federated pretraining.
Fine-tuning uses 10 epochs of AdamW, learning rate $1\times10^{-6}$, weight decay $1\times10^{-2}$, cosine scheduling with a 1-epoch warm-up, and label smoothing of 0.2. The best checkpoint is selected according to validation AUC.}
\end{minipage}
\end{table*}

\clearpage

\noindent{\scriptsize\bfseries Panel B. Labelled-data scaling and aggregate metric summaries.}
\vspace{2pt}
{\tiny\raggedright
Summarizes method behavior across labelled-data settings.
\textbf{B1}, fold-averaged trends for classification and calibration metrics; \textbf{B2}, extended metric heatmaps separating higher-is-better classification metrics from lower-is-better calibration/uncertainty metrics; \textbf{B3}, the accuracy--calibration trade-off, where points closer to the upper-left indicate stronger discrimination with better calibration.\par}
\vspace{-3pt}

\begin{center}
\includegraphics[width=1.0\textwidth,keepaspectratio]{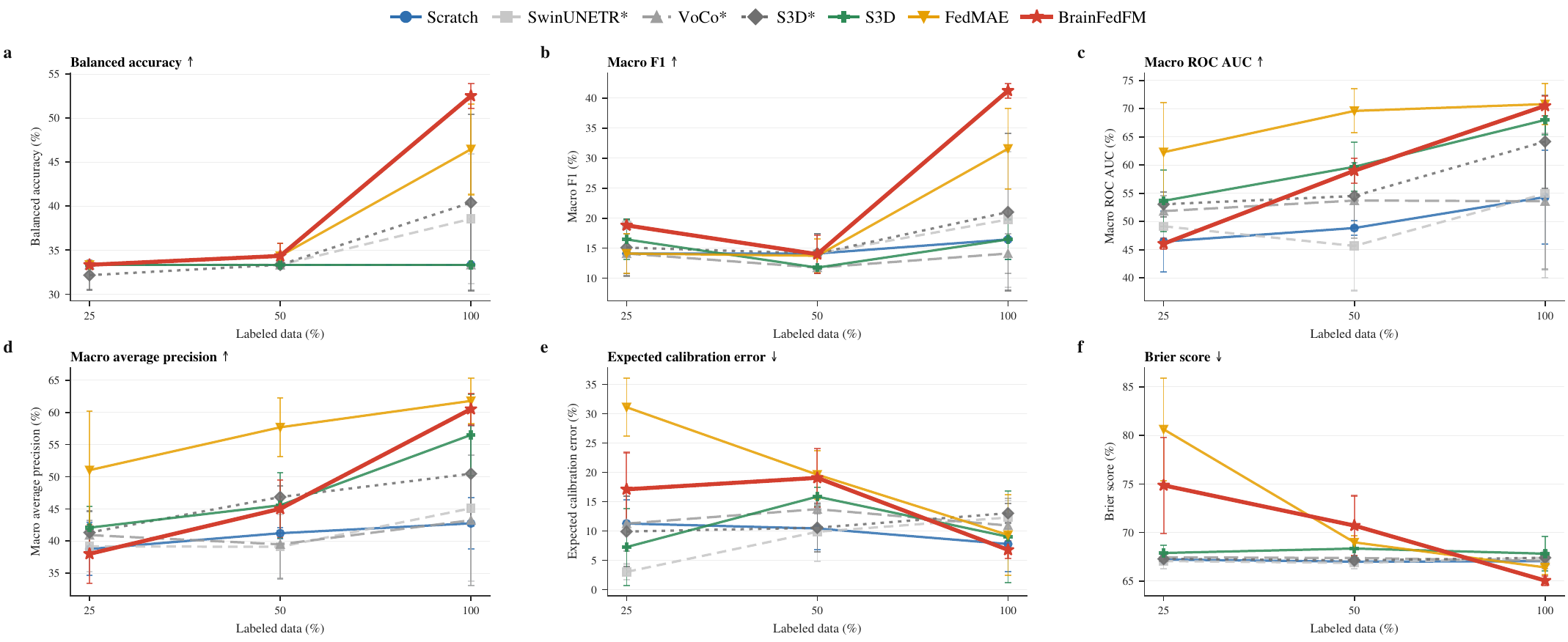} \par\vspace{3pt}
\includegraphics[width=1.0\textwidth,keepaspectratio]{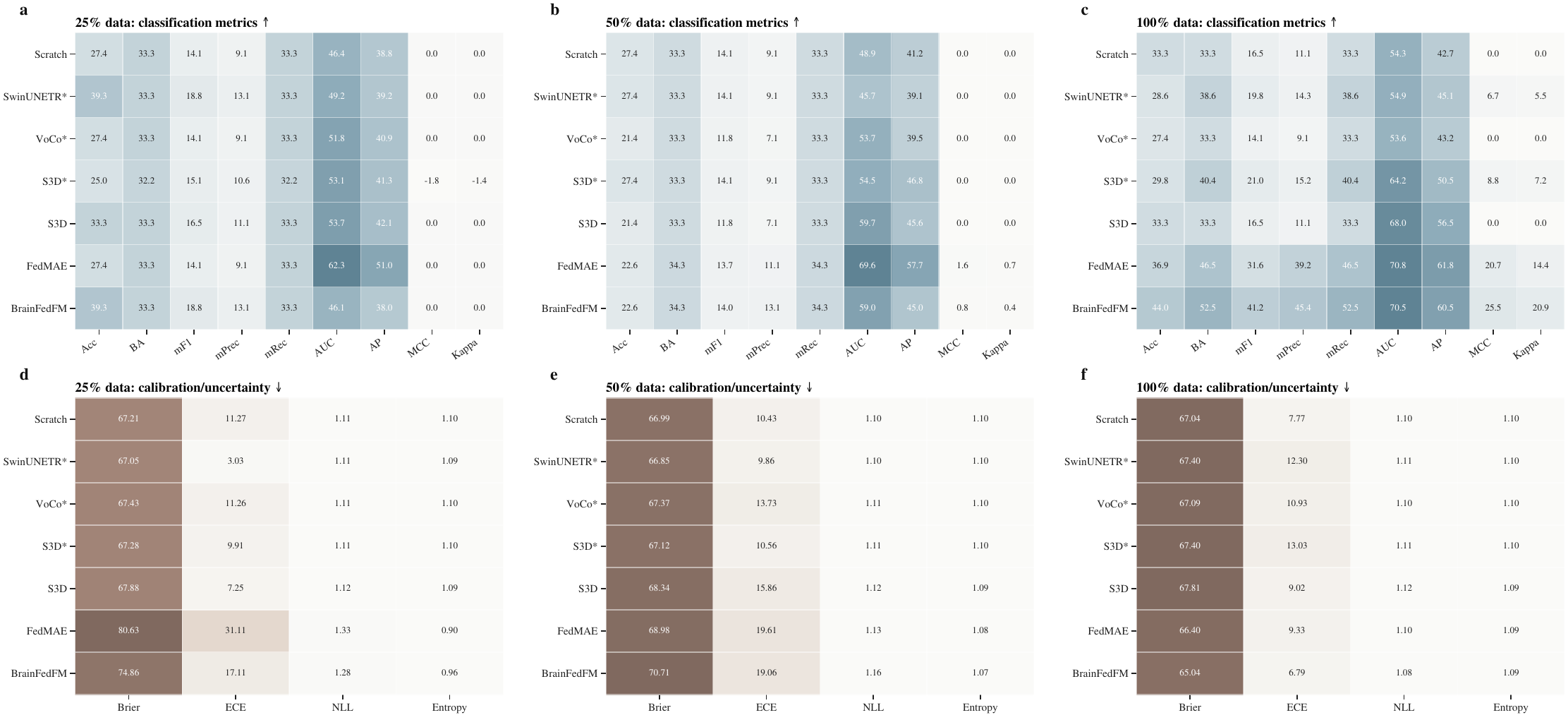} \par\vspace{5pt}
\includegraphics[width=0.935\textwidth,keepaspectratio]{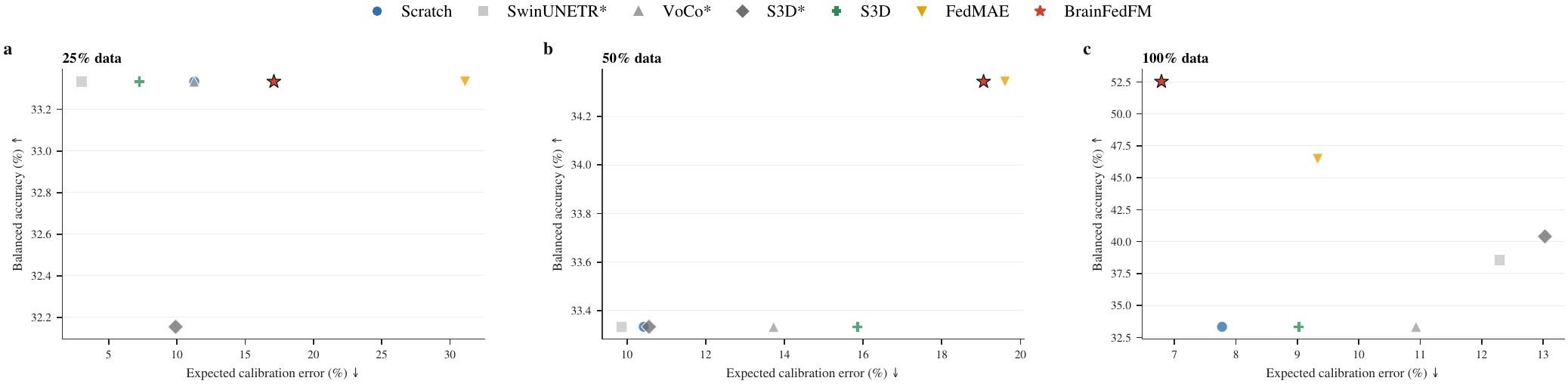}
\end{center}

\vspace{-2pt}

\noindent{\scriptsize\bfseries Panel C. Fold stability and full-data performance analyses.}
\vspace{2pt}
{\tiny\raggedright
Examines robustness and error modes beyond the aggregate scores. \textbf{C1}, fold-to-fold variability and average method ranking across balanced accuracy, macro F1, and macro ROC AUC; \textbf{C2}, probability-level analysis at 100\% labelled data, including micro-averaged ROC/precision--recall curves, reliability curves, and calibration scores; \textbf{C3}, row-normalized confusion matrices at 100\% labelled data, with rows denoting true psychiatric disorder classes and columns denoting predicted classes.\par}

\clearpage

\begin{center}
\includegraphics[width=0.84\textwidth,keepaspectratio]{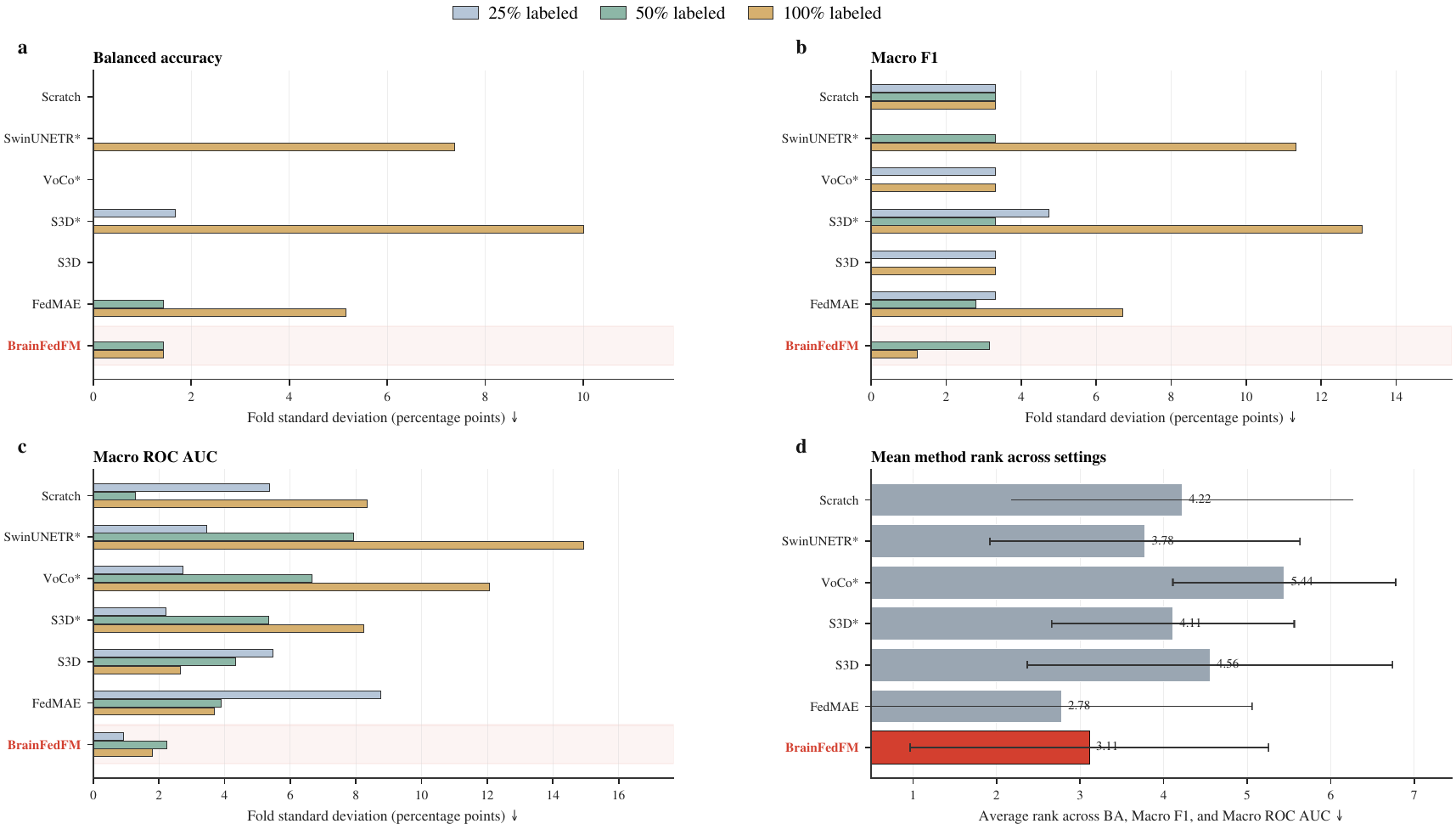}   \par\vspace{5pt}
\includegraphics[width=0.7889\textwidth,keepaspectratio]{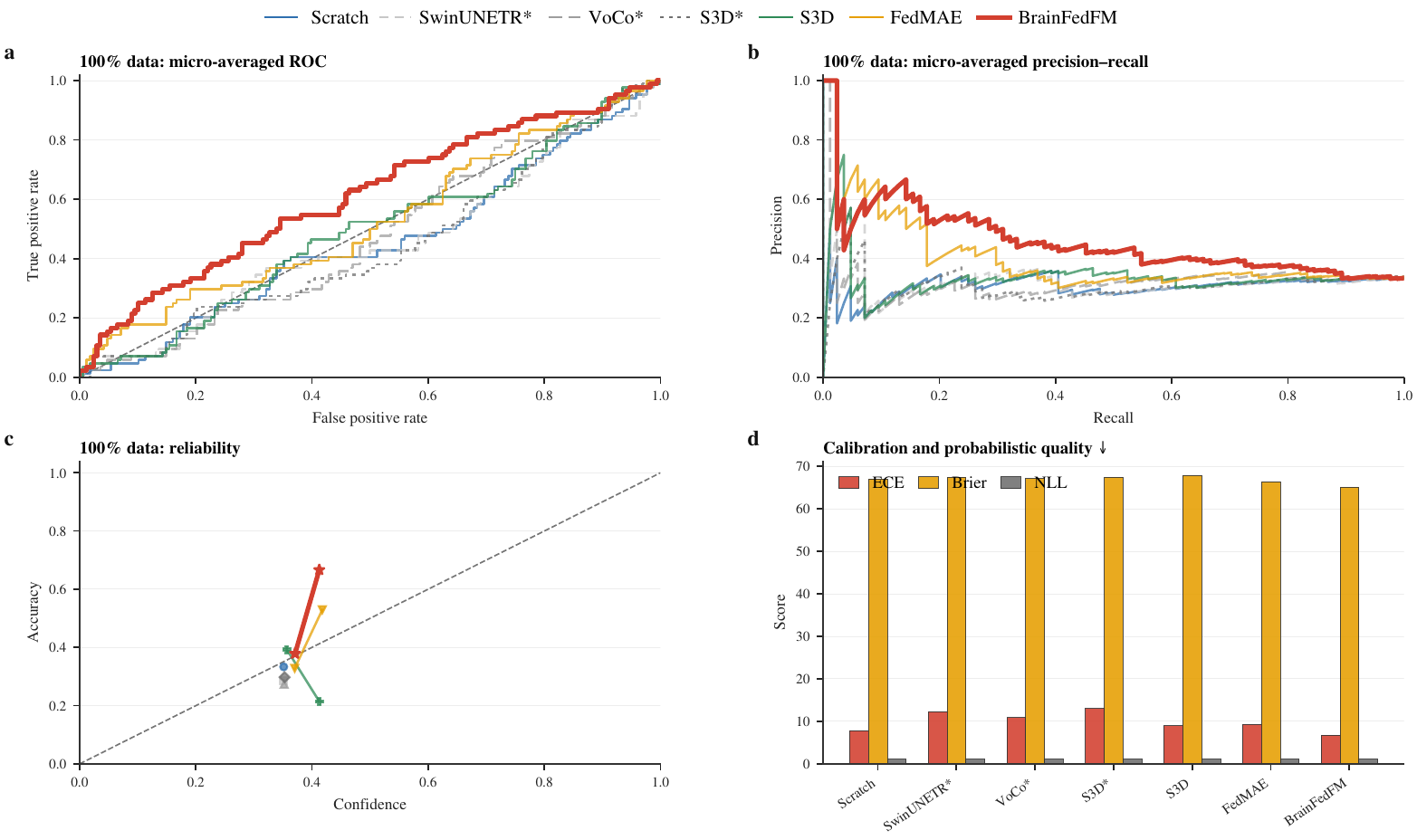}  \par\vspace{4pt}
\includegraphics[width=0.835\textwidth,keepaspectratio]{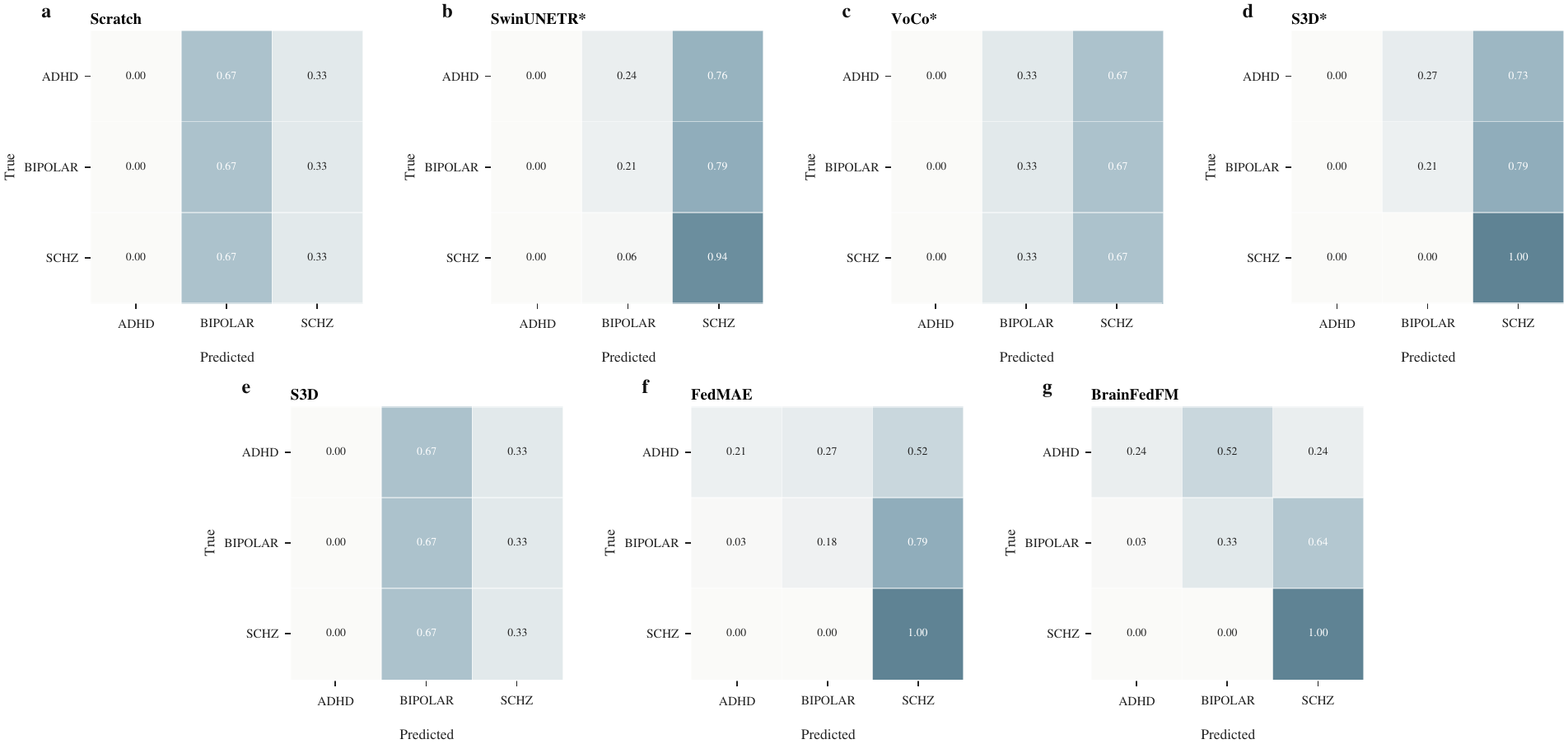}
\end{center}

\definecolor{pct100}{gray}{0.97}
\definecolor{pct75}{gray}{1.0}
\definecolor{pct50}{gray}{0.99}
\definecolor{pct25}{gray}{1.0}
\definecolor{overallgray}{gray}{1.0}
\definecolor{ourscol}{RGB}{244,248,252}

\begin{table*}[t]
\centering
\tiny
\caption[Multi-disorder psychiatric diagnosis task]{\textbf{Multi-disorder psychiatric diagnosis task.}
We evaluate foundation model performance on the SRPBS benchmark. This task aims to classify brain MRI scans into healthy control (HC), autism spectrum disorder (ASD), major depressive disorder (MDD), and schizophrenia spectrum disorder (SSD) categories.
Accurate multi-disorder psychiatric diagnosis is important for neuropsychiatric disease analysis, supporting consistent subject stratification, and cross-disorder characterization. The table summarizes performance under 25\%, 50\%, and 100\% labelled-data settings, which indicate the proportion of the labelled training--validation pool used for downstream fine-tuning and model selection; the held-out test set remains unchanged across all settings. We compare a randomly initialized scratch baseline, centralized pretrained foundation models, and federated pretrained foundation models.}
\label{tab:resultsrpbs}

\renewcommand{\arraystretch}{3.85}
{\scriptsize\bfseries Panel A. Summary under different labelled-data percentages.}
\vspace{5pt}

{\setlength{\tabcolsep}{3.38pt}

\resizebox{1.0\textwidth}{!}{
\begin{tabular}{
l c c c c c c
>{\columncolor{ourscol}\centering\arraybackslash}p{1.62cm}
}
\toprule
\multicolumn{2}{c}{\textbf{Brain sMRI FMs }}
& \multicolumn{4}{c}{Centralized Pretraining}
& \multicolumn{2}{c}{Federated Pretraining} \\
\cmidrule(lr){1-2}\cmidrule(lr){3-6}\cmidrule(lr){7-8}
\makecell[c]{Metric/Methods}
& Scratch
& \makecell[c]{SwinUNETR$^{\ast}$}  &\ VoCo$^{\ast}$ \ & \ S3D$^{\ast}$  & \ S3D 
& \makecell[c]{FedMAE} & \makecell[c]{BrainFedFM} \\
\midrule

\rowcolor{pct100}
\textbf{100\%  BA $\uparrow$}
& 39.87($\pm$0.448) & 39.03($\pm$0.741)  & 42.92($\pm$7.264) & 57.03($\pm$10.62) & 65.01($\pm$2.610) & 67.45($\pm$0.927)
& \cellcolor{ourscol} 65.92($\pm$0.688) \\
\rowcolor{pct100}
\textbf{100\%   mF1 $\uparrow$}
& 31.17($\pm$0.416) &  30.77($\pm$0.799) & 35.76($\pm$7.787) & 50.81($\pm$12.64) & 64.19($\pm$1.871) & 66.45($\pm$0.949)
& \cellcolor{ourscol} 65.39($\pm$0.551) \\
\rowcolor{pct100}
\textbf{100\%   AUC $\uparrow$}
& 81.70($\pm$1.016) & 83.21($\pm$0.914)  & 80.55($\pm$0.653) & 83.76($\pm$2.566) & 86.14($\pm$1.228) & 89.12($\pm$0.701)
& \cellcolor{ourscol} 87.72($\pm$0.831) \\
\midrule

\rowcolor{pct50}
\textbf{50\%  BA $\uparrow$}
& 29.08($\pm$3.071) & 27.85($\pm$4.029) & 38.10($\pm$2.478)  & 54.40($\pm$1.139) & 58.51($\pm$8.161)  &  66.95($\pm$1.359)
& \cellcolor{ourscol} 66.02($\pm$0.257) \\
\rowcolor{pct50}
\textbf{50\%  mF1 $\uparrow$}
& 20.01($\pm$5.436) & 15.45($\pm$8.676)  & 30.14($\pm$1.514) & 49.11($\pm$1.740) & 56.11($\pm$9.461) & 65.88($\pm$1.753)
& \cellcolor{ourscol} 65.25($\pm$0.416) \\
\rowcolor{pct50}
\textbf{50\% AUC $\uparrow$}
& 76.14($\pm$1.325) & 77.67($\pm$1.485)  & 78.89($\pm$1.258)  & 83.50($\pm$0.785) &  84.54($\pm$1.523) & 85.40($\pm$0.945)
& \cellcolor{ourscol} 85.35($\pm$0.715) \\
\midrule

\rowcolor{pct25}
\textbf{25\%  BA $\uparrow$}
& 26.95($\pm$2.759) & 30.85($\pm$4.559)  & 30.93($\pm$4.542) & 51.56($\pm$3.446) & 59.56($\pm$5.286) & 54.42($\pm$6.803)
& \cellcolor{ourscol} 57.41($\pm$7.203)\\
\rowcolor{pct25}
\textbf{25\%  mF1 $\uparrow$}
& 16.52($\pm$5.682) & 21.81($\pm$6.691)  & 22.21($\pm$7.043) & 46.38($\pm$1.925) & 55.70($\pm$7.690) & 53.60($\pm$8.919)
& \cellcolor{ourscol} 56.41($\pm$7.259) \\
\rowcolor{pct25}
\textbf{25\% AUC $\uparrow$}
& 72.93($\pm$2.105) &  76.13($\pm$2.507)  & 75.77($\pm$2.065) & 79.21($\pm$2.567) & 82.08($\pm$1.913) & 82.66($\pm$1.703)
& \cellcolor{ourscol} 83.00($\pm$1.157)\\
\bottomrule
\end{tabular}
}}

\vspace{5pt}

\begin{minipage}{1.0\textwidth}
\scriptsize
\textit{Notes. Values are mean($\pm$std) over three folds. BA, balanced accuracy; mF1, macro F1-score; AUC, macro one-vs-rest ROC AUC. Scratch uses random initialization; SwinUNETR$^{\ast}$, VoCo$^{\ast}$, and S3D$^{\ast}$ use publicly released centralized foundation model weights; S3D uses centralized pretraining on our pretraining corpus, FedMAE and BrainFedFM use federated pretraining.
Fine-tuning uses 10 epochs of AdamW, learning rate $1\times10^{-6}$, weight decay $1\times10^{-2}$, cosine scheduling with a 1-epoch warm-up, and label smoothing of 0.2. The best checkpoint is selected according to validation AUC.}
\end{minipage}
\end{table*}

\clearpage

\noindent{\scriptsize\bfseries Panel B. Labelled-data scaling and aggregate metric summaries.}
\vspace{2pt}
{\tiny\raggedright
Summarizes method behavior across labelled-data settings.
\textbf{B1}, fold-averaged trends for classification and calibration metrics; \textbf{B2}, extended metric heatmaps separating higher-is-better classification metrics from lower-is-better calibration/uncertainty metrics; \textbf{B3}, the accuracy--calibration trade-off, where points closer to the upper-left indicate stronger discrimination with better calibration.\par}
\vspace{-3pt}

\begin{center}
\includegraphics[width=1.0\textwidth,keepaspectratio]{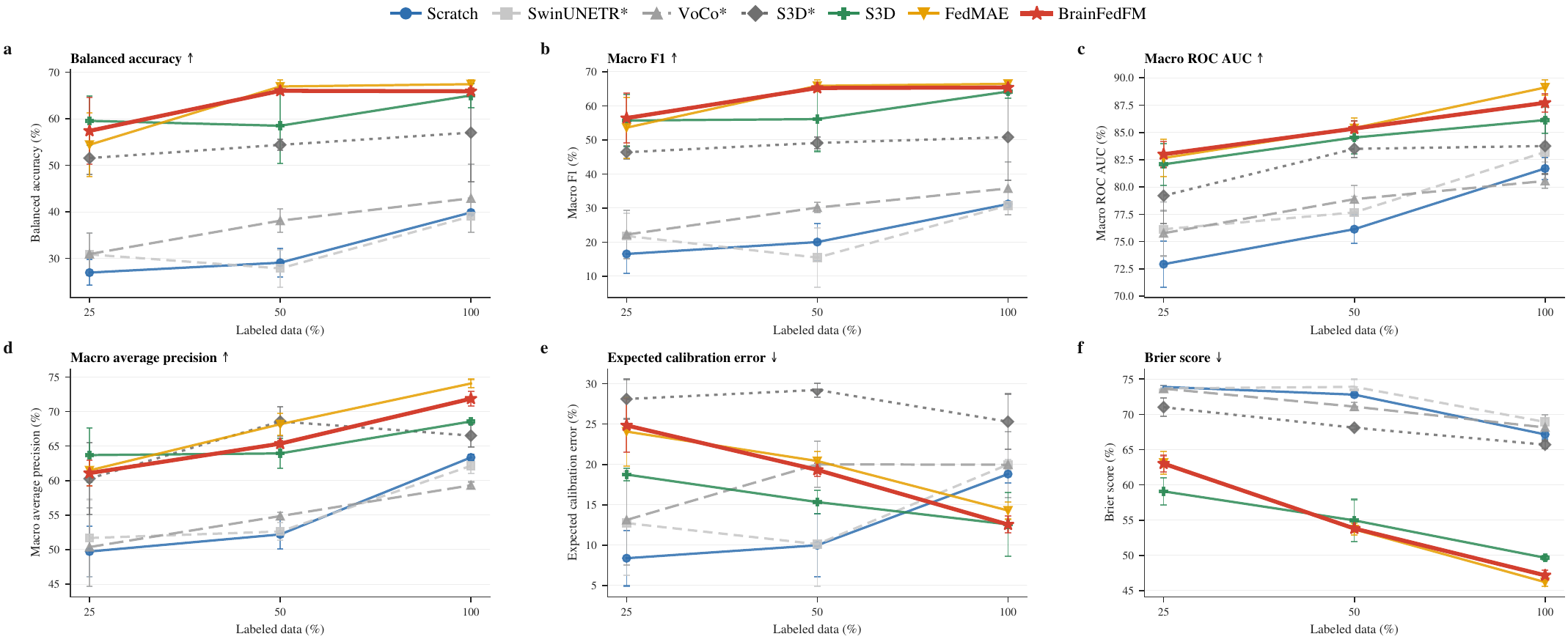} \par\vspace{3pt}
\includegraphics[width=1.0\textwidth,keepaspectratio]{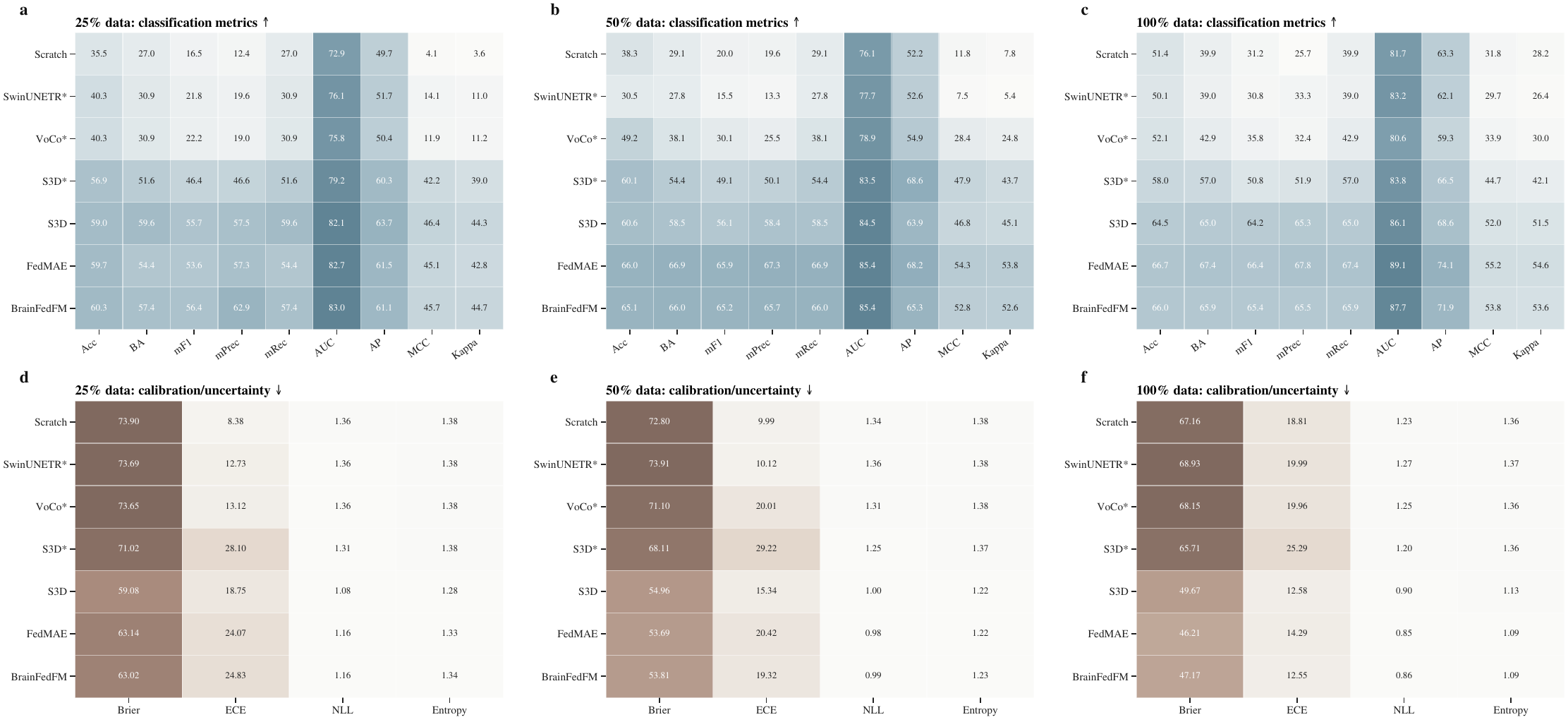} \par\vspace{5pt}
\includegraphics[width=0.935\textwidth,keepaspectratio]{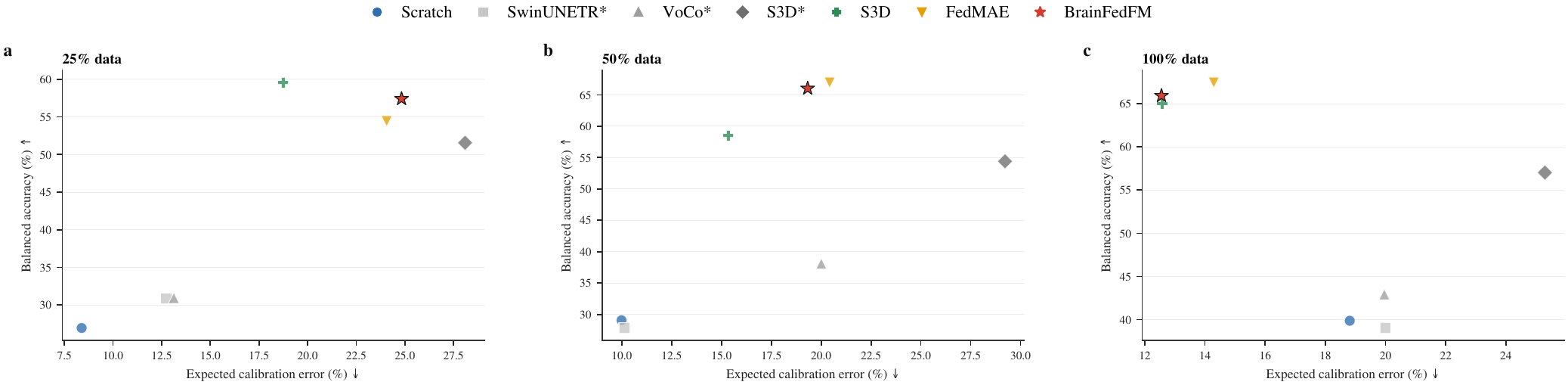}
\end{center}

\vspace{-2pt}

\noindent{\scriptsize\bfseries Panel C. Fold stability and full-data performance analyses.}
\vspace{2pt}
{\tiny\raggedright
Examines robustness and error modes beyond the aggregate scores. \textbf{C1}, fold-to-fold variability and average method ranking across balanced accuracy, macro F1, and macro ROC AUC; \textbf{C2}, probability-level analysis at 100\% labelled data, including micro-averaged ROC/precision--recall curves, reliability curves, and calibration scores; \textbf{C3}, row-normalized confusion matrices at 100\% labelled data, with rows denoting true psychiatric classes and columns denoting predicted classes.\par}

\clearpage

\begin{center}
\includegraphics[width=0.838\textwidth,keepaspectratio]{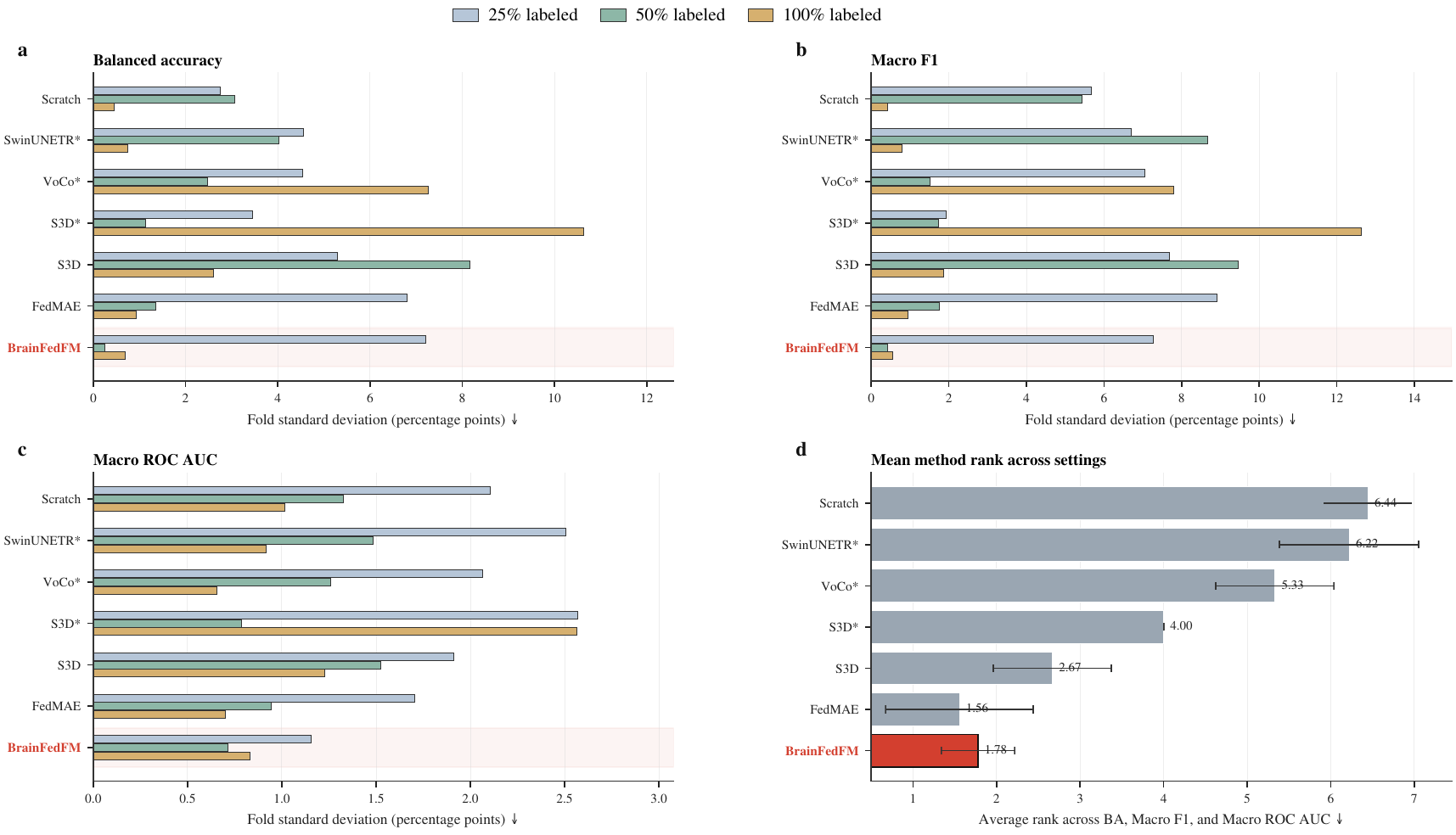}   \par\vspace{6pt}
\includegraphics[width=0.785\textwidth,keepaspectratio]{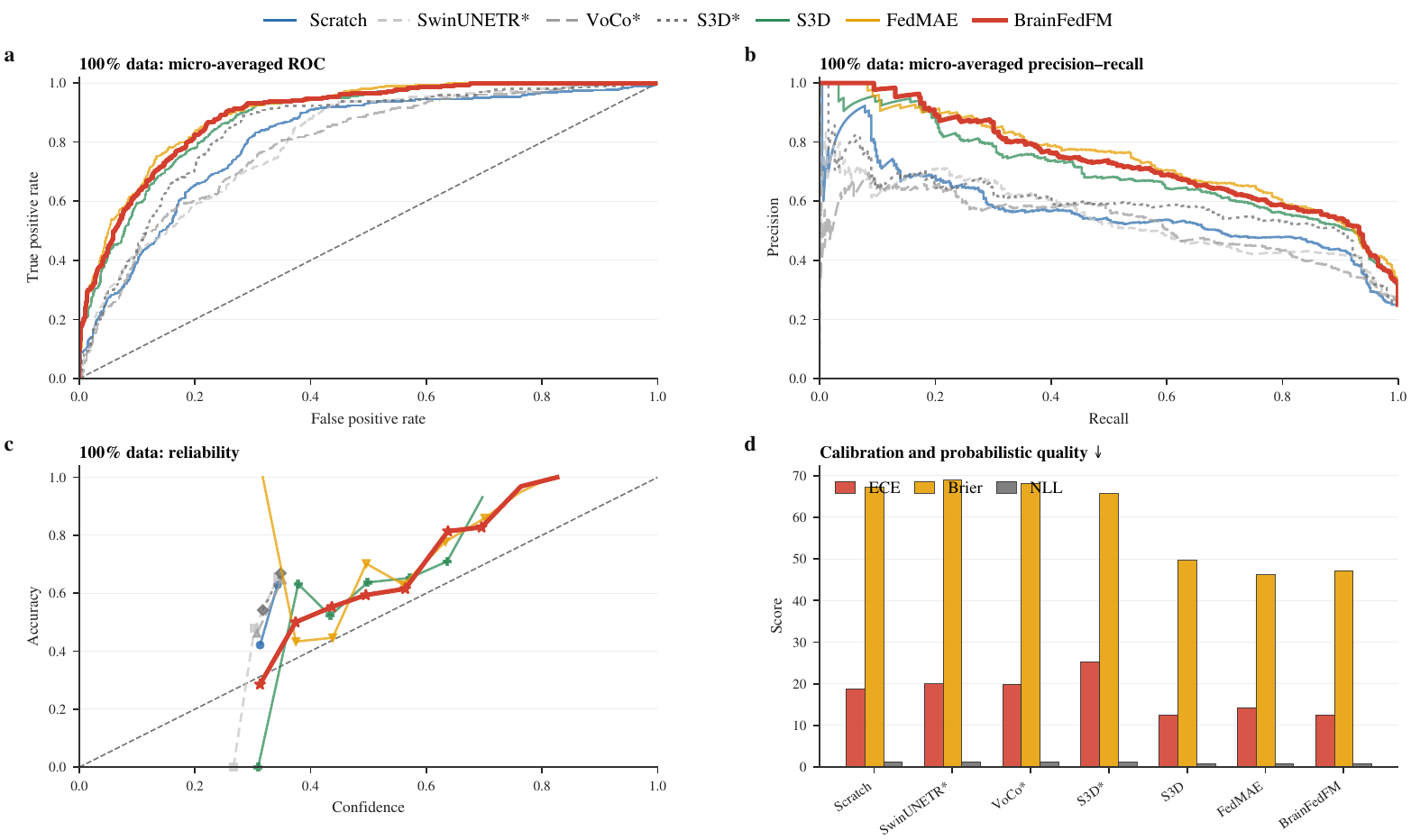}  \par\vspace{4pt}
\includegraphics[width=0.838\textwidth,keepaspectratio]{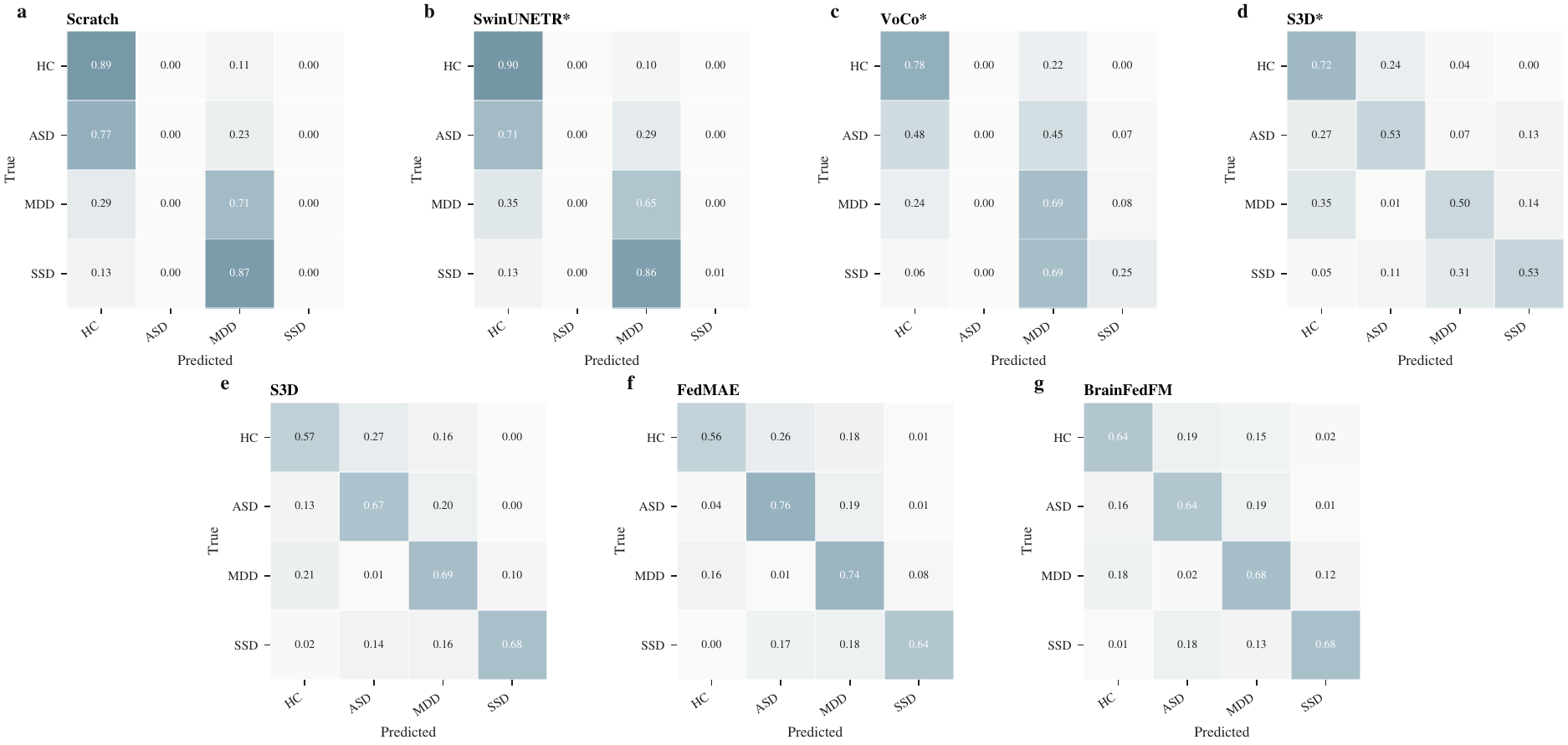}
\end{center}

\clearpage
\SupplementaryBibliography

\end{bibunit}

\end{document}